%% file: dissertation.tex
\documentclass[PhD]{dukethesis2006}
\pdfoutput=1 

\usepackage{amsmath}
\usepackage{amssymb}
\usepackage{amsthm}
\usepackage{array}
\usepackage{epsfig}
\usepackage{graphicx}
\usepackage{xy}

\usepackage[T1]{fontenc}
\usepackage{lmodern}
\usepackage[hidelinks]{hyperref}
\usepackage{multirow}
\usepackage{url}
\usepackage{subfigure}
\usepackage[dvipsnames]{xcolor}
\usepackage{bbding}
\usepackage{caption}

\usepackage{setspace}

\usepackage{algorithm}
\usepackage{algpseudocode}
\algtext*{EndWhile}
\algtext*{EndIf}
\algtext*{EndFor}
\algtext*{EndProcedure}
\usepackage[super]{nth}

\usepackage{mathtools}
\DeclarePairedDelimiter{\ceil}{\lceil}{\rceil}
\DeclarePairedDelimiter{\floor}{\lfloor}{\rfloor}

\usepackage{amsthm}
\newtheorem{thm}{Theorem}[section] 

\newcommand{\thistheoremname}{}
\newtheorem{genericthm}[thm]{\thistheoremname}

\newtheorem*{genericthm*}{\thistheoremname}
\newenvironment{namedthm*}[1]
  {\renewcommand{\thistheoremname}{#1}%
   \begin{genericthm*}}
  {\end{genericthm*}}
  
\DeclareMathOperator*{\argmin}{arg\,min} 

\usepackage[]{siunitx} 
\usepackage{bbm} 

\usepackage{refcount}

\author{Zhiyao Xie}
\advisor{Yiran Chen, Supervisor}
\member{Hai Li, Co-Supervisor}
\member{Jiang Hu}
\member{Jeffrey Derby}
\member{James Morizio}

\department{Electrical and Computer Engineering}
\title{Intelligent Circuit Design and Implementation with \\Machine Learning}

\begin{document}

\maketitle

\makeabstract

\Copyright

\include{txt/0_abstract}

\include{txt/0_acknowledgement}

\tableofcontents

\listoftables

\listoffigures


\graphicspath{{images/}, {image_power/}, {image_IRdrop/}, {image_net_time/}, {image_routability/}, {image_flowtuning/}}


\doublespacing

\input{txt/1_intro}

\input{txt/2_power}

\input{txt/3_net_time}

\input{txt/4_IRdrop}

\input{txt/5_routability}

\input{txt/6_flowtuning}

\input{txt/7_conclusion}


\newpage
\phantomsection
\addcontentsline{toc}{chapter}{Bibliography}

\singlespacing
\bibliographystyle{unsrt}
\begingroup

\endgroup


\biography

\doublespacing

Zhiyao Xie is a Ph.D. candidate at the Electrical and Computer Engineering Department of Duke University. He received his Bachelor's degree from the City University of Hong Kong in 2017. His research is mainly about efficient circuit design and innovative EDA methods, especially focused on novel ML algorithms in EDA. During his Ph.D. study, he worked as a Research Intern in multiple leading IC design or EDA companies, including Nvidia, Arm, Cadence, and Synopsys.

Zhiyao's first-authored work~\cite{xie2021apollo} received the Best Paper Award in MICRO'21. He also received the Best Research Poster Award at the Student Research Forum in ASP-DAC'22. He has authored or co-authored papers in multiple primer computer architecture and EDA conferences or journals, including MICRO~\cite{xie2021apollo}, ICCAD~\cite{chang2021auto, liang2020routing, xie2018routenet, xie2020fast}, DAC~\cite{pan2022towards}, ASP-DAC~\cite{xie2020fist, xie2021net, xie2020powernet}, DATE~\cite{huang2019routability}, and TCAD~\cite{xie2021timing}. After graduation, Zhiyao will join the ECE Department of Hong Kong University of Science and Technology (HKUST) as an Assistant Professor in Fall 2022.


\end{document}

%% file: txt/0_abstract.tex
\abstract

Electronic design automation (EDA) technology has achieved remarkable progress over the past decades. However, modern chip design is not completely automatic yet in general and the gap is not easily surmountable. For example, the chip design flow is still largely restricted to individual point tools with limited interplay across tools and design steps. Tools applied at early steps cannot well judge if their solutions may eventually lead to satisfactory designs, inevitably leading to over-pessimistic design or significantly longer turnaround time. While existing challenges have long been unsolved, the ever-increasing complexity of integrated circuits (ICs) leads to even more stringent design requirements. Therefore, there is a compelling need for essential improvement in existing EDA techniques. 

The stagnation of EDA technologies roots from insufficient knowledge reuse. In practice, very similar simulation or optimization results may need to be repeatedly constructed from scratch. This motivates my research on introducing more ``intelligence'' to EDA with machine learning (ML), which explores complex correlations in design flows based on prior data. Besides design time, I also propose ML solutions to boost IC performance by assisting the circuit management at runtime. 

In this dissertation, I present multiple fast yet accurate ML models covering a wide range of chip design stages from the register-transfer level (RTL) to sign-off, solving primary chip-design problems about power, timing, interconnect, IR drop, routability, and design flow tuning. Targeting the RTL stage, I present APOLLO, a fully automated power modeling framework. It constructs an accurate per-cycle power model by extracting the most power-correlated signals. The model can be further implemented on chip for runtime power management with unprecedented low hardware costs. Targeting gate-level netlist, I present Net$^\textbf{2}$ for early estimations on post-placement wirelength. It further enables more accurate timing analysis without actual physical design information. Targeting circuit layout, I present RouteNet for early routability prediction. As the first deep learning-based routability estimator, some feature-extraction and model-design principles proposed in it are widely adopted by later works. I also present PowerNet for fast IR drop estimation. It captures spatial and temporal information about power distribution with a customized CNN architecture. Last, besides targeting a single design step, I present FIST to efficiently tune design flow parameters during both logic synthesis and physical design.

%% file: txt/0_acknowledgement.tex
\acknowledgements
I would like to take this great opportunity to express my gratitude to everyone who has helped me during my Ph.D. study. It is impossible for me to get this Ph.D. degree without their tremendous help and support in these years.  

I first thank my advisor Prof. Yiran Chen and co-advisor Prof. Hai (Helen) Li. Five years ago, I was an undergraduate student with very limited research experience. They offered me the precious opportunity to join their CEI lab at Duke. They have provided me with so much priceless advice and support in both research and career development. Prof. Chen not only always has unique research insights, but also gives me full flexibility and strong support to explore new directions. He also managed to set up multiple great collaborations with other excellent researchers for me. 

Second, I want to thank all other committee members. Prof. Jiang Hu, who is both my committee member and a perfect collaborator, has provided tremendous great ideas and suggestions since my first research project. His expertise, patience, and diligence always impress me. 
Prof. Jeffrey Derby taught me advanced digital design and was on my committee since preliminary exam. 
Prof. James Morizio is both my committee member and my teacher on VLSI design methodologies. I also worked with him as the administrator of EDA tools at Duke. 

Also, I would like to thank my internships mentors and managers. I thank Haoxing (Mark) Ren, Yanqing Zhang, and Brucek Khailany in Nvidia Research. Besides being a great mentor, Haoxing gave me many great ideas since the beginning of my Ph.D. study. Brucek is a great manager and provides a good research environment. I thank Min Pan, Anand Rajaram, and Aiqun Cao in Synopsys. I got familiar with real industrial EDA tools with their help. I thank Xiaoqing Xu, Shidhartha Das, and Brian Cline in Arm Research. Xiaoqing mentored me with great patience and provided a lot of learning advice. Shidhartha is such a knowledgeable, experienced, and reliable mentor. I thank Jie Chen and Weibin Ding in Cadence. They provided my first internship opportunity in the EDA industry. I cannot list everyone here, but I thank all colleagues I have worked with in these companies. 

In addition, I thank my other collaborators, Ellas Fallons, Weiyi Qi, Rongjian Liang, Erick Carvajal Barboza, Guan-Qi Fang, Yu-Hung Huang, Shao-Yun Fang, Huanrui Yang, Ang Li, Chen-Chia Chang, Jingyu Pan, and Tunhou Zhang. Also, I hope to thank all members of our CEI lab at Duke University. I had a great time working in this lab and enjoy the high diversity of our research backgrounds, which makes many cross-discipline collaborations possible.  

In summary, I am extremely honored and lucky to have worked with so many excellent researchers, engineers, teachers, mentors, and friends. I am so glad that I made the right choice to pursue a Ph.D. at Duke University five years ago, when my undergraduate supervisor, Prof. Jun Fan contributed greatly to my decision.

Finally, I want to thank my mother Wenzhen Wu, my father Yong Xie, and my wife Jia Li, who married me one month ago. They always support me. The Ph.D. study is a long journey and I am so grateful to have them together with me.

 

%% file: txt/1_intro.tex
\chapter{Introduction}

\pagenumbering{arabic}

Integrated circuit (IC) is the foundation stone of the modern information society. Its complexity has been continuously growing in past decades, from circuits with merely hundreds of components to multi-billion-transistor processors or SoCs. Meanwhile, new types of designs keep emerging, from microwatt IoT devices to neural network accelerators~\cite{chen2020survey}. 
On the other hand, the pace of process technology scaling by Moore's Law~\cite{moore1965cramming}, a key enabler of performance gain, is evidently slowing down~\cite{theis2017end}. Driven by insatiable market needs for generational performance improvements, design companies are in increasingly greater demand for experienced manpower and stressed with unprecedented longer turnaround time. The nonrecurring engineering (NRE) cost associated with chip design also keeps skyrocketing accordingly~\cite{DesignCost}. Therefore, there is a compelling need for essential improvement on design efficiency through new methodologies and design automation techniques. This motivates a closer examination of existing design automation techniques.



The huge success of ICs in past decades largely hinged on the advance of electronic design automation (EDA) techniques, which handle the exponentially increasing design complexity for circuit designers. However, despite the adoption of latest commercial EDA tools, existing chip design flows are still not fully automatic in general, and the gap is not easily surmountable. For example, design steps in the flow are mostly restricted to individual tools or functions with very limited automatic coordination among them. Tools in early design stages cannot well judge if their solutions may eventually lead to satisfactory final quality of results (QoR), and a poor early solution cannot be detected until very late in the design cycle. Such disjointedness in the flow is traditionally mitigated by two common workarounds. The first option is to make pessimistic evaluations with heuristic methods, in order to ensure design closure at downstream design stages. Despite providing a fast solution, this leads to over-conservative designs with unnecessary circuit QoR loss. The second workaround is to iteratively adjust early solutions with real feedbacks from downstream stages. But considering each design iteration on complex designs may take days, the number of allowed trials can be very limited due to the stringent time-to-market requirement. 
As a result, manually explored solutions in the huge circuit design space with complex correlations may be far from the optimal solution. In summary, this workaround targets better design quality at the cost of extra turn-around time and human efforts, without any guarantee on QoR improvement or design closure. While these existing challenges have long been unsolved, the ever-increasing complexity of integrated circuits (ICs) leads to even more stringent design requirements and larger solution spaces. Driven by the compelling need for better design efficiency, we need fundamental changes in existing design methodologies! 





The stagnation of existing design methodologies roots from their weak capability of design knowledge extraction and reuse. Conventional EDA techniques may keep constructing solutions from scratch even if similar simulations or optimizations have already been performed previously, perhaps even repeatedly. To fundamentally improve this, I believe more ``intelligence'' should be introduced into existing design flows. This points to a main strength of machine learning (ML) – the capability to extract complex correlations between two separated parts based on prior knowledge. Such ``ML for EDA'' or ``ML for hardware design'' techniques have demonstrated great potential in revitalizing existing design methodologies.



\begin{figure*}[t]
\centering
\includegraphics[width=\textwidth]{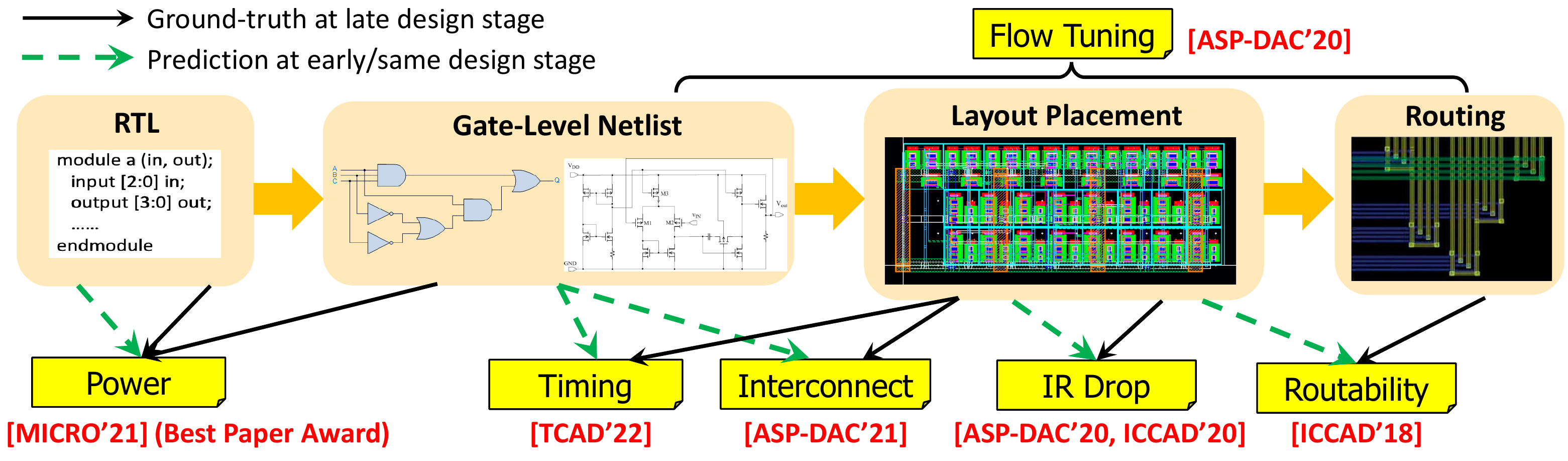}
\caption{An overview of my works presented in this dissertation.} 
\label{overall}
\end{figure*}

In this dissertation, I present multiple fast yet accurate ML for EDA methods, which cover a wide range of chip design stages from the register-transfer level (RTL) to sign-off, solving primary chip-design problems about power~\cite{xie2021apollo}, timing~\cite{xie2021timing}, interconnect~\cite{xie2021net}, IR drop~\cite{xie2020fast, xie2020powernet}, routability~\cite{xie2018routenet}, and design flow tuning~\cite{xie2020fist}. Figure~\ref{overall} shows a logical overview of my research.

Targeting the RTL stage, I present APOLLO~\cite{xie2021apollo}, a fully automated power modeling framework. It constructs a lightweight cycle-accurate power model by automatically extracting most power-correlated RTL signals as inputs. The model can be further implemented as an on-chip power meter for runtime power management. It unprecedentedly achieves high accuracy, fine temporal resolution, and low hardware implementation cost at the same time. 
Targeting gate-level netlist, I present Net$^\textbf{2}$~\cite{xie2021net} for early estimations on post-placement wirelength of each individual net. It further enables more accurate timing analysis~\cite{xie2021timing} without actual physical design information. Targeting circuit layout, I present PowerNet~\cite{xie2020powernet} for fast dynamic IR drop estimation on layouts. Both spatial and temporal information about current demand is captured with a customized convolutional neural network (CNN) architecture. It supports accurate cross-design estimations for both vertorless and vector-based IR drop. I also present RouteNet~\cite{xie2018routenet} for early routability prediction. It supports the routability prediction of mixed-size designs with different granularities. As the first deep learning-based routability estimator, some feature-extraction and model-design principles proposed in it are widely adopted by later works. Besides targeting individual design steps, I finally present FIST~\cite{xie2020fist} to efficiently tune design flow parameters during both logic synthesis and physical design,  optimizing the trade-off among power, performance, and area. It learns parameters' impact on design quality based on prior data. In the remainder of this chapter, I will introduce each work, then summarize other representative research efforts in ML for EDA.


\section{Power Modeling at RTL and Runtime}

Stringent energy-efficiency demands drive design decisions across the entire compute-spectrum, ranging from embedded applications, mobile computing to data-centers. As such, accurate power estimation is crucial for making prudent engineering trade-offs not only during CPU microarchitecture design~\cite{haj2016fine,jacobson2011abstraction,zhou2019primal,kim2019simmani} but also for runtime power management. The requirements on power estimation differ according to the target application. For instance, dynamic voltage and frequency scaling (DVFS) ~\cite{huang2012accurate,zoni2018powerprobe} is orchestrated by the system firmware and/or the operating system (OS), and hence requires {\it coarse-grained} temporal resolution in power-tracing, where each sample represents power for epochs that can be microseconds in duration.

In contrast, recent techniques for fast power management~\cite{godycki2014enabling,kasture2015rubik} and voltage boosting~\cite{hsu2015adrenaline} require fine-grained temporal resolution - for instance, a complete voltage boosting operation in ~\cite{hsu2015adrenaline} occurs in tens of nanoseconds. Similarly, voltage-noise effects such as $Ldi/dt$ noise develops in <10 cycles in modern high-performance CPUs. Therefore, quantifying the impact of fast voltage-noise and the efficacy of mitigation features such as adaptive-clocking~\cite{mair2017adaptiveclocking,bowmanJSSCAdaptiveClocking} require {\it fine-grained} temporal resolution in power-tracing~\cite{reddi2009voltage,webel2020proactive,grochowski2002}, where a sample exists for every CPU cycle (per-cycle temporal resolution).


{\bf Design-Time Power Modeling:} For fine-grained power-tracing, CPU design teams typically rely on industry-standard power analysis tools such as \cite{powerpro} to replay simulation vectors at the RTL or gate-level with back-annotated parasitics. Power is computed from the switching statistics of individual signal nets and the capacitive load that they drive. 
This approach is very accurate and serves as the signoff standard, but it comes with a very high computational cost. 
It does not scale for the analysis on long-running workloads and/or simulating the simultaneous execution of multiple CPU cores.


An alternative approach relies upon FPGA-based netlist emulation~\cite{emulation-platform} to address the speed impact of power estimation. In this approach, a simulation trace is generated from FPGA, then the extracted switching statistics are processed using power analysis EDA software~\cite{powerpro} to obtain power traces. 
However, per-cycle power tracing is still onerous using this approach due to the significant storage constraints on modern computer servers. 
Our own studies demonstrate storage requirements in excess of 200GB for a 17-million cycle simulation, leading to infeasible execution time using power analysis tools. Thus, this approach is typically restricted to coarse-grained temporal resolution where power tracing is averaged over millions of CPU cycles.

{\bf Runtime Power Estimation:} Previous works have demonstrated runtime regression models using hardware performance monitoring event-counters to guide OS-orchestrated DVFS~\cite{JosephISLPED01, bircher2007complete, walker2016accurate}. These models average counter-values that accumulate specific micro-architectural events, such as L2 cache misses and the number of retired instructions, across thousands or millions of CPU cycles. However, these events typically exhibit poor correlations to per-cycle micro-architectural activity. 
Furthermore, the process of averaging over long CPU cycles renders these approaches significantly inaccurate when fine-grained power tracing is required.

Recently, RTL-based runtime power monitoring with on-chip power meter (OPM) 
\cite{najem2016design,cremona2020automatic,pagliari2018all,
zoni2018powerprobe, zoni2018powertap} has been proposed to improve temporal resolution at the expense of dedicated hardware circuit. However, existing techniques struggle to simultaneously achieve high resolution and low hardware area overhead. For example, 
the work in \cite{najem2016design} restricts area overhead
to 1.5-4\%, but its highest temporal resolution is 2500 clock cycles. A recent work~\cite{zoni2018powertap} improves resolution to 100 cycles, but with significant area overhead (4-10\%). Thus, there are undesirable trade-offs between accuracy, speed, temporal-resolution, and on-chip hardware overhead that render the prior art unsuitable for fine-grained power estimation.

In Chapter~\ref{chapter:power}, I present APOLLO, a unified power modeling framework addressing both the design-time and runtime challenges with a consistent model structure. The centerpiece of APOLLO is a new power proxy selection technique based on minimax concave penalty (MCP) regression. It enables per-cycle power tracing for benchmarks executing over millions of CPU cycles. For runtime monitoring, it provides per-cycle accurate power estimation with $0.2\%$ area overhead. APOLLO is the first power monitoring technique with cycle-accuracy and sub-$1\%$ area overhead. Moreover, the proxy selection process in APOLLO is fully automated and thereby extensible to new designs. Compared to PRIMAL~\cite{zhou2019primal}, a recent machine learning approach, APOLLO reaches similar accuracy but is orders of magnitude faster. APOLLO also significantly outperforms Simmani~\cite{kim2019simmani}, another state-of-the-art work, on both accuracy and computation speed. Moreover, APOLLO achieves both fine-grained temporal resolution and lower hardware overhead than \cite{pagliari2018all}, a recent OPM technique.


\section{Net Length and Timing Modeling at Netlist}

In modern VLSI design, logic synthesis plays a critical role by mapping design RTL into netlists with logic gates. Previous studies~\cite{xie2020fist} show that different logic synthesis solutions can result in $3\times$ difference in power and more than one clock cycle difference in slack at the sign-off stage. 
As the design complexity keeps increasing, logic synthesis may not generate the netlist with the highest quality, because it lacks a credible prediction on the QoR of synthesized netlists at subsequent design stages like placement and routing (P\&R). For example, estimated slacks from the synthesis tool can be largely different from sign-off static timing analysis (STA) results. To alleviate such poor predictability at the early stage, more design iterations are required to reach an optimized design quality, thus largely increase the overall turnaround time.

To improve the design predictability, a recent industrial trend among commercial EDA flows~\cite{Fusion2,ispatial} takes an ambitious goal to explicitly address the interaction between logic synthesis and layout. Commercial synthesis tools~\cite{Genus} provide increasingly better support on physical-aware logic synthesis by directly integrating both placement and optimization engines from the physical design tool~\cite{Innovus} into its logic synthesis process. 
Such a trend in the EDA industry has demonstrated the importance of predictability at early design stages and its large impact on the final chip quality, but this solution is costly. Directly invoking placement and optimization engines during synthesis can be highly time-consuming. 

Besides invoking core engines at downstream design stages, in recent years, ML techniques have been widely adopted to improve the predictability in the chip design flow. However, a large portion of these ML methods only focus on post-placement predictions. Predictions at earlier stages on netlist are more challenging due to the absence of placement information. 
Existing estimators~\cite{bodapati2001prelayout,fathi2009pre, liu2012neural} on net length, a fundamental design information related to both power and timing, still cannot achieve very high accuracy. Recent ML techniques tend to only estimate the overall wirelength of a netlist~\cite{liu2012neural} or lengths of a few selected paths~\cite{hyun2019accurate} for better accuracy, rather than predicting the length of each individual net. However, during synthesis, the knowledge of individual net sizes can help to identify potentially long-wire nets in any path and guide transformations focusing on them. 
Also, due to the absence of individual net length information, the wire load cannot be accurately estimated, making accurate timing prediction also extremely challenging.
To the best of our knowledge, detailed pre-placement ML estimator on timing, one of the most important design objectives, is still not available. 
In summary, individual net length and timing are two important and correlated design objectives that are difficult to predict before placement. In Chapter~\ref{chapter:net_time}, I address the problem by a pre-placement prediction flow with estimators on both net length and timing.

\section{Fast IR Drop Modeling on Layout}

Dynamic IR drop describes the deviation of the power supply level from its specification caused by localized power demand and switching patterns. It must be restricted in order for a circuit to meet its timing target and function properly. As such, it is vitally important to verify if IR drop
satisfies design constraints and identify constraint violation regions, a.k.a. hotspots. As chip complexity continues to grow, IR drop evaluation becomes increasingly challenging.

In industrial designs, dynamic IR drop estimation is often obtained from simulation-based commercial tools, which are accurate but very time-consuming. 
ML-based approaches~\cite{yamato2012fast, ye2014chip, lin2018ir, fang2018machine} have been explored to achieve faster estimation. These works predict dynamic IR drop of each cell through features such as cell positions, timing windows, path resistance, etc. with supervised machine learning techniques.

A major weakness shared by most previous works is that they are not ``design-independent'', i.e., {\em transferable} to new designs that are not seen in its training dataset. They need to train a new model for each distinct design. 
In addition, most prior works only focus on vector-based analysis, ignoring vectorless IR drop. For dynamic IR drop, the peak IR drop in the design can be analyzed either using vectorless analysis or vector-based analysis using simulation patterns from value change dump (VCD) files. Vectorless IR drop analysis is highly desirable for IR mitigation during physical design for two main reasons. First, for a large chiplet, vector-based IR drop analysis requires a huge number of simulation patterns to cover most regions and thus can be unbearably slow. Second, designers are unable to obtain accurate power simulation patterns early in the design process. For large industrial designs, multiple teams work on different RTL units in parallel and the overall simulation patterns change throughout the design process. Vectorless IR drop provides a faster and earlier estimation in this case. However, accurate estimation of vectorless IR drop is more difficult than vector-based due to the increased diversity in switching activity distribution. 


In Chapter~\ref{chapter:ir_drop}, I present a CNN-based method PowerNet, which provides a transferable ML model for both vectorless and vector-based IR drop estimations. I emphasize more on vectorless estimation, considering its higher difficulty and usability. PowerNet addresses these challenges by its innovative preprocessed features and CNN architecture. In previous works \cite{fang2018machine}, design dependent features such as coordinates and timing information of each cell are directly fed into the ML model. Since locations and timing do not directly cause IR drop, fitting a model on these features directly tends to introduce the {\em overfitting} problem, making the model inaccurate on unseen designs. 
Instead, design-dependent information should be preprocessed to correlate with IR drop before feeding to ML models. It is known that IR drop directly correlates with cell power consumption. Therefore, PowerNet carefully incorporates these design-dependent features into power maps. 
Then it utilizes an innovative CNN architecture to capture the maximum transient IR drop. 

\section{Early Routability Modeling on Layout}

Every chip design project must complete routing without design rule violation before tapeout. However, this basic requirement is often difficult to be satisfied especially when routability is not 
adequately considered in early design stages. In light of this fact,
routability prediction has received serious attention 
in both academic research and industrial tool development.
Moreover, routability is widely recognized as a main objective for cell placement. 

In industrial designs, fast trial global routing is often employed for routability prediction at placement stage~\cite{GLARE}. The ``fast'' here is relative to full-fledged global router that generates
solutions for further detailed routing. Such trial global routing is still too slow from the routability prediction point of view, as it is called many times within placement engine. Probabilistic prediction~\cite{LouISPD01,WestraISPD04} and other fast alternatives~\cite{spindler2007fast} have been developed. However, their sacrifice on accuracy is quite significant and trial global routing is still the de facto standard despite its costly runtime~\cite{GLARE}.

In addition to forecasting overall routability, one also needs to predict locations of design rule checking (DRC) hotspots where routability optimization engines can be applied to fix them.
Evidently, predicting hotspot locations is much more difficult than forecasting overall routability, which is often indicated by design rule violation (DRV) count. In this case, even global routing is not accurate enough~\cite{chan2017routability} due to complicated design rules imposed upon design layout for manufacturing. Overall, global routing is neither fast enough for overall routability forecast nor accurate enough for pinpointing DRC hotspots.

To find accurate yet fast routability prediction approach, people explored machine learning techniques. 
Multivariate adaptive regression spline (MARS) and support vector machine (SVM)-based routability forecast without using global routing information is proposed in~\cite{chan2016beol}. However, this technique does not indicate how to handle macros, which prevail in modern chip designs and considerably increase the difficulty of routability prediction~\cite{zhou2015accurate}. In~\cite{chan2017routability}, an SVM-based method is introduced for predicting locations of DRC hotspots. It shows accuracy improvement compared to global routing on testcases without macros. Cases without macros allow this method to use a set of small regions in a circuit as a training set and predict routability of other regions of the same circuit. Such convenience disappears when macros present, as a large region (often entire layout) of a circuit needs to be ``receptive'' to the ML model for accurate predictions.

In Chapter~\ref{chapter:routability}, I present a routability estimator named RouteNet to solve two problems: 1. Fast routability forecast for cell placement in terms of the number of design rule violations (\#DRV) such that a few relatively routable placement solutions can be identified among many candidate solutions; 2. Prediction of DRC hotspot locations such that the few identified solutions can be proactively modified to prevent design rule violations. In both problems, we will consider macros, which are prevalent in modern industrial designs. The approach is built upon CNN, which has not been investigated for routability prediction before this method is proposed.


\section{Design Flow Tuning}

Modern industrial chip design flows are immensely complex. A design flow might have multiple steps, each step might have multiple functions and each function can be configured with many parameters. Consequently, industrial flows may have hundred-thousand lines of scripts and are configured with thousands of parameters.

The impact of parameter settings on overall design quality is phenomenal, thus industrial design teams will tune flow parameters as best as they can. Flow parameters are usually tuned manually based on designers' experiences. Because industrial design flows would take several hours or days to run on large designs, the manual parameter tuning process can be very time-consuming, especially for novice designers. Consequently, design turn-around time is stretched long or design quality is compromised with an inadequate exploration of parameters.
 
Therefore, automatic design flow parameter tuning is highly desirable. However, due to the difficulty of collecting vast amounts of design flow data for implementing synthesis and physical design flows, there are few published methods in this area before this work. A genetic algorithm (GA)-based flow tuning method is proposed in \cite{ziegler2016synthesis}, where genetic algorithm explores different parameter settings to find the optimal one without learning 
the effect of different parameters. This would suffer from the need to run more samples to find a good solution. The work of \cite{ziegler2016scalable} then introduces a customized learning approach to predict possible parameter settings for the next sampling iteration. Both works are highly customized to a company's in-house flow without many details disclosed and thus difficult to generalize.

In Chapter~\ref{chapter:flow}, I present a Feature-Importance Sampling and Tree-Based (FIST) method to conduct design flow parameter tuning. FIST learns the impact of parameters from previously well-explored designs and fully utilizes such information in its sampling process. Some works in design space exploration (DSE) also introduced prior knowledge transfer. For example, \cite{papamichael2015nautilus} improves genetic algorithm by guiding DSE with expertise from IP authors. However, this technique requires human knowledge, while FIST learns the prior knowledge automatically and transfers the learning to new designs. Furthermore, FIST leverages an efficient ML model XGBoost \cite{chen2016xgboost} and proposes a dynamic model adjustment method to overcome the overfitting problem in the early stages of parameter tuning.

\section{Summary of Works on ML for EDA}
\label{sec:related}

In recent years, ML for EDA or hardware design has become a trending topic~\cite{huang2021machine, rapp2021mlcad}. Besides works~\cite{xie2021apollo, xie2021timing, xie2021net, xie2020fast, xie2020powernet, xie2018routenet, xie2020fist} presented in this dissertation, many other research contributions on this topic also worth mentioning. We can observe ML models being applied to almost all design stages of a typical VLSI design flow. For high-level synthesis (HLS), models have been proposed for fast result estimation~\cite{dai2018fast, makrani2019pyramid} and design space exploration~\cite{liu2013learning, liu2016efficient}. Many power models~\cite{xie2021apollo, zhou2019primal, kim2019simmani, zhang2020grannite} are also proposed in early design stages. At logic synthesis, ML models are proposed for chip quality prediction~\cite{yu2018developing, neto2019lsoracle} and optimization~\cite{hosny2020drills, neto2021slap}.
During physical design, more works perform predictions or optimizations on almost all important design metrics, including timing~\cite{barboza2019machine, kahng2015si}, macro placement~\cite{mirhoseini2021graph}, routability~\cite{xie2018routenet, chen2020pros, huang2019routability, chang2021auto, pan2022towards}, IR drop~\cite{xie2020powernet, ho2019incpird, zhou2020gridnet}, clock tree quality~\cite{lu2019gan}, crosstalk~\cite{liang2020routing}, 3D integration~\cite{lu2020tp}, etc. Also, many ML models have been developed for design verification~\cite{katz2011learning, fine2003coverage}, design for testability (DFT)~\cite{ma2019high}, and lithography problems~\cite{yang2018layout, yang2019gan, ye2019lithogan}. Besides the methods applied at specific design stages, the design flow tuning~\cite{kwon2019learning, xie2020fist} also attracted considerable attention in ML for EDA. 



ML methods are of course not only limited to digital designs. For analog design, similarly, various models have been developed for topology design~\cite{kunal2020general, li2016analog}, device sizing~\cite{wang2020gcn, hakhamaneshi2019bagnet}, pre-layout estimation~\cite{ren2020paragraph, kunal2020gana}, layout evaluation~\cite{liu2020towards, li2020customized}, layout generation~\cite{xu2019wellgan, liu2020closing, zhu2019geniusroute}, and analog design testing~\cite{stratigopoulos2008error}.

Besides being a hot research topic in academia, ML-based estimators also gained popularity in the EDA industry. Recent versions of commercial tools already support the construction of ML models on delay~\cite{Innovus} or congestion predictions~\cite{ICC2}, providing improved PPA or faster convergence after invoking the ML models in their tools~\cite{Innovus, ICC2}. 
In addition, EDA vendors have provided ML models for design space exploration or design flow tuning, named DSO.ai~\cite{dsoai} and Cerebrus~\cite{cerebrus}.

Among all these ML applications targeting both digital or analog designs, almost all popular ML techniques have been applied, covering supervised, unsupervised, and reinforcement learning algorithms. In most recent works, a strong trend towards neural network (NN)-based algorithms, especially deep learning techniques, can be observed~\cite{rapp2021mlcad}. In summary, ML for EDA has an impressive impact in both EDA academia and industry. We have strong reasons to believe ML models will play more important roles in design automation in the future.

%% file: txt/2_power.tex
\chapter{Power Modeling at RTL and Runtime}
\label{chapter:power}

\section{Background}

\begin{table*}[th!]
  \centering
    \caption{Comparison among various power modeling approaches. The percentage numbers are hardware overheads measured in area.}
      \label{prev_works}
         \hspace{-.07in}
  \resizebox{\linewidth}{!}{ \renewcommand{\arraystretch}{1.3} 
  \begin{tabular}{|c|c|c|c|c|c|}
    \hline
    \textbf{Methods (Hardware}  &  \textbf{Demonstrated}  & \textbf{Model} & \textbf{Temporal} & \textbf{PC / Proxy}  & \textbf{Cost or}  \\
    \textbf{Overhead in Area\%)} &  \textbf{Application}  &  \textbf{Type} &  \textbf{Resolution}  &  \textbf{Selection}   & \textbf{Overhead}  \\
    \hline
    \hline
    \cite{brooks2000wattch, jacobson2011abstraction, lee2006accurate, li2009mcpat, rethinagiri2014system}  \ 
                & \multirow{3}{*}{Design-time}   & \multirow{1}{*}{Analytical}   & \multirow{1}{*}{>1K cycles} &  \multirow{1}{*}{N/A}   &   \multirow{1}{*}{Low}  \\
    \cline{1-1} \cline{3-6} 
    \cite{zhang2020grannite}  \
                &     & \multirow{4}{*}{Proxies}       & \multirow{2}{*}{>1K cycles}   &  \multirow{3}{*}{Automatic}  &  High \\
    \cline{1-1}  \cline{6-6}
    \cite{bogliolo2000regression, shao2014aladdin}    \ 
                &  software model    &    &   &  \multirow{3}{*}{or no selection} & Medium \\
    \cline{1-1} \cline{4-4} \cline{6-6} 
    \cite{zhou2019primal}  \
                &        &        & \multirow{2}{*}{Per-cycle}   &   &   High \\
    \cline{1-1} \cline{6-6}
    \cite{brooks2003new, lee2015dynamic, kumar2019learning, ye2000design, wu1998cycle}  \
                &     &        &         &        &   Medium \\
    \hline
    \hline
    \cite{coburn2005power} \ \ \textbf{(300\% overhead)}  &  \multirow{3}{*}{Design-time}  &  \multirow{4}{*}{Proxies}  &  \multirow{2}{*}{Per-cycle} &   \multirow{3}{*}{Automatic} &  High  \\
     \cline{1-1} \cline{6-6}
     
   \cite{yang2015early} \ \ \textbf{(16\% overhead)} \  
                &    \multirow{3}{*}{FPGA emulation}   &        &      &    &  \multirow{3}{*}{Medium}  \\

    \cline{1-1} \cline{4-4}
    \cite{kim2019simmani}     &        &       & $\sim$100s cycles  &    &  \\
    \cline{1-1} \cline{4-5}
    \cite{sunwoo2010presto}   &     &        &  Per-cycle  &  Manual+auto   &   \\
    \hline
    \hline
    \cite{bellosa2000benefits, bertran2010decomposable, bircher2007complete, gilberto2005power, goel2010portable, huang2012accurate, isci2003runtime} \           &  \multirow{6}{*}{Runtime monitor}   & \multirow{2}{*}{Event}     & \multirow{2}{*}{>1K cycles}   &   \multirow{3}{*}{Manual}  &   \multirow{3}{*}{Low}  \\
     \cite{oboril2015high, pricopi2013power, rodrigues2013study, sagi2020lightweight, singh2009real, walker2016accurate, JosephISLPED01} &   &  \multirow{2}{*}{counters}  &   &  &  \\
    \cline{1-1} \cline{4-4} 
    \cite{kalyanam2017power} \ 
                &      &       & $\sim$100s cycles    &         &     \\
    \cline{1-1} \cline{3-5} \cline{6-6}
    \cite{cremona2020automatic} \textbf{(2-20\%)}, \cite{najem2016design} \textbf{(1.5-4\%)}  \
                &      & \multirow{3}{*}{Proxies}      & \multirow{2}{*}{>1K cycles}   & \multirow{3}{*}{Automatic}  &   \multirow{3}{*}{Medium}  \\
    \cite{pagliari2018all} \textbf{(7\%)}  &   &   &    &    &       \\
    \cline{1-1} \cline{4-4} 
         \cite{zoni2018powertap} \textbf{(4-10\%)}, \cite{zoni2018powerprobe} \textbf{(7\%)}   \
                &     &       & $\sim$100s cycles  &    &   \\
    \hline
    \hline
    \multirow{2}{*}{APOLLO \ \textbf{(0.2\% overhead)} }   \
                & Design-time model     & \multirow{2}{*}{Proxies}      & \multirow{2}{*}{Per-cycle}  &  \multirow{2}{*}{Automatic}  & \multirow{2}{*}{Low}\\
                & Runtime monitor    &         &           &        &  \\
    \hline
  \end{tabular}
  } 
\end{table*}


Power is a primary design objective and power modeling is an extensively studied topic. Table~\ref{prev_works} summarizes representative power estimation approaches, which can be categorized into design-time power models and runtime on-chip power meters.

\textbf{Design-Time Power Models:} Many design-time approaches~\cite{brooks2000wattch,li2009mcpat,lee2006accurate,jacobson2011abstraction, rethinagiri2014system} construct analytical models for micro-architectural power estimation by collecting statistics from performance simulators~\cite{binkert2011gem5, binkert2006m5}. 
Wattch~\cite{brooks2000wattch} is an architectural dynamic power simulation tool using a linear model, and McPAT~\cite{li2009mcpat} integrates power, area, and timing in a modeling framework. Each functional unit is characterized and attributed a power value when activated. Multiple active units are then added together to compute the overall power~\cite{grochowski2002}. However, this approach cannot handle internal variations in power consumption due to data- and control-dependent variations in workload. Therefore, these models are preferably used as an average over thousands or millions of CPU clock cycles. 
Additionally, inaccuracies have been observed~\cite{xi2015quantifying, lee2015powertrain, rethinagiri2014system} for McPAT on new designs.

Design-time models on selected RTL power-proxies are employed to perform power simulations. 
Early works~\cite{bogliolo2000regression, wu1998cycle, ye2000design} construct macro-models to abstract power estimations for small circuit modules with thousands of gates. 
In recent years, machine learning (ML) techniques are exploited. Lee, et al.~\cite{lee2015dynamic} adopt gradient boosting and Kumar, et al.~\cite{kumar2019learning} apply a decision tree model to every component of a simple microprocessor. PRIMAL~\cite{zhou2019primal} predicts per-cycle power by processing transitions of all registers with the convolutional neural network (CNN). GRANNITE~\cite{zhang2020grannite} makes use of graph neural network~\cite{kipf2017semi} to estimate the average power of each workload. Although the ML approach achieves significant speedup compared with accurate commercial tools~\cite{powerpro}, it can be prohibitively expensive (computationally) for per-cycle simulation on industry-standard CPU designs. Evidently, these techniques are intended for simulation-level power-tracing and are too expensive for runtime on-chip monitoring.

FPGA emulation~\cite{coburn2005power, kim2019simmani, yang2015early, sunwoo2010presto} is a popular approach to accelerating power simulations for large designs. We use the term ``emulation'' in a broad sense to include techniques that use of FPGA at design-time. In reality, there are various ways to do so, which may be named differently in other literature. Perhaps the first power emulation work is \cite{coburn2005power}, which has 300\% hardware overhead. Another work~\cite{yang2015early} employs singular value decomposition (SVD), which can be computationally expensive. 
Both \cite{coburn2005power} and \cite{yang2015early} are demonstrated only at block-level designs. A microprocessor-level application of FPGA emulation is Simmani~\cite{kim2019simmani}, whose temporal resolution is 128 clock cycles. PrEsto~\cite{sunwoo2010presto} achieves cycle-accuracy, but its hardware cost is quite significant, e.g., it consumes more than 50\% of LUTs on Xilinx Virtex-5 LX330 to simulate ARM Cortex-A8 processor design. 
Moreover, its proxy selection process is not completely automated.

\textbf{Runtime On-Chip Power Meters (OPMs):} Analog power sensors~\cite{bhagavatula2012low, bhagavatula2013power} can provide accurate power estimation at runtime. However, they require ADCs that consume a large area overhead.
A popular runtime approach is to estimate power dissipation according to performance counters~\cite{bellosa2000benefits, bircher2007complete,bertran2010decomposable, goel2010portable, huang2012accurate, isci2003runtime, JosephISLPED01, pricopi2013power, rodrigues2013study, singh2009real, sagi2020lightweight, walker2016accurate, gilberto2005power}. Since these counters already exist in industrial-grade microprocessor designs, they can be treated as free and the associated area overhead is minimum.    
However, counter-based methods typically rely on architects' knowledge of a specific design to define representative hardware events. This limits existing methods to well-studied microprocessors and hinders automatic migration to new designs.
For example, \cite{isci2003runtime, bellosa2000benefits, bircher2007complete} exclusively targets Intel Pentium{\textsuperscript{\textregistered}} processors, \cite{kalyanam2017power} is exclusively aimed at the Qualcomm Hexagon 680 DSP, and both works of \cite{pricopi2013power} and \cite{walker2016accurate} target ARM Cortex-A7 and Cortex-A15 processors. 
Moreover, counter-based power monitors monitor micro-architectural events that manifest several cycles after the causal trigger event. Therefore, they are poorly correlated with recent pipeline activity and are therefore restricted to coarse-grained temporal resolutions.

Compared to counter-based techniques, proxy-based power monitors are much more friendly to automation and applicable to multiple designs~\cite{najem2016design,cremona2020automatic,pagliari2018all,zoni2018powerprobe, zoni2018powertap}. 
Existing proxy-based techniques suffer from the conflict between low silicon-area overhead and fine-grained temporal resolution. Some of them \cite{najem2016design, cremona2020automatic, pagliari2018all} are coarse-grained with the temporal resolution of thousands of cycles. Their area overhead ranges from 1.5\% to 20\% over the baseline.

Recent methods~\cite{zoni2018powerprobe, zoni2018powertap} improve temporal resolution to 100 cycles. To limit extra overhead from improved resolution, they restrict proxies mostly to primary I/O signals of design modules at selected hierarchy level, significantly reducing the freedom of proxy selection and the underlying power model. Even with this restriction, their area overhead is still $>4\%$~\cite{zoni2018powerprobe, zoni2018powertap}. In \cite{kalyanam2020proactive}, a manually-designed digital power meter technique is introduced to address voltage-droop in DSP engines. This technique takes advantage of predictable dataflow patterns that are not available for general-purpose CPUs. 
In \cite{webel2020proactive}, the authors describe a voltage-noise mitigation strategy that combines power proxies with critical path monitors. The work does not formally describe the creation or the accuracy of power proxies in detail. Further, it is unclear whether the methodology is easily portable across designs.

\begin{figure}[!h]
\centering
\includegraphics[width=0.9\textwidth]{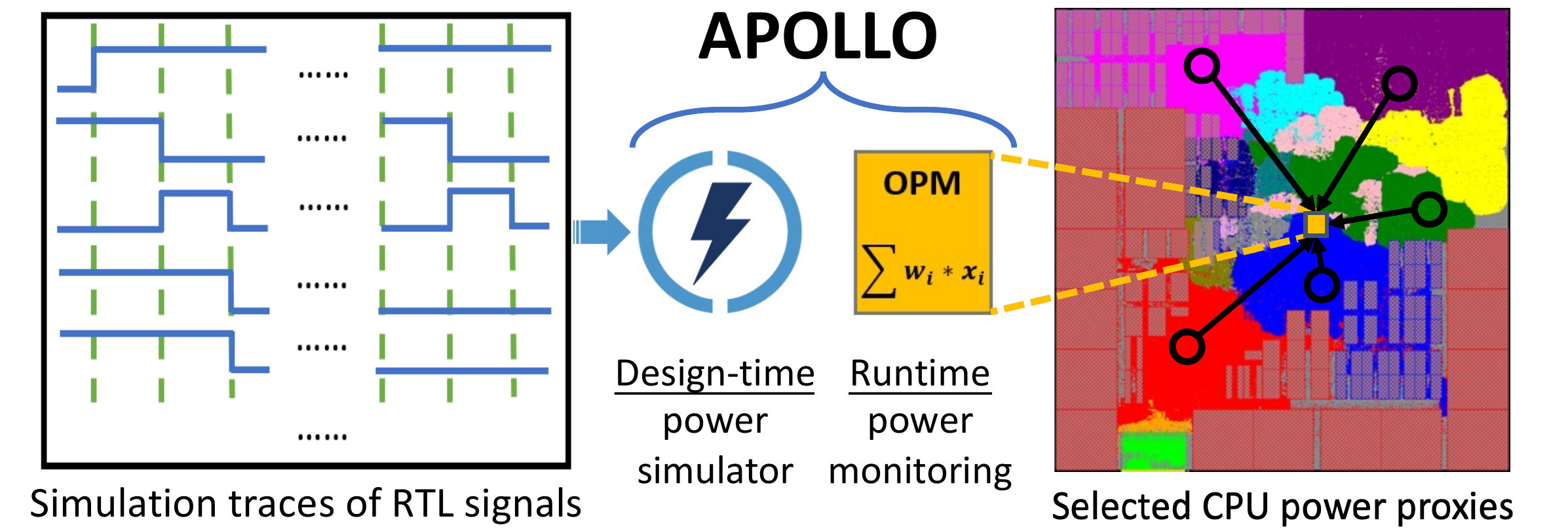}
\caption{APOLLO provides a design-time power simulator and a runtime on-chip power meter (OPM) based on a consistent model.}
\label{introduct}
\end{figure}

\textbf{Position of APOLLO:} APOLLO is a unified RTL-stage power modeling framework addressing both the design-time and runtime challenges within a consistent model structure, as shown in Figure~\ref{introduct}, with Neoverse{\texttrademark} N1 as an example. 
Although proxy-based techniques have been intensively studied, APOLLO distinguishes itself from previous methods by a new proxy selection technique based on the MCP algorithm. Different from other fully-automatic signal selection methods~\cite{cremona2020automatic, kim2019simmani, lee2015dynamic, najem2016design, pagliari2018all, wu1998cycle, zoni2018powertap, zoni2018powerprobe}, the selection technique in APOLLO allows flexible selection from any combination of signals~(unlike \cite{cremona2020automatic, lee2015dynamic, zoni2018powertap, zoni2018powerprobe}), performs supervised selection~(unlike \cite{kim2019simmani}), reduces correlations between proxies~(unlike \cite{pagliari2018all}), and proves to be scalable to a large number of candidate signals~(unlike \cite{najem2016design, wu1998cycle}).
In contrast to~\cite{webel2020proactive}, the APOLLO framework is a fully automated framework that simultaneously achieves accurate power estimation with per-cycle temporal resolution and generates low-cost silicon implementation with $<1\%$ power/area overhead. This is not obviously achievable by published previous works. Furthermore, APOLLO is proven on commercial million-gate CPUs (Neoverse N1 and Cortex-A77), thus indicating scalability to real-world applications.

\section{APOLLO Methodology}
\label{sec:method}

The total power consumption in a CMOS circuit is contributed by switching and leakage components. Leakage power is determined by the junction temperature and the threshold voltage of transistors. Since it is relatively invariant to code-execution, leakage power measurement is generally not relevant to runtime power management. 
Similarly, leakage power can be easily estimated using EDA tools~\cite{powerpro} at design-time. Therefore, APOLLO focuses on modeling the switching component of the total power. 

The switching component can be further broken down into dynamic power due to code-dependent charging/discharging of gate/wire-capacitance, short-circuit power during slow signal slews, and glitch power. In practice, power due to glitches and the short-circuit power is much smaller than dynamic power~\cite{kim2019simmani}, and all three components correlate with signal transitions. 
The dynamic power at each cycle $i$ can be measured by the summation of power consumed at the capacitance of all toggling gates and wires as

\vspace{-0.1 in}
\begin{equation}
    Power_{\ dyn} [i] = \frac{1}{2} V^2 \sum_{g \ \in \  \{toggling\ gates\}} C_{g}  \label{eq:dynmaic} 
       \vspace{-0.05 in}
\end{equation}
Equation (\ref{eq:dynmaic}) does not include the ``frequency'' component since it is expressed in per-cycle terms. While this approach has signoff-level accuracy, it is computationally intensive and does not scale to workload-execution timescales on large designs with fully annotated parasitics. Since toggling activities of gates are highly correlated with each other, Equation (\ref{eq:dynmaic}) can be reasonably approximated by a simpler linear model based on $Q$ selected proxies as 
\vspace{-0.05 in}
\begin{equation}
   Power_{\ dyn} [i] \approx \frac{1}{2} V^2 \sum_{j \ \in \  \{Q\ power\ proxies\}} w^{\ unscaled}_{j} \label{eq:dynmaic_approx}
   \vspace{-0.05 in}
\end{equation}

Note that the equation in (\ref{eq:dynmaic_approx}) is a measure of the {\it power-demanded} by the CPU from the power-delivery network (PDN) {\it before} it is modulated by the PDN response.  Hence, the voltage can be viewed as a constant, and by scaling the weights, we reach the simpler final model as Equation (\ref{eq:power_estimator}). In equivalent terms, Equation (\ref{eq:dynmaic_approx}) can also be viewed as a measure of the CPU current demand.

APOLLO targets the high model accuracy with low computation and implementation cost by automatically identifying representative RTL proxies from a large number of highly-correlated RTL signals and building a lightweight power model. Given a design with $M$ RTL signals $S_M$, APOLLO selects a subset $S_Q \subset S_M$ with $Q=|S_Q|\ll M$, as power proxies, and $Q$ is the number of proxies. Then it builds a linear power estimator based on $S_Q$. For per-cycle power tracing,   
\vspace{-0.05in}
\begin{equation}
p[i] = 
 \sum_{j=1}^Q w_j \cdot x_j[\,i] 
~~~~\text{for the } i^{\,th} \text{ clock cycle,} \label{eq:power_estimator}
\vspace{-0.05 in}
\end{equation}
where $x_1[i], x_2[i], ..., x_Q[i] \in \{0,1\}$ are input features indicating the togglings or transitions of $Q$ proxies in the $i^{\,th}$ clock cycle, $w_1, w_2, ..., w_Q \in \mathbb{R}^+$ are trainable weights, and $p[i]$ is the predicted power of the same cycle.

Selecting power proxies $S_Q$ from $S_M$ with $Q\ll M$ can greatly accelerate power simulation, reduce data volume for emulation-assisted power analysis, and lower hardware cost for runtime OPM. The choice of $Q$ controls the trade-off between accuracy and efficiency. Although linear power models have been widely used in the past, our proxy selection technique distinguishes APOLLO from previous methods.

Given $N$ cycles of simulation traces, the $N$ ground-truth labels $y[1], y[2], ..., y[N] \in \mathbb{R}^{+}$ are per-cycle power values generated from a commercial RTL power analysis flow~\cite{powerpro}, where back-end parasitics are annotated to the RTL design but netlist-level details are abstracted out for flow acceleration in our experiment. It shall be noted that APOLLO applies to an arbitrary method of ground-truth power data collection.

\subsection{Automatic Training Data Generation}

\begin{figure}[!tb]
\centering
    \subfigure[]{\includegraphics[height=.34\textwidth]{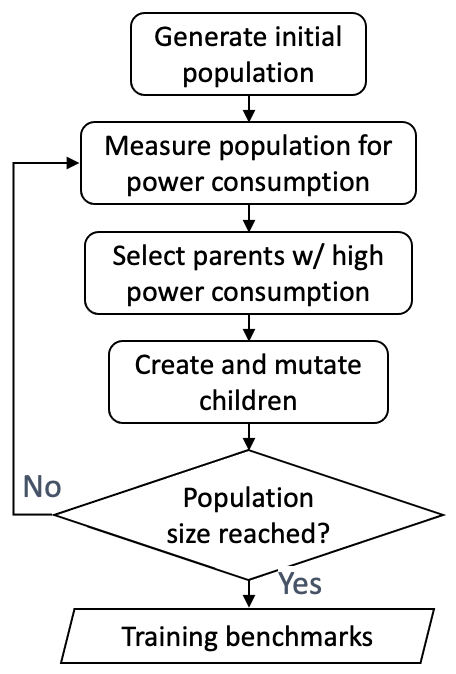}\label{traing_data_gen_a}}
    \hspace{.35in}
    \subfigure[]{\includegraphics[height=.34\textwidth]{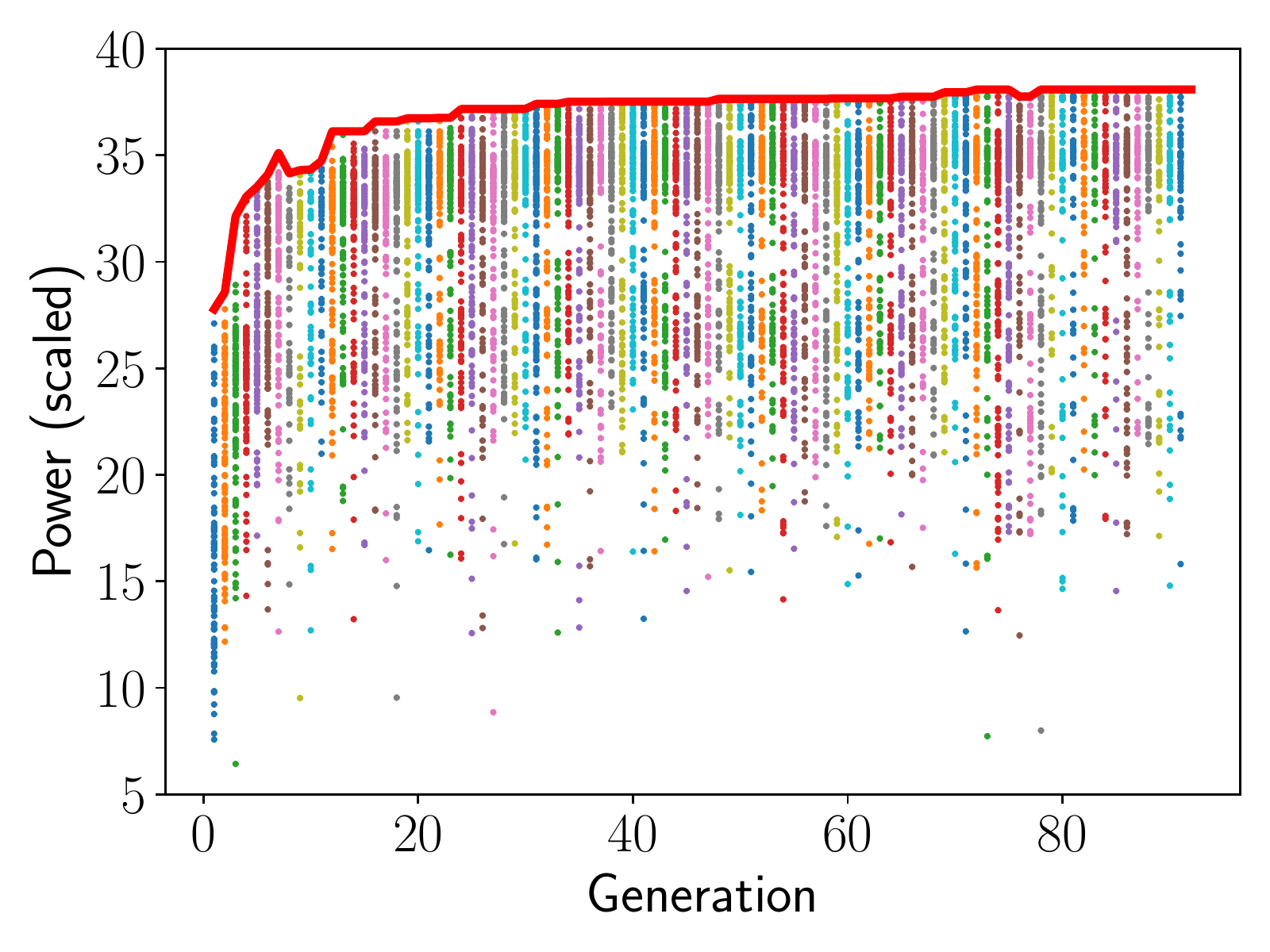}\label{training_data_gen_b}}
    \vspace{-.1in}
    \caption{Training data generation. (a) GA-based generation flow. (b) A diverse set of training micro-benchmarks with a wide range of measured power.}
    \label{training_data_gen}
\end{figure}

The APOLLO framework starts with automatic training data generation for the target design as shown in Figure~\ref{traing_data_gen_a}. Generated micro-benchmarks are then replayed with EDA tools to generate power labels.

Previous automatic proxy selection methods~\cite{zhou2019primal, kim2019simmani, yang2015early, pagliari2018all, zhang2020grannite} mainly adopt three categories of training data: 1) random stimuli, 2) realistic workloads, 3) handcrafted ISA tests or micro-benchmarks. However, for 1), previous studies lack details on how to automatically generate a large number of random stimuli with sufficient diversity for an arbitrary design. For 2), realistic workloads typically include redundant patterns and cannot efficiently cover the full range of power consumption, especially high power consumption scenarios. For 3), it is particularly challenging to generate a diverse training set using manually-developed power benchmarks, even with expert micro-architectural knowledge. 


We circumvent these practical engineering challenges by auto-generating the training set of micro-benchmarks using a genetic algorithm (GA)-based framework~\cite{hadjilambrou2019gest} that is micro-architecture agnostic. Our benchmark-generation flow starts with an initial population of randomly generated micro-benchmarks (referred to as ``individual'') created with a constrained set of instructions. The average power of each micro-benchmark is then measured using the EDA tool~\cite{powerpro}. The ones with highest power are selected as so-called ``parents'' that are then paired together (crossover) and mutated to create ``child'' instruction sequences for the next generation and so on. The power measurements of all generated micro-benchmarks are shown in Figure \ref{training_data_gen_b} across multiple generations. The GA-based optimization loop is primed to generate the worst-case power-consuming benchmark, or a power-virus, as indicated by the envelope of the scatter plot. 


As the optimization converges to the worst-case power virus, successive generations favor higher power-consuming benchmarks. However, early generations naturally favor those lower power-consuming benchmarks as the algorithm is yet to identify higher power-consuming instruction sequences. A combination of low and high power-consuming benchmarks across generations naturally creates a rich diversity of benchmarks spanning a large range (>$5\times$ ratio between the maximum and minimum individuals) of power consumption. 

\subsection{Features and Labels Collection}

\begin{figure}[!tb]
\centering
    \subfigure[Design RTL and simulation trace]{\includegraphics[height=.26\textwidth]{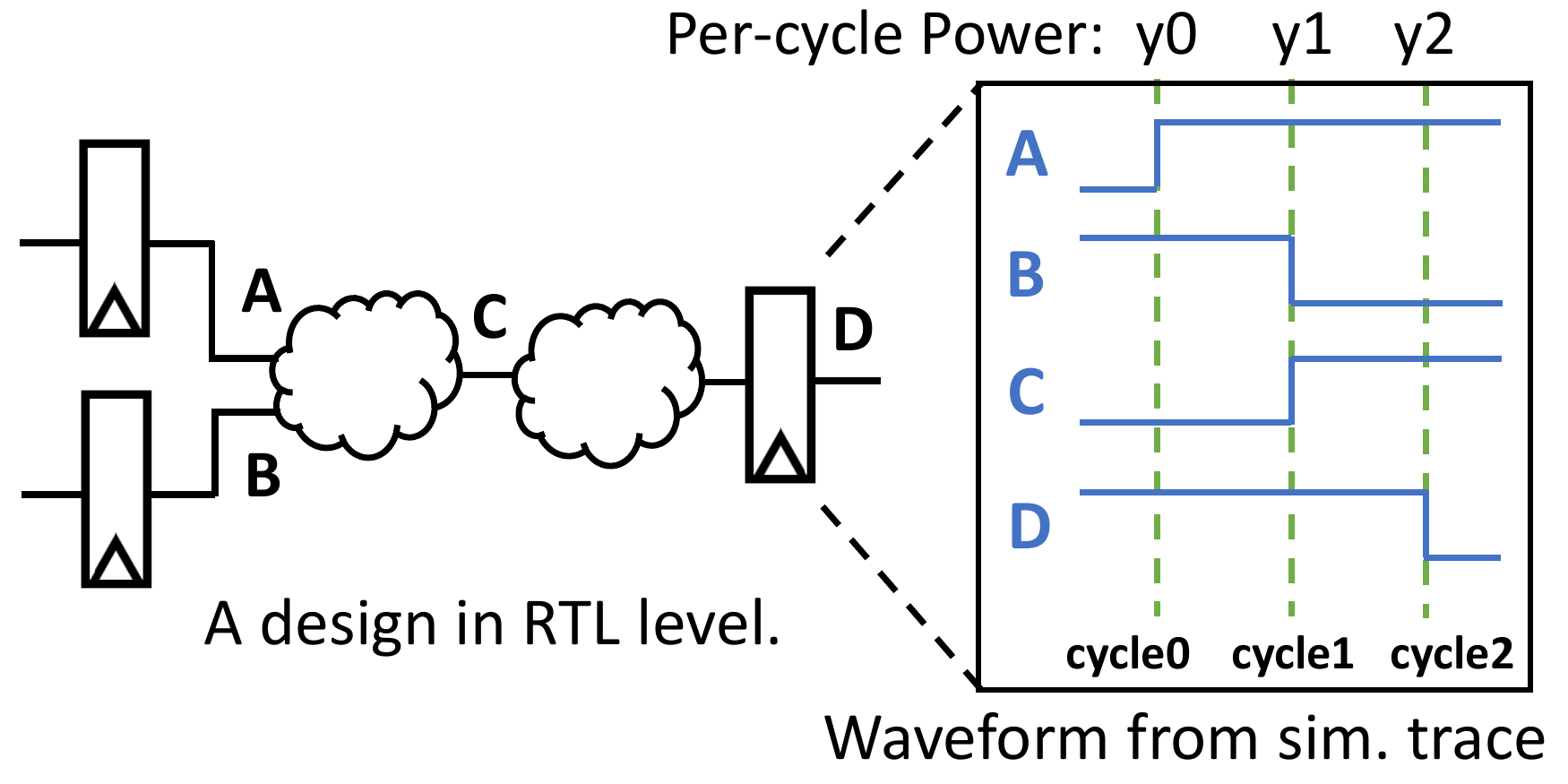}\label{feature_a}}
    \hspace{.1in}
    \subfigure[Feature and label]{\includegraphics[height=.26\textwidth]{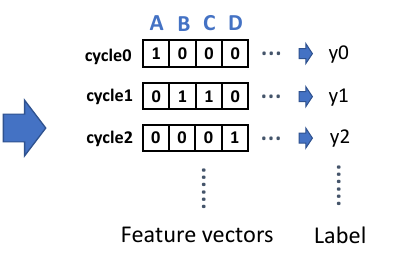}\label{feature_b}}
    \caption{Feature and label collection based on $M$ RTL signals and $N$ cycles of simulation traces.}
    \label{feature_cons}
\end{figure}  

Figure~\ref{feature_cons} shows the procedure to construct features from the RTL-simulation traces and labels from power simulation results. 
As Equation (\ref{eq:dynmaic}) shows, per-cycle toggling activities reflect the net transitions and directly correlate with power consumption. 
At each cycle, for each RTL signal, either a rising or falling edge in the simulation trace is set to 1 as features, while no toggling is set to 0. 
As such, each RTL signal contributes to one element in the feature vector. 

For $M$ RTL signals and $N$ cycles of simulation traces, the raw input feature vectors are $\mathbf{x} \in \{0, 1\}^{N \times M}$, and the input vectors with only $Q$ selected proxies are denoted as $x \in \{0, 1\} ^ {N \times Q}$. The corresponding label is $N$ per-cycle power consumptions $y \in \mathbb{R}^N$ simulated with the EDA tool~\cite{powerpro}.

\subsection{ML-Based Power Proxy Selection}
\label{section:selection}

\begin{figure}[!t]
\centering
    \subfigure[]{\includegraphics[height=0.29\textwidth]{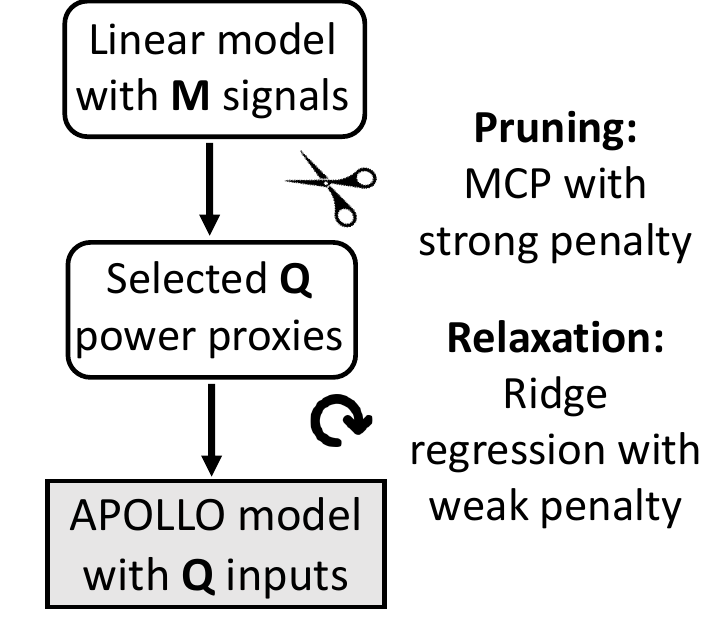}
    \label{APOLLO_training}}
    \hspace{.3in}
    \subfigure[]{\includegraphics[height=0.29\textwidth]{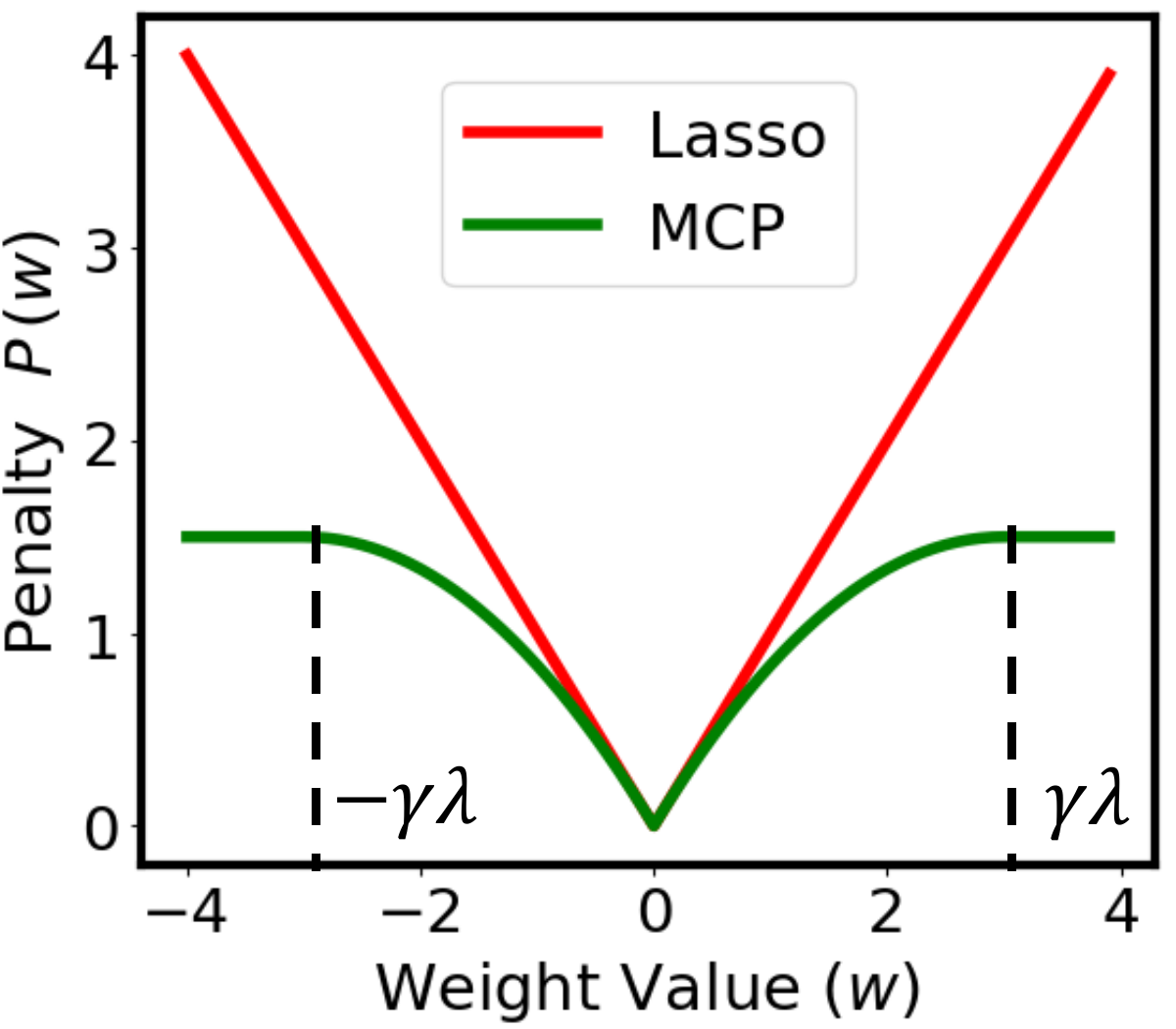}
    \label{penalty}}
\vspace{-.15in}
\caption{APOLLO model construction. (a) Model construction process. (b) Penalty terms of MCP and Lasso.}
\end{figure}

Once raw features and labels are collected, we go through the steps in Figure \ref{APOLLO_training} to construct the APOLLO model. The key step is to select $Q$ representative power proxies. It starts with building a temporary linear power model $p' = \sum_{j=1}^M w'_j \cdot \mathbf{x}_j$ with all $M$ RTL signals in raw input features. This linear model is not trained only to minimize the prediction error in the training dataset. Instead, when minimizing the prediction error during training, the model simultaneously shrinks all weights $w'_1, w'_2, ..., w'_M$ so that the majority of weights eventually become zero, i.e., the model becomes sparse. Then only those RTL signals associated with non-zero weight terms are selected as power proxies. This procedure is also referred to as pruning. Such sparsity-inducing training is realized by applying a penalty term $\mathcal{P}$ in the loss function to penalize weights. Equation~\eqref{eq:loss} shows the loss function, which consists of both the ordinary prediction error ($\mathcal{L}$) measured in mean squared error and the penalty term ($\mathcal{P}$). 

\vspace{-0.5in}
\begin{align}
    \text{Loss} &= \mathcal{L} + \mathcal{P} = \frac{1}{N}\sum_{i=1}^N(y[i] - p'[i])^2 + \sum_{j=1}^M P_{penalty}(w'_j)  \label{eq:loss}
    \vspace{-.2in}
\end{align}

The sparse linear model is constructed by adopting sparsity-inducing penalty terms. 
The most widely adopted penalty term for sparsity is Lasso~\cite{tibshirani1996regression}, defined as 
\vspace{-0.1in}
\begin{equation}
    P_{Lasso}(w'_j) = \lambda |w'_j|  \label{eq:lasso}\\
    \vspace{-0.3in}
\end{equation}

This Lasso penalty shrinks all weights at the same rate decided by the hyper-parameter $\lambda$, which is the penalty strength. However, to ensure $Q \ll M$, we need to set a very large penalty strength $\lambda$ such that the majorities of weights shrink to zero. As a result, when most small weights shrink to zero and their associated terms are pruned out, the weights of remaining terms are penalized too much to provide accurate power predictions. Based on such an inaccurate model, the selected power proxies are not representative enough.

To overcome the aforementioned limitation, APOLLO adopts the MCP~\cite{zhang2010nearly} as the penalty term, which is defined by 
\vspace{-0.05in}
\begin{equation}
    P_{MCP}(w'_j, \gamma > 1) = 
  \begin{cases} 
  \lambda |w'_j|  - \frac{{w'_j}^2}{2\gamma} & \text{if } |w'_j| \leq \gamma \lambda \\
  \frac{1}{2} \gamma \lambda ^2  & \text{if } |w'_j| > \gamma \lambda  \\
  \end{cases} \label{eq:mcp}
\end{equation}

The hyper-parameter $\gamma$ in MCP sets a threshold ($\gamma\lambda$) between large and small weights.
Figure~\ref{penalty} visualizes both $P_{Lasso}$ and $P_{MCP}$ with $\lambda=1$ and $\gamma=3$. 
The absolute derivative of a penalty term indicates the weight shrinking rate during training~\cite{parikh2014proximal}. 
Since $|\partial P_{Lasso} / \partial w'_j| = \lambda$, all weights shrink at the same rate $\lambda$ in Lasso. In comparison, the absolute derivative of MCP penalty is given by 
\begin{equation}
    \left|\frac{\partial P_{MCP}(w'_j, \gamma > 1)} {\partial w'_j}\right| = 
  \begin{cases} 
  \lambda  - \frac{|w'_j|}{\gamma} & \text{if } |w'_j| \leq \gamma \lambda \\
  0  & \text{if } |w'_j| > \gamma \lambda  \\
  \end{cases} \label{eq:dmcp}
\end{equation}
Compared with the uniform shrinking rate for $P_{Lasso}$, large weights with values $>\gamma\lambda$ in MCP do not shrink at all, since derivatives of their penalty terms are zero. For weights with values $<\gamma\lambda$, smaller weights shrink faster. As such, MCP leaves large weights unpenalized and thereby benefits the prediction accuracy of the generated power model.
In our experiment, this MCP-based model is efficiently optimized by adopting the coordinate descent method~\cite{wright2015coordinate} and the proximity operator of MCP~\cite{wen2018survey}.
The penalty strength $\lambda$ can be adjusted to control the number of selected proxies $Q$.

\subsection{Final Model Construction}
\label{section:construction}

After power proxy selection by pruning with MCP, we have trained a temporary model $p' = \sum_{j=1}^M w'_j \cdot \mathbf{x}_j$ with $Q$ selected proxies $S_Q$ and corresponding non-zero weight terms $w'_j$. This temporary model can already provide rather accurate predictions. However, although MCP protects larger weights, many remaining weights are still penalized by the large penalty strength $\lambda$ to a certain extent. To further boost the model accuracy, we train a new linear model $p = \sum_{j=1}^Q w_j \cdot x_j$ from scratch with only selected power proxies $S_Q$. 
In this new linear model, the ordinary L2 penalty, i.e., ridge penalty~\cite{hoerl1970ridge}, is applied, with a much weaker penalty strength compared with the $\lambda$ used in the previous proxy selection step. This weak ridge penalty is applied to reduce overfitting.

As shown in Figure~\ref{APOLLO_training}, this step is named \emph{relaxation} and generates the final APOLLO power model. During the previous proxy selection step, to shrink most weights to zero, the penalty term $\mathcal{P}$ dominates the loss, and the prediction error $\mathcal{L}$ is less optimized. This \emph{relaxation} can be viewed as a fine-tuning stage to better optimize $\mathcal{L}$. Since L2 is not sparsity-inducing, the number of proxies $Q$ remains unchanged. 


\subsection{Multi-Cycle Power Modeling}
\label{subsec:APOLLO-multi-cycle}

\begin{figure}[!tb]
\centering
\includegraphics[width=0.85\textwidth]{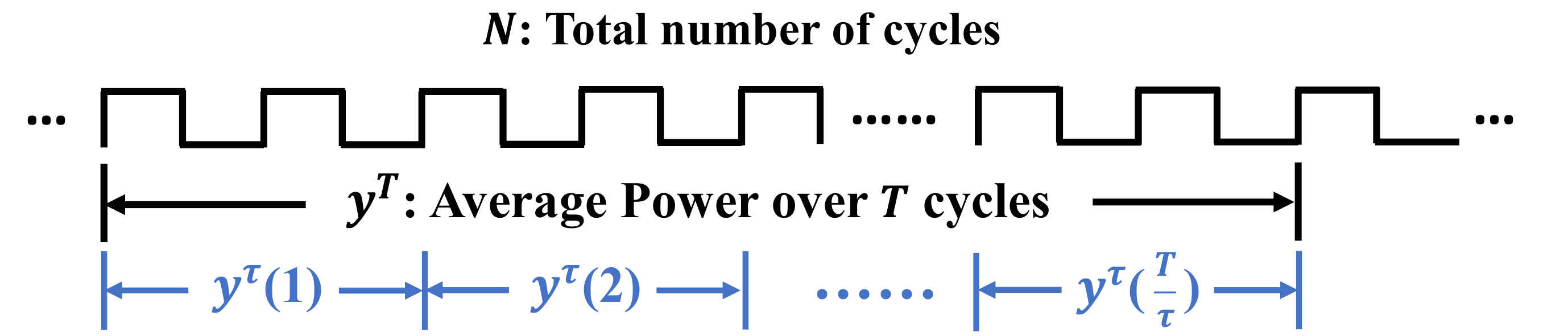}
\vspace{.1in}
\caption{Multi-cycle APOLLO --- Label is $y^T$, with a measurement window of $T$ cycles. Label at each interval is $y^\tau$, with selected interval of $\tau$ cycles.}
\label{multi-cycle}
\end{figure}

In previous subsections, we construct the APOLLO model for per-cycle power tracing. Such fine-grained temporal resolution enables applications like voltage droop mitigation. In this sub-section, we generalize the APOLLO model to larger time-window sizes. Like Figure~\ref{multi-cycle} shows, this multi-cycle model estimates the average power over a time window with $T$ cycles, which is chosen to be a power of two for ease of efficient hardware implementation.

A straightforward multi-cycle solution is to directly use the average of $T$ per-cycle power predictions $p^T$ over the $T$-cycle window\footnote{We use the superscript on a variable to denote the average of the variable over a timing window with multiple cycles.}. It uses the same per-cycle model for any $T$. Such an approach captures details of individual clock cycles but neglects correlations among different clock cycles.
Alternatively, one can average the transitions over $T$ cycles and generate a $T$-cycle power estimation based on the average toggling rate. 
However, this approach loses useful information such as cycle-details that can be particularly helpful when $T$ becomes large. 
In addition, the model developed by this approach is dependent on the varying $T$. 
In Section~\ref{subsec:accuracy}, we show that both the average-prediction and the average-input approaches fail to provide an accurate and robust solution.

We introduce a multi-cycle estimation technique that overcomes the weakness of the aforementioned approaches. A time window of $T$ is divided into multiple intervals of $\tau$ cycles. The values of
$T$ and $\tau$ are selected such that $T$ is integer multiples of $\tau$. 
An example is shown in Figure \ref{multi-cycle}. 
During the model construction and training, for each $\tau$-cycle interval $k$, we measure both the average toggling activities $\mathbf{x}^{\tau}_1(k),\allowbreak ...,\allowbreak \mathbf{x}^{\tau}_M(k) \in \mathbb{R}$ and the average power $y^{\tau}(k) \in \mathbb{R}$ over the $\tau$ cycles\footnote{We use parentheses and brackets to differentiate the indices of intervals and cycles.}. Based on these raw inputs and labels, we execute the same training procedure as the per-cycle model to select $Q$ power proxies $S_Q$ with features $x_1^\tau, ..., x_Q^\tau$. The result is a $\tau$-cycle model denoted as APOLLO$^\tau$, whose weights are denoted as $\omega_1,...,\omega_Q$. It is to be noted that the construction of APOLLO$^\tau$ is independent of $T$, and its performance is controlled by selecting an appropriate $\tau$ value as a hyper-parameter before training. 

At the inference stage, there are $\frac{T}{\tau}$ intervals in a time window. As Figure~\ref{multi-cycle} shows, the final prediction $p^T\in \mathbb{R}$ at each $T$-cycle window is the average over these $\frac{T}{\tau}$ predictions from the APOLLO$^{\tau}$ model:

\vspace{-0.3in}
\begin{align}
    & p^T = \frac{1}{T/\tau} \ \sum_{k=1}^{T/\tau} p^{\tau}(k)  \ \text{, where } p^\tau(k) = \sum_{j=1}^Q \omega_j \cdot x^{\tau}_j(k) \label{eq:multi}
\end{align}

Here, the input $x_j^{\tau}(k) \in \mathbb{R}$ for each interval is a real number instead of a binary. If directly implemented on hardware, this requires $Q$ counters and multipliers like previous OPMs~\cite{najem2016design, cremona2020automatic,  zoni2018powertap,  zoni2018powerprobe}. In contrast, the toggling in each cycle is a binary number and thus the per-cycle model can be implemented by AND gates instead of multipliers. To avoid multipliers for on-chip implementation of the multi-cycle model, we rearrange the inference process in Equation (\ref{eq:multi}) as below: 

\vspace{-0.6in}
\begin{flalign*}
    &\qquad \text{Take the first interval $p^\tau(1)$ when $k=1$ as example:} && \nonumber\\
    &\qquad\qquad p^\tau(1) = \sum_{j=1}^Q \omega_j \cdot \textcolor{blue}{x^{\tau}_j(1)} = \sum_{j=1}^Q \omega_j \cdot \textcolor{blue}{\frac{1}{\tau}\sum_{i=1}^{\tau} x_j[i]} = \textcolor{blue}{\frac{1}{\tau}}  \textcolor{blue}{\sum_{i=1}^{\tau}} \sum_{j=1}^Q  \omega_j \cdot \textcolor{blue}{x_j[i]} \nonumber \\
\end{flalign*}
\vspace{-1in}
\begin{flalign}
    &\qquad \text{Thus,  } \ \  p^T = \frac{1}{T/\tau} \ \sum_{k=1}^{T/\tau} p^{\tau}(k) = \frac{1}{T} \sum_{i=1}^{T} \sum_{j=1}^Q  \omega_j \cdot  x_j[i] && \label{eq:multi_final}
    \vspace{-0.1in}
\end{flalign}

In Equation~(\ref{eq:multi_final}), the weights are multiplied with binary numbers instead of real numbers.
This new inference process can be regarded as predicting $T$-cycle average power according to per-cycle toggles. As such, it takes per-cycle details, considers correlations among multiple cycles, and hence overcomes the drawbacks of aforementioned approaches.
Interestingly, $\tau$ is no longer needed in inference. By setting $T$ to be power of 2, the division in Equation~(\ref{eq:multi_final}) can be realized by directly discarding the $\log_2(T)$ lowest bits. 
Therefore, the on-chip implementation of this multi-cycle model can reach low hardware overhead like the per-cycle model.

\section{Application of the Power Modeling Framework}

\subsection{Design-Time Power Analysis}
\label{sec:APOLLO-ebpf}

\begin{figure}[t]
\centering
    \subfigure[]
    {\includegraphics[height=.32\textwidth]{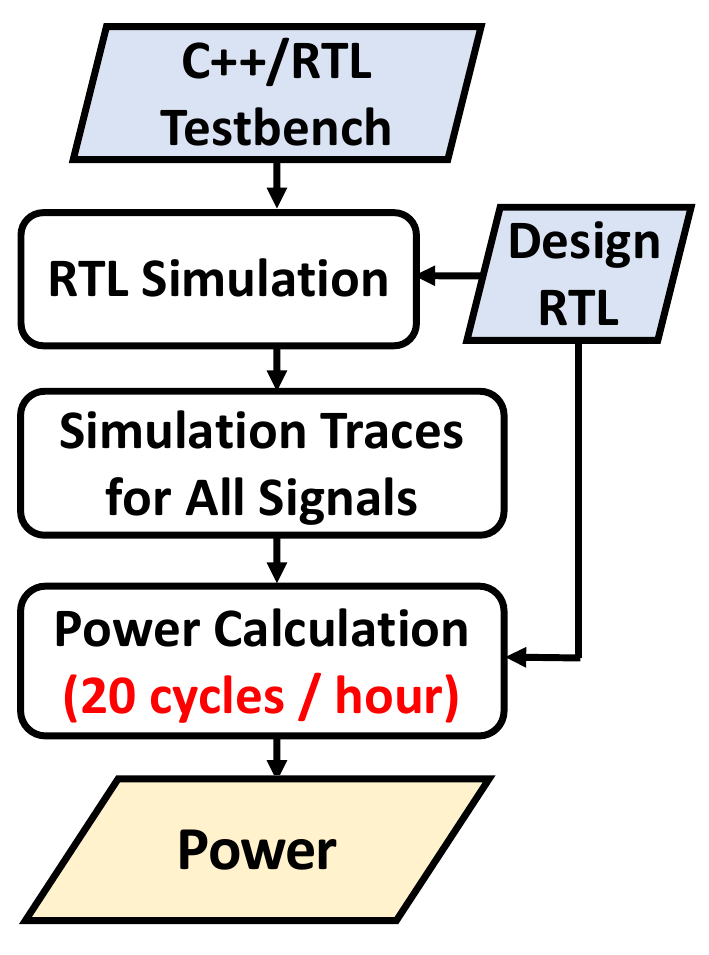}\label{tradition_flow}}
    \hspace{0.2in}
    \subfigure[]
    {\includegraphics[height=.32\textwidth]{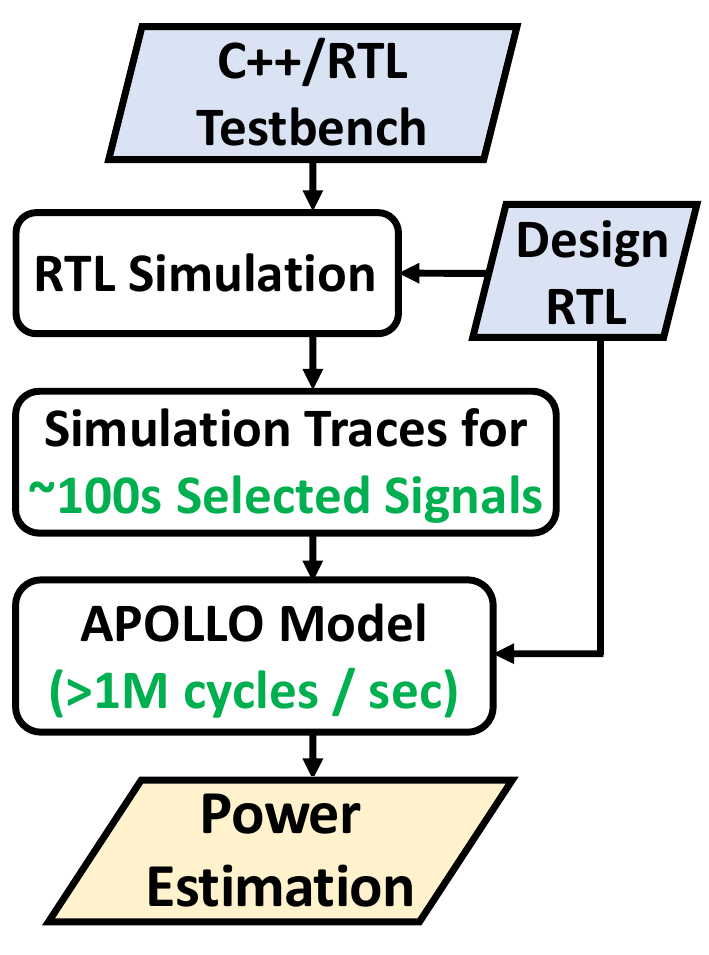}\label{APOLLO_flow}}
    \hspace{0.2in}
    \subfigure[]
    {\includegraphics[height=.32\textwidth]{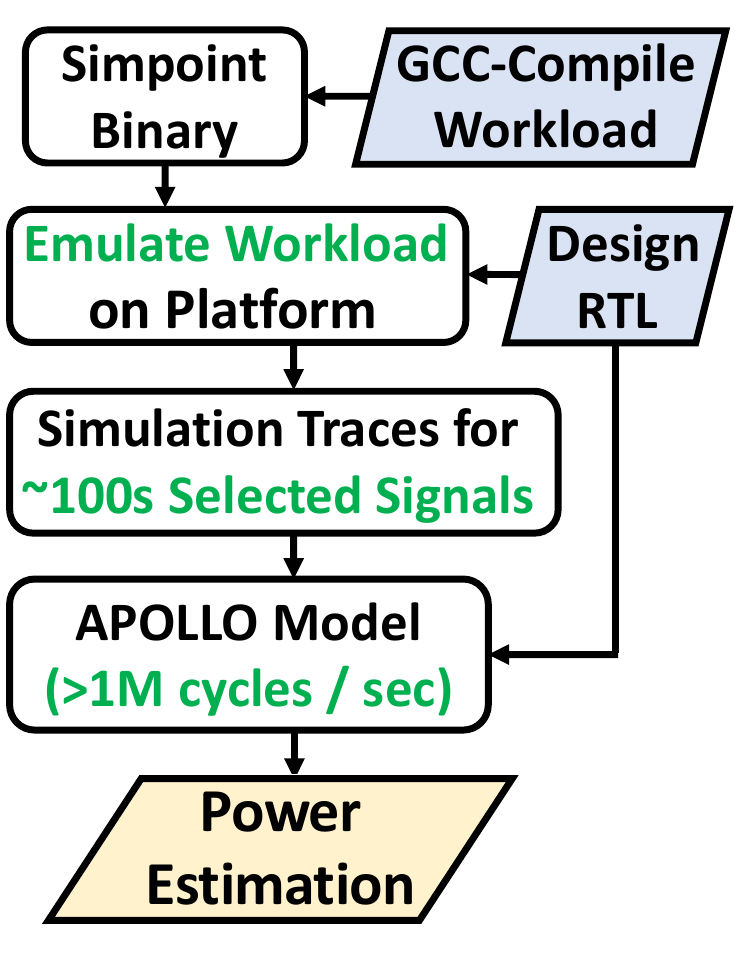}\label{EBPF}}
    \vspace{-.1in}
    \caption{Design-time power analysis flows. (a) Commercial analysis. (b) APOLLO-based analysis. (c) APOLLO with emulator-assisted analysis.} 
    \label{flows}
\end{figure}

A typical conventional design-time power analysis flow is shown in Figure~\ref{tradition_flow}. It generates simulation traces for all signals in VCD or FSDB file format through RTL simulation, then performs power calculation with simulation tools using these traces. Such a flow is very time-consuming. One major bottleneck is the last step of power calculation, which is extremely slow for large designs. 

To accelerate this process, we incorporate the APOLLO model into the flow as shown in Figure~\ref{APOLLO_flow}, where the number of signals to be traced is greatly reduced and the last step of power calculation is replaced by APOLLO. APOLLO can infer power for millions of cycles within seconds. This APOLLO-assisted power simulation flow works well for cases where RTL simulation time is reasonable.

For long-running benchmarks, the RTL simulation step in Figure~\ref{APOLLO_flow} becomes an execution bottleneck.  
We propose to overcome this using an emulator-assisted power analysis flow as shown in Figure~\ref{EBPF}. 
In this flow, millions of benchmark cycles are emulated on a commercial platform~\cite{emulation-platform} with dedicated hardware to generate million-cycle simulation traces in minutes.
In the absence of APOLLO, the power-simulation flow requires switching details of all nets to be dumped. For a large industrial-scale design, this can easily exceed hundreds of GB leading to storage and memory capacity issues during power analysis. Traditionally, this problem is circumvented by estimating power at a coarse-grained temporal resolution, e.g., thousands of clock cycles. With APOLLO, the data is reduced by several orders of magnitude by only collecting the toggling activities of $Q$ power proxies, and a cycle-accurate estimation for the emulator-assisted flow is enabled.

\subsection{Runtime On-chip Power Meter}

APOLLO provides an accurate and fine-resolution runtime OPM with low hardware cost.
The OPM implements a linear model with $Q$ power proxies as the input, which is a binary vector at each cycle, e.g., $x[i] \in \{0, 1\}^{Q}$. 
All weights are quantized into $B$-bit fixed-point values, which can be configured to accommodate potential model re-training using sign-off or hardware measurement power values.

\begin{figure}[!tb]
\centering
\includegraphics[width=0.8\textwidth]{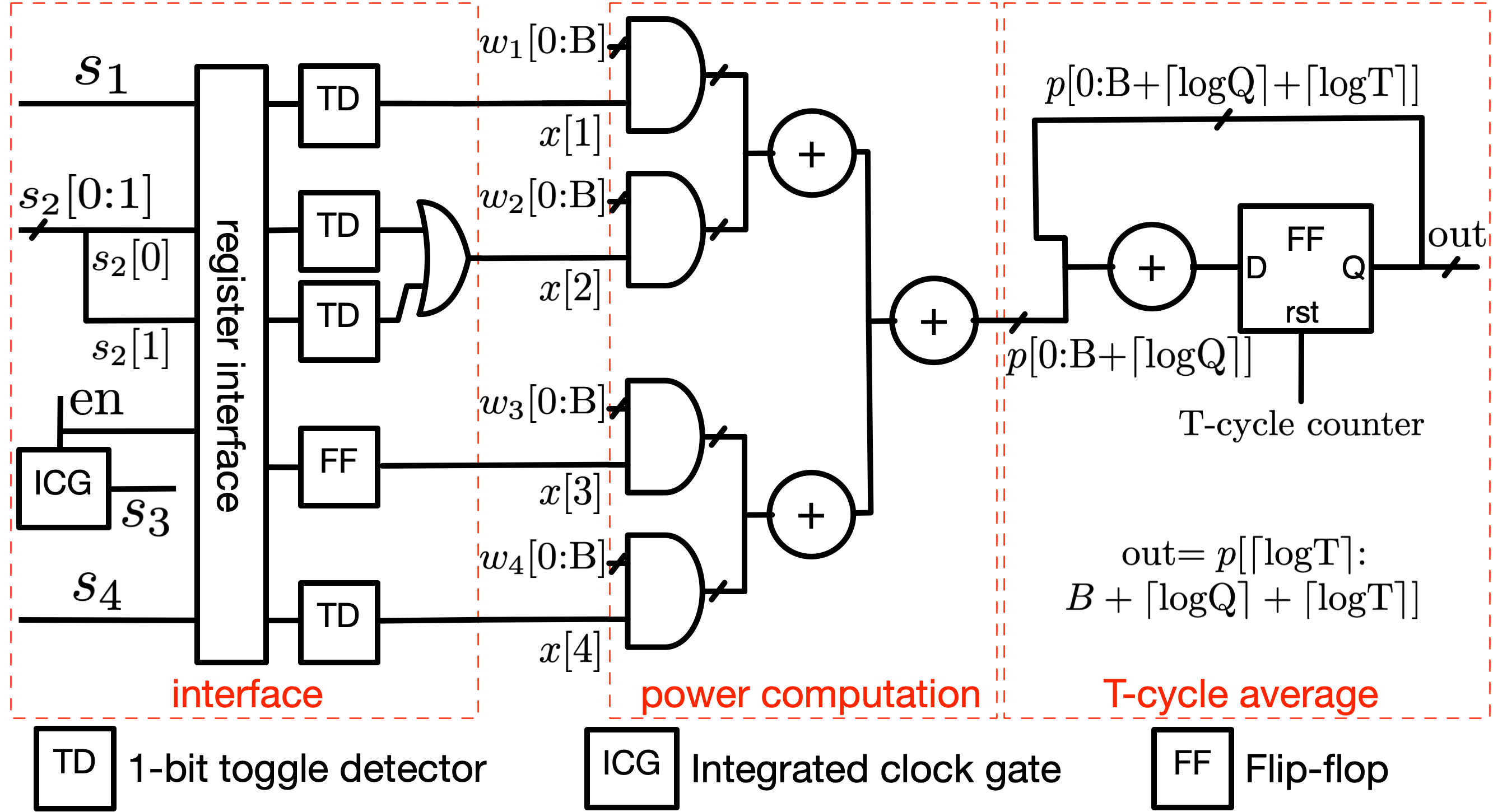}
\caption{OPM integration with the CPU design. }
\label{opm_design}
\end{figure}

The APOLLO-OPM is fully integrated with the microprocessor. Figure~\ref{opm_design} shows the OPM consists of three components, i.e., ``interface'', ``power computation'' and ``T-cycle average''. 
The ``interface'' latches the input signals using the register interface and then extracts per-cycle toggling activities as single-bit values for each power proxy. 
The register interface minimizes the data path timing impact from OPM on the original design. 
The ``interface'' takes power proxies, i.e., $s_1$, $s_2[0:1]$, $s_3$ and $s_4$ as inputs, which are further categorized into three cases:
(1) 1-bit signal ($s_1$ and $s_4$). A 1-bit toggle detector ``XOR''s the monitored signal with its registered version to determine whether a toggle occurred. 
(2) Bus signal ($s_2[0:1]$). We set up 1-bit signal interface for each bit of the bus. An extra OR gate determines whether the entire bus signal toggles.
(3) Gated clock signal ($s_3$). A gated clock signal ($s_3$) toggles twice during one clock cycle. Instead of using a 1-bit toggle detector, we automatically trace the clock enable signal ($en$), which is directly latched using a flip-flop to determine whether gated clock signal toggles at the same cycle as other power proxies.

The ``power computation'' component calculates the intermediate values from the quantized weights, i.e., $w_j[0:B]$, and per-cycle toggling values, i.e., $x[j]$. 
The bit width of these power values is extended to $B+\lceil$logQ$\rceil$ to ensure the full precision addition.
After intermediate values are computed on a cycle-by-cycle basis, a ``T-cycle average'' component computes the average power over T cycles using flip-flops and adders. 
The flip-flop reset is controlled by a T-cycle counter, which resets the value of output, i.e., $out$, every T cycles. 
Similarly, the bit width of intermediate values is extended to $B+\lceil$logQ$\rceil+\lceil$logT$\rceil$ to guarantee full precision addition. 
The output power value needs to be divided by $T$ according to Equation~(\ref{eq:multi_final}).
This is realized by dropping the lowest log$\lceil$T$\rceil$ bits as T is set to be the power of 2.

The OPM structure in Figure~\ref{opm_design} is applicable to both per-cycle and multi-cycle power model, due to the linear model structure discussed in Equation(\ref{eq:multi_final}).
The OPM is implemented with generic templates (configurable in $B$, $Q$ and $T$) in C++ using the Catapult HLS tool~\cite{catapult-hls} and synthesized into gate-level netlist using Design Compiler~\cite{design-compilier}. 

The key to the low-cost implementation is two-fold. First, the APOLLO only selects $<0.05\%$ RTL signals as power proxies.
Secondly, calculation of the per-cycle power only requires a conditional accumulation of the proxy weights depending upon whether they toggled or not. As such, only a set of AND gates and adders, instead of multipliers, are needed for the computation. 
The hardware implementation cost of the APOLLO model is much lower than previous approaches, such as Simmani~\cite{kim2019simmani}. 
Most previous OPMs require a counter and multiplier for each proxy, which incurs a much larger area cost.
Furthermore, although APOLLO-OPM may include different sets of trained weights from per-cycle and multi-cycle power model, they share the same hardware structure, which allows greater flexibility and configurability compared to previous studies.

\section{Evaluation}

\subsection{Experimental Setup}

\begin{table}[!h]
      \centering
        \caption{Designer-handcrafted Testing Benchmarks}
        \label{tbl:testbench}
      \renewcommand{\arraystretch}{1.2}
        \begin{tabular}{ |c||c|c|c|c| } 
        \hline
         Name   &  dhrystone & maxpwr\_cpu & dcache\_miss &  saxpy\_simd \\
         \hline 
         Cycles &  1222    &  600   &   654   & 1986    \\
         \hline
         \hline
         Name   &  maxpwr\_l2 &  icache\_miss & cache\_miss &  daxpy  \\
         \hline
         Cycles &    1568    &  800  &  600   &   1600    \\
         \hline
         \hline
         Name   &  memcpy\_l2  &  throttling\_1 &  throttling\_2  & throttling\_3 \\
         \hline
         Cycles &  3000  &  1100   & 1100     & 1100  \\
         \hline
        \end{tabular}
\end{table}

In our experiments, micro-benchmarks used in model training and testing are kept strictly different and separate. Through the automatic training data generation, $>1,000$ random micro-benchmarks are obtained in 4 days to cover a wide range of average power consumption, among which around 300 micro-benchmarks are selected to form the training set with a uniform power distribution. 20\% of the training data are selected to form a validation set for parameter tuning. Unlike the training data, which are automatically generated, the testing data are from 12 representative micro-benchmarks handcrafted by CPU designers corresponding to various use cases, as shown in Table~\ref{tbl:testbench}. They cover both low- and high-power consumption regions. The three micro-benchmarks named `throttling' reflect applying different throttling schemes~\cite{arm-neoverse-n1-trm} to the microprocessor. The simulation trace lengths $N$ for training and testing are approximately 30,000 and 15,000 cycles on Neoverse N1, respectively.

All experiments are firstly performed on the Neoverse N1~\cite{arm-neoverse-n1, christy20208}, a microprocessor for a wide range of cloud-native server workloads executing at world-class performance and efficiency. To verify the robustness of APOLLO on different designs, we further test on Cortex-A77~\cite{arm-cortex-a77-trm}, a high-performance energy-efficient microprocessor targeting mobile and laptop devices. 5,000 cycles of training data and 2,000 cycles of testing data are generated for Cortex-A77.
The numbers of RTL signals $M$ are $>5\times 10 ^5$ and $>1\times 10^6$ for Neoverse N1 and Cortex-A77, respectively. 

The RTL simulation is performed using VCS\textsuperscript{\textregistered}~\cite{vcs} and the ground-truth power is simulated by PowerPro\textsuperscript{\textregistered}~\cite{powerpro} based on a commercial 7nm technology setup. 
All ML models are implemented with Python v3.7. For baseline methods, CNN-based models are based on Pytorch v1.5~\cite{paszke2019pytorch}, and other models are implemented with scikit-learn v0.22 \cite{scikit-learn}. 
For APOLLO, we implement the MCP algorithm and the coordinate descent algorithm with NumPy~\cite{harris2020array}. 
During training, the MCP regressor converges within 200 iterations, with the threshold of unpenalized weights set to $\gamma=10$. The overall proxy selection and model training time of APOLLO and all baseline methods are within three hours, which is affordable.

All accuracies are measured on the testing data. Metrics include the coefficient of determination ($R^2$)~\cite{nagelkerke1991note}, the normalized root mean squared error (NRMSE), and the normalized mean absolute error (NMAE), defined as follows. The $\bar{y}$ is the average over all $N$ labels $y[i]$.

\begin{equation*}
 \text{NRMSE} = \frac{1}{\bar{y}} \sqrt{\frac{\sum_{i=1}^N (y[i] - p[i])^2}{ N } },\ \text{NMAE} = \frac{\sum_{i=1}^N | y[i] - p[i] |}{\sum_{i=1}^N y[i] }  \\   
\end{equation*}


\begin{table}[!h]
    \caption{Comparisons with Baseline Methods}
    \label{baseline}
      \centering
      \resizebox{0.8\linewidth}{!}{ \renewcommand{\arraystretch}{1.3}
        \begin{tabular}{ |c|c|c|c|c|c| } 
        \hline
         \multirow{2}{*}{Works} &  Simmani    &  PRIMAL  & PCA  &  Lasso &  \multirow{2}{*}{APOLLO}   \\
         & \cite{kim2019simmani} & \cite{zhou2019primal}  &  \cite{zhou2019primal}  &  \cite{pagliari2018all} &    \\
         \hline \hline
        Proxies Selection & K-means    & \XSolidBrush & \XSolidBrush  &  Lasso   &  MCP  \\ 
        \hline
         Pre-Processing   & Polynomial & \XSolidBrush &  PCA    &  \XSolidBrush  &  \XSolidBrush  \\
         \hline
          ML Model    &  Elastic Net  &  CNN    & Linear  &  Linear  &  Ridge \\ 
          \hline
        \end{tabular}
        }
\end{table}

For experimental comparisons, it is difficult to exhaust the significant body of previous researches for various target designs and application scenarios. Our solution is to compare the accuracy of APOLLO with representative approaches that target the highest accuracy with a high level of acceptable computation complexity. These complex non-linear methods~\cite{zhou2019primal, kim2019simmani} prove to outperform simple linear models adopted in most runtime approaches. We also compare with a recent runtime technique~\cite{pagliari2018all} which uses a sparsity-induced algorithm. Table \ref{baseline} shows comparisons with Simmani~\cite{kim2019simmani},  PRIMAL~\cite{zhou2019primal}, and Pagliari et al.~\cite{pagliari2018all}. 
For Simmani, signals are clustered with K-means algorithm and power proxies are selected from different clusters. 
After that, toggling activities of both the $Q$ power proxies and the $Q^2$ 2$^{\text{nd}}$ order polynomial terms are adopted as potential model features. The adopted elastic net model is a linear model with a combination of both Lasso and Ridge penalties, where the power measurement window size $T$ is a hyperparameter tuned to improve model accuracy.

PRIMAL~\cite{zhou2019primal} targets accurate design-time simulation on software with several methods, among which the CNN produces best results and is adopted for comparison. It uses all flip-flop signals as input proxies without any selection. As the number of flip-flops is at least one order of magnitude greater than typical values of $Q$, the simulation/emulation cost of PRIMAL is much higher than APOLLO. Moreover, the use of CNN makes it impractical for runtime OPM.
Another method proposed by PRIMAL~\cite{zhou2019primal} is principal components analysis (PCA). It shall be noted that dimension reduction techniques like PCA still require the toggling activities of all candidate signals as the initial input during inference. This is computationally expensive and fundamentally different from proxy selections. Pagliari et al.~\cite{pagliari2018all} adopt Lasso regression, the most widely-used sparsity-inducing method, for proxy selection and model construction.
For previous methods considering only flip-flop signals as input features, to avoid underestimation of their accuracy, we implement them with all RTL signals as input features for a fair comparison. This is expected to generate better accuracy than limiting proxies only to flip-flop signals.

\subsection{Accuracy of APOLLO}
\label{subsec:accuracy}

\begin{figure}[!h]
\begin{minipage}[t]{0.5\linewidth}
    \centering
    \includegraphics[height=0.75\textwidth]{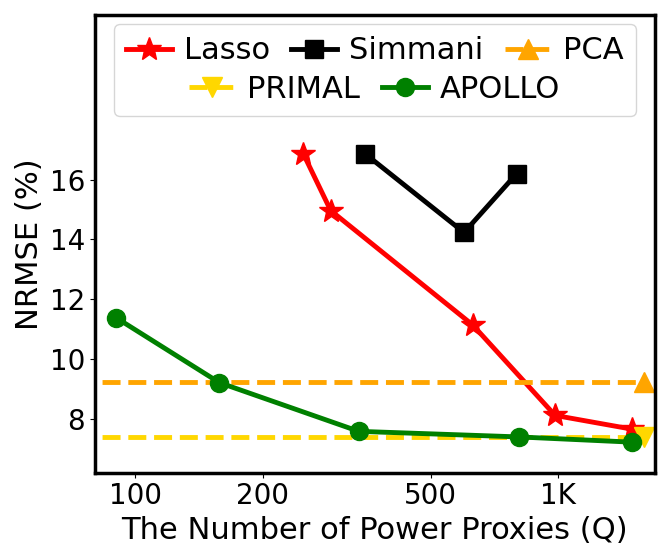}
\end{minipage}
\hspace{-2mm}
\begin{minipage}[t]{0.5\linewidth} 
    \centering
    \includegraphics[height=0.75\textwidth]{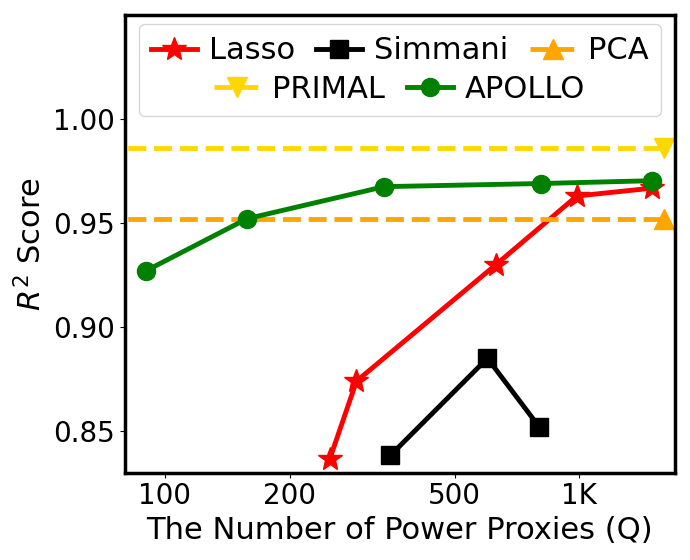}
\end{minipage}    
\vspace{-.2in}
\caption{Per-cycle power accuracy vs. number of proxies for per-cycle power prediction (Neoverse N1).}
\label{select_sigs}
\end{figure}

For per-cycle power estimation, APOLLO is compared with other methods in Figure~\ref{select_sigs}, which measures the trade-off between $Q$ and corresponding prediction accuracy on Neoverse N1. The previous Lasso-based method~\cite{pagliari2018all} and Simmani~\cite{kim2019simmani} are also applied to the per-cycle estimation for a fair comparison.
Both CNN in PRIMAL and the PCA model are represented by horizontal lines since their $Q=M$ in this comparison.
APOLLO achieves $\text{NRMSE}<10\%$ and $R^2 > 0.95$ with $Q\approx150$, which is less than $0.03\%$ of total RTL signals in Neoverse N1. It shows similar NRMSE when comparing PRIMAL with APOLLO at $Q=500$. In contrast, the NRMSE of Simmani and Lasso is higher than 12\% even with $Q=500$. This explains why the previous Lasso-based method~\cite{pagliari2018all} and Simmani~\cite{kim2019simmani} restrict their applications to coarse-grained temporal resolution.

\begin{figure*}[!h]
\centering
\includegraphics[width=\textwidth]{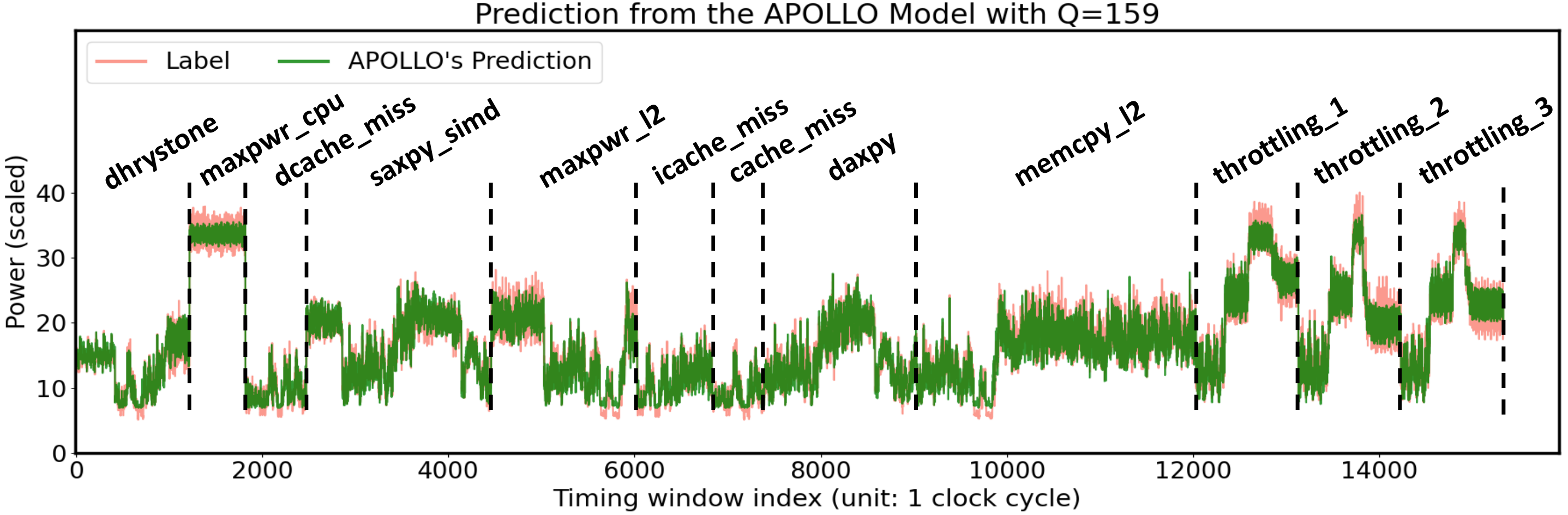}
\vspace{-.1in}
\caption{Evaluation of an APOLLO model with $Q=159$ (Neoverse N1).}
\label{visuala}
\end{figure*}

We provide a detailed evaluation of the APOLLO model with $Q=159$, which obtains $\text{NRMSE} =9.4\%$ and $R^2=0.95$.
Figure~\ref{visuala} illustrates prediction $p$ and label $y$ as power traces on the $15,000$-cycle testing dataset, covering all 12 handcrafted micro-benchmarks. APOLLO's prediction overlaps well with the ground truth for distinctive patterns from different benchmarks.
We also measured the accuracy in NRMSE and NMAE for each individual micro-benchmark. The NMAE is less than $10\%$ for all benchmarks. 

The APOLLO method can enable relative power comparisons across microarchitecture configurations, since it leads to generally unbiased power predictions that neither consistently over-estimate nor under-estimate a microarchitecture. Such unbiased predictions originate from the rich diversity in our automatically generated training data, covering both low- and high-power benchmarks of each design. As Figure~\ref{visuala} shows, \emph{averaged} predicted and ground-truth power are close for all testbenches on Neoverse N1. The averaged ground truth is $16.9$ and the prediction is $16.8$, showing merely $0.6\%$ difference (similar for Cortex-A77). Thus,  microarchitectural comparisons can be made easily if the relative difference in the power consumption exceeds this small error bar.

\begin{figure}[!h]
\begin{minipage}[t]{0.5\linewidth} 
    \includegraphics[height=0.7\textwidth]{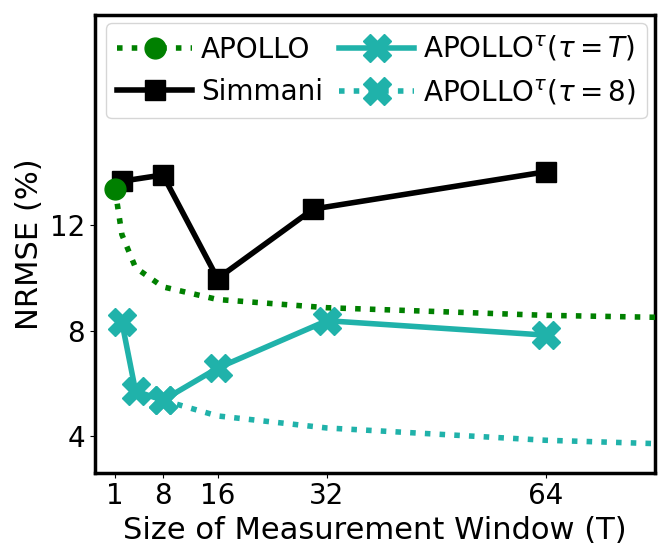}
    \end{minipage}
    \hspace{-.1in}
   \begin{minipage}[t]{0.5\linewidth}  
    \includegraphics[height=0.7\textwidth]{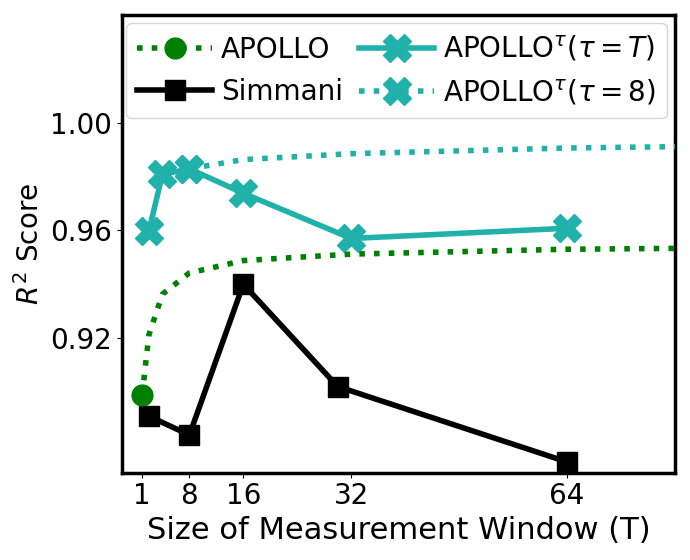}
    \end{minipage}
\vspace{-.2in}
  \caption{$T$-cycle accuracy vs. window size ($T$) for multi-cycle prediction (Neoverse N1). --- $Q=200$ for Simmani, $Q=70$ for APOLLO methods.}
  \label{multi}
\end{figure}

Figure~\ref{multi} estimates power over measurement windows with $T$ cycles. 
Previous multi-cycle model Simmani~\cite{kim2019simmani} is trained and validated for different $T$ values $\{4, 8, 16, 32, 64\}$. 
APOLLO in Figure~\ref{multi} stands for the simple average over $T$ per-cycle predictions. 
The green dotted line means predictions of various $T$ values are all averaged from the same \emph{per-cycle} APOLLO model. In comparison, several multi-cycle APOLLO$^\tau$ models with interval sizes $\tau=T=\{4, 8, 16, 32, 64\}$ are trained. Results show that $\tau=8$ provides the best accuracy. We thus choose $\tau=8$ for multi-cycle model and the dotted line is from APOLLO$^\tau$($\tau=8$) for all $T$ values. Notice that $Q=200$ for Simmani, while all APOLLO-based models keep $Q=70$. 
In Figure~\ref{multi}, the simple average of per-cycle APOLLO is already more accurate than Simmani for all $T$ values using around one-third of proxies. 
The multi-cycle APOLLO$^\tau$ with $\tau=8$ further improves NRMSE by $5\%$. This supports our claim in Section \ref{subsec:APOLLO-multi-cycle}, indicating that both simple average of per-cycle model ($\tau=1$) and directly averaging inputs for any $T$ ($\tau=T$) fails to provide the most accurate and robust solution.

\begin{figure}[!th]
\begin{minipage}[t]{0.5\linewidth}
    \centering
    \includegraphics[height=0.7\textwidth]{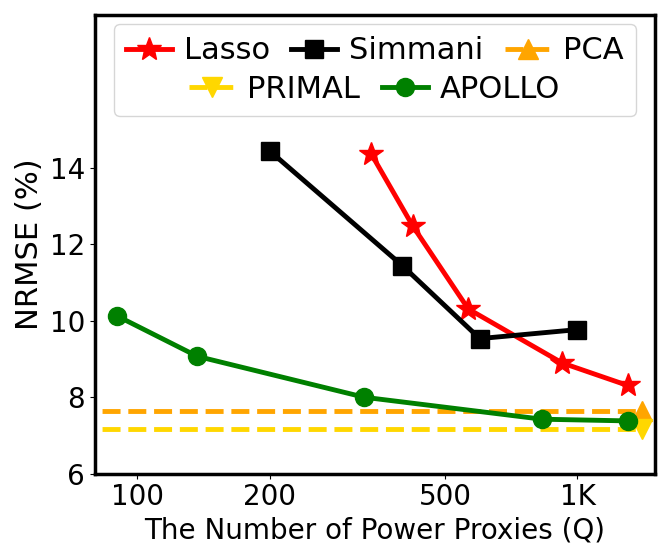}
\end{minipage}
\hspace{-.1in}
\begin{minipage}[t]{0.5\linewidth} 
    \centering
    \includegraphics[height=0.7\textwidth]{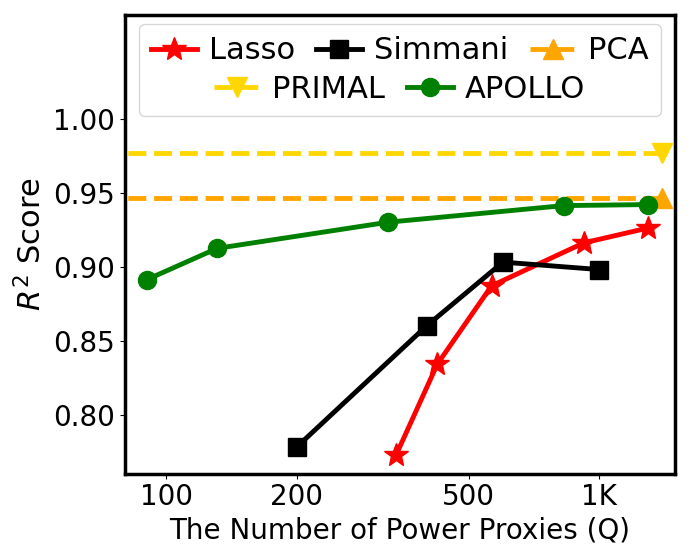}
    \hspace{-2mm}
\end{minipage}    
\vspace{-.2in}
\caption{Per-cycle power accuracy vs number of proxies for per-cycle power prediction (Cortex-A77).}
\label{other}
\end{figure}

To verify that APOLLO generalizes well on different designs, we measure the per-cycle accuracy on Cortex-A77. 
The comparisons are shown in Figure~\ref{other}. 
Similar to the trend in Figure~\ref{select_sigs}, APOLLO achieves NRMSE $=8\%$ when $Q\approx300$, which is less than 0.03\% of total RTL signals in Cortex-A77, while Simmani and Lasso show NRMSE $>10\%$ with $Q=500$. In addition, APOLLO obtains comparable NRMSE with the CNN in PRIMAL when $Q=500$. 

\subsection{Model Discussion}
\label{sec:discussion}

\begin{figure}[t]
\hspace{-2mm}
\begin{minipage}[t]{0.5\linewidth}
    \captionsetup{width=0.86\textwidth}
    \centering
    \includegraphics[height=0.7\textwidth]{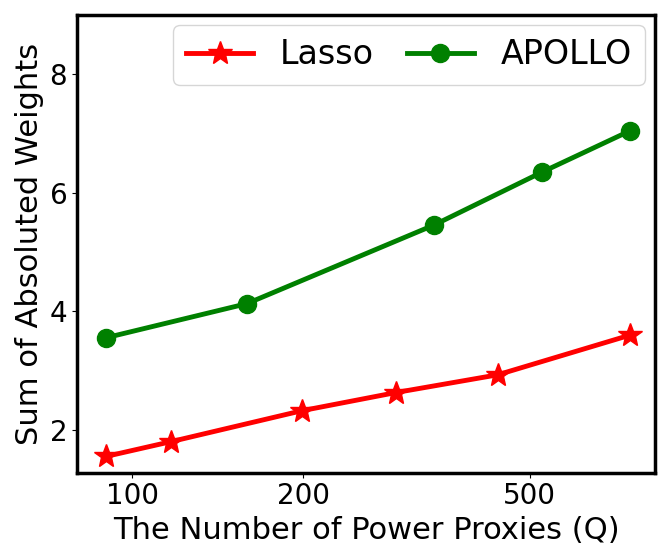}
    \vspace{-.08in}
    \caption{Sum of all absolute weights.}
    \label{fig:total_weight}
\end{minipage}
\begin{minipage}[t]{0.5\linewidth} 
    \captionsetup{width=0.86\textwidth}
    \centering
    \includegraphics[height=0.7\textwidth]{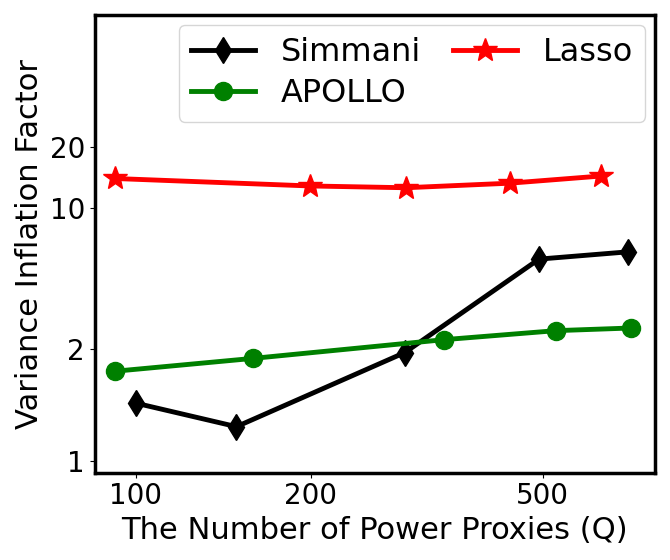}
    \vspace{-.08in}
    \caption{Variance inflation factors (VIF).}
    \label{vif}
\end{minipage}        
\vspace{-.05in}
\end{figure}

We provide insights into APOLLO's high-quality predictions from two additional perspectives. 
First, with the same $Q$, the MCP adopted by APOLLO allows large weights compared with the Lasso.
This is verified in Figure~\ref{fig:total_weight}, which reports the summation of all $Q$ absolute weights in each model.
Second, the correlation among the selected power proxies can jeopardize the generalization of models. Figure~\ref{vif} shows the average variance inflation factor (VIF)~\cite{xuegong2000introduction}, which quantifies the correlation among proxies for each method.
APOLLO shows a much lower VIF than Lasso regression. 
By shrinking weights with different rates, the MCP tends to treat correlated RTL signals differently so that correlated ones are not selected simultaneously as proxies. 
Another observation is that Simmani also achieves low VIF by selecting power proxies from different clusters. However, since the clustering-based selection is unsupervised, the correlation between power proxies and the label is not as directly optimized as APOLLO. Simmani is not covered in Figure~\ref{fig:total_weight} as it is not a linear model and its weights are not comparable with APOLLO/Lasso.

We further categorize the $Q$ APOLLO-extracted proxies based on the RTL signal properties:
1) determine whether a proxy is a gate clock signal;
2) for a non-clock RTL signal, determine which functional unit it belongs to. 
Figure~\ref{power_proxy_dist} shows the distributions of the 159 power proxies for Neoverse N1 CPU based on the aforementioned RTL signal properties. 
39 power proxies are gated clock signals, which means APOLLO captures the major contributor, i.e., clock network, of the dynamic power consumption.
Furthermore, with the APOLLO model, the weights of the gated clock signals provide useful insights into the power-hungry clock gating structure, which sets guidelines for designers to further optimize clock power. 
APOLLO model also captures significant power contributors, such as ``Vector Execution'' (19 out of 159), ``Issue'' (36 out of 159), and ``Load Store'' (28 out 159). 
These power proxies are critical indicators to enhance the throttling schemes and mitigate CPU maximum power consumption~\cite{arm-neoverse-n1-trm}.  

\subsection{Hardware Prototype of APOLLO-OPM}
\label{subsec:APOLLO-pmu-hw}

\begin{figure}[!tb]
\centering
\vspace{-.1in}
    \subfigure[]{\includegraphics[height=0.27\textwidth]{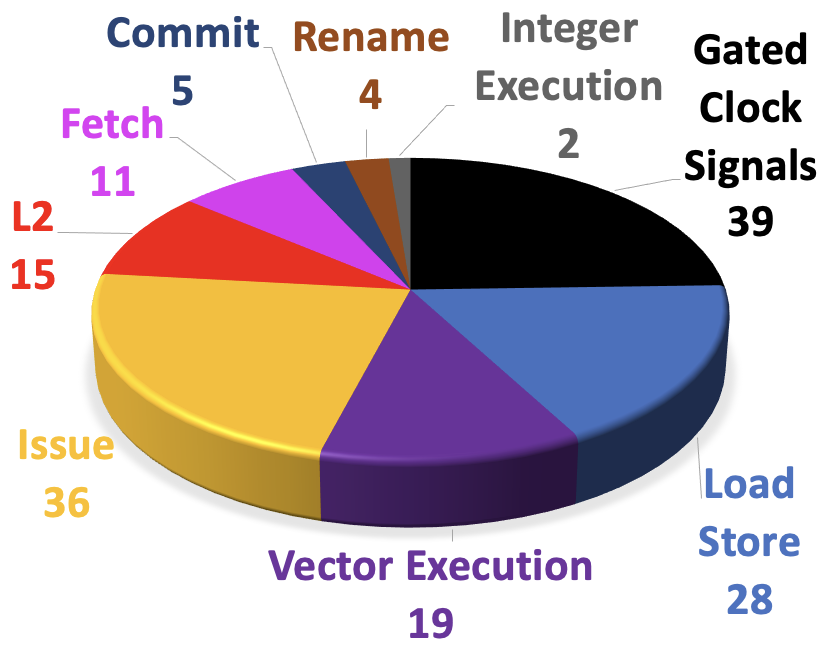}
    \label{power_proxy_dist}}
    \hspace{.2in}
    \subfigure[]{\includegraphics[height=0.3\textwidth]{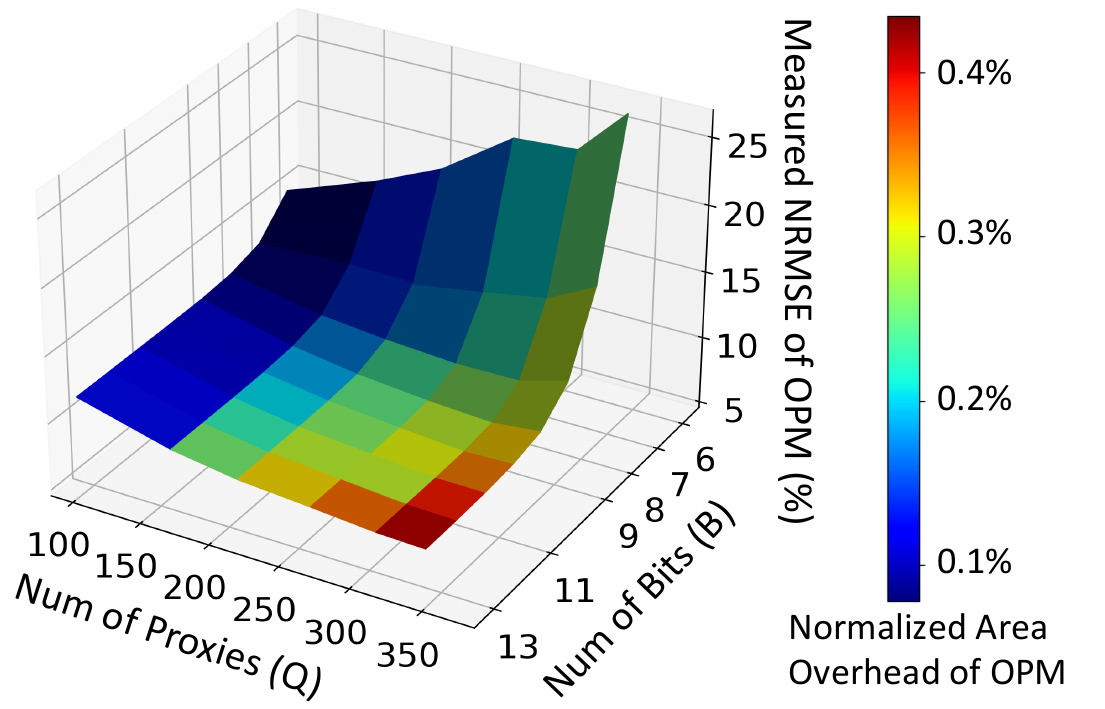}
    \label{area}}
\vspace{-.1in}
\caption{(a) Distribution of extracted proxies from Neoverse N1. (b) Trade-off between the area overhead and accuracy (NRMSE) of the OPM.}
\end{figure}

We synthesize the APOLLO model as an OPM under the same target frequency and 7nm technology as Neoverse N1 CPU. 
The model accuracy is measured in NRMSE and the cost is quantified by area overhead. 
The trade-off between accuracy and area normalized by the total gate area of Neoverse N1 is shown in Figure~\ref{area}. 
By varying the number of selected proxies $Q$ and the number of bits $B$ used for weight quantization, such trade-off curve is explored to help determine appropriate values for $Q$ and $B$. Although we are exploring the area and accuracy trade-off using a per-cycle power model, our automated OPM generation accommodates the average power computation over $T$ cycles and the only extra hardware cost is one $B+\lceil$logQ$\rceil+\lceil$logT$\rceil$-bit flip flop and adder.
To evaluate the accuracy of this implementation, we simulate our hardware solution with the 15,000-cycle testing data of Neoverse N1.
According to Figure~\ref{area}, both $Q$ and $B$ have a considerable impact on accuracy and area. 
For all $Q$ values, the accuracy loss is high for $B < 9$ and becomes negligible when $B > 10$. 
Thus, our strategy is to keep $B\approx 10$ and vary $Q$ to generate different solutions. Specifically, with 10-bit weights, 
the quantization leads to $<0.1\%$ NRMSE increase compared with the APOLLO model on software at design-time. 
For an OPM with $B=10$ and $Q=159$, its total gate area is only $0.2\%$ of the gate area of Neoverse N1. It has a latency of 2 cycles.

OPM overheads are analyzed using physical implementation estimations with the overall Neoverse N1 CPU, for the OPM placement region at a central location within the CPU floorplan, bounded as illustrated in Figure \ref{introduct}. Individual proxies routed from different blocks to the centralized OPM require buffering that incurs area and power overheads. On the Neoverse N1 CPU, we budget a single clock cycle to account for the latency of routing multiple proxies to the OPM by registering all inputs at the OPM interface (Figure \ref{opm_design}), at the expense of an extra cycle latency. 

Driving the proxies to the centralized OPM requires high-strength buffers that contribute an additional 0.4$\%$ power overhead. The OPM circuitry itself consumes 0.5$\%$ power overhead, leading to an overall power overhead of 0.9$\%$ compared to the baseline CPU power at 3GHz in a commercial 7nm technology. In comparison, the reported power overheads of all previous proxy-based runtime monitors are $1.9-14\%$~\cite{cremona2020automatic}, $2.7-4\%$~\cite{najem2016design}, $5.7\%$~\cite{pagliari2018all}, $10\%$~\cite{zoni2018powertap}, and $4.7\%$~\cite{zoni2018powerprobe}. The total area overhead remains negligible ($<0.5\%$).

\subsection{Application Scenarios}

\begin{figure}[!tb]
    \centering
    \includegraphics[width=\textwidth]{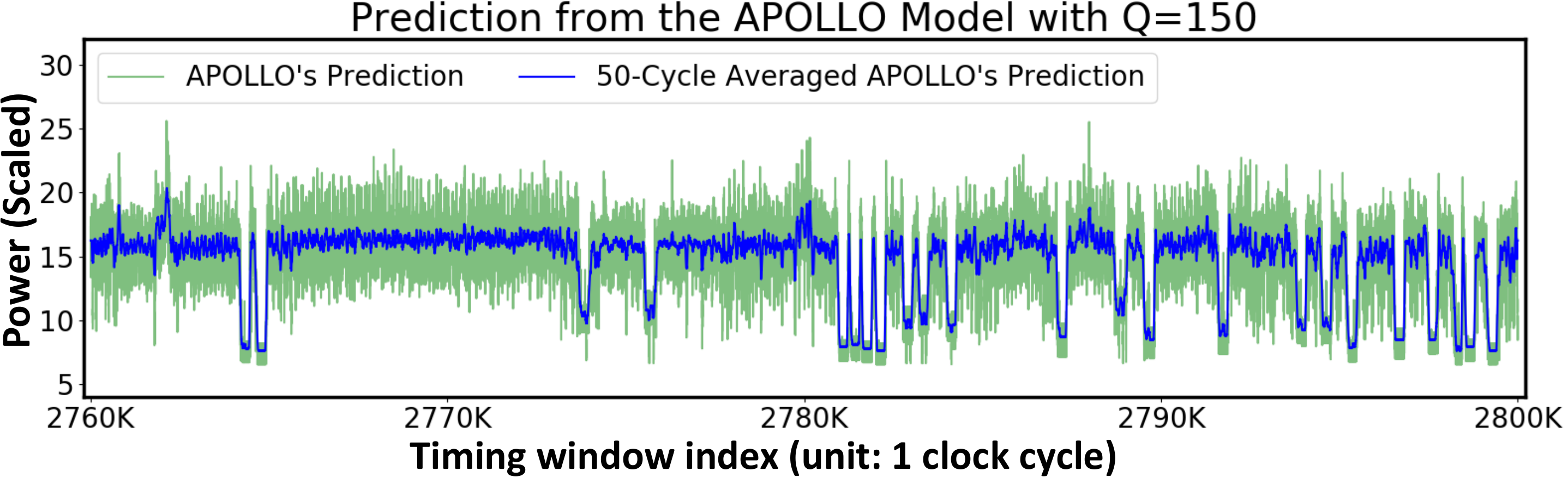}
    \caption{A portion (40,000 over 17 million cycles) of power estimation from the APOLLO-integrated emulator-assisted power analysis (Neoverse N1).}
    \vspace{-.05in}
    \label{visual_EBPF}
\end{figure}

\textbf{{Design-Time Power Introspection}} We described in Section \ref{sec:APOLLO-ebpf} how the APOLLO model can be integrated into an emulator-assisted workload simulation framework. By only recording the toggle trace of $Q=150$ power proxies, the size of a simulation trace with $N=$ 17 million cycles on Neoverse N1 is reduced to only 1.1 GB. The entire trace is generated on Palladium\textsuperscript{\textregistered} Z1 emulation platform~\cite{emulation-platform} within 3 minutes. This capability enables accurate generation of power trace spanning >10M processor cycles within minutes, enabling unprecedented design-time power introspection. Figure~\ref{visual_EBPF} illustrates this in the power trace generated for the ``hmmer'' benchmark from the SPEC2006 on the Neoverse N1. We show only a portion (40,000 cycles) of the whole trace to illustrate distinct transitions in the CPU power and current demand.

Achieving this using EDA tools is computationally infeasible for industry-scale CPU designs. We estimate the inference time on one billion cycles, covering $1/3$ of a second in chip runtime for the 3GHz Neoverse N1. With a linear model, APOLLO inference only takes one minute with $Q < 500$. In comparison, the CNN model in PRIMAL takes months and the PCA takes around one week, since both algorithms do not perform proxies selection. As for Simmani, since it takes approximately $Q^2$ polynomial terms as input, its inference time can increase quadratically with $Q$. It may take Simmani days for inference of a billion cycles when $Q=1000$.

\begin{figure}[!tb]
\centering
\includegraphics[width=0.8\textwidth]{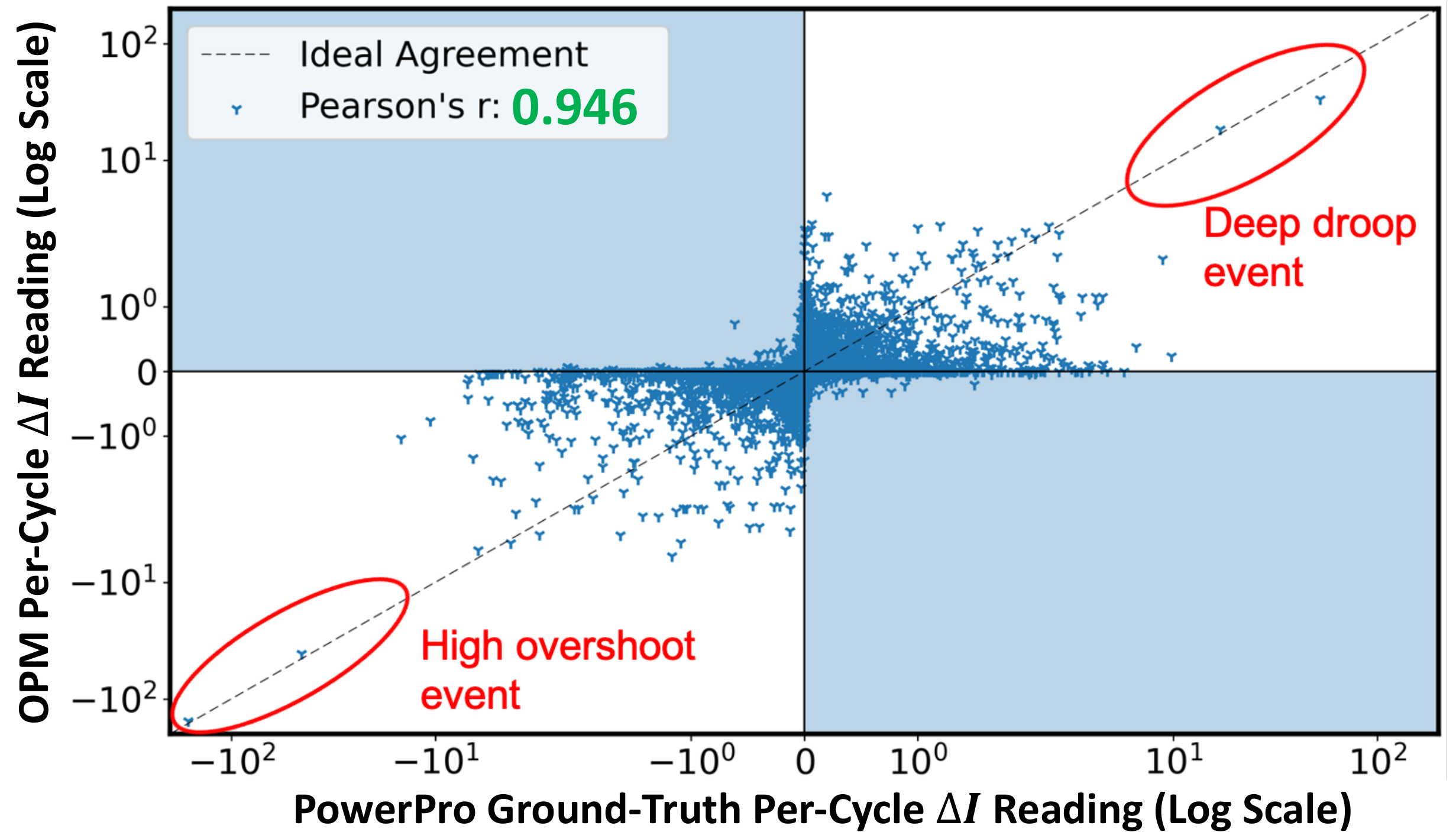}
   \caption{Voltage droop analysis based on per-cycle power on Neoverse N1, showing OPM prediction versus ground-truth (scaled to arbitrary units).} 
      \label{fig:VoltageDroop}
\end{figure}

\textbf{Runtime Proactive Ldi/dt Mitigation} Using the APOLLO-OPM's per-cycle estimation capability, it is possible to predict $Ldi/dt$ voltage-droop events ahead of time before their actual occurrence at a low cost\footnote{In~\cite{reddi2009voltage}, authors describe an online training approach where a voltage-emergency signature is dynamically learned to predict future noise events. This approach requires a checkpoint and recovery mechanism for initial failures when no signature has been learned. This approach is onerous to implement in industrial CPU designs. Correctness in presence of corner cases is difficult to guarantee.}.
We intend to develop this further in our future work, but here we provide a brief conceptual description of how this can be realized using the OPM. 
The differentiation ($di/dt$) operator in continuous time is equivalent to the differencing ($\Delta I$) in discrete time. 
We plot both the OPM readings on Neoverse N1 and the ground truth $\Delta I$ samples (scaled to arbitrary units) from PowerPro~\cite{powerpro} in a scatter plot in Figure~\ref{fig:VoltageDroop}.
Note that the plot is in log-scale to cover a wide data range with visibility to details, magnifying the uncorrelated samples that are actually small in magnitude. The Pearson's correlation~\cite{lee1988thirteen} between OPM and the ground truth reaches 0.946, indicating a high correlation.

The points in the bottom-right and the top-left quadrants indicate samples where OPM estimations depart significantly from the ground truth. The signal magnitudes recorded in these quadrants are near the origin (indicating small-magnitude delta current) as a consequence of the OPM accuracy.
Points in the top-right quadrant indicate cycles where there is an increased current demand relative to the previous cycle. Such cycles are typically precursors to voltage-droop events. The bottom-left quadrant indicates a drastic reduction in current demand leading to potential voltage-overshoots. 
For the samples in deep droop and overshoot regions, APOLLO OPM correlates well with the ground truth. This indicates that the OPM can accurately estimate CPU current transients, and thus enable circuit-level mitigation schemes such as adaptive-clocking to engage prior to the development of voltage-droop.  

Proactive droop mitigation using proxies has been proposed in prior art~\cite{webel2020proactive, kalyanam2020proactive}. In \cite{webel2020proactive}, authors describe a combination of pipeline event indicators and digital power-proxies for droop-event indication. However, the technique for creating this proxy is not formally described. The work of \cite{kalyanam2020proactive} describes proactive mitigation on the Hexagon DSP engine. DSP engines are data-plane dominated, in contrast with CPUs that are control-plane dominated. As such, manual design for CPU power-proxies is significantly harder, particularly when fine-grained temporal resolution is necessary.

\section{Summary}

Power introspection is increasingly important in modern high-performance CPU designs, for both design-time optimization and runtime management. This has particular significance in many-core infrastructure SoCs in ultra-scaled technology nodes. Within a unified framework, APOLLO bridges an important technology gap by providing both cycle-accurate design-time power simulation and low-overhead on-chip power metering. We demonstrate that by monitoring $<0.05\%$ RTL signals, the OPM achieves $R^2 > 0.95$ with <$1\%$ area/power overhead when integrated with the Neoverse N1 CPU core. 

The future research can be focused on two directions. Firstly, the margin reduction can be further developed and quantified using proactive $Ldi/dt$ mitigation with OPM. Secondly, the APOLLO design-time model can be translated into higher abstraction models (C/C++ instead of RTL), thereby integrating performance simulation with power-tracing. Ultimately, the APOLLO capability may enable the development of new mechanisms for smarter power and thermal management in future SoCs. The framework is extensible to diverse compute engines and is therefore a compelling addition to the microarchitects' toolbox.

%% file: txt/3_net_time.tex
\chapter{Net Length and Timing Modeling at Netlist}
\label{chapter:net_time}

\section{Background}

{\bf Net Length Estimation:} For state-of-the-art semiconductor manufacturing technology nodes, interconnect is a dominating factor for integrated circuit (IC) performance and power, e.g., it can contribute to over 1/3 of clock period~\cite{hyun2019accurate} and about 1/2 of total chip dynamic power~\cite{MagenSLIP04}. Interconnect characteristics are affected by almost every step in a design flow, but not explicitly quantified and optimized until the layout stage. 
Therefore, previous academic studies attempted to address the interconnect effect in design steps prior to layout, e.g., layout-aware synthesis~\cite{PedramDAC91,PedramICCAD91}. 
To achieve such a goal, an essential element is to enable fast yet accurate pre-layout net length prediction, which has received significant research attention in the past~\cite{liu2012neural, hyun2019accurate, bodapati2001prelayout,fathi2009pre,hu2003wire,kahng2005intrinsic,liu2004pre}. Some works \cite{bodapati2001prelayout,fathi2009pre} pre-define numerous features describing each net, then a polynomial model is built by fitting these features.
The work of \cite{liu2012neural} estimates wirelength by artificial neural networks (ANN), but it is limited to the total wirelength on an FPGA only, which is easier to estimate than individual net length. The mutual contraction (MC) \cite{hu2003wire} estimates net length by checking the number of cells in every neighboring net. The intrinsic shortest path length (ISPL) \cite{kahng2005intrinsic} is an interesting heuristic, which finds the shortest path between cells in the net to be estimated, apart from the net itself. The idea in \cite{liu2004pre} is similar to \cite{kahng2005intrinsic} in measuring the graph distance between cells in the netlist. The recent work \cite{hyun2019accurate} on wirelength prediction can only estimate the wirelength of an entire path instead of individual nets, and it relies on the results from virtual placement and routing.

Although net length prediction has been extensively studied previously, we notice a major limitation in most works.
That is, they only focus on the local topology around each individual net with an over-simplified model. In other words, when estimating each net, usually their features only include information from nets one or two-hop away. The big picture, which is the net's position in the whole netlist, is largely absent. However, a placer normally optimizes a cost function defined on the whole netlist. It is not likely to achieve high accuracy without accessing any global information. Some previous models indeed attempt to embrace global information like the number of 2-pin nets in an entire circuit~\cite{bodapati2001prelayout,fathi2009pre}, or a few shortest paths~\cite{kahng2005intrinsic}, but such information is either too sketchy~\cite{bodapati2001prelayout,fathi2009pre} or still limited to a region of several hops~\cite{kahng2005intrinsic}. 
Since the global or long-range impact on individual nets is much more complex than local circuit topologies, it can hardly be captured by simple models or models with only human-defined parameters that cannot learn from data. 
To solve this, we propose a new approach, called Net$^\mathbf{2}$, based on graph attention network~\cite{velivckovic2017graph}.
Its basic version, Net$^\mathbf{2f}$, intends to be fast yet effective. 
The other version, which emphasizes more on accuracy and is
denoted as Net$^\mathbf{2a}$, 
captures rich global information with a highly flexible model
through circuit partitioning. 

Recently, deep learning has generated a huge impact on many applications where data is represented in Euclidean space. However, there is a wide range of applications where data is in the form of graphs. Machine learning on graphs is much more challenging as there is no fixed neighborhood structure like in images. All neural network-based methods on graphs are referred to as graph neural networks (GNN). The most widely-used GNN methods include graph convolution network (GCN)~\cite{kipf2016semi}, graphSage (GSage)~\cite{hamilton2017inductive}, and graph attention network (GAT) \cite{velivckovic2017graph}. They all convolve each node's representation with its neighbors' representations, to derive an updated representation for the central node. Such operation essentially propagates node information along edges and thereby topology pattern is learned.

Similarly, in EDA, circuit designs are embedded in Euclidean space after placement, which inspired many CNN-based methods~\cite{xie2018routenet,fang2018machine,xie2020powernet}. But before placement, a circuit structure is described as a graph and spatial information is not yet available. Till recent years, GNN is explored for EDA applications~\cite{ma2019high,zhang2019circuit}.  
The work in \cite{ma2019high} predicts observation point candidates with a model similar to GSage~\cite{hamilton2017inductive}. Graph-CNN~\cite{zhang2019circuit} predicts the electromagnetic properties of post-placement circuits. This method is limited to very small-scale circuit graphs with less than ten nodes. Overall, GNN has great potential but is much less studied than CNN in EDA. Based on GAT, the fast version of our Net$^\mathbf{2}$ is $1000\times$ faster than placement, and the accurate version of Net$^\mathbf{2}$ significantly outperforms plug-in use of existing GNN techniques.

{\bf Timing Estimation:} In digital circuit design, timing is a primary design objective that needs to be considered since very early design stages. A fast and accurate pre-placement timing estimator can essentially benefit design automation by providing early and high-fidelity feedback to synthesis solutions or during the timing-driven placement.
However, accurate timing estimation is extremely challenging before placement, largely due to the absence of wire length information. It is highly difficult to estimate the impact from wires when locations of all cell instances have not been fixed. In some commercial tools~\cite{Innovus}, the timing engine ignores or under-estimates the wire load before placement. As a result, they fail to correlate well with the post-placement timing report. 


ML techniques are also proposed for timing prediction. But due to aforementioned challenges, almost all existing ML-based timing estimators~\cite{barboza2019machine, kahng2013learning, kahng2015si, han2014deep} are only applied after placement for sign-off timing analysis. Barboza et al.~\cite{barboza2019machine} reduce the pessimism in the pre-routing timing report from current commercial tools. The work of \cite{kahng2013learning} makes predictions with its incremental STA tools and \cite{kahng2015si} predicts sign-off timing based on non-SI (signal integrity) analysis. 
Besides these timing estimators, some ML-based flow tuning methods~\cite{xie2020fist} optimize their flow for better timing. They typically treat the design as a black box by training one separate model for each design, and only predict the overall quality like WNS (worst negative slack) without providing any detailed timing predictions on each net or path. Compared with our method which predicts the delay at every individual net, this type of black-box predictions are significantly more coarse-grained, less challenging, and apply to fewer scenarios.

In this work, we propose to address the absence of wire information and provide an accurate pre-placement timing estimator with our knowledge from net length estimation. Both features and predictions of our net size estimator Net$^\mathbf{2}$ are selected as input to timing prediction. Different from a representative timing estimator~\cite{barboza2019machine}, which incorporates both gate and wire delays to a net and does not differentiate multiple input pins of the same cell, we estimate the delay of every individual cell arc and net arc. To accomplish this, in our timing estimator, we construct two separate timing models for cell arc and net arc with different input features.

\section{Methodology}

\subsection{Problem Formulation}

\begin{figure}[!bt]
  \centering
    \includegraphics[width=0.8\columnwidth]{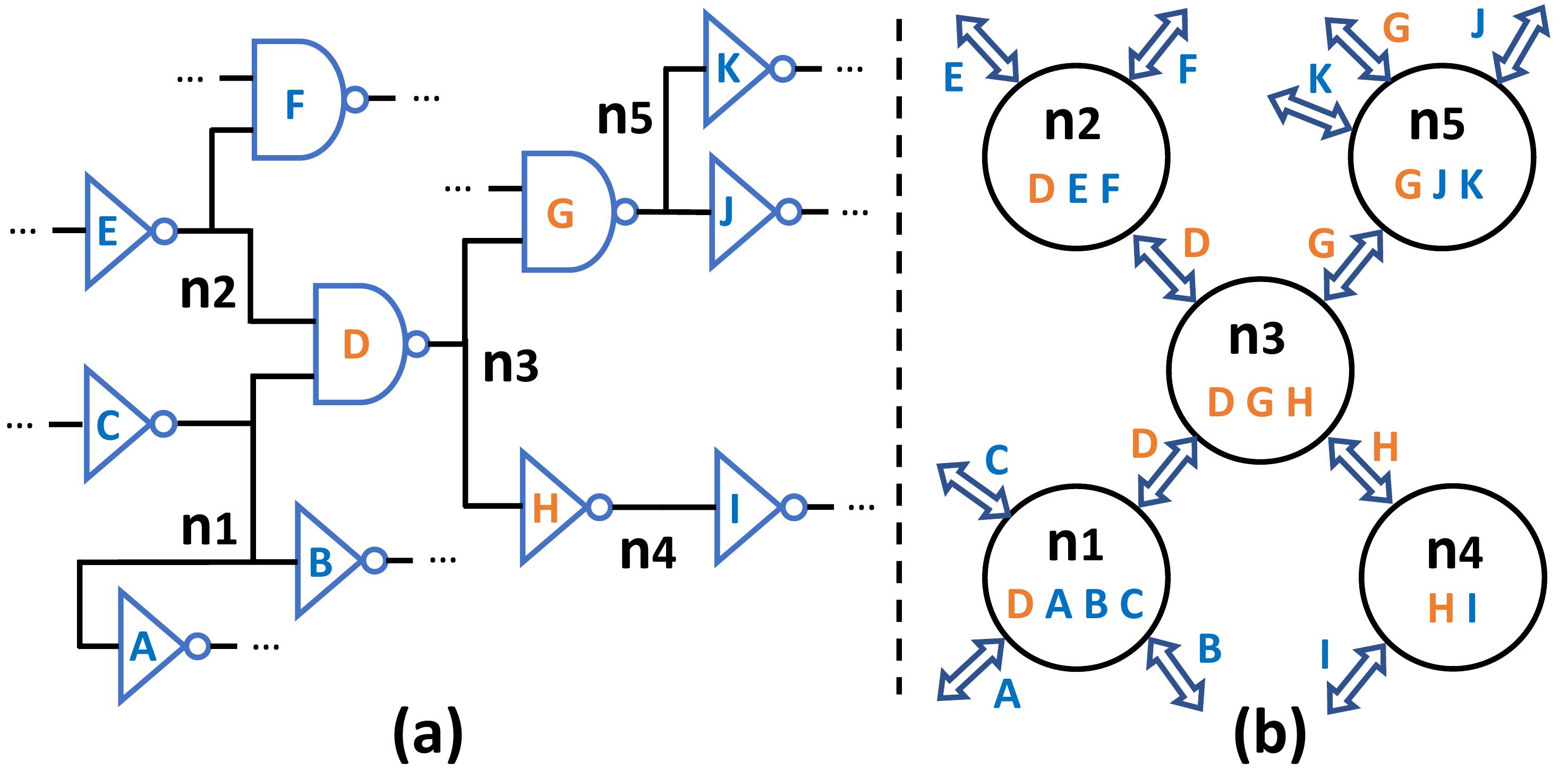}
  \caption{Convert netlist to graph. (a) Example netlist. (b) Corresponding graph.}
  \label{method_circuit}
\end{figure}

The major target in this work is to predict the size of each net with pre-placement features. The net length $L^k$ of each individual net $n_k$ is the label for training and prediction. The net length is the half perimeter wirelength (HPWL) of the bounding box of the net after placement. The features of each net are based on the connection information derived from the circuit netlist. These features include information about each analyzed net's driver, sinks, fan-in size, fan-out size, and the number of neighbors. In addition, our method directly processes the netlist as a graph to capture global information of the whole circuit design.

We define terminologies of relevant features with the example in Figure~\ref{method_circuit}, and commonly-used notations throughout this paper are all summarized in Table~\ref{tbl:notation}. Figure~\ref{method_circuit}(a) shows part of a netlist, including five nets \{$n_1$, $n_2$, $n_3$, $n_4$, $n_5$\} and 11 cells \{$c_A$, $c_B$, ..., $c_K$\}. Now we focus on net $n_3$, which touches 3 cells \{$c_D$, $c_G$, $c_H$\} and is referred to as a \emph{3-pin net}. Its \emph{driver} is cell $c_D$; its \emph{sinks} are cells \{$c_G$, $c_H$\}. We denote the area of $n_3$'s driver cell as $a_{dri}^3$. Net $n_3$'s \emph{fan-ins} $N^3_{in} =$ \{$n_1$, $n_2$\} ; its \emph{fan-outs} $N^3_{out} =$ \{$n_4$, $n_5$\}. Its \emph{fan-in size} is 2, denoted as $|N^3_{in}| = 2$. Its \emph{fan-out size} (number of sinks) is 2, denoted as $|N^3_{out}| = 2$. Every net can have only one driver but multiple sinks. Thus, the number of cells $= 1 + |N^3_{out}| = 3$ for this net. Net $n_3$'s one-hop neighbors include both its fan-in and fan-out: $\mathcal{N}(n_3) = N_{in}^3 \cup N_{out}^3 = $ \{$n_1$, $n_2$, $n_4$, $n_5$\}. The number of its neighbors is also known as the degree of $n_3$: $deg(n_3) = |\mathcal{N}(n_3)| = 4$.

\begin{table}[!t]
  \centering
  \caption{Notations Commonly Used in this Chapter}
  \label{tbl:notation}
  \resizebox{\linewidth}{!}{ \renewcommand{\arraystretch}{1.3}  
  \begin{tabular}{l | l  | l | l | l | l }
 	\hline
 	\multirow{2}{*}{Notation}  &  \multirow{2}{*}{Description} &  \multirow{2}{*}{Notation} & 
 	 \multirow{2}{*}{Description}    &  \multirow{2}{*}{Notation} & 
 	 \multirow{2}{*}{Description}  \\
 	  &     &     &     &     &     \\
	\hline
	$n_k$             & net or node    &  $O_k$  & node features  &  $P[c_H]$    &  cell partition IDs    \\
	\hline
	\{$c_G$, $c_H$\}  & cells    &  $E_{b \rightarrow k}$  & edge features   &  $M[n_k]$   &  node partition IDs  \\
	\hline
	$a^k_{dri}$ &   driver's area  &  $\mathcal{N}(n_k)$ & 1-hop neighbors  & $N^k_{in}$, $N^k_{out}$  &  fan-in/fan-out nets  \\
	\hline
	$h^{(t-1)}_k$, $h^{(t)}_k$ & GNN embeddings    &   $deg(n_k)$       &  degree of net & $|N^k_{in}|$, $|N^k_{out}|$      &  fan-in/fan-out size        \\
	\hline
	$W^{(t)}$, $\theta^{(t)}$ & learnable weights    &   $L^k$          &  net length label       &  $g(\,), \sigma(\,)$ & activation functions  \\
	\hline
	  C2.A$\Rightarrow$C2.Z &  timing arc   & C2.A  &  pin A of cell C2            & $std(\,)$, $\mu(\,)$ & std deviation, mean      \\
	\hline
	$c_{kb}$ or $c_{bk}$ &   cell on $n_b \rightarrow n_k$   &  $\backslash$ & exclude from list  &  $[\,\,\,||\,\,\,]$ & concat to one list  \\ 
	\hline
	$n_b \rightarrow n_k$ & directional edge &  Time$^\mathbf{f/a}$ & timing estimator  &	Net$^\mathbf{2f/2a}$ &  net size estimator   \\
	\hline
  \end{tabular}
  }
\end{table}


To apply graph-based methods, we convert each netlist to one directed graph. Different from most GNN-based EDA tasks, net length prediction focuses on nets rather than cells. Thus we represent each net as a node, and use the terms \emph{node} and \emph{net} interchangeably. For each net $n_k$, it is connected with its fan-ins and fan-outs through their common cells by edges in both directions. The common cell shared by both nets on that edge is called its \emph{edge cell}. For example, in Figure \ref{method_circuit}(b), net $n_3$ is connected with nets $n_4$ and $n_5$ through its sinks $c_G$ and $c_H$; it is connected with nets $n_1$ and $n_2$ through its driver $c_D$. The edges through edge cell $c_G$ is denoted as $n_3 \rightarrow n_5$ and $n_5 \rightarrow n_3$. The edge cell $c_G$ can also be referred to as $c_{35}$ or $c_{53}$. We differentiate edges in different directions because we will assign different edge features to $n_3 \rightarrow n_5$ and $n_5 \rightarrow n_3$. 

An important concept throughout this paper is global and local topology information. We use the number of hops to denote the shortest graph distance between two nodes on a graph. The \emph{information} of each net refers to its number of cells and driver's area. Local information includes the information about the estimated net itself, or from its one to two-hop neighboring nets. In contrast, global information means the pattern behind the topology of the whole netlist or the information from nets far away from the estimated net $n_k$. Here we define the `far away' of global information as at least three-hop away from the analyzed net. This is beyond the scope of several previous methods~\cite{hu2003wire, fathi2009pre}. By performing clustering/partitioning, the global information can incorporate the information from the whole netlist, reaching the furthest net. The range of neighbors that can be accessed by each model is referred to as the model's \emph{receptive field}.

\subsection{The Overall Flow}

\begin{figure}[!t]
  \centering
    \includegraphics[width=0.8\columnwidth]{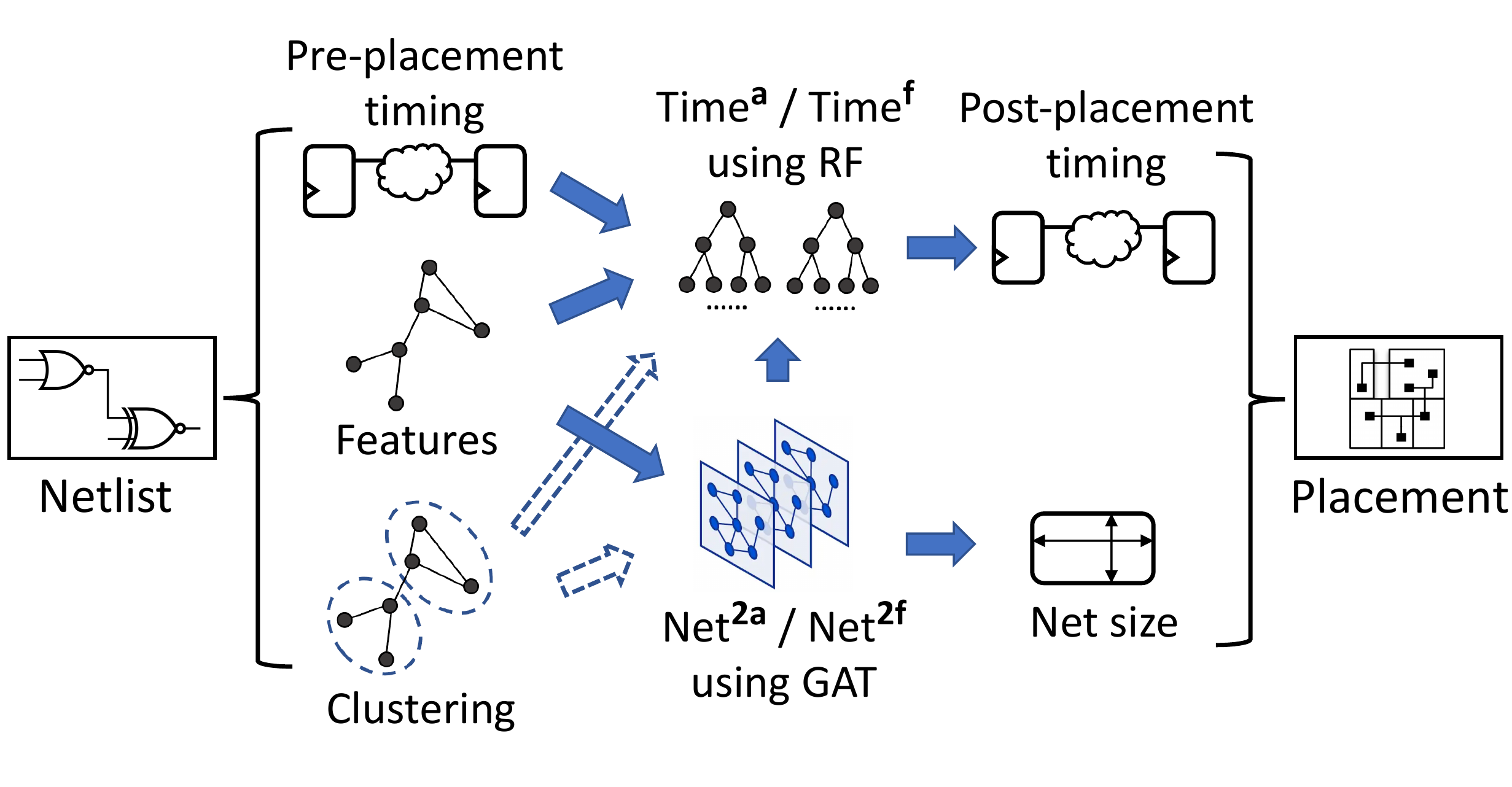}
  \caption{The net size and timing prediction flow.}
  \label{net:flow}
\end{figure}

Figure~\ref{net:flow} shows the overall pre-placement flow for both individual net size and timing predictions. It is applied before layout and predicts post-placement design objectives. Prediction results can benefit optimization and evaluation for both synthesis and placement. For our net length estimator Net$^{\mathbf{2}}$, we develop a fast version and an accuracy-centric version named Net$^{\mathbf{2f}}$ and Net$^{\mathbf{2a}}$, respectively. As Figure~\ref{net:flow} shows, both versions of Net$^{\mathbf{2}}$ extract features directly from the netlist, while Net$^{\mathbf{2a}}$ further captures global information by performing clustering on the circuit netlist. As for timing prediction, we also provide both accurate and fast versions of timing estimators, named Time$^{\mathbf{a}}$ and Time$^{\mathbf{f}}$. The dashed blue arrows in Figure~\ref{net:flow} mean the arrows only hold for accuracy-centric versions of our methods, like Net$^\mathbf{2a}$ and Time$^\mathbf{a}$. It indicates that the clustering/partitioning information is only utilized by Net$^\mathbf{2a}$ and Time$^\mathbf{a}$, providing higher accuracy at the cost of extra runtime for clustering. Besides features used by net size prediction, the pre-placement timing report from commercial EDA tools is also used as the input. The timing estimators also utilize the information from net size predictions as important input features.


\subsection{Node Features on Graph}
\label{netsize_node}

\begin{algorithm}[!tb]
\caption{Graph Generation with Node Features}
\label{alg1}
  \textbf{Input}: Basic features \{$|N_{in}^k|$, $|N_{out}^k|$, $a_{dri}^k$\}, net length label $L^k$, \\ \hspace*{15mm}the fan-in nets $N_{in}^k$ and fan-out nets $N_{out}^k$ of each net $n_k$. \\
  \textbf{Generate Node Features}:
\begin{algorithmic}[1]
\For{each net $n_k$}
\State $in_{in} =$ [\,], $in_{out} =$ [\,], $a_{all} = [a_{dri}^k]$, $out_{in}=$ [\,], $out_{out} =$ [\,] 
\For{each net $n_i \in N_{in}^k$, each net $n_o \in N_{out}^k$} \label{line:it_i}
\State $in_{in}$.add ($|N_{in}^i|$) \textbf{;} $in_{out}$.add ($|N_{out}^i|$) \textbf{;} $out_{in}$.add ($|N_{in}^o|$) 
\State $out_{out}$.add ($|N_{out}^o|$) \textbf{;} $a_{all}$.add($a_{dri}^o$) \label{line:it_oa}
\EndFor
\State $O_k$ = \{$|N_{in}^k$|, $|N_{out}^k|$, $a_{dri}^k$, $\sum {a_{all}}$, $\sum{out_{in}}$, $\sum{out_{out}}$, $\sum in_{in},   \newline
\hspace*{20mm} \sum in_{out}$, $std(out_{in})$, $std(out_{out})$, $std(in_{in})$, $std(in_{out})$\}\label{line:node_f}
\EndFor
\end{algorithmic} 
\textbf{Build Graph}:
\begin{algorithmic}[1]
\State Initiate a graph $G$. Each net is a node.
\For{each net $n_k$}
\State For node $n_k$ in $G$, set $O_k$ as node feature, $L^k$ as label.
\For{each net $n_b \in N_{in}^k \cup N_{out}^k$}
\State Add directed edge $n_b$ \textrightarrow $n_k$.
\EndFor
\EndFor
\end{algorithmic} 
  \textbf{Output}: Graph $G$ with node features $O$ and label $L$.
\end{algorithm}

Algorithm \ref{alg1} shows how we build a directed graph and generate features for each node with a given netlist.
On average, a net with more large cells tends to be longer. Thus, the most basic net features include the net's driver's area, fan-in and fan-out size \{$|N^k_{in}|$, $|N^k_{out}|$, $a_{dri}^k$\}. Feature $\sum a_{all}$ is the sum of areas over all cells in $n_k$. It is calculated by including the drivers of all $n_k$'s fan-outs in line \ref{line:it_oa}, which are the sinks of $n_k$. Besides these basic features, we capture the more complex impact from neighbors. As shown in line \ref{line:it_i}, we go through all neighbors of $n_k$ to collect their fan-in and fan-out sizes. The summation $\sum$ and standard deviation $s()$ of these neighboring information are added to node features $O_k$ in line \ref{line:node_f}.

\subsection{Edge Features}

In Algorithm \ref{alg1}, node features $O_k$ include up to two-hop neighboring information. The receptive field of the GNN method itself depends on the model depth, which is usually two to three layers. Thus the model can reach as far as four to five-hop neighbors, which is already more than previous works. To achieve a good trade-off between accuracy, speed, and computation cost, our fast-version model Net$\mathbf{^{2f}}$ adopts this conservative and efficient setting to reach as far as five hops. But for the accuracy-centric Net$\mathbf{^{2a}}$, it goes way beyond that to capture more global information from the whole graph.


To capture global information, we use an efficient multi-level partitioning method hMETIS \cite{karypis1999multilevel} to divide one netlist into multiple clusters/partitions. The partition method minimizes the overall cut between all clusters, which provides a global perspective. In this paper, we use the terms \emph{partition} and \emph{cluster} interchangeably. 
We denote the partition result as $M$. Each net $n_k$ is assigned a cluster ID $M[n_k]$, which denotes the cluster it belongs to. To capture more information, on the same netlist, we also build a hyper-graph $HG_c$ by using cells as nodes, such that we can also assign cells to clusters. In $HG_c$, each hyper-edge corresponds to a net. Similarly, the partition result on $HG_c$ is denoted as $P$. Each cell $c_k$ is assigned a cluster ID $P[c_k]$. Notice that $HG_c$ is only used to generate cluster ID for each cell.

Cluster IDs are not directly useful by themselves. What matters in this context is the difference in cluster IDs between cells and nets. Algorithm \ref{alg2} shows how the cluster information is incorporated into GNN models through novel edge features $F_0, F_1, F_2, f_3$. The most important intuition behind this is: for a high-quality placement solution, on average, the cells assigned to different clusters tend to be placed far away from each other.

\begin{algorithm}[!t]
\caption{Define Edge Features on Graph}
  \textbf{Input}: Cell cluster ID $P[c_k]$ for each cell $c_k$, net cluster ID $M[n_k]$ and the 
  \\\hspace*{15mm}neighbors $\mathcal{N}(n_k)$ of each net $n_k$. Directed graph $G$. \\
\label{alg2}
\vspace{-.2in}
\begin{algorithmic}[1]
\Function{measureDiff}{$c_{bk}$, $n_b$, $c_{ok}$, $n_o$} 
    \State $f_0 = 1 - (P[c_{bk}] == P[c_{ok}])$  \label{line:f0}
    \State $P_b = [P[c] $ for $ c \in n_b] $ // cluster IDs for $n_b$'s cells \label{line:f1s}
    \State $P_o = [P[c] $ for $ c \in n_o] $ // cluster IDs for $n_o$'s cells
    \State $P_{b\_not\_o} = P_b \backslash P_o$ $\qquad$ // IDs in $P_b$ but not in $P_o$
    \State $P_{o\_not\_b} = P_o \backslash P_b$ $\qquad$ // IDs in $P_o$ but not in $P_b$
    \State $f_1 = \frac{|P_{b\_not\_o}|}{|P_b|} + \frac{|P_{o\_not\_b}|}{|P_o|}$ // percent of different IDs \label{line:f1e}
    \State $f_2 = 1 - (M[n_b] == M[n_o])$
    \State return [$f_0$, $f_1$, $f_2$]
\EndFunction
\State
\For{each net $n_k$}
\For{each net $n_b \in \mathcal{N}(n_k)$}
\State $F_0$ = [\,], $F_1$ = [\,], $F_2$ = [\,]
\State Cell $c_{bk}$ is the edge cell on $n_b$ \textrightarrow $n_k$
\State Other neighbors $N_{other}^k = \mathcal{N}(n_k) \backslash \{n_b\}$
\For{each net $n_o \in N_{other}^k$}
\State Cell $c_{ok}$ is the edge cell on $n_o$ \textrightarrow $n_k$
\State $f_0$, $f_1$, $f_2$ = \textproc{measureDiff} ($c_{bk}$, $n_b$, $c_{ok}$, $n_o$)
\State $F_0$.add($f_0$) \textbf{;} \ \ $F_1$.add($f_1$) \textbf{;} \ \  $F_2$.add($f_2$)
\EndFor
\State  $f_3 = 1 - (M[n_b] == M[n_k])$
\State $E_{b \rightarrow k}$ = \{$\sum F_0$, $\mu (F_0)$, $\sum F_1$, $\mu (F_1)$, $\sum F_2$, $\mu (F_2)$, $f_3$\}\label{line:edge_comb}
\State Set $E_{b \rightarrow k}$ as the feature of edge $n_b \rightarrow n_k$ in $G$.
\EndFor
\EndFor
\end{algorithmic} 
  \textbf{Output}: Graph $G$ with edge features $E$.
\end{algorithm}

In Algorithm \ref{alg2}, we design the edge features to quantify the source node's contribution to the target node's length. The contribution here means the source net is ``pulling'' the edge cell far away from other cells in the target net. The edge features measure such ``pulling'' strength. When the edge cell is ``pulled'' away, the target net results in a longer length. In Algorithm \ref{alg2}, for edge $n_b$ \textrightarrow $n_k$, function \textproc{measureDiff} measures the difference in assigned clusters between node $n_b$ and every other neighboring node $n_o$, which indicates the distance between $c_{bk}$ and $c_{ok}$. If the distance between edge cell $c_{bk}$ and every other cell $c_{ok}$ in $n_{k}$ is large, it means $c_{bk}$ is placed far away from other cells in net $n_k$. In this case, edges features $F_0, F_1, F_2, f_3$ are large. That is why edge features imply how strong the edge cell is ``pulled'' away from the target node.

\begin{figure}[!tb]
  \centering
    \includegraphics[width=0.8\columnwidth]{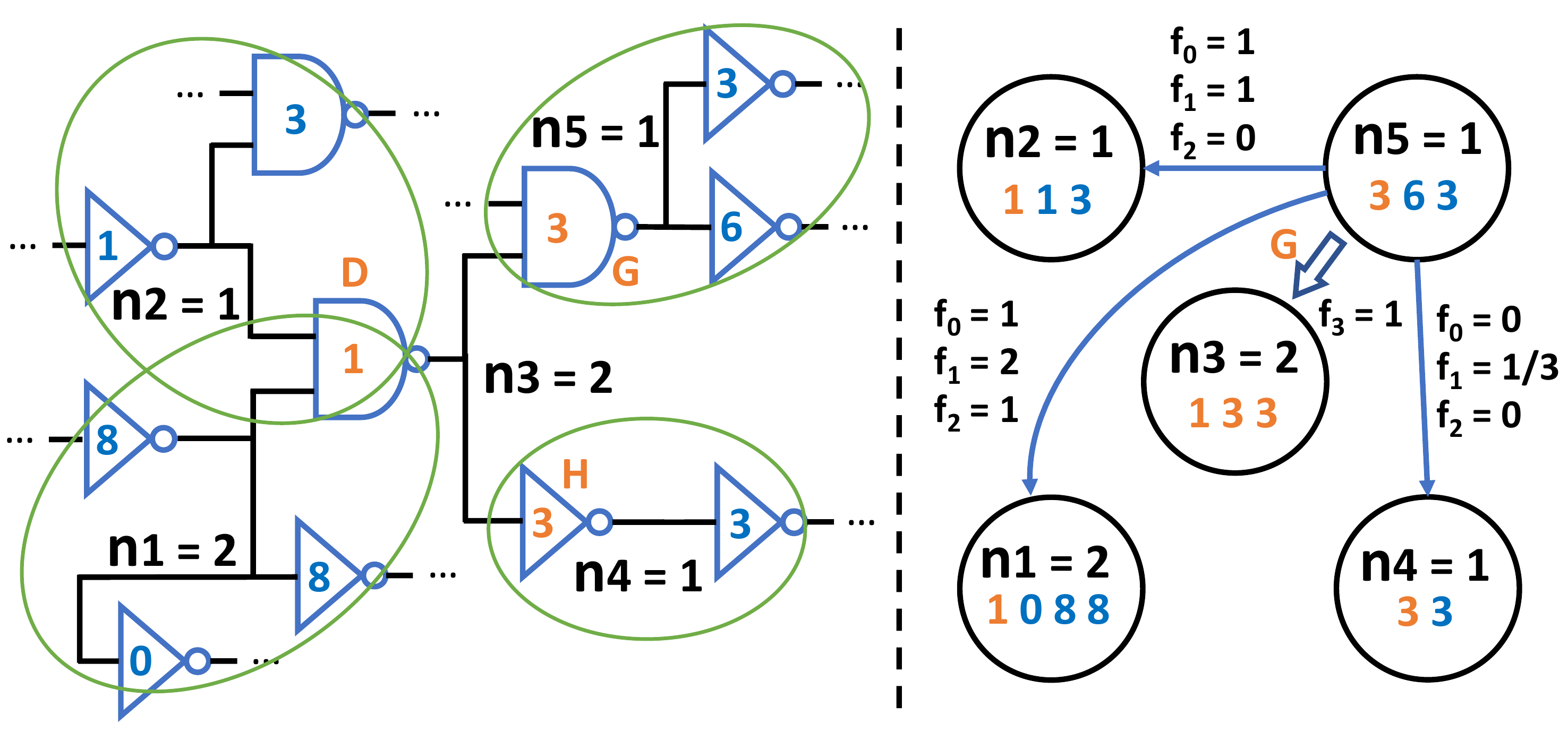}
    	\vspace{-3mm}
  \caption{Define edge features by partition results.}
    \vspace{-3mm}
  \label{method_cluster}
\end{figure}

Figure \ref{method_cluster} shows an example of Algorithm \ref{alg2} using the netlist same as Figure \ref{method_circuit}. The number on each cell or net is the cluster ID assigned to it after partition. Figure \ref{method_cluster} measures the edge features of edge $n_5$ \textrightarrow $n_3$, representing how strongly edge cell $c_G$ is pulled by $n_5$ from both cells \{$c_D$, $c_H$\} in $n_3$. To calculate this, we measure the distance between $c_G$ and $c_H$ by \textproc{measureDiff}($c_G$, $n_5$, $c_H$, $n_4$) in Algorithm \ref{alg2}; and the distance between $c_G$ and $c_D$ by \textproc{measureDiff}($c_G$, $n_5$, $c_D$, $n_1$) and \textproc{measureDiff}($c_G$, $n_5$, $c_D$, $n_2$).

Take \textproc{measureDiff}($c_G$, $n_5$, $c_H$, $n_4$) as an example to show how it measures distance between $c_G$ and $c_H$. As shown in the line \ref{line:f0} of Algorithm~\ref{alg2}, feature $f_0$ measures the difference in $c_G$ and $c_H$' cluster IDs, $f_0 = 1 - (P[c_G] == P[c_H]) = 1 - (3 == 3) = 0$. Feature $f_1$ measures the difference in all cells between $n_5$ and $n_4$. As shown from line \ref{line:f1s} to \ref{line:f1e}, $P_5 = [3, 6, 3]$ and $P_4 = [3, 3]$. Then $P_{5\_not\_4} = [6]$ and $P_{4\_not\_5} = []$. They are normalized by the number of cells $|P_5| = 3$ and $|P_4| = 2$, in order to avoid bias toward nets with many cells. Thus, $f_1 = \frac{1}{3} + \frac{0}{2} = \frac{1}{3}$. Feature $f_2$ measures the difference between $n_5$ and $n_4$, $f_2 = 1- (M[n_5] == M[n_4]) = 1 - (1==1) = 0$. As this example shows, we only measure whether cells / nets have the same cluster IDs, and the order of IDs does not matter.

After measuring the difference in cluster ID between $c_{G}$ and all other cells in $n_3$, for the edge $n_5$ \textrightarrow $n_3$, $F_0 = [1, 1, 0]$; $F_1 = [2, 1, \frac{1}{3}]$; $F_2 = [1, 0, 0]$. $f_3$ measures the difference between $n_5$ and $n_3$, $f_3=1$. This example shows how we incorporate global information from partition into edge features. Actually, we generate multiple different partitioning results $M$, $P$ by requesting different number of clusters. That results in multiple different \{$F_0$, $F_1$, $F_2$, $f_3$\}. All these different edge features are processed in line \ref{line:edge_comb} and concatenated together as the final edge features $E_{b \rightarrow k}$.

\subsection{GNN and Net$^\mathbf{2}$ Models}

\setlength{\belowdisplayskip}{3pt} \setlength{\belowdisplayshortskip}{3pt}
\setlength{\abovedisplayskip}{-5pt} \setlength{\abovedisplayshortskip}{-5pt}

This section introduces how GNN models are applied on the graph G we build. GNN models are comprised of multiple sequential convolution layers. Each layer generates a new embedding for every node based on the previous embeddings. For node $n_k$ with node features $O_k$, denote its embedding at the $t^{\text{th}}$ layer as $h_k^{(t)}$. Its initial embedding is the node features $h_k^{(0)} = O_k$. Sometimes the operation includes both neighbours and the node itself, we use $n_{\beta}$ to denote it: $n_{\beta} \in \mathcal{N}(n_k) \cup \{n_k\}$. In each layer $t$, GNNs calculate the updated embedding $h_k^{(t)}$ based on the previous embedding of the node itself $h_k^{(t-1)}$ and its neighbors $h_b^{(t-1)} | n_b\in \mathcal{N}(n_k)$.

We show one layer of GCN, GSage, and GAT below. Notice that there exist other expressions of these models. The two-dimensional learnable weight at layer $t$ is $W^{(t)}$. In GAT, there is an extra one-dimensional weight $\theta^{(t)}$. The operation $[\: ||\: ]$  concatenates two vectors into one longer vector. Functions $\sigma$ and $g$ are sigmoid and Leaky ReLu activation function, respectively.

On GCN (with self-loops), $\mathcal{F}_{GCN}^{(t)}$ \cite{kipf2016semi} is:

\begin{gather*}
h_k^{(t)} = \sigma (\sum_{n_{\beta}\in {\mathcal{N}(n_k) \cup \{n_k\}}} a_{k\beta} W^{(t)} h_\beta^{(t-1)})\\
\text{where   } a_{k\beta} = \frac{1}{\sqrt{deg(k) + 1} \sqrt{deg(\beta) + 1}} \in \mathbb{R}
\end{gather*}

On GSage, $\mathcal{F}_{GSage}^{(t)}$ \cite{hamilton2017inductive} is:
\vspace{6pt}
\begin{gather*}
  h_k^{(t)} = \sigma ( W^{(t)} [h_k^{(t-1)} || \frac{1}{deg(k)} \sum_{n_b\in {\mathcal{N}(n_k)}}  h_b^{(t-1)}]) 
\end{gather*}

On GAT, $\mathcal{F}_{GAT}^{(t)}$ \cite{velivckovic2017graph} is:
\vspace{8pt}
\begin{gather*}
  h_k^{(t)} = \sigma (\sum_{n_\beta\in {\mathcal{N}(n_k) \cup \{n_k\}}} a_{k\beta} W^{(t)} h_\beta^{(t-1)}) \qquad \qquad \\
\end{gather*}
  \vspace{-.5in}
\begin{align*}
  \text{where   } a_{k\beta} &= softmax_\beta (r_{k\beta})  \text{ \,\; over $n_k$ and its neighbors,} \\
  r_{k\beta} &= g (\theta^{(t)\intercal} [ W^{(t)} h_\beta^{(t-1)} || W^{(t)} h_k^{(t-1)} ]) \in \mathbb{R}
\end{align*}

Here we briefly discuss the difference between these methods. GCN scales the contribution of neighbors by a pre-determined coefficient $a_{k\beta}$, depending on the node degree. GSage does not scale neighbors by any factor. In contrast, GAT uses learnable weights $W$, $\theta$ to firstly decide node $n_\beta$'s contribution $r_{k\beta}$, then normalize the coefficient $r_{k\beta}$ across $n_k$ and its neighbors through a softmax operation. Such a learnable $a_{k\beta}$ 
leads to a more flexible model. For all these GNN methods, the last layer's output embedding $h_k^{(t)}$ is connected to a multi-layer ANN.

The node convolution layer of the Net$^\mathbf{2}$ is based on GAT, considering its higher flexibility in deciding neighbors' contribution $a_{k\beta}$. Thus node convolution layer is $\mathcal{F}_{GAT}^{(t)}$. In the final embedding, we concatenate the outputs from all layers, instead of only using the output of the final layer like most GNN works. This is a customization, by which the embedding includes contents from different depths. The shallower ones from the first few layers include more local information, while the deeper ones from the last few layers contain more global information. Such an embedding provides more information for the ANN model at the end and may lead to better convergence. The idea of combining shallow and deep layers has inspired many classical deep learning methods in Euclidian space \cite{he2016deep, ronneberger2015u}, but it is not widely applied in GNNs for node embeddings. After three layers of node convolution, the final embedding for each node is $[h_k^{(1)} || h_k^{(2)} || h_k^{(3)}]$. Without partitioning, this is the embedding for our fast solution Net$^\text{\textbf{2f}}$.

In order to utilize edge features, here we define our own edge convolution layers $\mathcal{E}$ as customization. For each directed edge $n_b$ \textrightarrow $n_k$, we concatenate both target and source nodes' features $[O_k || O_b]$ together with its edge features $E_{b\rightarrow k}$ as the input of edge convolution. Combining node features when processing edge features enables $\mathcal{E}$ to distinguish different edges with similar edge features. The output embedding is:

\vspace{-.2in}
\begin{gather*}
e_{k\_sum} = \sum_{n_b \in \mathcal{N}(n_k)} W_2 W_1 [O_k || E_{b\rightarrow k} || O_b] \\
e_{k\_mean} = \frac{1}{deg(k) }e_{k\_sum}
\end{gather*}

The two two-dimensional learnable weights $W_1$ and $W_2$ can be viewed as applying a two-layer ANN to the concatenated input. We choose two-layer ANN rather than one-layer here because the input vector $[O_k || E_{b\rightarrow k} || O_b]$ is long and contains heterogeneous information from both edge and node. We prefer to learn from them with a slightly more complex function. After the operation, both $e_{k\_sum}$ and $e_{k\_mean}$ are on nodes. Then, we add an extra node convolution using the output from edge convolution as input. This structure learns from neighbors' edge embeddings $e_{b\_sum}, e_{b\_mean}$.

\vspace{-.1in}
\begin{equation*}
h_k^{(e)} = \mathcal{F}_{GAT}^{(e)} ([e_{k\_sum} || e_{k\_mean}], [e_{b\_sum} || e_{b\_mean}])
\end{equation*}

Inspired by the same idea in Net$^\mathbf{2f}$, we combine the contents from all layers for our accurate solution Net$^\mathbf{2a}$. Its final embedding is $[h_k^{(1)} || h_k^{(2)} || h_k^{(3)} || e_{k\_sum} || e_{k\_mean} || h_k^{(e)}]$. For both Net$^\mathbf{2f}$ and Net$^\mathbf{2a}$, their final embeddings are then connected to an ANN.

\subsection{Timing Prediction Method}

\label{sec:timing_method}

This section introduces our timing prediction method in detail. The timing estimator is constructed and applied to directly predict the delay of each individual timing arc. Then based on the inference result, we further obtain arrival time, required arrival time, and slack of each circuit node by traversing the graph with predicted delay values. Similar to the Net$^\mathbf{2}$ model, we provide both fast and accuracy-oriented versions for timing prediction, named Time$^\mathbf{f}$ and Time$^\mathbf{a}$, respectively.

\begin{figure}[!b]
  \centering
    \includegraphics[width=0.74\columnwidth]{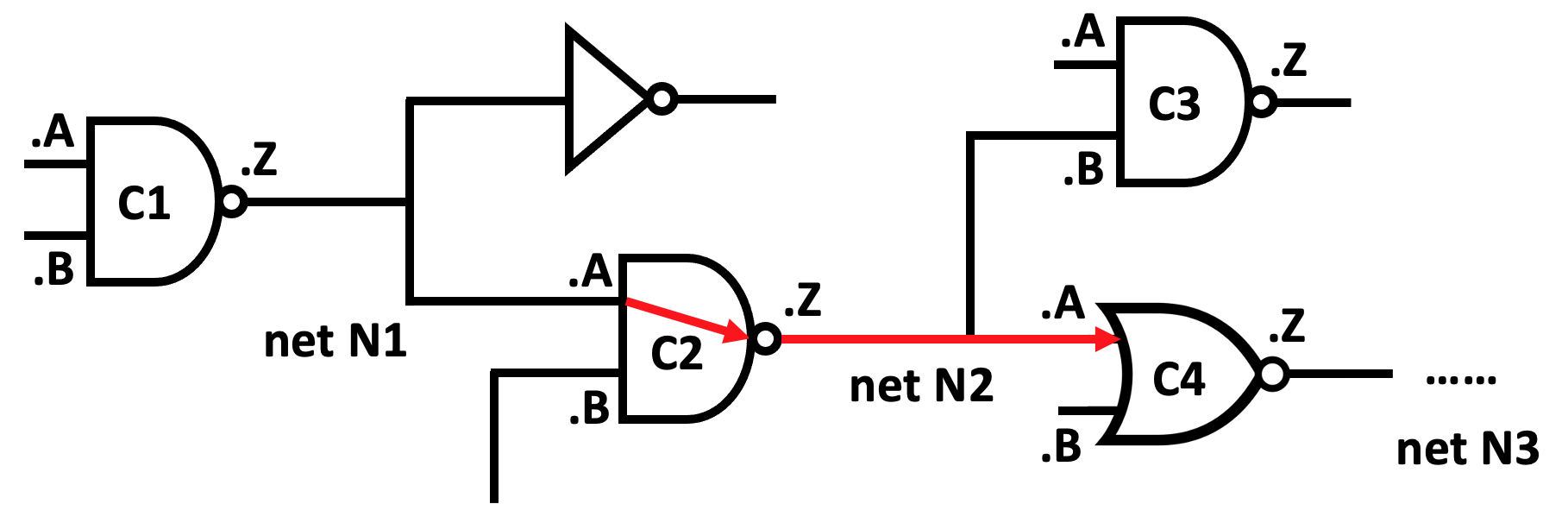}
    \caption{An example to illustrate timing prediction algorithm.}
  \label{example_timing}
\end{figure}

\begin{table}[!t]
  \centering
  \caption{Pre-Placement Features for Timing Prediction}
  \label{tbl:feature_timing}
      \resizebox{0.75\linewidth}{!}{ \renewcommand{\arraystretch}{1.2}
  \begin{tabular}{| l |}
 	\hline
 	  \textbf{For each cell arc (C2.A$\Rightarrow$C2.Z in cell C2, for example)}  \\ 
 	\hline 
 	\hline
 	    Pre-placement delay of the arc itself: C2.A$\Rightarrow$C2.Z \\
 	\hline
	    Source pin information: capacitance, slew, slack at C2.A\\
	\hline
	    All net-size-relevant features of the following net: net N2  \\
	\hline
	    Predicted size of previous net: net N1   \\
	\hline
    \multicolumn{1}{c}{}  	\\
	\hline
 	   \textbf{For each net arc (C2.Z$\Rightarrow$C4.A in net N2, for example)}  \\ 
    \hline
    \hline
	    Source pin information: max capacitance, slew, slack at C2.Z \\
	\hline 
	    Sink pin information: capacitance at C4.A  \\
    \hline
        All net-size-relevant features of the net: net N2  \\ 
    \hline
        Predicted size of the following net: net N3  \\
    \hline

  \end{tabular}
  \vspace{-3mm}
  }
 \end{table}

We take the simplified circuit in Figure~\ref{example_timing} to demonstrate our timing estimator, which predicts the delay of every timing arc. The timing arc, as the basic component of a timing path, can be categorized into cell arc and net arc. Each cell arc is between an input pin and output pin of a cell, and each net arc is between the driver pin and load pin of a net. Considering their different properties, in our timing estimator, two separate timing prediction models are constructed to handle these two types of timing arcs. For each timing arc, we denote the pin from which it originates as the source pin, and the pin at which it ends as the sink pin. For example, the timing arc C2.A$\Rightarrow$C2.Z means a cell arc from source pin C2.A to the sink pin C2.Z.


For each cell, the cell-arc timing model predicts the post-placement delay of all its cell arcs. Take cell C2 in Figure~\ref{example_timing} as an example, the model predicts delays of both C2.A$\Rightarrow$C2.Z and C2.B$\Rightarrow$C2.Z. This is essentially different from the timing model in a representative post-placement timing estimator~\cite{barboza2019machine}, which assumes the delays of all cell arcs in the same cell are the same. This approximation in~\cite{barboza2019machine} may lead to inaccuracies, considering the input slews and diffusion capacitances seen by input pins of the same cell can be different. Our observation in experiments shows the cell delays at C2.A$\Rightarrow$C2.Z and C2.B$\Rightarrow$C2.Z can differ a lot and thus distinguishing all cell arcs helps to achieve higher accuracy. Similarly, for each net, the net-arc timing model predicts the post-placement delay of each net arc. In net N2, for example, there are two net arcs C2.Z$\Rightarrow$C4.A and C2.Z$\Rightarrow$C3.B. In addition, our timing estimator takes the worst delay between rising and falling as the ground truth, without constructing separate models for rising and falling edges. This avoids doubling the required timing models and simplifies the timing analysis through traversals.

Table \ref{tbl:feature_timing} summarizes selected features for cell arcs and net arcs, with examples on C2.A$\Rightarrow$C2.Z and C2.Z$\Rightarrow$C4.A in Figure \ref{example_timing}, respectively. All these features in Table \ref{tbl:feature_timing} are from two main sources, as summarized below.

\begin{itemize}
\item All relevant slew, delay, and slack information from the pre-placement timing report.  
\item All relevant net and cell information that can be derived from the netlist. It includes the global information captured by performing clustering on the netlist. 
\end{itemize}

Both cell-arc and net-arc models are constructed based on the existing timing report and the netlist information used in net size prediction. We can also view the prediction procedure as improving the inaccurate pre-placement timing report by incorporating net size information into our ML-based timing model. 
Specifically, a detailed explanation of all selected features is elaborated as follows.

\begin{itemize}
\item \textbf{Pre-placement delay:} Although the pre-placement timing report fails to evaluate wire load accurately for delay measurement, it still shows a generally acceptable correlation with ground truth. Thus the pre-placement delay of the predicted cell arc itself is adopted as an important input feature. Notice that the pre-placement delay of net arcs is set to zero in some commercial layout tools~\cite{Innovus}, thus the delay of the net arc itself is not used as a feature of C2.Z$\Rightarrow$C4.A. 
\item \textbf{Capacitance at the input pin of cells:} For the same type of cell, the capacitance at the cell input pin is usually proportional to the cell's driving strength. For cell arcs, a larger capacitance at each arc's source pin indicates the larger driving strength and thus smaller delay. For net arcs, a larger capacitance at each arc's sink pin indicates a higher load seen by the wire. 
\item \textbf{Pre-placement slew at the source pin:} The slew, or named transition time, also significantly affects the delay. Thus the pre-placement slew at the source pin of both types of arcs is adopted as features. 
\item \textbf{Detailed net size information:} The net size of net N2 is a determining factor of the delay. For both cell and net arcs in this example, it is directly proportional to the wire load seen by the C2.Z pin. A larger wire load at C2.Z takes longer to charge/discharge, leading to a larger cell-arc delay. For the net arc, the net size is also proportional to the wire length from C2.Z to C4.A. For the fast timing estimator Time$^\mathbf{f}$, both node features and Net$^\mathbf{2f}$-predicted size of the net N2 are included as features. For the accurate version Time$^\mathbf{a}$, besides using predictions from Net$^\mathbf{2a}$, the clustering-related information of this node is also included as features. 
\item \textbf{Brief net size information:} For the cell arc like C2.A$\Rightarrow$C2.Z, its previous net N1 affects the input slew at the source pin C2.A. For the net arc like C2.Z$\Rightarrow$C4.A, its following net N3 affects where cell C4 is placed, and thus affects the distance between C2.Z and C4.A. Since their impact on the arcs is less than the net N2, we only adopt the brief net size information, which is corresponding net size estimators' predictions on these nets as features. 
\end{itemize}

Based on extracted features of the two different types of timing arcs, we develop one cell-arc model and one net-arc model, both based on the random forest (RF) algorithm~\cite{breiman2001random}. Tree-based ML algorithms are good at handling largely distinct types of features, which include slew, delay, capacitance, cell/net number, and clustering information in this case. Instead of directly predicting the ground-truth post-placement delay of each arc, our model is actually trained to predict the \emph{incremental delay}, which is the difference between pre-placement and the ground-truth post-placement timing. Then the final predicted delay is the summation of both pre-placement delay and the prediction of the incremental delay. This strategy, not adopted in previous ML-in-EDA works~\cite{barboza2019machine}, helps the model to directly capture wire-load-induced delay based on the pre-placement timing report.

\section{Evaluation}

\subsection{Experimental Setup}

\begin{table}[!h]
  \centering
  \caption{ Number of Nets in Designs}
  \label{tbl:designs}
  \resizebox{0.9\linewidth}{!}{
  \begin{tabular}{| l | c c  | c c | c  c |}
 	\hline
 	\multirow{2}{*}{Benchmark}  &  \multirow{2}{*}{Design} &  \multirow{2}{*}{\# Net} &  \
 	 \multirow{2}{*}{Design} &  \multirow{2}{*}{\# Net} &  \multirow{2}{*}{Design} &  \multirow{2}{*}{\# Net}  \\
 	  &  &  &  &   & & \\
	\hline
	\multirow{2}{*}{ISCAS'89}      &   s13207 &  4 K  &  s35932 & 31 K  &   s38417 &  26 K  \\
	                               &   s38584 & 18 K  &  s5378  &  4 K  &  s9234\_1 & 4 K \\
 	\hline
	\multirow{3}{*}{ITC'99}  &  b14  & 22$\,$K   &  b15  & 12$\,$K &  b17  & 40$\,$K   \\
                             &  b18  & 115$\,$K  &  b19  & 225$\,$K  &  b20  & 39$\,$K  \\
                             &  b21  &  39$\,$K  &  b22  & 58$\,$K   &      &       \\
	\hline
	Faraday                  & DMA   &  42$\,$K & DSP & 73$\,$K & RISC  &  98$\,$K   \\ 
	\hline
	\multirow{4}{*}{OpenCores}  &  systemcaes & 13$\,$K  & wb\_dma    & 6$\,$K  &  systemcdes &  4$\,$K    \\ 
	                            & des         & 6$\,$K   &  ethernet   & 71$\,$K  & mem\_ctrl   & 10$\,$K  \\ 
	                            &  pci        & 25$\,$K  & spi         & 4$\,$K   &  tv80       & 13$\,$K  \\ 
	                            & usb\_funct  & 24$\,$K  &  vga\_lcd   & 106$\,$K & wb\_conmax  & 87$\,$K   \\ 
	\hline
	\multirow{2}{*}{ANUBIS}     &  DLX        & 19$\,$K & ALPHA      & 41$\,$K &  FPU        & 36$\,$K   \\ 
	                            & mor1k      & 178$\,$K   & OR1200  & 847$\,$K  &    &  \\ 
	\hline
	Gaisler    &  leon2   & 835$\,$K & leon3mp & 640$\,$K  &  netcard & 551$\,$K         \\
	\hline
  \end{tabular}
  }
\end{table}

To thoroughly validate our algorithms, we constructed a comprehensive dataset by including 37 different designs with largely varying sizes. All 37 designs are synthesized with Synopsys Design Compilier\textregistered \ ~\cite{design-compilier} in 45nm NanGate Library~\cite{NanGate}, and then placed by Cadence Innovus\texttrademark \ v17.0~\cite{Innovus}. When testing ML models on each design, we train the model only on the \emph{other} 36 designs in the dataset to prevent information leakage. Thus, all accuracy numbers measure the performance on new designs completely \emph{unseen} to the existing model. These accuracies reflect how well the model generalizes to each of the 37 designs in our dataset. The detail of each design is shown in Table~\ref{tbl:designs}. They are collected from various benchmarks, including ISCAS'89~\cite{brglez1989combinational}, ITC'99~\cite{corno2000rt}, ANUBIS~\cite{possignolo2017anubis}, and other selected designs from Faraday, OpenCores and Gaisler in the IWLS'05~\cite{albrecht2005iwls}. To ensure all designs in the experiment are representative, we discard tiny designs with less than 3K nets. As shown in Table~\ref{tbl:designs}, the size of these designs ranges from 4K to 800K nets. We set the clock period of all designs to be 1.5ns and thus most designs result in a negative worst slack. This mimics a common design scenario where designers target high performance and rely on the timing estimator to address negative slacks on critical paths.

All GNNs are built with Pytorch 1.5~\cite{paszke2019pytorch} and Pytorch-geometric~\cite{Fey/Lenssen/2019}. The partition on graphs is performed by hMETIS~\cite{karypis1999multilevel} executable files. The RF models are developed based on the random forest regressor in scikit-learn~\cite{scikit-learn}. The experiment is performed on a machine with a Xeon E5 processor and an Nvidia GTX 1080 graphics card.

Hyper-parameter values are decided during parameter tuning. This is accomplished by testing combinations of hyper-parameters on a much smaller validation dataset constructed for parameter tuning. This smaller validation dataset may comprise netlists only from one benchmark like ITC’99, in order to make the testing faster and allow us to test how well the model generalizes on other designs in the whole dataset. Here we introduce the best hyper-parameters after parameter tuning. They target to achieve a good trade-off between bias and variance, making the model sufficiently flexible while not too complex.
For all GNN methods, we use three layers of GNN with two layers ANN. The attention head number of GAT is two. The size of each node convolution output is 64. The size of edge convolution output is twice of the input size $[O_k||E_{b \rightarrow k}||O_b]$. The size of the first-layer ANN is the same as its input embedding, and the size of the second-layer ANN is 64. A batch normalization layer is applied after each GNN layer for better convergence. Because of the difference in graph size, each batch includes only one graph, and the training data is shuffled during training. We use stochastic gradient descent (SGD) with a learning rate 0.002 and momentum factor 0.9 for optimization. GNN models converge in 250 epoches. For all RF models in timing prediction, we set the number of tree-based estimators to be 80 and the maximum depth of each estimator to be 12. Other parameters are left the same as default settings.

When partitioning each netlist, we generate seven different cell-based partitions $P$ by requesting the number of output clusters to be the number of cells divided by 100, 200, 300, 500, 1000, 2000, and 3000. Because different partitions are generated in parallel, the overall runtime depends on the slowest one. Similarly, we generate three net-based partitions $M$ by requesting the cluster number to be the number of nets divided by 500, 1000, and 2000. These cluster numbers are achieved by tuning during experiments, which provides good enough coverage over different cluster sizes.
 
Representative previous methods MC~\cite{hu2003wire}, ISPL~\cite{kahng2005intrinsic}, and Poly~\cite{fathi2009pre} are implemented for comparisons. As for traditional ML models, besides the polynomial model proposed in previous work~\cite{fathi2009pre}, we implement a three-layer artificial neural network (ANN) model using node features $O$. Here we summarize the receptive field of all methods. MC is limited to one-hop neighbors, while Poly and ANN can reach two-hop neighbors. The receptive field of ISPL varies among different nodes. According to \cite{kahng2005intrinsic}, ISPL for most nets is within several hops. In comparison, all GNNs and Net$^\mathbf{2f}$ can access five-hop neighbors. Net$^\mathbf{2a}$ measures the impact from the whole netlist.


We evaluate our methods with various metrics, including mean absolute error (MAE), correlation coefficient R, and coefficient of determination $R^2$. Given a list of ground-truth labels $\{y_i\}$ and corresponding predictions $\{p_i\}$ with length $N$, these metrics are defined below. The $\bar{y}$ and $\bar{p}$ represent the average of labels and predictions, respectively. 

\begin{align*}
& R = \frac{\sum_{i=1}^N (y_i - \bar{y}) (p_i - \bar{p}) }{\sqrt{\sum_{i=1}^N (y_i - \bar{y})} \sqrt{\sum_{i=1}^N} (p_i - \bar{p})} \\[3mm]  
&\text{MAE} = \frac{\sum_{i=1}^N |y_i - p_i|}{ N }  \quad \quad \ \  R^2 = 1 - \frac{ \sum_{i=1}^N (y_i - p_i) ^2}  { \sum_{i=1}^N ( y_i - \bar{y} ) ^2} 
\vspace{2mm}
\end{align*}

In addition, for classification tasks, we evaluate the accuracy with Receiver Operating Characteristic (ROC) curve, measured based on 
true positive rate (TPR) and false positive rate (FPR). 
For classification tasks, a threshold is applied to turn the raw prediction into binary. A higher threshold will lead to higher TPR, but also higher FPR; otherwise the vice. The ROC curve thus indicates the trade-off between TPR and FPR when varying the threshold, and a larger area under curve (AUC) indicates higher accuracy. The AUC ranges from 0 to 1, with AUC = 0.5 indicating the accuracy of random guessing.

\subsection{Net Length Prediction Result}

\begin{figure*}[!h]
  \centering
    \subfigure[20 bins generated according to labels.] 
    {\includegraphics[width=\textwidth]{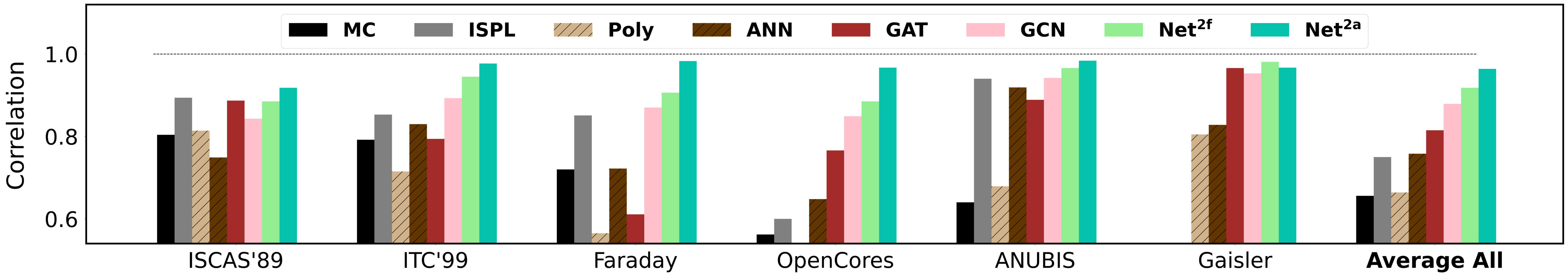}\label{bins_ra} } \\
    \subfigure[20 bins generated according to predictions.]
    {\includegraphics[width=\textwidth]{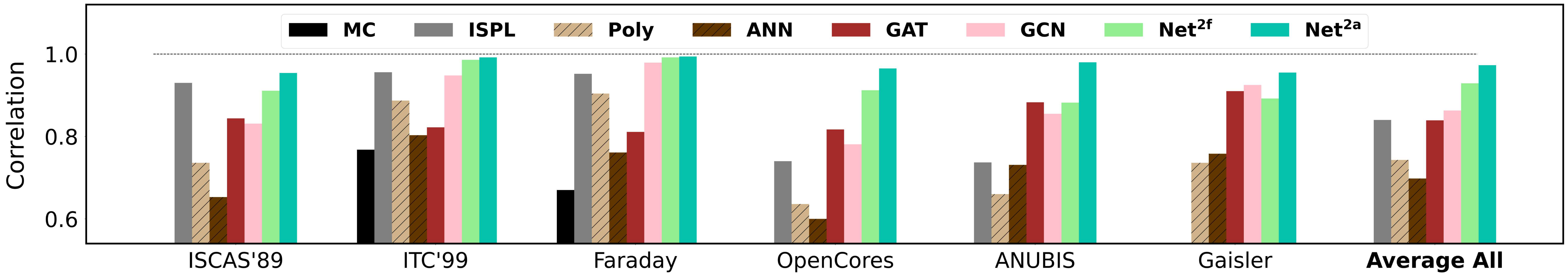}\label{bins_rb} }   
  \caption{The correlation coefficient $R$ between net length prediction and label. Averaged over designs in each benchmark.}
  \label{bins_r}
\end{figure*}

We first measure the correlation between prediction and ground truth on all nets in each netlist in Figure~\ref{bins_r}, with a classical criterion used in many net length estimation works~\cite{fathi2009pre,kahng2005intrinsic,liu2004pre}. For each netlist, we firstly calculate a range of net length [$L_{0\%}$, $L_{95\%}$]. It means from the shortest net length to the 95 percentile largest net length. The top 5\% longest nets are excluded to prevent an extraordinarily large range. Then the calculated range is partitioned into 20 equal bins, and the average of both predictions and labels in each bin is calculated. After that, the correlation coefficient $R$ between these 20 averaged predictions and labels is measured and reported. To make fair comparisons, we calculate the range [$L_{0\%}$, $L_{95\%}$] and define such 20 bins using both labels and predictions, as shown in Figure~\ref{bins_ra} and \ref{bins_rb}, respectively. Figure \ref{bins_r} reports the correlation coefficient averaged over designs from the same benchmark. In addition, the `Average All' bars in Figure \ref{bins_r} show the averaged $R$ over netlists from all 37 designs.

Figure~\ref{bins_ra} and \ref{bins_rb} show highly similar trend of averaged accuracy, indicating that Net$^\mathbf{2a}$ > Net$^\mathbf{2f}$ > GCN/GAT > ANN. The correlations of GNN methods are significantly higher than previous methods with a limited receptive field including MC, ANN, ISPL, and Poly. Then Net$^\mathbf{2f}$ outperforms GAT and GCN with its residual connection. By capturing the global information, Net$^\mathbf{2a}$ performs the best on all benchmarks, with an average accuracy $R = 0.964$.

\begin{table*}[!p]
  \centering
  \caption{Accuracy comparison. (a) Net length prediction evaluated with long nets identification. (b) Timing prediction evaluated with arc delay.}
  \label{tbl:10LongestNet}
      \renewcommand{\arraystretch}{1.1}
  \setlength{\tabcolsep}{1.2mm}
  \hspace*{-.28in}\resizebox{1.08\linewidth}{!}{
  \begin{tabular}{|c | c c c c | c c p{0.8cm} p{0.9cm} || c c | c c | c c | c c | }
 	\hline
 	  \multirow{4}{*}{Design} &  \multicolumn{8}{c||}{\multirow{2}{*}{(a) Long Nets Identification Accuracy (AUC\%)}} &  \multicolumn{8}{c|}{\multirow{2}{*}{(b) Arc Delay Prediction Accuracy}} \\
 	  &   \multicolumn{8}{c||}{ }  &  \multicolumn{8}{c|}{ }  \\
 	  \cline{2-17}
 	   &  \multirow{3}{*}{MC} & \multirow{3}{*}{ISPL}  & \multirow{3}{*}{Poly}  & \multirow{3}{*}{ANN} &  \multirow{3}{*}{GAT}  & \multirow{3}{*}{GCN}  & \multirow{3}{*}{Net$^\mathbf{2f}$} &  \multirow{3}{*}{Net$^\mathbf{2a}$}  &  \multicolumn{2}{c|}{report-} &  \multicolumn{2}{c|}{\multirow{2}{*}{~onlyTime~}} &  \multicolumn{2}{c|}{\multirow{2}{*}{~Time$^\mathbf{f}$~}} &  \multicolumn{2}{c|}{\multirow{2}{*}{~Time$^\mathbf{a}$~}}   \\
 	   &     &   &  &    &  &  &  & & \multicolumn{2}{c|}{timing} &  &  &  &  &  &  \\
 	   &     &   &  &    &  &  &  &  & R  & R$^2$ & R & R$^2$ & R & R$^2$   &  R & R$^2$\\
	\hline
	    s13207     & 51.6   &  70.7   &  51.8  & 77.8 & 81.0  &   80.5 &    82.1  &  89.0  &  1.00   &  1.00   & 0.99  & 0.98  &  1.00   &  0.99  &  1.00  &  1.00      \\
	    s35932     & 71.6   &  79.5   &  72.5  & 73.8 & 77.7  &   76.8 &    81.8  &  88.8  &  0.94   &  0.85   & 0.95  & 0.91  &  0.95   &  0.91  &  0.97  &  0.94    \\ 
	    s38417     & 74.0   &  81.0   &  73.0  & 71.0 & 72.0  &   74.9 &    78.4  &  85.5  &  0.99   &  0.97   & 0.98  & 0.94  &  0.99   &  0.97  &  0.99  &  0.98    \\
	    s38584     & 68.8   &  86.2   &  71.0  & 81.2 & 80.1  &   82.2 &    85.8  &  91.4  &  0.98   &  0.95   & 0.99  & 0.96  &  0.99   &  0.97  &  0.99  &  0.98    \\
	    s5378      & 64.5   &  79.8   &  67.4  & 64.8 & 74.7  &   80.7 &    79.0  &  87.8  &  1.00   &  0.99   & 0.99  & 0.96  &  1.00   &  0.99  &  0.99  &  0.98    \\
	    s9234\_1   & 67.5   &  68.5   &  64.0  & 69.9 & 77.5  &   81.0 &    80.2  &  91.6  &  1.00   &  0.99   & 0.99  & 0.96  &  1.00   &  0.99  &  0.99  &  0.98    \\
 	\hline    
	     b14       & 70.5   &  77.9   &  71.3  & 74.8 & 74.7  &   76.3 &    81.2  &  90.8  &  0.92   &  0.74   & 0.92  & 0.85  &  0.94   &  0.87  &  0.95  &  0.90    \\
	     b15       & 67.9   &  65.5   &  67.3  & 71.7 & 73.0  &   72.3 &    78.3  &  88.5  &  0.98   &  0.92   & 0.97  & 0.92  &  0.97   &  0.94  &  0.98  &  0.96    \\
         b17       & 72.4   &  69.0   &  66.3  & 74.0 & 70.2  &   75.5 &    78.7  &  88.8  &  0.96   &  0.87   & 0.95  & 0.89  &  0.97   &  0.93  &  0.97  &  0.95    \\
b18*       & 73.2   &  69.1   &  72.4  & 70.6 & 71.9  &   75.4 &    78.9  &  89.6  &  0.96   &  0.88   & 0.96  & 0.91  &  0.96   &  0.90  &  0.98  &  0.95    \\
b19*       & 72.0   &  69.2   &  69.4  & 65.3 & 71.8  &   76.0 &    76.5  &  88.6  &  0.96   &  0.86   & 0.92  & 0.83  &  0.92   &  0.73  &  0.97  &  0.93     \\
         b20       & 74.2   &  77.5   &  71.6  & 75.2 & 72.4  &   75.9 &    83.8  &  91.8  &  0.92   &  0.78   & 0.93  & 0.86  &  0.95   &  0.89  &  0.96  &  0.93     \\
         b21       & 74.8   &  76.6   &  73.3  & 77.4 & 75.0  &   78.1 &    84.8  &  92.2  &  0.92   &  0.77   & 0.93  & 0.86  &  0.95   &  0.90  &  0.96  &  0.92     \\
         b22       & 73.7   &  76.5   &  72.4  & 75.4 & 75.4  &   77.6 &    85.2  &  90.2  &  0.91   &  0.74   & 0.92  & 0.85  &  0.94   &  0.88  &  0.95  &  0.90     \\
	\hline 
	     DMA       & 59.3   &  65.4   &  62.1  & 68.1 & 70.8  &   75.8 &    78.1  &  89.3  &  0.91   &  0.74   & 0.92  & 0.84  &  0.92   &  0.84  &  0.94  &  0.89      \\ 
	     DSP       & 67.2   &  71.3   &  65.9  & 67.6 & 67.3  &   72.2 &    74.0  &  88.8  &  0.90   &  0.72   & 0.90  & 0.81  &  0.91   &  0.83  &  0.94  &  0.88      \\ 
	     RISC      & 68.4   &  67.4   &  65.8  & 70.0 & 67.5  &   68.0 &    72.5  &  88.3  &  0.94   &  0.84   & 0.95  & 0.89  &  0.95   &  0.89  &  0.97  &  0.93      \\ 
	\hline     
	    systemcaes & 75.5   &  63.5   &  78.9  & 71.8 & 68.3  &   72.3 &    80.7  &  89.5  &  0.90   &  0.69   & 0.90  & 0.81  &  0.94   &  0.87  &  0.95  &  0.91    \\ 
	    wb\_dma    & 53.4   &  51.6   &  46.0  & 72.3 & 84.2  &   75.9 &    88.9  &  93.4  &  0.97   &  0.90   & 0.97  & 0.94  &  0.98   &  0.95  &  0.98  &  0.96    \\ 
	    systemcdes & 39.3   &  66.5   &  56.0  & 79.8 & 86.7  &   86.9 &    89.6  &  95.8  &  0.98   &  0.92   & 0.97  & 0.90  &  0.98   &  0.93  &  0.99  &  0.98     \\ 
	    des        & 55.9   &  50.6   &  60.9  & 59.5 & 79.0  &   79.6 &    81.1  &  88.5  &  0.97   &  0.89   & 0.95  & 0.88  &  0.97   &  0.93  &  0.98  &  0.95    \\  
	    ethernet   & 65.6   &  67.0   &  65.7  & 66.5 & 78.8  &   76.4 &    80.5  &  85.1  &  0.84   &  0.58   & 0.85  & 0.73  &  0.89   &  0.79  &  0.91  &  0.83     \\ 
	    mem\_ctrl  & 57.6   &  61.8   &  57.7  & 64.6 & 77.8  &   75.8 &    80.5  &  88.6  &  0.98   &  0.92   & 0.97  & 0.93  &  0.98   &  0.95  &  0.98  &  0.96     \\ 
	    pci        & 70.7   &  69.3   &  71.2  & 71.6 & 79.3  &   75.8 &    82.2  &  91.3  &  0.97   &  0.89   & 0.94  & 0.87  &  0.97   &  0.94  &  0.98  &  0.96     \\ 
	    spi        & 57.0   &  57.4   &  49.6  & 65.6 & 77.3  &   79.7 &    82.7  &  88.1  &  0.98   &  0.94   & 0.98  & 0.95  &  0.99   &  0.97  &  0.99  &  0.97    \\ 
	    tv80       & 74.2   &  63.9   &  71.6  & 66.3 & 68.9  &   68.6 &    74.4  &  87.3  &  0.96   &  0.87   & 0.96  & 0.91  &  0.97   &  0.93  &  0.97  &  0.94    \\ 
	    usb\_funct & 69.6   &  68.6   &  70.9  & 72.7 & 77.5  &   73.1 &    79.7  &  92.8  &  0.95   &  0.86   & 0.95  & 0.91  &  0.96   &  0.92  &  0.98  &  0.95    \\ 
vga\_lcd*  & 67.2   &  55.1   &  77.5  & 79.9 & 86.3  &   88.2 &    88.7  &  93.4  &  0.73   &  0.38   & 0.83  & 0.71  &  0.85   &  0.69  &  0.88  &  0.76    \\ 
	    wb\_conmax & 42.6   &  72.2   &  44.0  & 56.9 & 66.7  &   70.1 &    71.4  &  88.5  &  0.94   &  0.81   & 0.94  & 0.88  &  0.94   &  0.87  &  0.96  &  0.92    \\ 
	\hline     
    	  DLX      & 71.1   &  73.1   &  59.2  & 73.0 & 79.5  &   80.6 &    82.8  &  91.0  &  0.94   &  0.81   & 0.93  & 0.87  &  0.95   &  0.90  &  0.97  &  0.94    \\ 
    	  ALPHA    & 63.6   &  75.6   &  71.4  & 79.9 & 81.4  &   78.9 &    85.8  &  92.2  &  0.91   &  0.74   & 0.93  & 0.85  &  0.94   &  0.88  &  0.96  &  0.91    \\ 
    	  FPU      & 61.6   &  82.7   &  63.9  & 74.5 & 71.7  &   74.3 &    80.4  &  89.9  &  0.98   &  0.95   & 0.97  & 0.93  &  0.98   &  0.94  &  0.99  &  0.97    \\ 
 mor1k*    & 71.3   &  62.1   &  82.0  & 85.1 & 86.4  &   86.7 &    89.1  &  94.6  &  0.33   &  -0.20  & 0.56  & 0.22  &  0.48   &  0.10  &  0.60  &  0.27    \\ 
 OR1200*   & 67.9   &  N/A    &  84.1  & 82.7 & 89.0  &   90.1 &    93.8  &  96.1  &  0.53   &   0.10  & 0.80  & 0.48  &  0.75   &  0.35  &  0.81  &  0.51    \\ 
	\hline     
 leon2*    & 67.2   &  N/A    &  83.1  & 90.2 & 92.4  &   94.2 &    94.9  &  97.0  &  0.05   &  -0.25  & 0.55  & 0.27  &  0.70   &  0.43  &  0.80  &  0.65    \\
 leon3mp*  & 69.8   &  N/A    &  80.4  & 82.1 & 80.5  &   81.1 &    82.4  &  89.2  &  0.50   &  -0.09  & 0.52  & -0.47 &  0.62   &  0.30  &  0.74  &  0.50    \\
 netcard*  & 73.7   &  N/A    &  80.6  & 73.1 & 80.5  &   80.2 &    84.0  &  90.1  &  0.35   &  -0.33  & 0.55  & -0.05 &  0.58   &  0.30  &  0.71  &  0.49    \\
  \hline
  \hline
Large (*)  &  \multirow{2}{*}{70.3}  &  \multirow{2}{*}{63.9}  &  \multirow{2}{*}{78.7} &  \multirow{2}{*}{78.6} &  \multirow{2}{*}{82.4} &  \multirow{2}{*}{84.0} &  \multirow{2}{*}{86.0} &  \multirow{2}{*}{92.3}  &  \multirow{2}{*}{0.55}  &  \multirow{2}{*}{0.17}  & \multirow{2}{*}{0.71} & \multirow{2}{*}{0.37}  & \multirow{2}{*}{0.73} & \multirow{2}{*}{0.48} &  \multirow{2}{*}{0.81} & \multirow{2}{*}{0.63} \\ 
Average  &    &    &   &   &  &  &   &   &  &  &   &   &  & & &   \\
\hline  \hline
\textbf{Average}   & \textbf{66.1}  &  \textbf{69.1}  &  \textbf{67.9}  &  \textbf{72.9}  &  \textbf{76.9}   & \textbf{78.0}   & \textbf{82.0} & \textbf{90.4} &  \textbf{0.86}  &  \textbf{0.70}  &   \textbf{0.89}  &   \textbf{0.78}   &  \textbf{0.91}   &  \textbf{0.82}  &  \textbf{0.94}  &  \textbf{0.87} \\ 
	\hline
  \end{tabular}
    \vspace{-6mm}
  }
\end{table*}

Besides correlation, we also measure the quality of net length estimators by how well they identify long nets in each circuit. Longer nets generally tend to contribute more wire load, and thus leaves a larger space for improving both wire-induced delay and the total wirelength. We believe identifying long nets is helpful for timing-related operations including timing-driven placement. Table~\ref{tbl:10LongestNet}(a) shows the accuracy in identifying the top 10\% longest nets. For each netlist, the 10\% longest nets are labeled as true, and the accuracy is measured in ROC curve's area under curve (AUC). Models capturing only one or two-hop neighbors, like MC and Poly, perform the worst. On average, ISPL outperforms MC and Poly with AUC $\approx$ 0.69. Notice that for large designs with more than 500 thousand nets, our implemented ISPL takes days of runtime, which is much slower than placement and too time-consuming in our experiment. Thus, we denote `N/A' for these designs in Table~\ref{tbl:10LongestNet}(a) and omit them when measuring the average accuracy for ISPL. Using our proposed node features, the ANN achieves AUC $\approx$ 0.73. In comparison, graph methods like GCN (AUC $\approx$ 0.78) and GAT (AUC $\approx$ 0.77) perform significantly better by learning with a larger receptive field reaching five-hop neighbors. By combining shallow and deep embeddings, Net$^\mathbf{2f}$ achieves AUC $\approx$ 0.82. Net$^\mathbf{2a}$ achieves AUC $\approx$ 0.90 by learning more global information from clustering on edge features with its edge convolution layer. The trend Net$^\mathbf{2a}$ > Net$^\mathbf{2f}$ > GAT > ANN clearly decompose the contribution of different component of our ML algorithm. The good accuracy can be attributed to convolution of node features introduced in GAT, our customization of residual connections in Net$^\mathbf{2f}$, and the global information in Net$^\mathbf{2a}$. 

Besides the average accuracy over all designs, we also count the average accuracy over 8 large designs with more than 100 K nets in Table~\ref{tbl:10LongestNet}. These large designs are marked with asterisks (*). The accuracy trend Net$^\mathbf{2a}$ > Net$^\mathbf{2f}$ > GAT > ANN remains the same for large designs. Also, the accuracy in net length prediction does not degrade when measured on large designs only.

\subsection{Timing Prediction Result}

For timing estimators, we first evaluate the accuracy in predicting the delay of each timing arc.  
Table~\ref{tbl:10LongestNet}(b) measures the delay of all arcs in the same netlist with both correlation coefficient $R$ and coefficient of determination $R^2$.
For a biased estimator, which means its predictions are consistently higher or lower than the ground-truth labels, it may achieve high $R$ but much lower $R^2$ if it well correlates with the label. 


In Table~\ref{tbl:10LongestNet}(b), the report\_timing is the pre-placement timing report from the timing engine from a representative commercial tool\footnote{According to the license agreement, we should not disclose the name of vendor's tool when making direct comparisons with it.}. Before placement, due to the absence of wire length information, the timing engine tends to underestimate wire load in its timing report. As a result, the reported delay values of all arcs are consistently smaller than the ground-truth post-placement report. In other words, the pre-placement report is biased towards more optimistic predictions, which is especially undesired in timing analysis since it under-estimates timing violations. Such a bias is reflected in its low averaged $R^2=0.70$, but the bias does not affect the correlation $R=0.86$. When averaged over all 37 designs, the fast timing estimator Time$^\mathbf{f}$ is 0.05 higher in $R$ and 0.12 higher in $R^2$ than the report from the commercial EDA tool. The accurate version, Time$^\mathbf{a}$, further achieves $R=0.94$ and $R^2=0.87$. The improvement in $R$ means both Time$^\mathbf{f}$ and Time$^\mathbf{a}$ not only fix the bias in pre-placement timing reports but also improve the correlation.  

In Table~\ref{tbl:10LongestNet}(b), to analyze the contributions from different features, we also include an extra baseline named `onlyTime', referring to only using timing-related features listed in Section~\ref{sec:timing_method} as input features of the timing model. This baseline measures whether timing input features alone are enough for accurate timing estimations. This `onlyTime' outperforms the report\_timing with $R=0.89$ and $R^2=0.78$, but is still less accurate than Time$^\mathbf{f}$, and the gap is even larger compared with Time$^\mathbf{a}$. This gap shows the contribution purely from the net-length-related predictions. In addition, we further measured the accuracy of a timing model using the ground-truth net-length as an input feature. The averaged accuracy turns out to be $R = 0.98$ and $R^2 = 0.96$. This can be viewed as an upper limit of the current Time$^\mathbf{f/a}$ model, assuming perfect net length predictions are available.

\begin{figure}[!p]
  \centering
    \vspace{-.3in}
    \includegraphics[width=0.82\columnwidth]{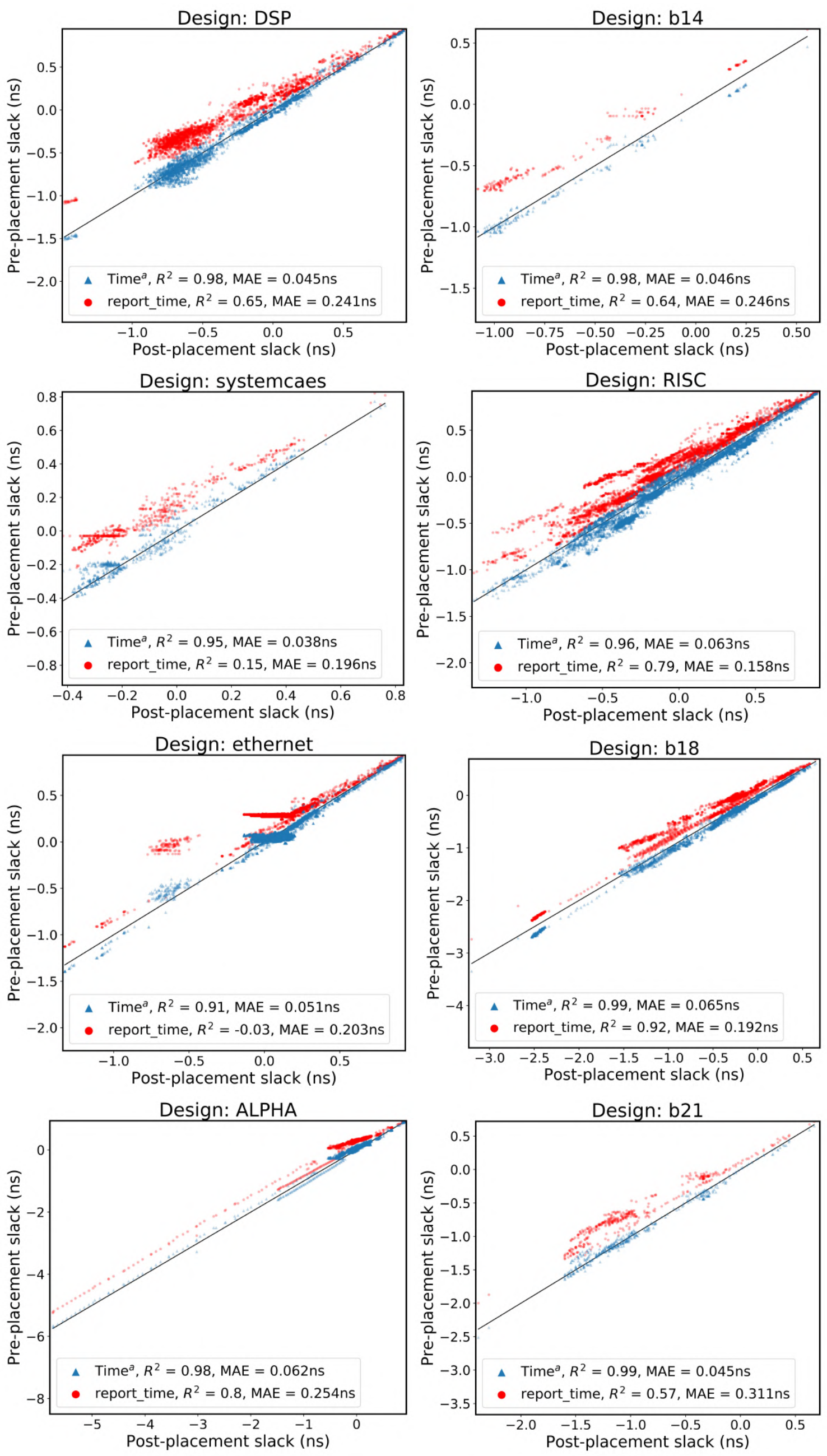}
    	\vspace{-1mm}
  \caption{Examples on pre-placement slack prediction.}
  \label{slack_example}
\end{figure}

We observe that the arc-delay prediction accuracy in Table~\ref{tbl:10LongestNet}(b) is lower for large designs with more than 100 K nets, like leon2, netcard, and OR1200. But as our analysis of Table~\ref{tbl:10LongestNet}(a) has demonstrated, net length prediction accuracy for large designs is not worse than average. Our study shows that in these large designs, there are much more very-long nets, which cause dominating wire-induced incremental delay. It means the gap between pre-placement and pose-placement timing is significantly larger and more difficult to predict, thus an inaccurate net length estimation causes a larger penalty to timing prediction accuracy. This is validated by the poor accuracy of `report\_timing' by the commercial tool on these large designs. Although our Time$^\mathbf{a}$ performs not as well on large designs, it more significantly outperforms the commercial tool baseline on large designs. As Table~\ref{tbl:10LongestNet}(b) shows, Time$^\textbf{a}$ outperforms `report\_timing' by 0.08 (= 0.94 - 0.86) in correlation $R$ when averaged over all designs, while by as large as 0.26 (= 0.81 - 0.55) in $R$ for large designs.


Based on the prediction on all delay arcs, we perform the PERT~\cite{chang2003statistical} traversal, which is widely used in STA, to measure the arrival time, required arrival time, and slacks. Figure~\ref{slack_example} shows predictions versus labels on calculated slacks for both pre-placement timing report and the estimator Time$^\mathbf{a}$. Eight representative designs are presented, and each subplot measures all slacks in the netlist. Similar to the trend of arcs delay, slacks from the pre-placement timing report are consistently higher than ground-truth, thus are biased towards optimism. For each design in Figure~\ref{slack_example}, the timing estimator Time$^\mathbf{a}$ achieves a much higher accuracy when measured in $R^2$ and absolute errors. Both averaged and median accuracies on slack prediction over all 37 designs are shown in Table~\ref{tbl:slack}. The averaged accuracy is more affected by less accurate predictions thus the median accuracy is higher. The trend in accuracy remains the same, showing Time$^\mathbf{a}$ > Time$^\mathbf{f}$ > report\_timing. On average, the Time$^\mathbf{a}$ achieves high $R^2=0.91$, indicating that high correlation and low bias are achieved simultaneously. Its mean absolute error is $0.11$ns, which reduces the error in pre-placement timing report by more than $50\%$ and is less than $10\%$ of the clock cycle.

\begin{table}[!h]
  \centering
  \caption{Pre-placement Path Slack Prediction Accuracy} 
  \label{tbl:slack}
\renewcommand{\arraystretch}{1.2}
  \begin{tabular}{|c | c c | c c | c c |}
 	\hline
 	   &   \multicolumn{2}{c|}{report\_timing} &  \multicolumn{2}{c|}{Time$^\mathbf{f}$} &
 	   \multicolumn{2}{c|}{Time$^\mathbf{a}$} \\
 	  &  Error  & R$^2$ & Error & R$^2$ & Error & R$^2$\\
	\hline
	\hline
Mean   &  0.38 ns  &  0.39  &  0.16 ns   &  0.86  &  0.11 ns  &  0.91 \\
  \hline
  Median   &  0.18 ns  &  0.77  &  0.07 ns   &  0.95  &  0.05 ns  &  0.97 \\
	\hline
  \end{tabular}
\end{table}

\begin{figure}[!h]
  \centering
    \includegraphics[width=0.8\columnwidth]{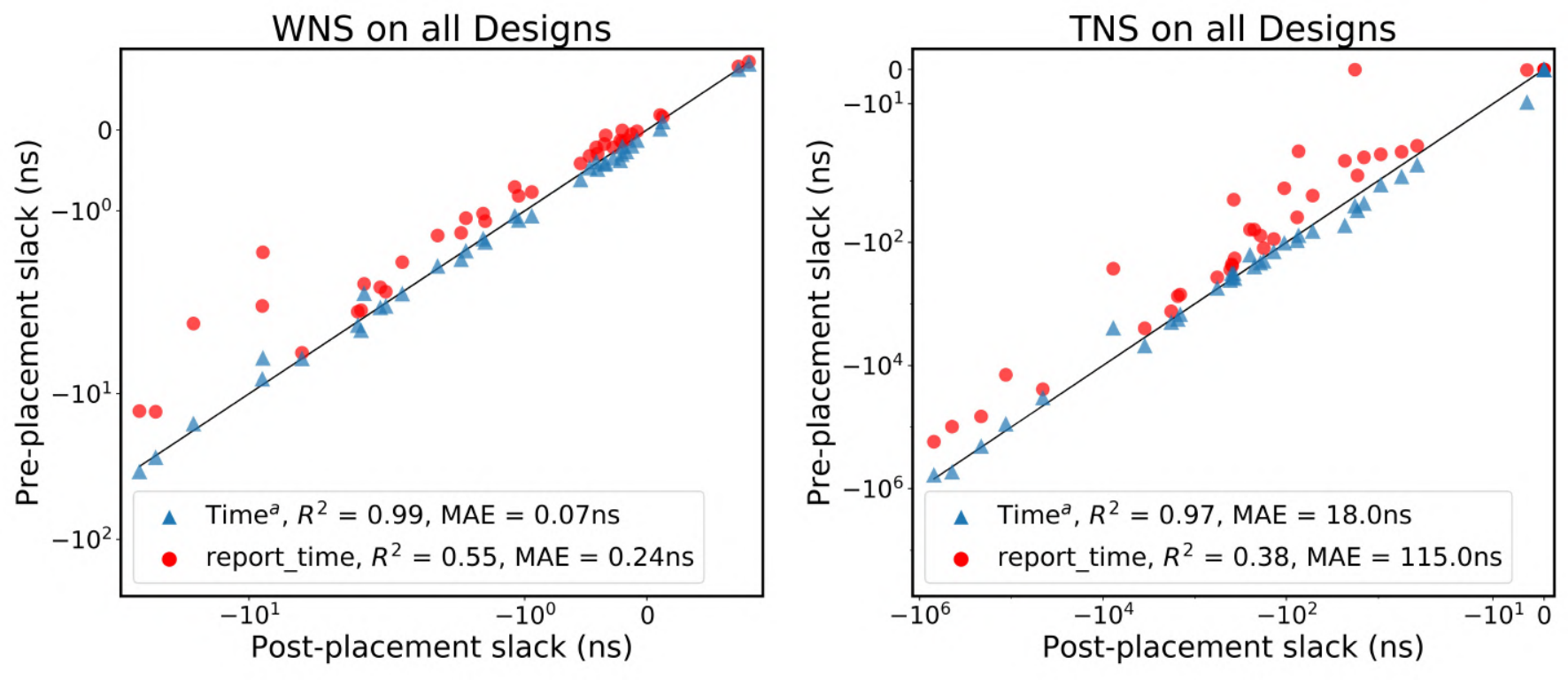}
    	\vspace{-1mm}
  \caption{WNS (left) and TNS (right) of all designs.}
  \label{TNS_WNS}
\end{figure}

According to all slacks calculated by traversing delay predictions, we can easily measure the total negative slack (TNS) and worst negative slack (WNS) of each netlist. The TNS and WNS of all designs are presented in Figure~\ref{TNS_WNS}. For the small portion of netlists with all slacks positive, we set the WNS to the lowest positive slack and leave the TNS to be zero. Each point in Figure~\ref{TNS_WNS} represents the TNS/WNS of one netlist. The estimator Time$^{\mathbf{a}}$ maintains its high accuracy in TNS and WNS prediction. Considering this correlation is measured on all designs and each testing design is completely unseen by the model, the result proves that the performance of Time$^{\mathbf{a}}$ is robust on all 37 tested designs in our experiment.

\begin{figure}[!h]
  \centering
    \includegraphics[width=0.8\columnwidth]{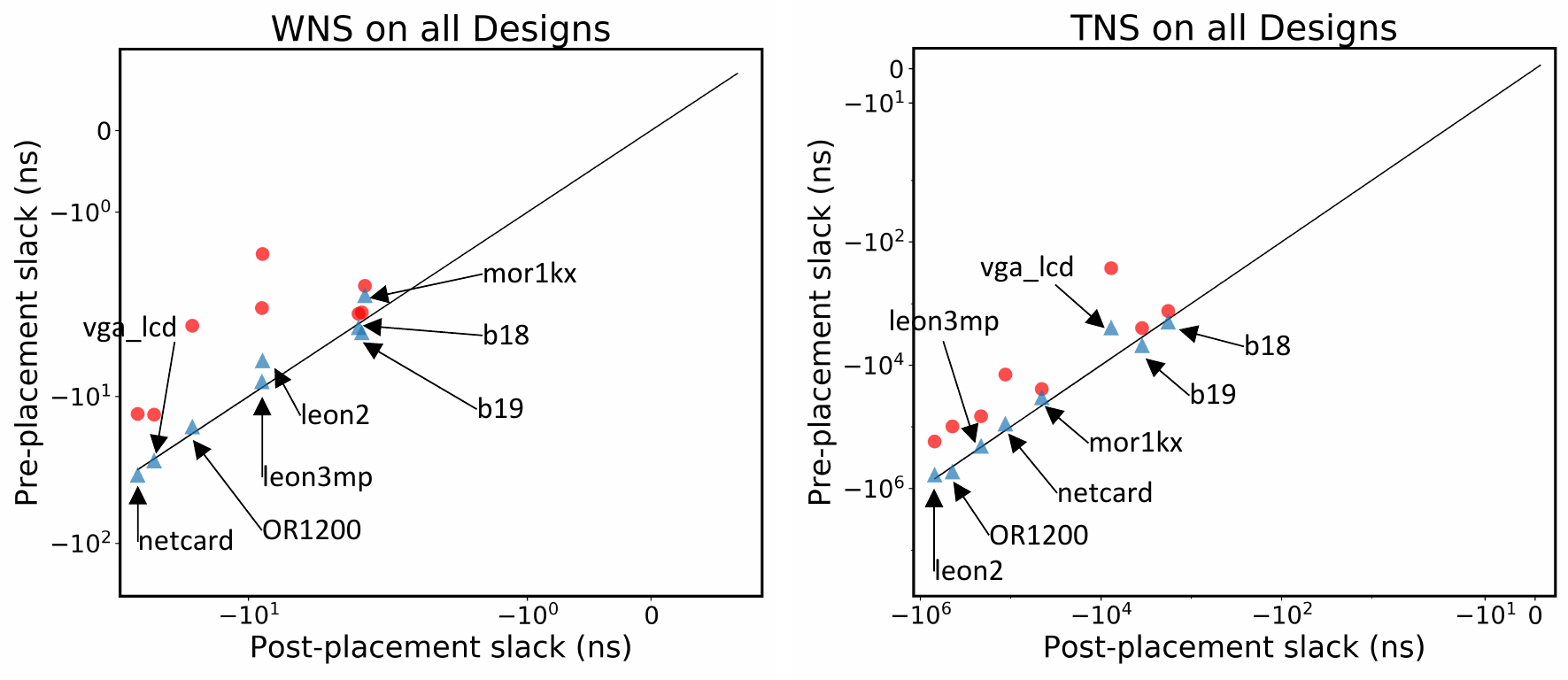}
    	\vspace{-1mm}
  \caption{WNS (left) and TNS (right) on only large designs.}
  \label{TNS_WNS_large}
\end{figure}

To show our timing model's performance on large designs more clearly, we pick those 8 largest designs with more than 100K nets and only show the TNS/WNS predictions on these designs in Figure~\ref{TNS_WNS_large}. The arrow with design name text in Figure~\ref{TNS_WNS_large} points to the prediction from Time$^\mathbf{a}$, and the corresponding evaluation from report\_timing shares the same ground-truth in the x-axis. Compared with other designs, WNS/TNS of large designs is more negative. The estimation from report\_timing is close to ground-truth for designs `b18' and `b19', but significantly more optimistic for designs `vga\_lcd', `netcard', `OR1200', and `leon2'. In comparison, our Time$^\mathbf{a}$ gives rather accurate predictions to all these large designs.

\subsection{Runtime Comparison}
\label{subsec:runtime}

\begin{table*}[!h]
  \centering
  \vspace{1mm}
  \caption{Detailed Runtime Comparison on Representative Designs (In Seconds)}
  \label{tbl:runtime}
  \setlength{\tabcolsep}{1mm}
  \renewcommand{\arraystretch}{1.3}
  \resizebox{\linewidth}{!}{
  \begin{tabular}{| c   c  | c  c  | c  c  | c  c | c  c | c  c | }
 	\hline
 	\multirow{2}{*}{Design} &  \multirow{2}{*}{\# Net} & \multirow{2}{*}{Place} 
 	& \multirow{2}{*}{Partition} & Net$^\mathbf{2f}$  & Net$^\mathbf{2a}$  &  Net$^\mathbf{2f}$ &  Net$^\mathbf{2a}$  & Time$^\mathbf{f}$ & Time$^\mathbf{a}$  & Time$^\mathbf{f}$ & Time$^\mathbf{a}$  
 	 \\
 	     &   &     &    &  Infer  & Infer & Speedup  & Speedup &  Infer  & Infer  & Speedup  & Speedup  \\
 	\hline
 	b15     &  12 K   &  30     &  1.6   &  0.03  &  0.03  &  1K$\times$  &  18$\times$  &  0.08 &   0.08  &  0.3K$\times$  &  18$\times$   \\ 
 	b21     &  39 K   &  128    &  7.1   &  0.05  &  0.05  & 2.6K$\times$ &  18$\times$  &  0.25 &   0.27  &  0.4K$\times$  &  17$\times$  \\ 
    mor1k   &  178 K  &  1174   &  64.8  &  0.11  &  0.25  & 11K$\times$  &  18$\times$  &  0.97 &   1.1   &  1.1K$\times$  &  18$\times$  \\ 
    netcard &  551 K  &  5517   &  313   &  0.48  &   1.0  & 11K$\times$  &  18$\times$  &  2.9  &   5.8   &  1.6K$\times$  &  17$\times$  \\ 
    leon3mp &  640 K  &  7180   &  283   &  0.54  &   1.0  & 13K$\times$  &  25$\times$  &  3.8  &   5.9   &  1.7K$\times$  &  25$\times$  \\ 
    OR1200  &  847 K  &  11353  &  427   &  0.67  &   1.5  & 17K$\times$  &  26$\times$  &  5.8  &   9.4   &  1.8K$\times$  &  26$\times$  \\ 
    leon2   &  835 K  &  11544  &  428   &  0.67  &   1.4  & 17K$\times$  &  27$\times$  &  5.0  &   9.8   &  2.0K$\times$  &  26$\times$  \\ 
	\hline
 	\end{tabular}
 	}
\end{table*}

Table~\ref{tbl:runtime} shows the runtime of placement, net length estimators Net$^\mathbf{2f/2a}$, and timing estimators Time$^\mathbf{f/a}$. We report the runtime separately for multiple representative designs, which cover a large range of design sizes from 12K to 800K nets in the netlist. For a fair comparison, the runtime of placement includes the placement algorithm only, without any extra time for file I/O, floorplanning, or placement optimization. The inference of Net$^\mathbf{2f/2a}$ requires one Nvidia GTX 1080 graphics card, and other runtimes are performed with CPU only.

As Table~\ref{tbl:runtime} shows, Net$^\mathbf{2a}$ takes slightly longer inference time than Net$^\mathbf{2f}$ for its extra edge convolution layer. The overall runtime of Net$^\mathbf{2a}$ includes both partition and inference. Partitioning contributes the majority of Net$^\mathbf{2a}$'s runtime. Net$^\mathbf{2a}$ is more than $>15\times$ faster than placement. The runtime of Net$^\mathbf{2a}$ can be potentially improved by using coarser but faster partition $P$ and $M$, especially on larger designs. Without partition, Net$^\mathbf{2f}$ is $>1000\times$ faster than placement. For timing estimators, similarly, Time$^\mathbf{a}$ takes longer inference time than Time$^\mathbf{f}$ since it takes more input features. The Time$^\mathbf{a}$ is $>15\times$ faster and the Time$^\mathbf{f}$ is $>1000\times$ faster than the placement for not-too-small designs. The runtime comparison between different designs in Table~\ref{tbl:runtime} shows that the speedup of Net$^\mathbf{2f/2a}$ and Time$^\mathbf{f/a}$ is more significant for larger designs. It validates the scalability of our method. 

\begin{table}[!h]
  \centering
  \vspace{1mm}
  \caption{Synthesis Runtime Measurement (In Seconds)} 
  \label{tbl:runtimeSyn}
  \renewcommand{\arraystretch}{1.2}
  \setlength{\tabcolsep}{3mm}
  \resizebox{0.68\linewidth}{!}{
  \begin{tabular}{| c | c | c| }
	\hline
	Traditional Synthesis &  Placement &  Partition \\
	\hline
	167   &   128 & 9.4 \\
	\hline
	Physical-aware Synthesis & \multicolumn{2}{c|}{ Physical-aware Synthesis } \\
	(with Fast Placement) &  \multicolumn{2}{c|}{ (with Complete Placement) } \\
	\hline
	282   &   \multicolumn{2}{c|}{414} \\
	\hline
  \end{tabular}
  }
  \vspace{-2mm}
\end{table}

Besides comparisons with placement, to gain more insights on the whole VLSI design flow, we also evaluate the runtime of both traditional logic synthesis and physical-aware synthesis from an industry-standard commercial synthesis tool\footnote{According to the license agreement, we should not disclose the name of vendor's tool when making direct comparisons with it.} in Table~\ref{tbl:runtimeSyn}. It is measured on the design b21. 
As introduced, physical-aware synthesis explicitly addresses the interaction between synthesis and layout with the cost of extra runtime. The commercial synthesis tool we use offers two options for physical-aware synthesis, one using a fast placement method and the other using the complete placement to provide feedback on the backend implementations. As shown in Table~\ref{tbl:runtimeSyn}, the two versions of physical-aware logic synthesis take 115 and 247 more seconds than traditional logic synthesis. This extra runtime is close to the time spent on placement. This verifies our claim that compared with ML-based solutions, the physical-aware synthesis is more  time-consuming.

As for the model training time, it takes around 30 minutes to train the Net$^\mathbf{2a}$ model, and less than 10 minutes to train the RF-based timing estimator Time$^\mathbf{a}$.

\section{Summary}

In this chapter, we first present Net$^\text{\textbf{2}}$, a graph attention network method customized for individual net length estimation. It includes a fast version Net$^\text{\textbf{2f}}$ which is 1000$\times$ faster than placement, and an accuracy-centric version Net$^\text{\textbf{2a}}$ which extracts global information and significantly outperform all previous net length estimation methods. Based on net length predictions, we further develop a pre-placement timing estimator, which achieves significantly better correlations with ground truth compared with the pre-placement timing report from commercial tools.

%% file: txt/4_IRdrop.tex
\chapter{Fast IR Drop Modeling on Layout}
\label{chapter:ir_drop}

\section{Background}

IR drop describes the deviation of a power supply level from its specification that occurs when current flows through power grids.  It must be restricted in order for a circuit to meet its timing target and function properly. As design and manufacturing technologies advance, the increased current load further exaggerates IR drop violations, which become a critical concern for both VLSI design and test~\cite{tehranipoor2010power}. Note that the power supply noise (PSN) is sometimes also loosely referred to as IR drop. But the PSN actually comprises both $Ldi/dt$ and the IR drop \cite{chen1997power}. The $Ldi/dt$, or named voltge droop component is an inductive effect caused by rapid current changes through power grids. This chapter focuses only on the IR drop problem, without considering the $Ldi/dt$.

In order to meet IR drop constraints, designers need to estimate and mitigate IR drop throughout design stages from placement to signoff in multiple iterations. It may also be measured during post-silicon verification. Obtaining an accurate estimation of IR drop through simulation-based commercial tools is very time consuming~\cite{Redhawk, fang2018machine}. Thus, IR drop mitigation guided by frequent IR drop simulations is computationally costly and hampers the overall design turnaround time. To speed up this process, a fast yet accurate IR drop estimator becomes a critical need, and ML techniques provide promising solutions.

\begin{figure}[!h]
  \centering
    \includegraphics[width=0.7\columnwidth]{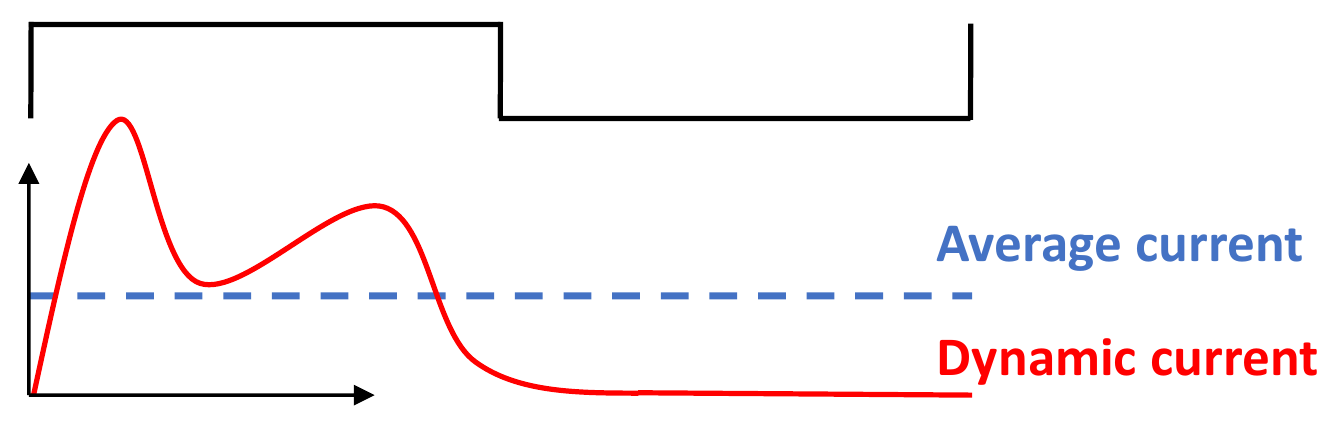}
  \captionsetup{justification=centering}
  \caption{Static and dynamic analysis on current.}
  \label{static}
\end{figure}

IR drop estimators can be classified into two major categories, based on whether they estimate static IR drop or dynamic IR drop.
Figure \ref{static} shows a comparison between the average current and dynamic current in one clock cycle~\cite{nithin2010dynamic}. 
The static IR drop analyses in most commercial tools only measure the average current drawn from power grids without considering switching activities~\cite{Redhawk, nithin2010dynamic}. It is widely used to identify the weakness of a power delivery network (PDN) at an early design stage when switching vectors are not available. There have been many traditional methods for fast power grids analysis \cite{zhao2002hierarchical, qian2005power, chen2001efficient, zhuo2008power}. In contrast, dynamic IR drop captures the peak transient current value based on switching activities. Thus, it is a more strict constraint and much more difficult to predict \cite{nithin2010dynamic}. The significant difference between static and dynamic IR drop leads to distinctive problem settings and corresponding ML solutions. 

\begin{table}[!h]
  \centering
  \caption{Comparison Among Different IR Drop Estimators}
  \label{tbl:t_prev}
  \begin{tabular}{|l | c c|}
 	\hline
	ML Methods & Model & Design Independent   \\  
	\hline
	\cite{yamato2012fast} (ITC 12) &    Linear Regression  &  No \\
	\cite{ye2014chip} (VTS 14)  &    SVM  &   No \\
	\cite{dhotre2017identification} (ATS 17) &  Clustering  & Unsupervised  \\
	\cite{lin2018ir} (VTS 18) &  ANN   &   No \\
	\cite{fang2018machine} (ICCAD 18) &  XGBoost   &   No \\
	\hline
	PowerNet & Max-CNN & Yes \\ 
	\hline
  \end{tabular}
\end{table}

Our work named PowerNet focuses on the estimation of dynamic IR drop. Related ML-based IR drop estimators are summarized in Table \ref{tbl:t_prev}. These prior works~\cite{yamato2012fast, ye2014chip, lin2018ir, fang2018machine} are not ``design-independent'', i.e., {\em transferable} to new designs that are not seen in its training dataset. Here we require the \emph{new} design to be different from training designs at the netlist level. Thus the following examples are not viewed as cross-design: 1) Models trained and tested with different layout implementation of the same netlist. 2) Models trained and tested on multiple designs, which appear in both training and testing set. As a result, previous works require training a new model for each distinct design. Some work~\cite{yamato2012fast} even dedicates one model for every single cell. Training a new model with new labels entails a long simulation and training time, which defeats the original purpose of fast estimation. The only exception is \cite{dhotre2017identification}, which is based on unsupervised learning and does not learn any previous knowledge.

\section{Methodology}

\subsection{Problem Formulation}

This work aims at detecting locations of IR drop hotspots. Hotspots are regions where IR drop is greater than a specified threshold. To estimate IR drop, every design is tessellated into an array of tiles, each of which is an $l\times l$ square. The tile size $l$ controls the granularity of our solution. In this way, a design with the size of $W \times H$ is represented as a $w \times h$ matrix, where $w = W/l$ and $h = H/l$. The IR drop at each tile is the mean value of IR drop of all cells within it. Then IR drop for the whole design is $IR \in \mathbf{R} ^{w \times h}$. The ground-truth $IR$ is also referred to as label in this paper. As for input features, different types of power dissipation values are calculated for each tile. We refer to each $w \times h$ power matrix as a power map. Essentially, power maps are the distribution of power density. PowerNet $F$ tries to give the closest estimation $F^*$ on $IR$ based on all $G$ different power maps $\{P_{map1} \,... \,P_{mapG}\}$. 

\begin{gather*} 
F:  \{P_{map1}\in \mathbf{R} ^{w \times h} ... \, P_{mapG} \in \mathbf{R} ^{w \times h}\}  \rightarrow \mathbf{R} ^{w \times h} \\ 
F^* = \argmin_{F} Loss(F(\{P_{map1} ... \, P_{mapG}\}), IR).
\end{gather*}

\subsection{Feature Extraction}

According to Ohm's Law, excessive IR drop can be caused by either high current or high resistance. As is typical in state-of-the-art VLSI design, we assume a uniform power grid in the power delivery network (PDN), which means the resistance distribution across a whole design is also rather uniform. Thus in PowerNet, we choose not to spend extra time calculating resistance for each cell. For designs with a non-uniform PDN, each cell's power value can be scaled by its resistance. The influence of resistance is further elaborated in Section \ref{resistance}. When resistance is considered consistent, current becomes the only key issue in IR drop estimation. Since local power consumption is proportional to local current, PowerNet utilizes cell power as its input features.

For each cell $c$, we do not exhaust all possible features that seem to be relevant, which make the model too complex and overfit. Instead, we select features that prove to provide essential information for IR drop estimation. Hard macros are not included. Below are the details of all features and the labels extracted from them:
\begin{itemize}
\item \textbf{Power}: Three types of power values are extracted, including internal power ($p_i$), switching power ($p_s$), and leakage power ($p_l$).
\item \textbf{Signal arrival time}: The minimum and maximum signal arrival times to the cell within a clock cycle are extracted, denoted as $t_{min}$ and $t_{max}$, respectively. 
\item \textbf{Coordinates}: The cell locations after placement are extracted, including min and max x axis ($x_{min}, x_{max}$), and min and max y axis ($y_{min}, y_{max}$).
\item \textbf{Toggle rate}: It ($r_{tog}$) describes how often output changes with regard to a given clock input.
\item \textbf{IR drop}: It ($ir$) measures the difference between the nominal supply voltage and the actual voltage arrived at each cell. 
\end{itemize}

All of the above features are scalar values. For these power types, internal power $p_i$ means power dissipated by capacitance internal to each cell; switching power $p_s$ is power dissipated by the load capacitance at the output of the cell; leakage power $p_l$, which is relatively small in the experiment, is consumed by unintended leakage that does not contribute to function. Based on these basic power types, we can generate more power information for each cell: 

\begin{gather*} 
p_{sca} = (p_i + p_s)*r_{tog} + p_l\ , \quad p_{all} = p_i + p_s + p_l
\end{gather*}

Both $p_{sca}$ and $p_{all}$ reflect the overall power dissipated by cells, but $p_{sca}$ scales the overall power by toggle rate of each cell. PowerNet learns to combine the total power from these different sources of power dissipation.

\subsection{Preprocessing by Decomposition}

\begin{algorithm}[!t]
\caption{Preprocessing by Decomposition}
\label{alg4}
  \textbf{Input}: \{$p_i$, $p_s$, $p_l$, $t_{min}$, $t_{max}$, $x_{min}$, $x_{max}$, $y_{min}$, $y_{max}$, $r_{tog}$\} for every cell $c$. Design width $W$, height $H$, cell number $C$ and clock cycle $T$. Tile size $l$ and time window $t$. \\
  \textbf{Preprocess}:
\begin{algorithmic}[1]
\State $w = W/l$, $h = H/l$, $N = T/t$ 
\State Set $P_{i}$, $P_{s}$, $P_{sca}$, $P_{all}$ $\in \{0\} ^ {w\times h}$, $\{P_{t}[\,j] \in \{0\} ^ {w\times h} \mid j \in [1, N]\}$
\For{each cell $c \in [1,C]$}
\State $p_{sca} = (p_i + p_s) * r_{tog} + p_l $
\State $p_{all} = p_i + p_s + p_l$
\State $x_{n} = \floor{(x_{min}/l)}$, $x_{x} = \ceil{(x_{max}/l)}$, $y_{n} = \floor{(y_{min}/l)}$, $y_{x} = \ceil{(y_{max}/l)}$ \label{line:s_b}
\State $s = (x_{x} - x_{n}) * (y_{x} - y_{n})$
\State Set mask $M \in \{0\} ^ {w\times h}$, $M[x_n:x_x][y_n:y_x] = 1$
\State $P_{i}$ += $M* p_{i}/s$, $P_{s}$ += $M* p_{s}/s$, $P_{sca}$ += $M* p_{sca}/s$, $P_{all}$ += $M* p_{all}/s$  \label{line:s_e}
\For{each int $j \in [1,N]$}        \label{line:t_b}
\If {$t_{min} < j*t$ and $t_{max} > j*t$}
\State $P_{t}[\,j]$ += $M* p_{sca}/s$   \label{line:t_e}
\EndIf
\EndFor
\EndFor
\end{algorithmic} 
  \textbf{Output}: Power maps $P_{i}$, $P_{s}$, $P_{sca}$, $P_{all}$ $\in \mathbb{R} ^ {w\times h}$, \\
  \hspace*{17mm} Time-decomposed power maps $\{P_{t}[\,j] \in \mathbb{R} ^ {w\times h} \mid j \in [1, N]\}$ 
\end{algorithm}

After power is extracted, the IR drop seen at each cell is not just simply proportional to its own cell power but also depends on its neighborhood due to both spatial and temporal current distributions. Spatially, local current is proportional to the sum of power demand of all cells in a local region. Hence, the power of neighboring cells also contributes to IR drop of the analyzed cell. We amortize cell power into grid tiles by a space decomposition. This also motivates us to adopt a CNN model in PowerNet, which is inherently designed for learning scalable two-dimensional patterns. Even when considering spatial information, a region with high overall power demand may still not be IR drop hotspot.  This case arises when cells in the region do not switch at the same time. Such asynchronous switching disperses voltage drop into a larger timing window. As a result, maximum dynamic IR drop, i.e. the highest-transient voltage drop, can still be low. PowerNet measures such influence by time decomposition during preprocessing.

Algorithm \ref{alg4} shows our preprocessing method. It generates power maps based on cell information. For each design, two types of power maps are generated. The first type includes $\{P_{i}$, $P_{s}$, $P_{sca}$, $P_{all}\}$. They only go through spatial decomposition and do not carry timing information. The second type $\{P_{t}[\,j] \in \mathbb{R} ^ {w\times h} \mid j \in [1, N]\}$ goes through both a space decomposition and a time decomposition.

\begin{figure}[!h]
  \centering
    \includegraphics[width=0.7\textwidth]{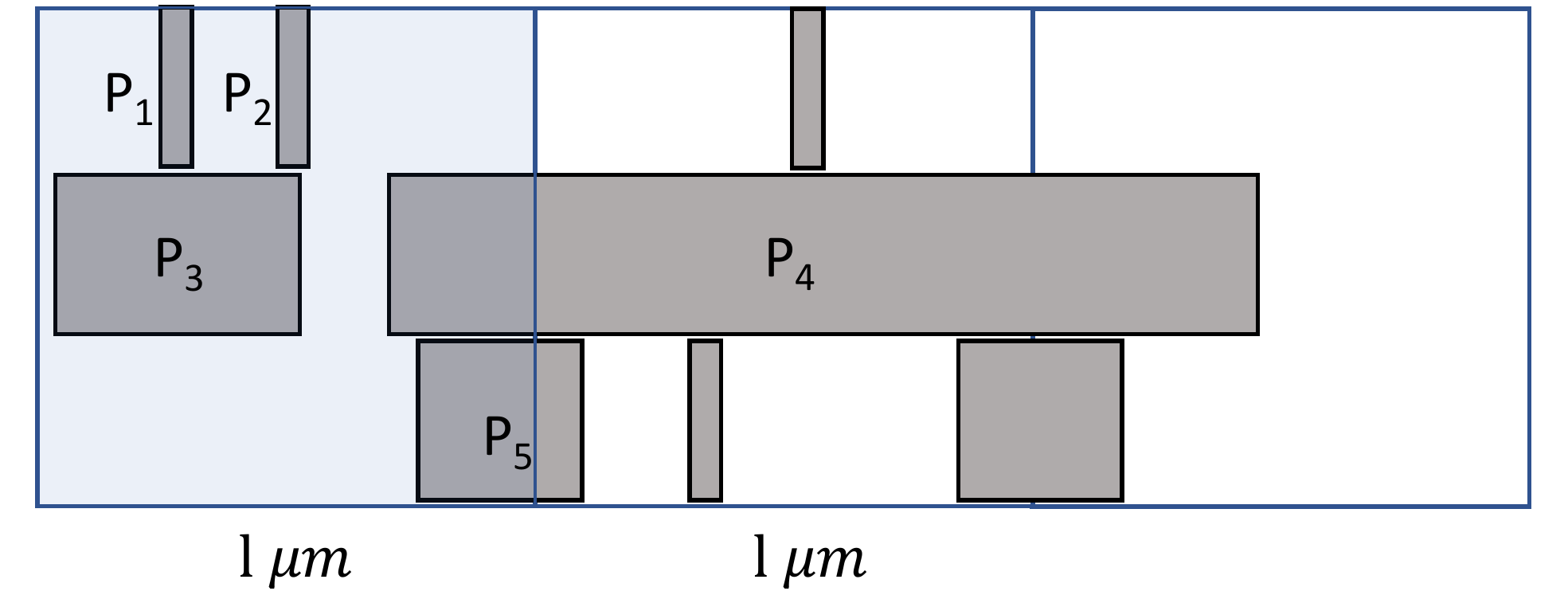}
    \vspace{-2mm}
  \caption{Space decomposition for IR drop estimation.}
  \label{spaceDecomp}
\end{figure}

As illustrated in Figure \ref{spaceDecomp}, space decomposition (Lines \ref{line:s_b} to \ref{line:s_e}) amortizes cell power into any grid tiles occupied by the cell. Assume the regular squares are grid tiles and grey rectangles are cells. $P_1$ to $P_5$ are cell power. For the leftmost highlighted tile, its power equals $P_1 + P_2 + P_3 + P_4/3 + P_5/2$. The long cell with power $P_4$ only contributes  one-third of its power to that of the highlighted tile, because altogether it overlaps with three tiles. Similarly, in line \ref{line:s_b} to \ref{line:s_e}, each cell contributes $p/s$, where $s$ is the number of overlapping tiles.

\begin{figure}[!h]
  \centering
    \includegraphics[width=.7\textwidth]{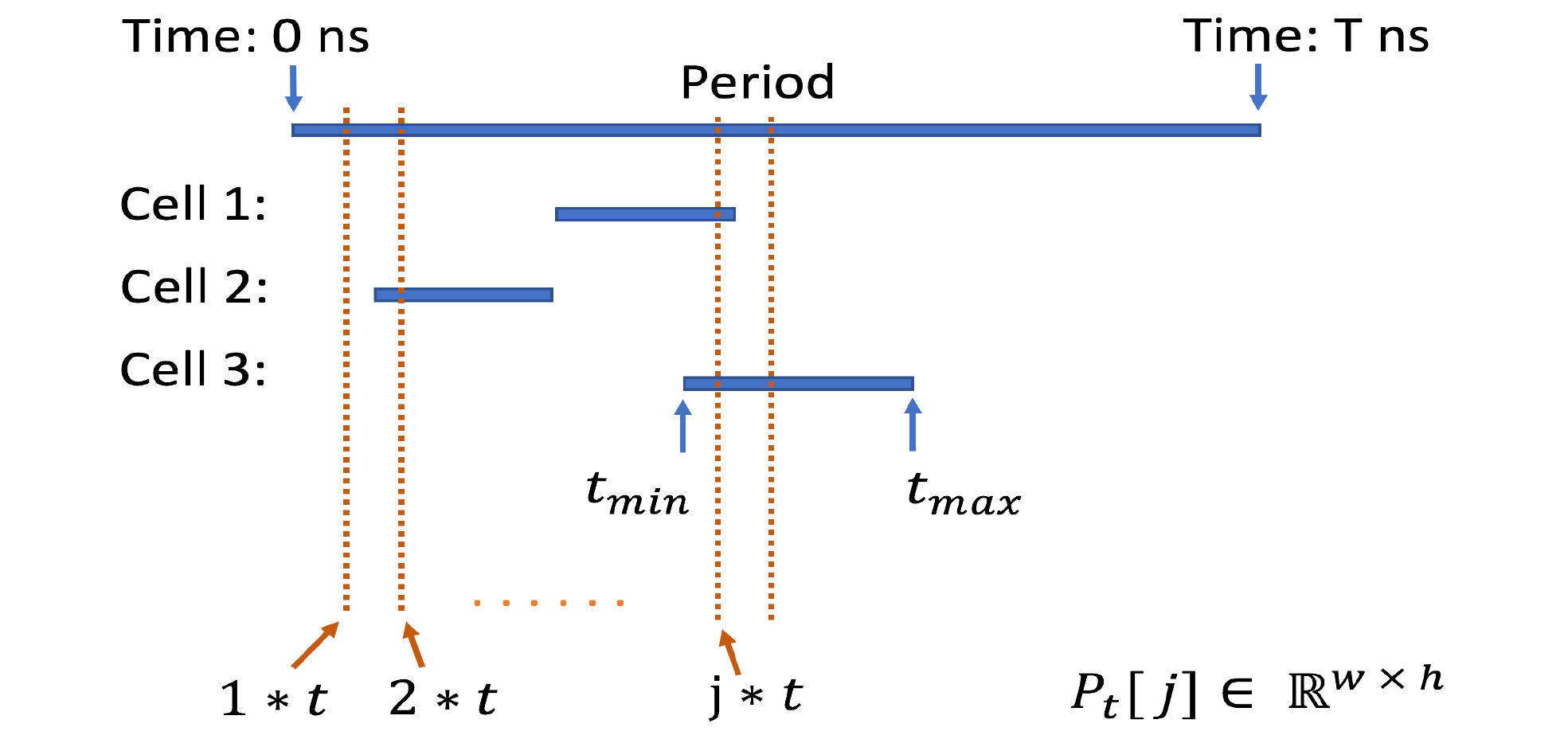}
  \caption{Time decomposition for IR drop estimation.}
  \label{timeDecomp}
\end{figure}

Lines \ref{line:t_b} to \ref{line:t_e} perform time decomposition. Every power map $P_t[\,j]$ corresponds to one time instant $j*t$. For each cell at $j*t$, it contributes power to a corresponding power map $P_t[\,j]$ only when $j*t$ falls between its signal arriving time $[t_{min}, t_{max}]$. In other words, only cells that can possibly switch at that instant are considered. Figure \ref{timeDecomp} demonstrates the mechanism. Vertical dashed lines are measured instants $1*t$ to $j*t$, and horizontal bars are the signal arrival time intervals of cells. In this example, only cells 1 and 3 will be counted for $P_t[\,j]$ and no cells are counted for $P_t[1]$.

\subsection{PowerNet Model}

\begin{algorithm}[!t]
\caption{Training of PowerNet}
\label{alg5}
  \textbf{Input}: IR drop label $IR \in \mathbb{R} ^ {w\times h}$, power $P_{i}$, $P_{s}$, $P_{sca}$, $P_{all}$ $\in \mathbb{R} ^ {w\times h}$, time-decomposed $\{P_{t}[\,j] \in \mathbb{R} ^ {w\times h} \mid j \in [1, N]\}$, input window size $k =2*k_h+1$, $k_h$ means half size. 
\begin{algorithmic}[1]
\Function{getInput}{$\,j, x, y$}
    \State Stack features $I = \{ P_{i}$, $P_{s}$, $P_{sca}$, $P_{all}$, $P_{t}[j] \} \in \mathbb{R} ^ {w\times h \times 5}$\label{line:l0}
    \State $I_{x,y} = I[x-k_h:x+k_h+1][y-k_h:y+k_h+1] \in \mathbb{R} ^ {k\times k \times 5}$ 
    \State return $I_{x, y}$
\EndFunction
\State
\State Initiate CNN model $f$: $\mathbb{R} ^ {k\times k \times 5} \rightarrow \mathbb{R}$, Loss function $J$
\For{$epoch \in [1,N_{epoch}]$}
\For{$x \in $ shuffle $([1, w])$, $y \in $ shuffle $([1, h])$}
\State $o_{max} = 0$
\For{each $j \in [1,N]$}                              \label{line:l1}
\State $I_{x, y} = $ \textproc{getInput} $ (\,j, x, y)$ \label{line:l2}
\State $o_{j} = f(I_{x, y})$
\If{$o_{max} < o_{j}$}
\State $o_{max} = o_{j}$
\EndIf
\EndFor
\State *Gradient Descent $f$ -= $\nabla J(o_{max}, IR[x, y])$
\EndFor
\EndFor
\end{algorithmic}
  \textbf{Output}: Trained CNN model $f$: $\mathbb{R} ^ {k\times k \times 5} \rightarrow \mathbb{R}$
\end{algorithm}

Algorithm \ref{alg5} shows how PowerNet $F$ handles power maps with its CNN model $f$. For each training epoch, it iterates through every tile $(x, y)$ in training designs. For every tile, it crops $k\times k$ input windows surrounding it from all relevant $w \times h$ power maps by the function \textproc{getInput}.

\begin{figure}[!t]
  \centering
    \includegraphics[width=.7\textwidth]{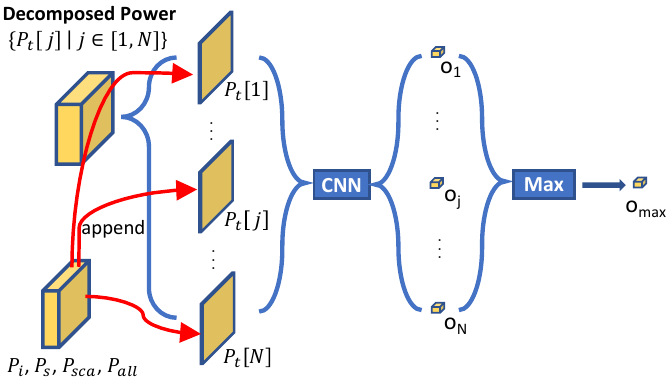}
  \caption{PowerNet structure.}
  \label{PowerNet}
\end{figure}

As shown in Lines \ref{line:l1} to \ref{line:l2} and Figure \ref{PowerNet}, for all $N$ time-decomposed power maps $\{P_{t}[\,j] \in \mathbb{R} ^ {w\times h} \mid j \in [1, N]\}$, they are processed separately by the same CNN model, together with all other common power maps $P_{i}$, $P_{s}$, $P_{sca}$, $P_{all}$. Hence, the input to the CNN is $\{P_{i}$, $P_{s}$, $P_{sca}$, $P_{all}$, $P_{t}[\,j] \}$ in Line \ref{line:l0}. It results in a total of $N$ CNN outputs $\{o_j \mid j \in [1, N]\}$. Then, the maximum output $o_{max} = Max(\{o_j \mid j \in [1, N]\})$ is the prediction result for the analyzed tile. This maximum structure highlights the only instant that leads to the peak IR drop. It guides CNN $f$ to learn such a pattern.

\begin{figure}[!t]
  \centering
    \includegraphics[width=.75\textwidth]{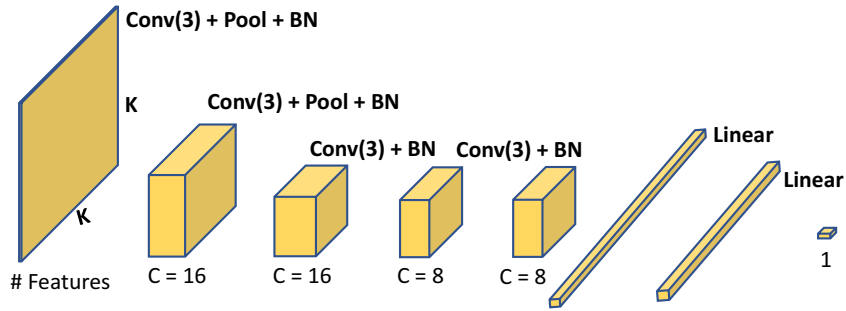}
  \caption{The CNN structure in PowerNet.}
  \label{CNN}
\end{figure}

Details of the CNN model $f$ in PowerNet is shown in Figure \ref{CNN}. There are four convolutional layers, two pooling layers and two fully connected layers. Size of convolution kernels is given in parentheses. The $C$ under each tensor gives the number of kernels defined in each convolutional layer. This CNN structure and hyperparameters like $N$, $k$ are tuned based on the performance during cross-validation. Choosing larger input $k$, more layers or kernels turns out to reduce model generalization and slow down the prediction, while a simpler structure underfits the data. Batch normalization (BN) \cite{batch_norm} is applied to accelerate model convergence. Adam method \cite{Adam} is used for optimization.
We adopt the mean absolute error between prediction and label (L1 loss) as loss function.

\section{Evaluation}
 
\subsection{Experiment Setup}

\begin{table}[!h]
\vspace{-2mm}
  \centering
  \caption{Designs Used in Experiment}
  \vspace{-2mm}
  \label{tbl:t1_ir}
  \begin{tabular}{l | c c c c | c c}
 	\hline
	Design   & D1 & D2 & D3 & D4 & MD1 & MD2   \\
	\hline
    \# cells (million) &  1.7 &  0.81  & 2.0  & 1.9 & 1.7 & 2.4 \\
    Hotspot	Portion    &  5.6\%  & 7.7\%  & 3.1\%  & 3.1\% & 0.65\% & 0.50\%  \\
	\hline
  \end{tabular}
\end{table}

In our experiment, we use six industrial designs in a sub-10nm technology node (Table \ref{tbl:t1_ir}) with an IR drop hotspot threshold of $56mV$, 6\% of the supply voltage ($0.94 V$). Features and IR drop labels are extracted after clock tree synthesis (CTS). Though tested at the CTS stage, PowerNet can also be applied to other stages. We perform vectorless analysis and use results from a commercial IR drop analysis tool as labels. We train the models and measure their accuracy on D1 to D4, then mitigate the IR drop of MD1 and MD2 with the estimation from PowerNet.
When testing estimation accuracy on D1 to D4, the ML model is trained only on data from the other three designs. It ensures that the tested design is totally {\em unseen} to the corresponding model, which eliminates the possibility of information leakage between the testing and training datasets.

We implemented CNN and tree-based XGBoost models similar to \cite{fang2018machine} as a comparison with PowerNet. Similar to \cite{fang2018machine}, two types of features are extracted, named  cell features and  map features. Cell features are one-dimensional and map features are two-dimensional.
For each cell $c$, cell features include its signal arrival time, coordinates, capacitance, unscaled overall cell power $p_{all}$, toggle rate $r_{tog}$ and cell type. The current of each cell is not included because it is not available in our design flow. Since voltages at different regions are all close to the supply voltage, current can be viewed as proportional to power. Then, local maps of both unscaled overall power $p_{all}$ and $r_{tog}$ around it are constructed as its map feature. Notice that compared with PowerNet, only one type of power $p_{all}$ is used. For the tree-based XGBoost model, all map features and cell features are directly used as model input. To fit into XGBoost, the two-dimensional map features are flattened into one dimension. For the CNN model, map features firstly go through three convolutional layers, each with 25, 25 and 50 filters. Then, the output of convolutional layers together with all cell features are fed into three fully connected layers, each with 512 neurons. A 0.4 dropout rate~\cite{srivastava2014dropout} is applied. Other hyper-parameters like optimizer or learning rate of baselines are carefully tuned for their best performance. They are trained and tested on the same designs as PowerNet for model comparison. 
 
All algorithms are implemented in Python. CNN-related models are built on PyTorch 1.0 \cite{paszke2017automatic}. For PowerNet, we set tile size $l=1 \,\si{\micro\meter}$, number of measured instants $N=50$, and an input window size $k=31$ in the experiment. Both training and testing are implemented on an 8-core CPU machine with 100 GB memory and one NVIDIA Tesla V100 GPU.

 \subsection{Accuracy and Speed Comparison}
 
 \begin{figure}[!h]
  \centering
  \centering
    \includegraphics[width=.8\textwidth]{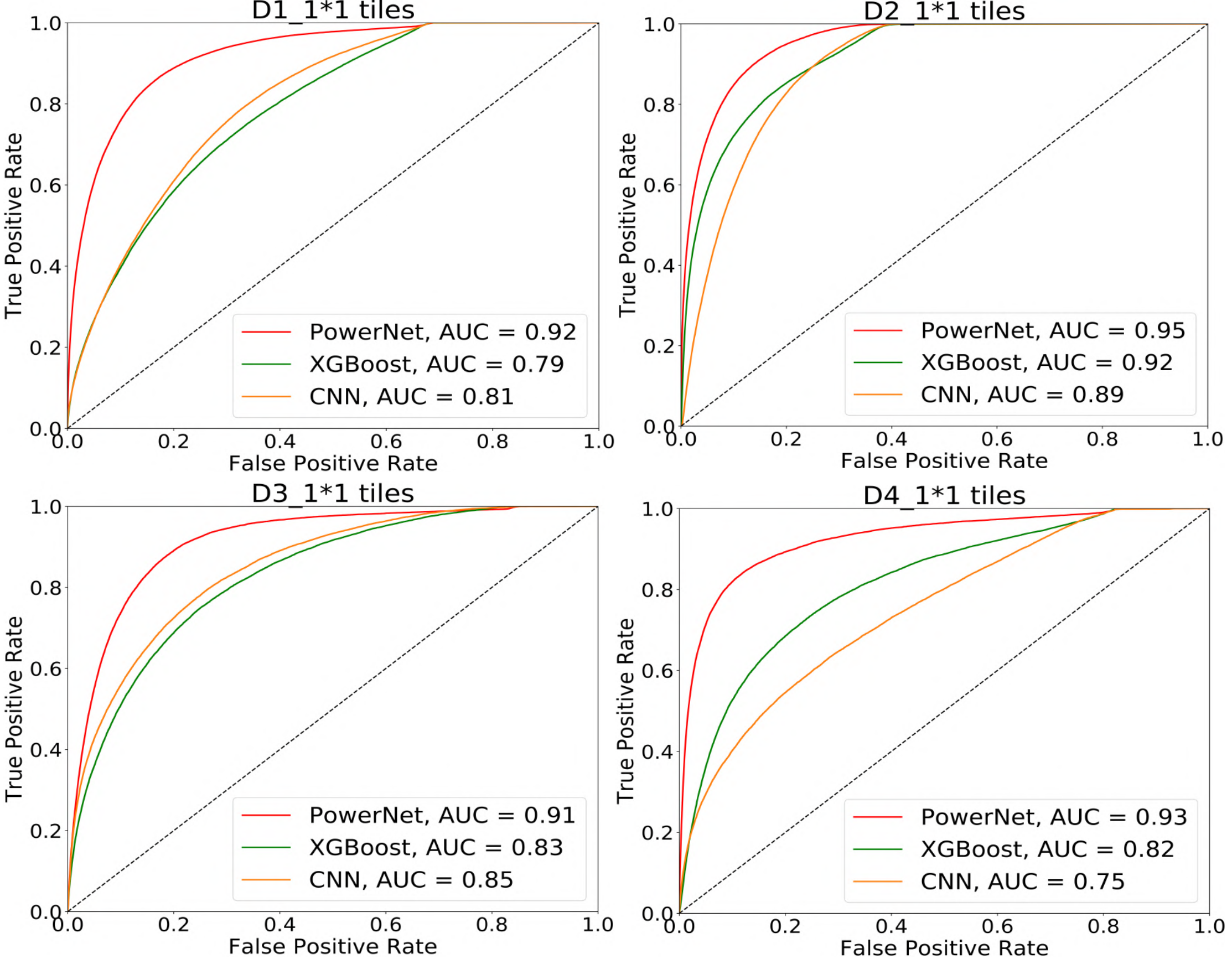}
    \vspace{-1mm}
  \caption{Accuracy comparison by ROC curve. Measured in $1\times 1$ tiles granularity.}
  \label{ROC_1}
\end{figure}

Figures \ref{ROC_1} shows the performance on different designs when measured in $1\times 1$ tiles. PowerNet achieves AUC higher than 0.9 for all designs. On average, the AUC for CNN, XGBoost and PowerNet are around 0.83, 0.84 and 0.93. 
Sometimes hotspots shown in $5l\times5l$ tile already provide sufficient information. In this case, measurements can be performed by tessellating both predictions and labels with a larger tile in $5l \times 5l$. In this case $IR \in \mathbf{R}^{\frac{w}{5}\times \frac{h}{5}}$. 
For $5\times5$ tiles, in our experiment, their AUC are around 0.86, 0.86 and 0.96. PowerNet's improvement in accuracy is $9\%$.

\begin{figure}[!t]
  \centering
    \includegraphics[width=\textwidth]{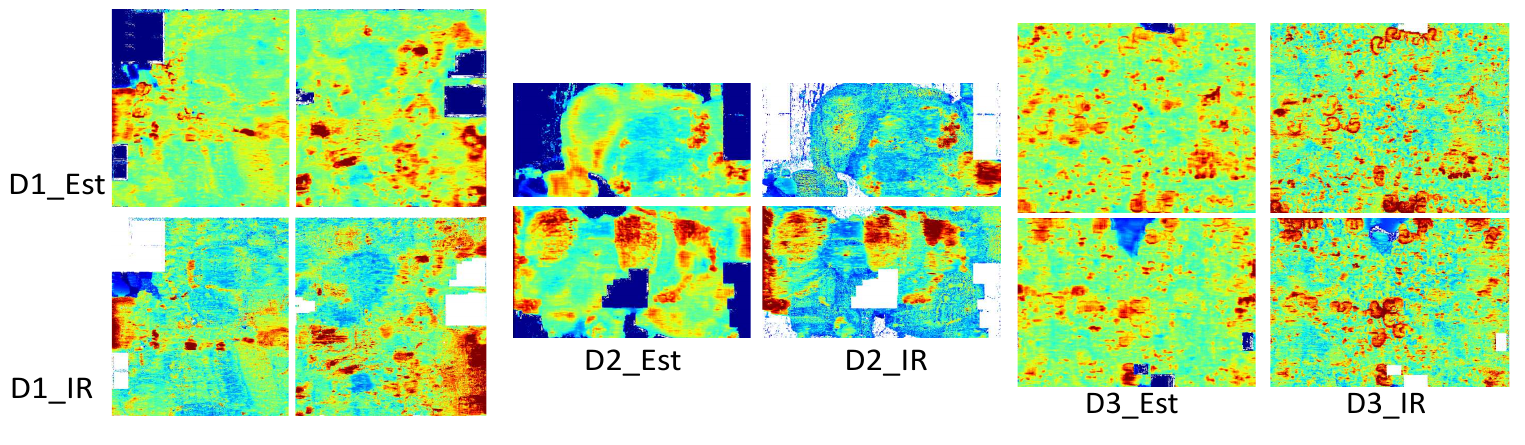}
  \caption{Visualization of IR drop estimation and ground truth.}
  \label{visual-min}
\end{figure}

Figure \ref{visual-min} shows visualizations of both ground truth and the prediction results from PowerNet. Only subsets of each design containing IR drop hotspot regions are displayed. Red color indicates higher values while blue corresponds to lower values and white means zero values. The white blocks in ground truth are usually the regions without any cells placed. The comparison shows that PowerNet can capture most IR drop hotspots.

 \begin{figure}[!tb]
  \centering
    \includegraphics[width=0.8\textwidth]{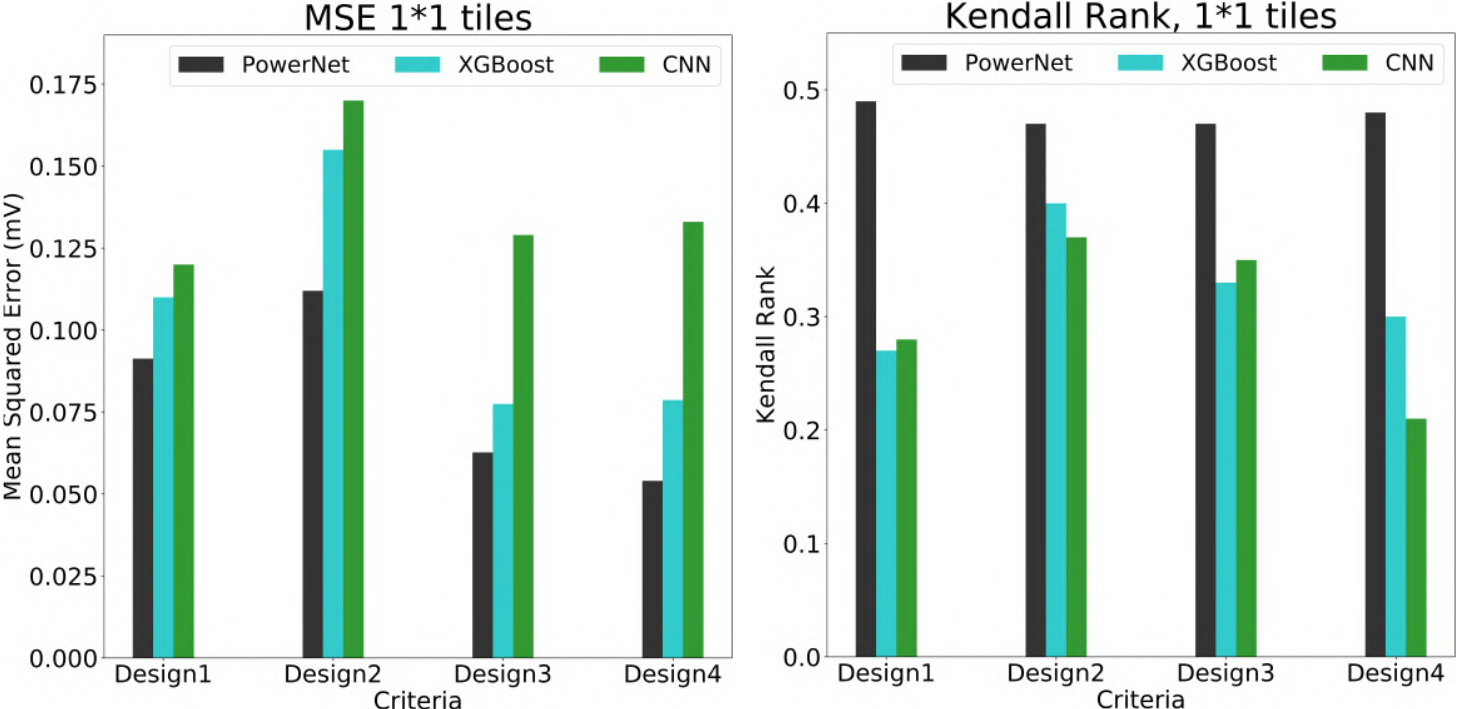}
  \caption{MSE and Kendall ranking coefficient on tiles by IR drop.}
  \label{kendall}
\end{figure}

Besides ROC curves, which reflect how well models recognize hotspots, we also measured how models fit and rank tiles according to their IR drop values in Figure~\ref{kendall}. The metrics are mean squared error (MSE) and Kendall rank coefficient~\cite{kendall1948rank} $\tau \in [-1, 1]$ between the estimation and ground-truth IR values for all tiles. A higher value of $\tau$ implies a more accurate ranking of tiles based on IR drop. The MSE and rank coefficients of PowerNet are consistently better than those of other ML methods. Note that a high MSE may be largely contributed by a consistent bias for all inferenced tiles, which means the model always gives a higher or lower prediction for all tiles in one design. In this case, it can still identify those higher-IR tiles or most serious hotspots if it ranks the IR value of tiles accurately.


The runtimes of the commercial IR drop analysis tool and ML inferences are measured on a design with around two million cells. Results are shown in Table \ref{tbl:time}. PowerNet achieves a $30\times$ speedup over the commercial tool. For a fair comparison, the 2.5 hour for the commercial tool only includes analysis time. Its overall runtime is more than 4 hours. Other ML methods are even faster than PowerNet, but are less accurate. PowerNet is slower than the baseline ML methods because its CNN $f$ generates $N$ outputs $o_j$ for each tile.

\begin{table}[!h]
  \centering
  \caption{Inference Time Comparison.}
  \label{tbl:time}
  \begin{tabular}{c | c c c c c}
 	\hline
Method     &  Commercial Tool   &  PowerNet  &  CNN  & XGBoost \\
	\hline
    Time   & 2.5 hour   &  5 min &  1.5 min  &  1.5 min \\
	\hline
  \end{tabular}
\end{table}

\subsection{IR Drop Mitigation in Design Flow}
We also integrated PowerNet into a design flow to mitigate the IR drop of MD1 and MD2. Based on PowerNet's estimation, we enhanced the local power grid (PG) in hotspot regions. 
Notice that the hotspot portions of MD1 and MD2 are much lower than D1 to D4 in the training set. This is because MD1 and MD2 were already close to tapeout and most serious IR drop problems were already fixed, making further IR drop mitigation even more challenging.

\begin{table}[!h]
  \centering
  \caption{Performance on IR Drop Mitigation}
  \label{tbl:t_ir_mit}
  \begin{tabular}{l | c c c c}
 	\hline
	\multirow{2}{*}{Design MD1} & Violated  & \#  & All     & Hotspot \\
	                            &  Cell     &    Hotspots   & IR (mV)  & IR (mV)\\
	\hline
	Before Mitigate  &  22185  &  5092  &   26.4   &  66.6   \\
	After  Mitigate  &  17052  &  3778  &   26.0   &  62.3   \\
	Improvement      &    23\% &   26\% &   0.4    &   4.3   \\
	\hline
	\multirow{2}{*}{Design MD2}  & Violated  & \#  & All    & Hotspot \\
	                             &  Cell     &    Hotspots   & IR (mV) & IR (mV)  \\
	\hline
	Before Mitigate &  31097  &  3627 &    31.4  &   62.2   \\
	After  Mitigate &  23941  &  2489 &    31.0  &   59.6   \\
	Improvement     &    23\% &  31\% &    0.4   &    2.6   \\
	\hline
  \end{tabular}
\end{table}

Table \ref{tbl:t_ir_mit} shows the IR drop mitigation result. We only add very thin PG straps (0.04 \si{\micro\meter}) at the PowerNet-estimated hotspots. This is the simplest and most basic fixing method. We choose such conservative fixing method to prevent occupying too many routing resources. ``All IR'' and ``Hotspot IR'' mean the averaged IR drop values among all tiles and all hotspots. After PG enhancement, the averaged IR drop for all tiles improves only 0.4 \si{\milli\volt}, which indicates that the modification on PG is very small. In comparison, when measured only on hotspots, IR drop improves 4.3 \si{\milli\volt} and 2.6 \si{\milli\volt}. It shows that PG enhancement is effective at the right places. With such a limited amount of modification in PG, 23\% of IR drop violation cells or around 30\% of hotspots are mitigated.

\subsection{Why PowerNet Performs Better}

We highlight four weaknesses of previous CNN and XGBoost baseline models that prevent them from outperforming PowerNet. First, unnecessary features can confuse ML models. If cell coordinates and time information are used as features but do not directly correlate with IR drop, a model can overfit to designs in the training set. Other features such as cell capacitance can be redundant when power is already provided. To verify this, we implemented an XGBoost model without cell coordinates or time information, and its averaged $1\times 1$ tiles AUC improved from 0.84 to 0.865. When we further removed cell capacitance from features, the averaged AUC remained at 0.865.
Second, different feature formats make the model inefficient. Notice that cell features are one dimensional but map features are two dimensional. For XGBoost, map features must be converted into one dimension, which loses spatial information. For CNN, cell features must be provided through a fully connected layer. In such an unusual CNN structure, cell features tend to be overwhelmed by more than 10,000 outputs from the 50 channels in the last convolutional layer. In comparison, PowerNet only uses two-dimensional features. 
Third, power information may not be fully utilized. When only overall power $P_{all}$ is chosen as a feature, the rich information from other power types $P_i, P_s, P_{sca}$ is lost. Advanced ML models like CNN are complex enough to learn patterns from different power types.
Fourth, time information is not well incorporated or captured. Baselines do not have features like the time-decomposed power maps in PowerNet to measure the worst transient local IR drop.

\begin{figure}[!h]
  \centering
    \includegraphics[width=0.7\textwidth]{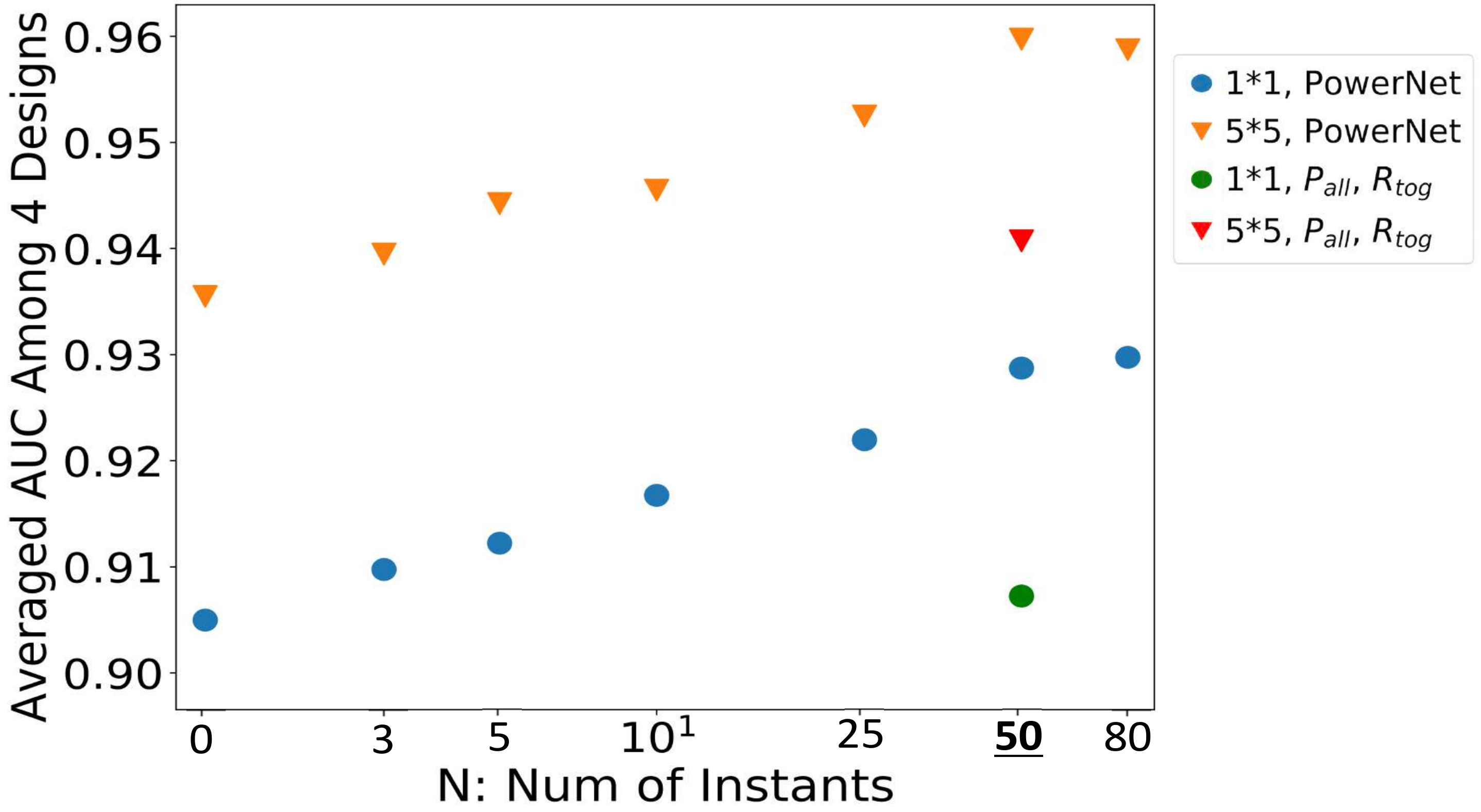}
  \caption{Effect of number of instants $N$ on performance.}
  \label{num_inst}
  \vspace{-1mm}
\end{figure}

Figure \ref{num_inst} isolates the contribution of including both time decomposition and multiple power types in variations of PowerNet. 
Average inference AUC accuracy over D1 to D4 is plotted on the Y-axis and the X-axis shows the number of sampled time instants. A higher $N$ means sampling more time instants and generating more corresponding power maps $P_t[\,j]$ within a given clock period $T$. For any region, more power maps better approximate its actual transient power.
The ``N=0'' indicates no time-decomposed power is adopted at all.
As expected, the time-decomposed power maps improve accuracy by capturing transient IR drop. When $N>50$, the improvement in accuracy by increasing $N$ diminishes.
Baseline models also differ from PowerNet by only using maps of features ($P_{all}$ and $r_{tog}$) instead of $\{P_{i}$, $P_s$, $P_{sca}$, $P_{all}\}$.  This variation is indicated as red and green marks in Figure \ref{num_inst}, where time-decomposed power maps $P_t[\,j]$ are kept the same for both variations. In addition to the $2.5\%$ accuracy improvement from time decomposition, adopting multiple types of power improves accuracy by more than $2\%$.

\begin{figure}[!t]
  \centering
    \includegraphics[width=0.83\textwidth]{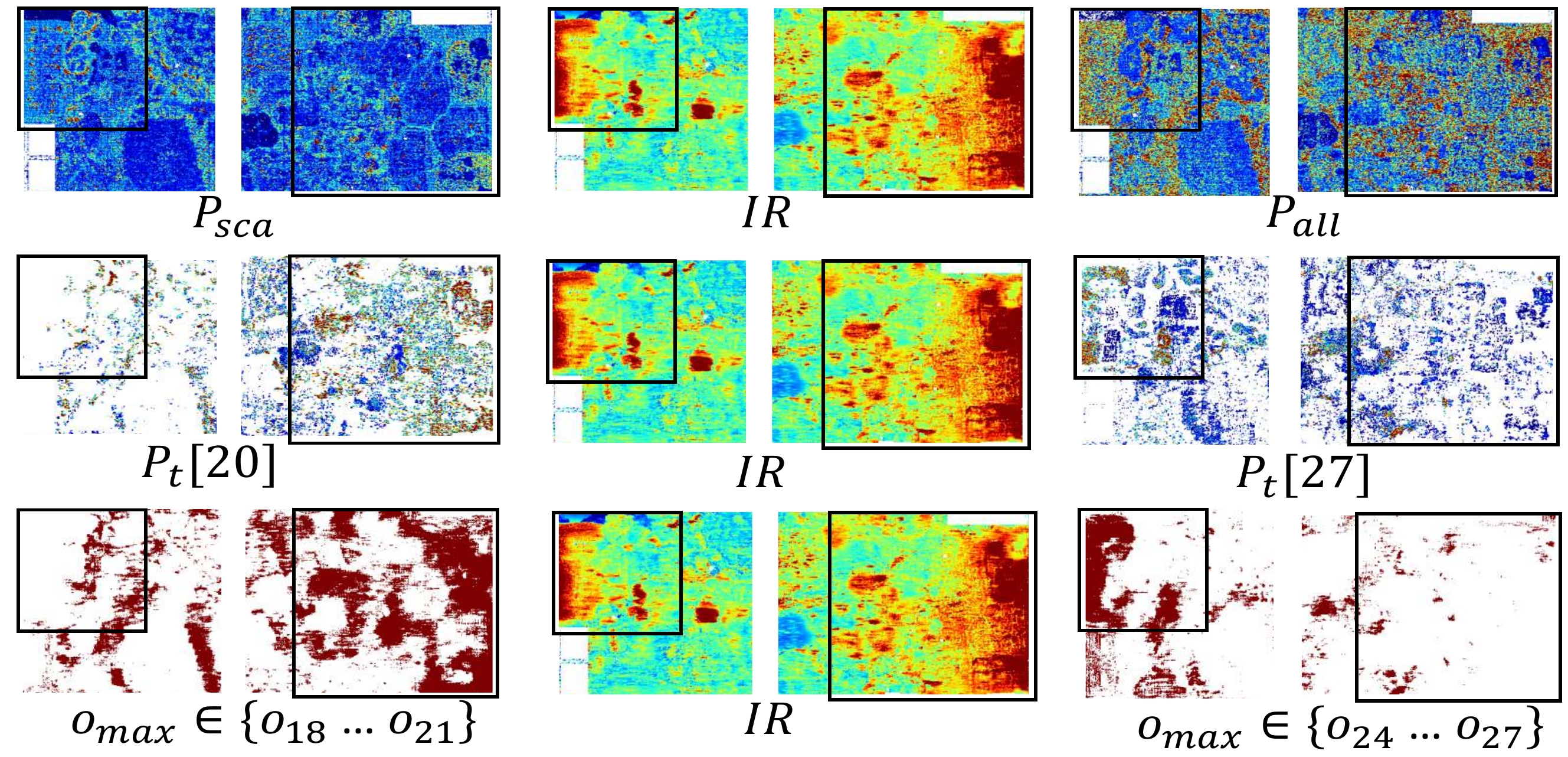}
    \vspace{-1mm}
  \caption{IR drop, power maps and maximum instant distribution of two regions from D1. Instants number $N=50$.}
  \vspace{4mm}
  \label{features}
    \includegraphics[width=0.83\textwidth]{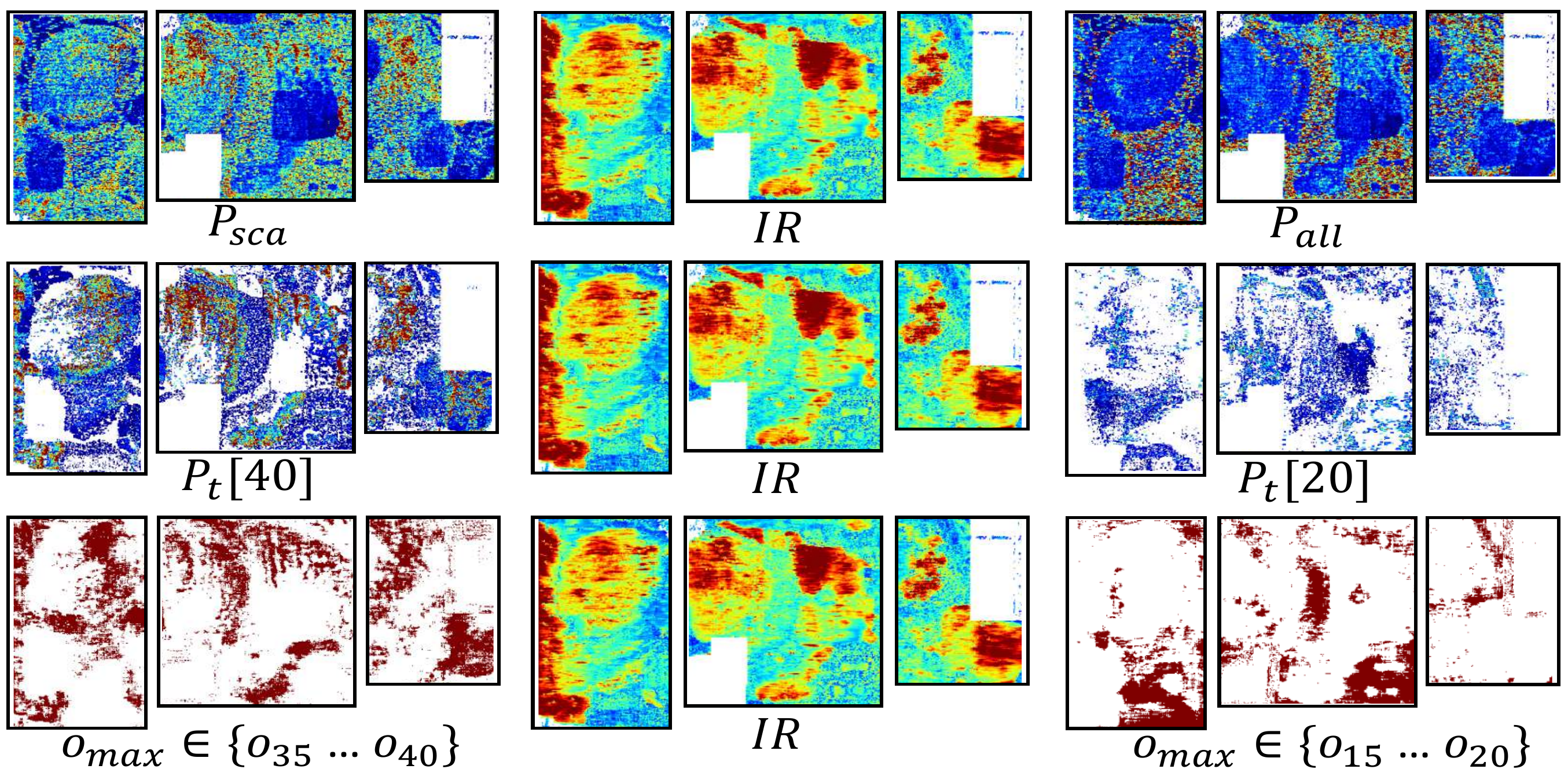}
    \vspace{-1mm}
  \caption{IR drop, power maps and maximum instant distribution of three regions from D2. Instants number $N=50$.}
  \label{features2}
\end{figure}

Figures \ref{features} and \ref{features2} show how the combination of space decomposition and time decomposition helps to explore the potential correlation between power maps and IR drop. It presents the visualization of $IR$, different power maps and maximum instant distribution for the local regions from both D1 and D2. Maximum instant refers to the time instant $j$ selected by maximum structure ($o_j = o_{max}$). The areas of interest are highlighted by black squares. They all contain strong IR drop hotspots. For both D1 and D2, it's difficult to observe much correlation in hotspots between their $\{P_{all}$, $P_{sca}\}$ and $IR$. That indicates training models without any time-decomposed power maps $P_t[\,j]$ can be difficult.

However, the correlation becomes much more clear when power maps $P_t[\,j]$ are provided. In D2, $P_t[40]$ and $IR$ share many common hotspots patterns in highlighted areas. In this case, $o_{40} = f(\{P_{i}$, $P_{s}$ $P_{sca}$, $P_{all}$, $P_{t}[40] \})$ is likely to accurately predict these common hotspot regions. However, another power map, $P_t[20]$, does not share as much hotspot patterns with $IR$. Its output $o_{20}$ may be less accurate. Considering that $P_t[20]$ is much weaker than $P_t[40]$ for most tiles in hotspot regions, we can reasonably assume $o_{40} > o_{20}$, or even $o_{max} = o_{40}$. This is verified by the maximum instant distribution. For every tile, we checked which instant is selected by the maximum structure. For almost all tiles at hotspot regions in D2, their $o_{max} \in \{o_{35}... o_{40}\}$. $P_t[40]$ indeed contributes more information than $P_t[20]$. It is the contribution of the more accurate $P_t[40]$ instead of $P_t[20]$ at these hotspot regions that finally gets captured by the maximum structure. 

Similarly, for D1, the highlighted region on the right correlates well with $P_t[20]$, and region on the left correlates with $P_t[27]$. 
Maximum instant distribution shows that $o_{max} \in \{o_{18}... o_{21}\}$ for most grids in the right region and $o_{max} \in \{o_{24}... o_{27}\}$ for most tiles in the left region.
Then the maximum structure will take $o_t[20]$ for grids on the right and $o_t[27]$ for tiles on the left. In this way, the hotspots caused by transient power at different instants can all be captured.

\subsection{Results Considering Other Factors}
\label{resistance}
\begin{table}[!b]
  \centering
  \caption{Inference Accuracy in ROC AUC (0.01*)}
  \label{tbl:resist}
  \begin{tabular}{l | c c c c | c}
 	\hline
     ML Methods  & D1   &  D2 &  D3 &  D4 & Ave  \\        
	\hline
	PowerNet ($1\times1$ tiles)  &  92.1  &  95.4 &  91.4 &  92.6 & 92.9  \\ 
	PRNet\;\;\;\;\;($1\times1$ tiles)  &  92.4  &  95.5 &  90.5 &  93.6 & 93.0  \\
	\hline
	PowerNet ($5\times5$ tiles)  &  95.4  &  96.7 &  94.8 &  97.0 & 96.0  \\ 
	PRNet\;\;\;\;\;($5\times5$ tiles)  &  95.7  &  96.8 &  93.2 &  97.5 & 95.8  \\
	\hline
  \end{tabular}
\end{table}

We measured the distribution of resistance in our benchmark design. Take D1 for example, the standard deviation in resistance across the whole design is only 2.8\si{\ohm}, $0.6\%$ of its average resistance. For such a uniform distribution, we chose not to spend extra time calculating per-cell resistance.  However, we did implement another variation of PowerNet where each cell's power is scaled with resistance, denoted as ``PRNet'' in Table \ref{tbl:resist}. ``Ave'' means accuracy averaged over all four designs. On average, the resistance-scaled PowerNet shows similar accuracy to the original one. This demonstrates that using per-cell resistance as a feature is not necessary for designs with uniform PDNs. By scaling power with resistance, ``PRNet'' can be further applied to designs with non-uniform PDNs.

\label{vector}

\begin{figure}[!t]
  \centering
    \includegraphics[width=0.85\textwidth]{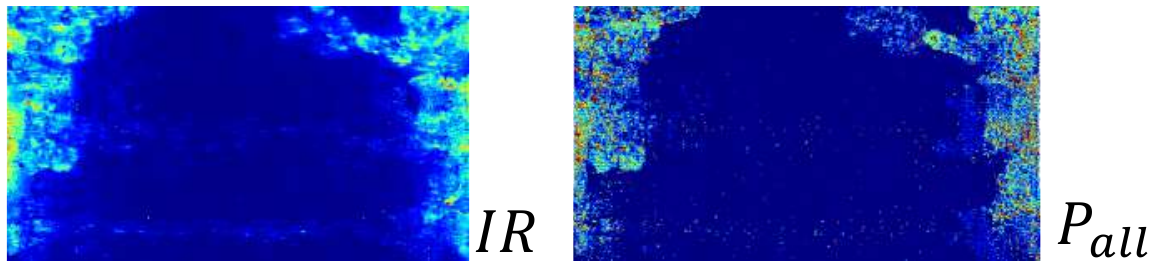}
  \caption{Power and IR drop of vector-based estimation.}
  \label{vector-min}
\end{figure}

We also measured PowerNet's performance on vector-based IR drop. 
The PowerNet architecture remains exactly the same, but cell power and IR drop are now collected when the commercial tool simulates IR drop with given simulation patterns.
Figure \ref{vector-min} shows 
the vector-based power map $P_{all}$ and label $IR$. Unlike the vectorless case in Figure \ref{features} or \ref{features2}, the power of a portion of activated cells is significantly higher than the others. As we mentioned, the correlation between power and IR drop value turns out to be very strong, which largely reduced the estimation difficulty.

We perform vector-based estimation on four other industrial designs VD1 to VD4. All models and procedures are the same as the vectorless case, except using cell power and IR drop from vector-based simulation. Table \ref{tbl:vector} shows vector-based estimation accuracy. As expected, all methods provide better estimation than vectorless scenario. But PowerNet still gives the best accuracy for every single design. The $1\%$ to $2\%$ improvement should not be underestimated when accuracy is already very high. To a certain extent, boosting accuracy from $98\%$ to $99\%$ means reducing half of the errors.

\begin{table}[!h]
  \centering
  \caption{Vector-Based Inference in ROC AUC (0.01*)}
  \label{tbl:vector}
  \begin{tabular}{l | c c c c | c c c c}
 	\hline
 	\multirow{2}{*}{ML Methods}   &   \multicolumn{4}{c|}{$1\times1$ tiles} &                                        \multicolumn{4}{c}{$5\times5$ tiles}    \\
	                  &  VD1  &  VD2 &  VD3 &  VD4 &   VD1 &  VD2 &   VD3 &  VD4  \\  
	\hline
	XGBoost           &  97  &  98 &  98 &  96 &   99 &  97 &   98 & 97 \\
	CNN               &  96  &  93 &  95 &  95 &   98 &  92 &   97 & 96  \\
	PowerNet          &  98  &  98 &  99 &  97 &  100 &  98 &  100 & 98  \\ 
	\hline
  \end{tabular}
\end{table}

\section{Summary}

In this chapter, we present a CNN-based dynamic IR drop estimator. Unlike existing ML works, our model is general and transferable to new designs. 
We validate the high accuracy of our approach on multiple industrial designs. It takes an order of magnitude less estimation time than commercial tools and significantly outperforms state-of-the-art ML methods in both vector-based and vectorless IR drop scenarios. 
The IR drop mitigation tool guided by our model reduces IR drop by more than 20\% with very limited PG modification.

%% file: txt/5_routability.tex
\chapter{Early Routability Modeling on Layout}
\label{chapter:routability}

\section{Background}


 

Design rule checking (DRC) verifies whether a specific layout meets the constraints imposed by the technology node for manufacturing. 
DRC violations can only be precisely measured after routing finishes, when the room for fixing DRC violations (DRVs) has become very limited. Besides mitigating DRVs manually, one option is engineering changing order (ECO), which tries to complete unrouted and partially routed nets while maintaining existing wires as much as possible. Another alternative is to delete part of existing wires and reroute them. But it is difficult for these minor post-routing modifications to fix all violations for layouts with poor routability. As a result, designers have to trace back to earlier stages, change their layout solution accordingly, and start a new design iteration. It can take many iterations to reach a DRV-clean layout, leading to a very long turnaround time.

To improve layout routability, a much better solution is to avoid DRVs with preventive measures in a proactive manner. This heavily relies on early routability prediction, which is difficult since the behavior of placement and routing engines in modern EDA tools is highly complex and rather unpredictable. One possible solution is to develop some fast trial routing algorithms, but it is hard to achieve ideal accuracy and speed at the same time. 
Another promising research direction nowadays is to learn from prior data by developing data-driven routability estimators with machine learning (ML) algorithms. A main strength of ML methods is the automatic extraction of complex correlations between separated design steps based on prior knowledge. Once the ML model has been trained, it can produce routability predictions in a very short time, without constructing solutions from scratch.

For routability prediction, there are different prediction granularities according to application scenarios. Some coarse-grained predictions only evaluate the overall routability of the whole layout. Such routability is usually measured with the total number of DRC violations, also named DRV count (\#DRV). Another similar metric is the total number of nets with DRC violations, also named violated net count. Such a coarse-grained prediction evaluates the whole layout and enables the identification of more routable layout solutions among many candidates. In comparison, fine-grained routability prediction tries to pinpoint the detailed locations with DRC violations. This supports modifying the layout at early stages to proactively prevent DRC violations.

Data-driven routability estimators are initially constructed with traditional ML models, including support vector machine (SVM)-based estimator~\cite{chan2016beol, chan2017routability}, multivariate adaptive regression spline (MARS)-based estimator~\cite{chan2016beol, zhou2015accurate}, and artificial neural network (ANN)-based estimator~\cite{tabrizi2019eh, tabrizi2018machine}. These ML models typically only process a limited number of input features. For fine-grained routability prediction on DRV locations, traditional ML methods are applied to make decisions based on a small cropped region with limited features from the layout. Such a small input region strongly limits the \textit{receptive field} (the field visible to the model) of these traditional ML methods. As a result, they cannot capture the \emph{global information} from a larger region of the layout. For example, the nets spanning a large region can largely affect the routability, thus should be captured by the model. 

\begin{figure}[!t]
  \centering
    \includegraphics[width=.77\textwidth]{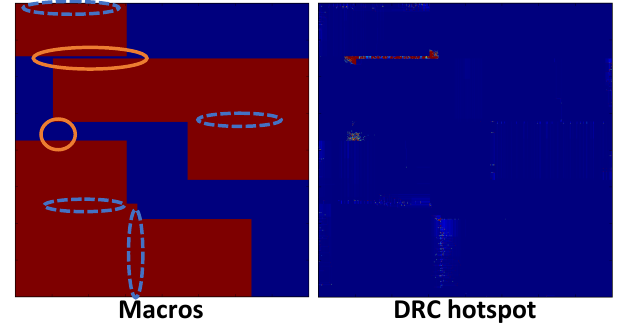}
  \caption{Macros and DRC hotspot distribution. All macros are red rectangles. Orange circles indicate regions with high density of DRC hotspots.} 
  \label{figure:macro}
\end{figure}

Another important global impact is from macros in mixed-sized designs. This is illustrated in Figure~\ref{figure:macro}, showing the distribution of DRC hotspots on a benchmark circuit with macros. The orange circles in Figure \ref{figure:macro} indicate a strong tendency for hotspots to aggregate at the small gap between neighboring macros. The remaining small number of hotspots, indicated by blue dashed circles, also locate sparsely around the edges of some macros. To detect DRC hotspots more precisely, such distribution pattern has to be captured, which requires global information about neighboring macros. In summary, a layout with macros is much less homogeneous than that without macros. Such homogeneity implies resemblance among different regions of a chip layout. Thus mixed-size designs with macros must have a much larger region, often entire layout, as a single training sample, in order for the routability estimator to capture the global view.




\section{Methodology}

\subsection{Problem Formulation}

We aim at solving two problems: 1. Early forecast of overall routability; 2. The prediction of DRC hotspot locations. In problem 1, the routability of a placement is evaluated by its DRV count denoted as \#DRV. The task is to fit a function $f_{\#DRV}$ that estimates ground-truth DRV count of a placement. In applications, $f_{\#DRV}$ is employed to select a few placements with relatively low \#DRV from many candidate placement solutions. In problem 2, DRC hotspots mean the specific locations with high density of DRVs. The task is to find a function $f_{hotspot}$ that detects most DRC hotspots in a placement. RouteNet solves problem 1 without performing any routing while its solution to problem 2 uses global routing as a feature. Since problem 2 is performed on a few relatively routable designs, its runtime constraint is less tight than problem 1.
In both tasks, the informative features about placements are input and DRV information is the prediction target. The ground-truth DRV information is also referred to as label.

To evaluate routability in terms of design rule violations, a layout (or placement solution) is tessellated into an array of grid cells, each of which is an $l\times l$ square. Then, a rectangular layout with size $W\times H$ is divided into $w\times h$ grid cells, where $w = W/l$ and $h=H/l$. In our experiments, $l$ is set to be the height of standard cells.

After design rule checking, the location and area of all DRC violations are reported. The overall DRV count for the $i$th placement solution is recorded as $y_i \in \mathbb{N}$. The density of DRV is calculated at a grid-level granularity. 
The DRV density in each grid cell is a summation of contributions from all violations covering this grid cell. As a result, for the $i$th placement with $w\times h$ grid cells, its DRV density is a two-dimensional matrix $Y_i \in \mathbb{R} ^ {w\times h}$. 
When the density of violations in a grid cell is higher than a threshold $\epsilon$, this grid cell is labeled as a DRC hotspot. 
Existence of DRC hotspots is a Boolean matrix $V_i \in \{0, 1\} ^ {w\times h}$, where $V_{i_{mn}} = \mathbbm{1} (Y_{i_{mn}} > \epsilon)$ for grid cell $(m, n)$.

\begin{figure}[t]
  \centering
    \includegraphics[width=\textwidth]{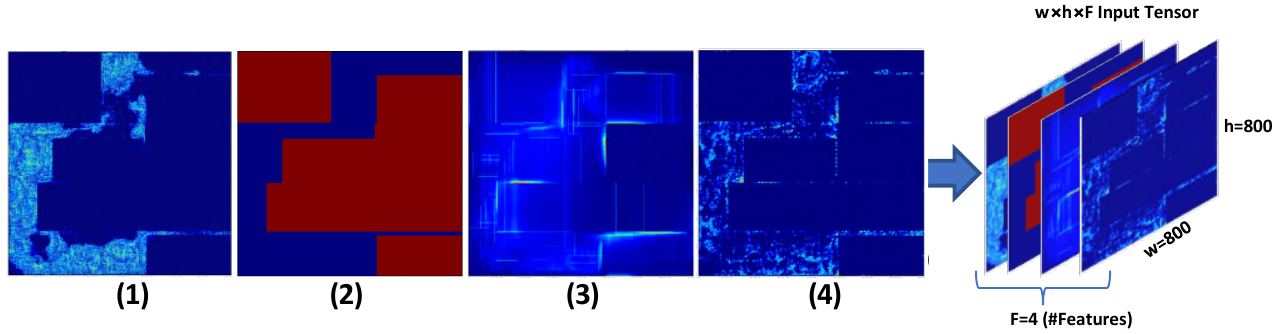}
  \caption{A 3D input tensor constructed by stacking 2D features, including (1) pin density, (2) macro region, (3) long-range RUDY, (4) RUDY pins.}
  \label{inputTensor}
\end{figure}

Similar to DRV density, densities of different informative features are calculated as the input of RouteNet. The $j$th feature of $i$th placement is $X_{ij} \in \mathbb{R} ^ {w\times h}$. If $F$ different features are generated, the input for the $i$th placement is $X_i \in \mathbb{R} ^ {w\times h\times F}$. Such $X_i$ is constructed by stacking all two-dimensional features $ \{ X_{ij} \mid j \in [1, F] \}$ together in a third dimension as shown in Figure~\ref{inputTensor}. Inputs for the two problems are not exactly the same. For example, since \#DRV prediction starts early, global routing information is not included its input $X_i^{\#DRV}$.
The two problems are formally stated as follows.

\noindent\textbf{Problem 1 (\#DRV prediction).} Find an estimator $f^{*}_{\#DRV}$ of DRV count in a placement:
\begin{align*}
& f_{\#DRV}: X_i^{(\#DRV)} \in \mathbb{R} ^ {w\times h \times F_1} \rightarrow y_i \in \mathbb{N} \\
& f_{\#DRV}^{*} = \argmin_{f} Loss(f (X_i^{(\#DRV)}), \: y_i) 
\end{align*}

\noindent\textbf{Problem 2 (Hotspot prediction).} Find a detector $f^{*}_{hotspot}$ of hotspots. It reports locations of all DRC hotspots in a placement:  

\begin{align*}
& f_{hotspot}: X_i^{(hotspot)} \in \mathbb{R} ^ {w\times h\times F_2} \rightarrow V_i \in \{0, 1\} ^ {w\times h} \\
& f_{hotspot}^{*} = \argmin_{f} Loss(f (X_i^{(hotspot)}), \: V_i) 
\end{align*}

\subsection{Feature Extraction}

\begin{figure}[t]
  \centering
  \includegraphics[width=0.7\textwidth]{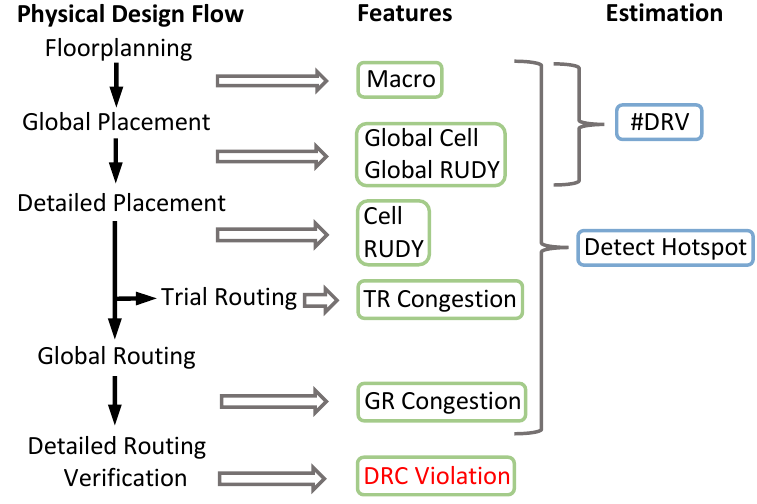}
  \caption{Feature extraction in physical design flow.}
  \label{flow}
\end{figure}

Figure \ref{flow} shows the input features extracted at different layout stages. One important feature we choose adopt is RUDY (Rectangular Uniform wire DensitY). It is employed as an input feature to our RouteNet as it partially correlates with routing congestion, is fast to obtain and can be directly represented as images that dovetail with RouteNet.

Given a cell placement, RUDY of a net is obtained by uniformly spreading the wire volume of this net into its bounding box.
For the $k$th net with bounding box $\{x_{min}^k,\, x_{max}^k,\, y_{min}^k,\, y_{max}^k\}$, its RUDY at location ($x$, $y$) is
defined as

\begin{gather*}
w^{k} = x_{max}^k - x_{min}^k, ~~~~~
h^{k} = y_{max}^k - y_{min}^k \\
c^k  = \begin{cases} 1     & x \in [x_{min}^k, x_{max}^k] \; , \; y \in [y_{min}^k, y_{max}^k]\\  0  & otherwise\\ \end{cases} \\
RUDY^k(x, y) \propto  c^k\;\frac{w^{k} + h^{k}}{w^{k} \times h^{k}}
\end{gather*}

RUDYs of all $K$ nets are calculated and superimposed on top of each other to provide a rough estimation of routing congestion as

\begin{equation*}
RUDY(x,y) = \sum_{k=1}^{K} RUDY^k(x, y)
\end{equation*}

Here is a detailed description of all features in Figure \ref{flow}:

\begin{itemize}
\item \textbf{Macro}: After floorplanning, the locations of all macros are fixed. Several features about macro are extracted: 1) The region occupied by macros. 2) The density distribution of macro pins in each metal layer.

\item \textbf{Global cell \& global RUDY}: After global placement, cell locations become available and then RUDY is calculated. 
Features calculated at this stage are denoted by {\em global cell} and {\em global RUDY}. After detailed placement, the information from global cell and global RUDY are recalculated by refined cell locations, denoted as {\em cell} and {\em RUDY}.

\item \textbf{Cell}: For both global cell and cell, two features are extracted: 1) The density distribution of cells. 2) The density distribution of cell pins.

\item \textbf{RUDY}: For both global RUDY and RUDY, three features are calculated: 1) Long-range RUDY. 2) Short-range RUDY. 3) RUDY pins.
  
  The original RUDY feature is decomposed into long-range and short-range RUDY. Long-range RUDY is from nets covering a distance longer than a threshold. 
  Similarly, short-range RUDY is for nets shorter than this threshold. Such decomposition is due to a stronger correlation between long-range RUDY and DRV than the shorter one. 
  To further capture such effect, RouteNet uses another feature named {\em RUDY pins}. 
  It is similar to pin density, while the contribution of each pin equals the long-range RUDY of the net it connects to.

\item \textbf{Congestion}: Two types of congestion are generated by placement and routing tool. The trial global routing, also denoted as trial routing, can be performed at the end of detailed placement. It produces an estimation of routing congestion, named {\em TR congestion}. Compared with it, the full-fledged global routing generates a more detailed congestion map, named {\em GR congestion}.

\item \textbf{DRC violation}: This is the prediction target and the label for training RouteNet. Its density is calculated in the same way as all other features. Most DRC violations only occupy very small regions.
\end{itemize}

For the $i$th placement, all above $F$ features $X_{ij} \in \mathbb{R} ^ {w\times h}$ are calculated independently and then combined together as one input tensor $X_i \in \mathbb{R} ^{w\times h\times F}$. 

\subsection{\#DRV Prediction}

\begin{algorithm}[!bth]
\caption{Algorithm of
RouteNet for
\#DRV Prediction}
\label{alg6}
\textbf{Input}: Number of training placements: $N$; Features: $ \{X_i \in \mathbb{R} ^ {w \times h\times 3} \mid i \in [1,N] \}$; \\\hspace*{14mm}Targets: $ \{y_i \in \mathbb{R} \mid i \in [1,N] \}$.\\
\textbf{Preprocess}: 
\begin{algorithmic}[1]
\For{each int $i \in [1,N]$ }
\State  Resize $ X_i \in \mathbb{R} ^ {w\times h \times 3}$ into $X_i^{\#DRV} \in \mathbb{R} ^ {224 \times 224 \times3}$
\EndFor
\State Find 25\%, 50\%, 75\% quantizes of $y_i$: $q_1$, $q_2$, $q_3$
\For{each int $i \in [1,N]$ }
\State $C_i \gets 0$
\For{each int $t \in [1,3]$ }
\If {$y_i > q_t$}
\State $C_i \gets t$, \textbf{break}
\EndIf
\EndFor
\EndFor
\State Form dataset $\{(X_i^{\#DRV},\: C_i) | i \in [1, N] \}$
\State Training set $\{(X_i^{\#DRV},\: C_i)\mid C_i =0 \:or \: C_i =3\}$ 
\end{algorithmic}
\textbf{Training}:
\begin{algorithmic}[1]
\State Get pretrained ResNet18 $f_{Res}: \mathbb{R} ^ {224 \times 224 \times3} \rightarrow \mathbb{R}^{1000}$
\State Replace output layer, s.t. $f_{\#DRV}: \mathbb{R} ^ {224\times224 \times3} \rightarrow \mathbb{R}$
\State Choose MSE as loss function, SGD for optimization 
\State Train $f_{\#DRV}$ with preprocessed dataset for $\sim$30 epoches
\end{algorithmic}
\textbf{Output}: $f_{\#DRV}$ estimating \#DRV level
\end{algorithm}

As an early routability forecast, the \#DRV prediction by RouteNet is performed before detailed placement starts. Details are shown in Algorithm \ref{alg6}. In order to convert \#DRV prediction to an image classification task, both input features and prediction target are preprocessed.

The range of \#DRV is broad for different placements of the same design.
At this early stage, two placements with a slight difference in \#DRV may not have substantially different patterns in their features, especially when routing information is absent.
As a result, RouteNet can wrongly capture such minor difference if the exact \#DRV is used as label.
To avoid this, we group placements into four \#DRV levels,
referred to as $c_0$, $c_1$, $c_2$, $c_3$, respectively, where $c_0$ corresponds to the class of placements with the least \#DRV and $c_3$ corresponds to class with the most \#DRV.

\begin{figure}[t]
  \centering
  \vspace{-1mm}
    \includegraphics[width=\textwidth]{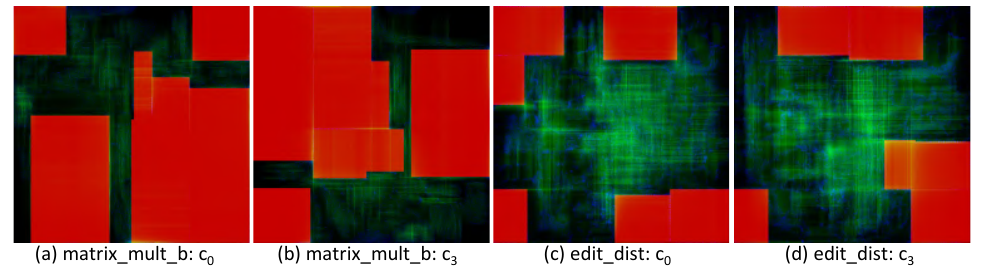}
  \caption{Input features for \#DRV prediction. Red: macro region; Green: global long-range RUDY; Blue: global RUDY pins.}
  \label{preImage}
\end{figure}

For all CNN models pretrained on dataset ImageNet, the dimension of input image $X_i$ is fixed to be $224\times 224\times 3$. This means the input images have $224\times 224$ pixels and $3$ channels (RGB). To utilize such pretrained model, the original input tensor $X_i \in \mathbb{R} ^ {w\times h\times F}$ needs to be resized into $X_i^{\#DRV} \in \mathbb{R} ^ {224\times 224 \times 3}$. To accomplish this, three features (macro, global long-range RUDY, global RUDY pins) are firstly selected to construct 3-channel input tensor in $\mathbb{R}^{w\times h \times 3}$. We choose them because they intuitively contain more global and general information than feature like pin density. After that, the 3-channel input is resized into $\mathbb{R} ^{224\times 224\times 3}$ by interpolation. Figure \ref{preImage} shows visualizations of preprocessed input, (a)(b) and (c)(d) are different placements with different levels of \#DRV for two benchmark circuits.

After the preprocessing step, transfer learning is applied to a pretrained 18-layer ResNet. The output layer is replaced to produce a single score. During the fine-tuning process, the weights in every layer are changeable.  Mean Square Error (MSE) is used as loss function and Stochastic Gradient Descent (SGD) is used for optimization. 

The dataset is randomly split into the training set and the validation set. For the training set, all data in classes $c_1$ and $c_2$ are removed, only classes $c_0$ and $c_3$ are kept. That is, only placements in highest \#DRV or lowest \#DRV levels will be used for training. Such removal proves to give better results than keeping all four classes. 


\subsection{DRC Hotspot Detection}

Different from \#DRV prediction, hotspot detection is more like an object detection task. The output can no longer be a simple score value or vector. Instead, we make it a two-dimensional density map, directly reflecting the existence of all hotspots in a placement. In this case, the size of output equals circuit size $f_{hotspot}(X_i) \in \mathbb{R} ^{w\times h}$. FCN enables such function format and accepts input with different $w$, $h$. As a result, different designs can be used for training and inference on exactly the same model.

\begin{figure}[t]
  \centering
    \includegraphics[width=0.8\textwidth]{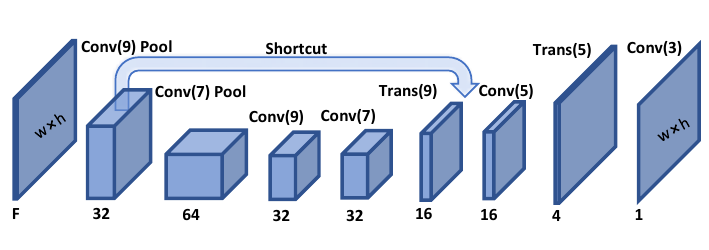}
  \caption{FCN architecture for hotspot detection.}
  \label{archi}
\end{figure}

Figure \ref{archi} shows the FCN architecture. It accepts input tensor with size $\mathbb{R} ^ {w\times h\times F}$ and produces a two-dimensional $\mathbb{R} ^ {w\times h}$ output. In this structure, a shortcut directly connects the \nth{2} layer to the \nth{7} layer, providing a shorter path from input to output. Two POOL layers downsize feature maps from $h\times w$ to $\frac{h}{4}\times \frac{w}{4}$ in the front, then two TRANS layers upsample the size back to $h \times w$.   Strides of  kernels in CONV  and TRANS layers  are set to 1 and 2, respectively. The kernel sizes are indicated by the number in parentheses, ranging from 3 to 9 grid cells.  Such pooling structure and large kernel size 
substantially enlarge the regions receptive to each grid cell in output. Compared with previous methods 
capturing features within a small region, more neighboring or global information is available for FCN and the result function $f_{hotspot}$ will be more complex.

During the training process, the DRV density $Y_i$ is used as label. DRV density is clipped by a threshold $c$ in Equation~\eqref{clip} to reduce the dominating effect of a few grid cells with very high DRV density. 
Batch normalization \cite{batch_norm} is applied to accelerate convergence in training. The Adam method~\cite{Adam} is used for optimization. 
The loss function is defined by a summation of pixel-wise Euclidean distance and L2 regularization in Equation \eqref{loss_function},

\begin{align}
Y_{i_{mn}}^{clip} &= min (Y_{i_{mn}}, c) \label{clip}\\
Loss =  \sum\limits^{N}_{i=1}\sum\limits^{w}_{m=1}\sum\limits^h_{n=1} ||f&_{hotspot}(X_{i_{mn}}) - Y_{i_{mn}}^{clip} ||_2 + \lambda ||W||_2 \label{loss_function}
\end{align}
where $\lambda$ is regularization coefficient and $W$ denotes all weights in FCN. By adding such L2 norm into loss function,  all weights are forced to decay towards zero. As a result, it reduces unnecessary weights and avoids overfitting.

\section{Evaluation}

\subsection{Experiment Setup}

\begin{table}[!h]
  \centering
  \caption{Circuit Designs Used in Experiment}
  \label{tbl:t1}
  \begin{tabular}{l | c c c c c}
 	\hline
	Circuit Name   & \#Macros  & \#Cells & \#Nets & Width (\si{\micro\meter}) & \#Placements \\
	\hline
	des\_perf       & 4  &   108288 & 110283	  & 900   & 600 \\
	edit\_dist      & 6  &   127413 & 131134	  & 800   & 300 \\
	fft             & 6  &   30625  & 32088	      & 800   & 300 \\
	matrix\_mult\_a & 5  &   149650 & 154284      & 1500  & 300 \\
	matrix\_mult\_b & 7  &   146435 & 151614      & 1500  & 300 \\
	\hline
  \end{tabular}
\end{table}

Five designs from ISPD 2015 benchmarks~\cite{ISPD2015} are used in the experiment. Table \ref{tbl:t1} shows their basic information. The shapes of all five designs are squares, whose sizes range from $800$ \si{\micro\meter} to $1500$ \si{\micro\meter}. Please note that the square shape of designs is not required. RouteNet accepts rectangular circuit design in any size. For each circuit design in our experiment, at least 300 different floorplans are generated by placing macros at different locations with the ``obstacle-aware macro placement'' algorithm~\cite{macroPlace}. Though placed differently, macros all tend to locate near the chip boundary in order to leave plenty of space at chip center region, where routing demand tends to be high.
Then, each floorplan is placed and routed by Cadence Encounter, a previous version of Cadence Innovus~\cite{Innovus}. The DRC violation information for each layout is recorded as label.

Both \#DRV prediction and hotspot detection methods of RouteNet are tested on all five designs. When each design is tested, the machine learning model is trained only on data from the other four designs. This ensures that the tested design is totally {\em unseen} to the corresponding model, which eliminates the possibility of information leak from the testing dataset to the training dataset.

All algorithms are implemented in Python. CNN is implemented based on PyTorch \cite{paszke2017automatic}. 
As references for comparison, SVM and Logistic Regression (LR)-based methods are implemented based on scikit-learn \cite{scikitlearn}. Training and inference of all methods are performed on a machine with 2.40 GHz CPU and one NVIDIA GTX 1080 graphics card.

\subsection{Overall \#DRV Prediction}

For comparison, the previous method~\cite{chan2016beol} is directly transferred to our benchmark as a reference. In this method, only the maximum value (across all grid cells) and the coefficients of variation of features are extracted for each placement. The same preprocessing is performed. 
Both LR and SVM with Radial Basis Function kernel are tested as classifiers. Compared with RouteNet, the input of this method $X_i^{Ref} \in \mathbb{R}^{2 \times F}$ contains much less feature information.

\begin{table}[!b]
  \centering
  \caption{\#DRV Prediction Comparison}
  \label{tbl:t12}
 \resizebox{0.9\linewidth}{!}{%
  \begin{tabular}{l | c c c c c | c c c c c}
 	\hline 
	\rule{0pt}{10pt}\multirow{3}{*}{\large{Circuit Name}}     &  \multicolumn{5}{c|}{\large $c_0$/$c_1$+$c_2$+$c_3$ accuracy (\%)}   & \multicolumn{5}{c}{\large Best rank in top 10}  \\
			   		 &	\multirow{2}{*}{SVM} & \multirow{2}{*}{LR}  & \multirow{2}{*}{TR}  & \multirow{2}{*}{GR}   &  \multirow{2}{*}{\shortstack{Route \\ Net}}	&  \multirow{2}{*}{SVM} & \multirow{2}{*}{LR}  & \multirow{2}{*}{TR}  & \multirow{2}{*}{GR} &  \multirow{2}{*}{\shortstack{Route \\ Net}}			\\
			   		 &&&&&& & \\
	\hline
	\rule{0pt}{11pt}\large{des\_perf}        & \large52 & \large69 & \large80  & \large77 &	\large80  	&	\large\nth{87} & \large\nth{15}  &\large\nth{2}  &  \large\nth{1}   &\large\nth{2} 	\\
	\rule{0pt}{10pt}\large{edit\_dist}       & \large64 & \large61 & \large78  & \large77 & \large76  	&	\large\nth{17} & \large\nth{17}  &\large\nth{3}  &  \large\nth{3}   &\large\nth{2} 	\\
	\rule{0pt}{10pt}\large{fft}              & \large59 & \large56 & \large73  & \large70 &	\large75  	& 	\large\nth{6} &  \large\nth{6}   &\large\nth{2}  &  \large\nth{33}  &\large\nth{1}	\\
	\rule{0pt}{10pt}\large{matrix\_mult\_a}  & \large68 & \large62 & \large78  & \large74 &	\large72  	&	\large\nth{30} & \large\nth{5}   &\large\nth{1}  &  \large\nth{1}   &\large\nth{5}	\\
	\rule{0pt}{10pt}\large{matrix\_mult\_b}  & \large58 & \large58 & \large76  & \large73 &	\large76  	&  	\large\nth{22} & \large\nth{93}  &\large\nth{4}  &  \large\nth{1}   &\large\nth{4}	\\
    \hline
	\rule{0pt}{11pt}\large{Average}  		 & \large60 & \large61 & \large77  & \large74 &	\large76  	&	\large\nth{32} & \large\nth{27}	& \large\nth{2}  &  \large\nth{8}  &\large\nth{3} \\
	\hline
  \end{tabular}
  }
\end{table}

The goal of \#DRV prediction is to select a small set of placements with low \#DRV from many candidates. 
Table \ref{tbl:t12} shows the performance in \#DRV prediction. The ``$c_0$/$c_1$+$c_2$+$c_3$'' accuracy checks the binary classification accuracy by treating all placements in $c_1$, $c_2$, $c_3$ as one class and  $c_0$  as the other. As indicated in Algorithm \ref{alg6}, $c_0$ means the lowest \#DRV level. This accuracy evaluates how different methods recognize placements with the lowest \#DRV level. The result shows that RouteNet has similar accuracy as trial routing (TR) and global routing (GR).

We are also interested in the quality of placements selected by each method. To evaluate this, we first rank all placements from same design in ascending order of \#DRV. Then for each method, the top ten placements predicted to have least \#DRV are selected.  Among such ten placements, the rank of the placements with least ground truth \#DRV is reported as ``Best rank in top 10'' in Table \ref{tbl:t12}. On average, RouteNet finds the \nth{3} best placement from hundreds of candidates within 10 selections. Again, it shows comparable performance with trial routing and global routing, and is much better than both LR and SVM.

\begin{figure}[tb!]
  \centering
    \includegraphics[width=0.6\textwidth]{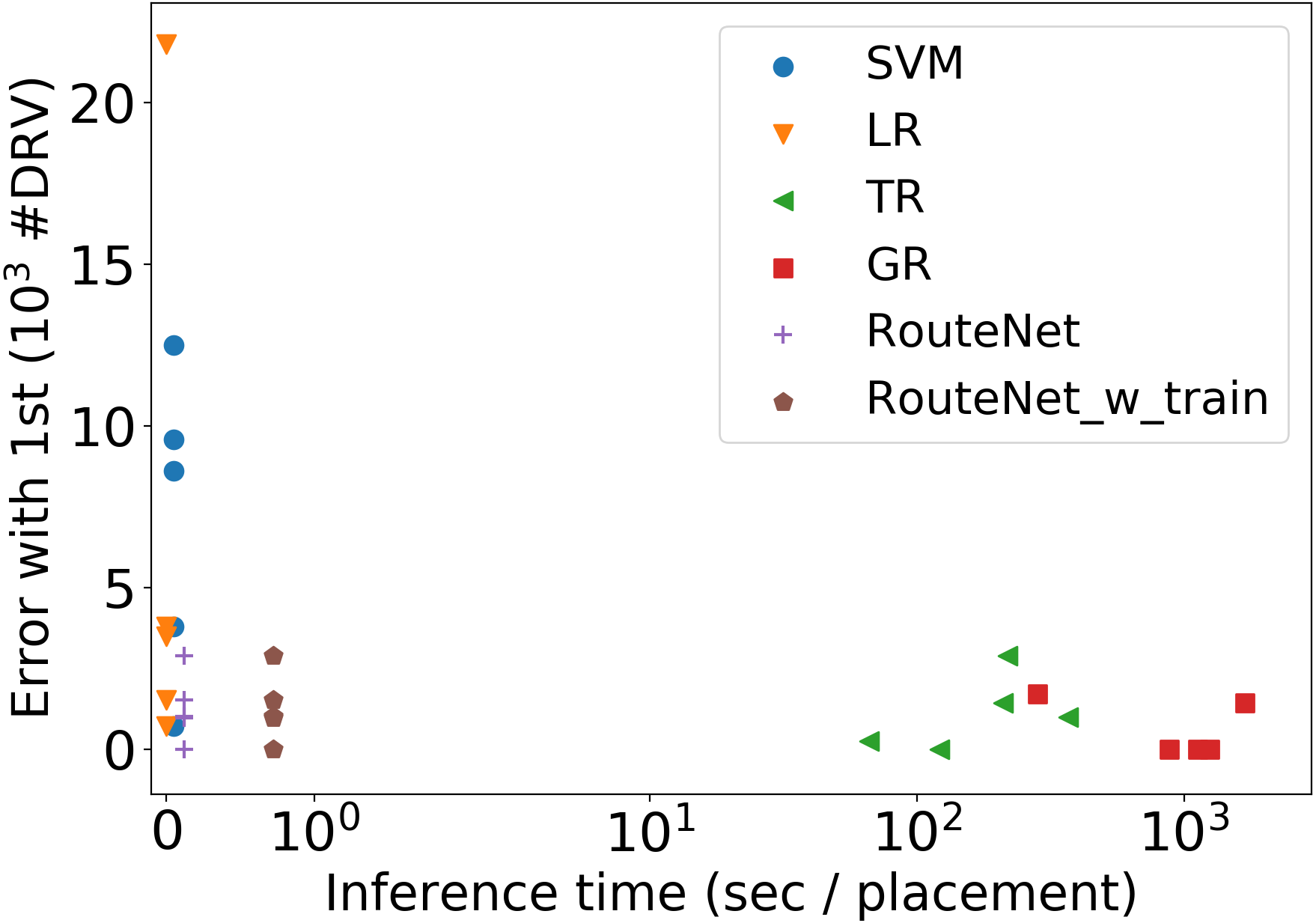}
  \caption{Trade-off between error with actually 1st-ranked placement and inference time in \#DRV prediction.}
  \label{MAE}
\end{figure}

When Table \ref{tbl:t12} only shows the rank of the best placement in 10 selections, Figure \ref{MAE} further indicates the gap between such ``best in ten'' and the actually 1st-ranked placement with least \#DRV. Such gap is denoted as error in \#DRV value. Each point represents the result of one design.
Besides accuracy, runtime is another essential factor to consider. Figure \ref{MAE} also shows the inference time for each method. Inference time is the overall time taken to predict one placement, starting at the end of global placement. In practice, RouteNet is trained in advance with other designs, so the training process costs no extra time during inference. But for reference, we still provide ``RouteNet\_w\_train'', which includes the training time of RouteNet.
In Figure \ref{MAE}, the results for RouteNet aggregate at the lower left corner, which means low inference time and high accuracy are achieved at the same time. By contrast, trial routing and global routing take substantially longer runtime to reach similar accuracy. LR and SVM, however, cannot guarantee low error though they are quite fast. Our RouteNet is the only fast and accurate method in \#DRV prediction. Even with training time included, the average inference time for one placement is still less than one second. 

\subsection{DRC Hotspot Detection}

For comparison, alternative methods similar to the previous work \cite{chan2017routability} are implemented. Features from each grid cell itself are extracted as its input, then grid cells are classified independently by either LR or SVM.

\begin{table}[!b]
  \centering
  \caption{Hotspot Detection Comparison}
  \label{tbl:t4_route}
  \begin{tabular}{l |  c | c c c c c}
  \hline
    \multirow{2}{*}{Circuit Name} &  \multirow{2}{*}{\shortstack{FPR \\ (\%)}}  &  \multicolumn{5}{c}{TPR (\%)}  \\ 
              &     &  TR & GR & LR     &   SVM  &  \textbf{RouteNet} \\
  \hline
  des\_perf         & 0.54  & 17 & 56 &  54  & 42  & \textbf{74} \\
  edit\_dist        & 1.00  & 25 & 36 &  38  & 28  & \textbf{64} \\
  fft               & 0.30  & 21 & 45 &  54  & 31  & \textbf{71} \\
  matrix\_mult\_a   & 0.21  & 13 & 30 &  34  & 12  & \textbf{49} \\
  matrix\_mult\_b   & 0.24  & 13 & 37 &  41  & 20  & \textbf{53} \\
  \hline
  Average           & 0.46  & 18 & 41 &  44  & 27  & \textbf{62} \\
  \hline
  \end{tabular}
\end{table}

Table \ref{tbl:t4_route} shows the accuracy in hotspot detection. TPR (True Positive Rate) and FPR (False Positive Rate) are used for evaluation. FPR describes the rate of grid cells being wrongly classified as hotspots. TPR, also named recall or sensitivity, describes the percentage of detected hotspots over all existing hotspots. By adjusting the decision threshold of prediction result, FPR and TPR change proportionally. For a fair comparison, we compare the TPR of all methods under the same FPR. The same decision threshold is used for all designs, which results in slightly different FPR among designs, but all under $1\%$.

\begin{figure}[!tb]
  \centering
    \includegraphics[width=0.92\textwidth]{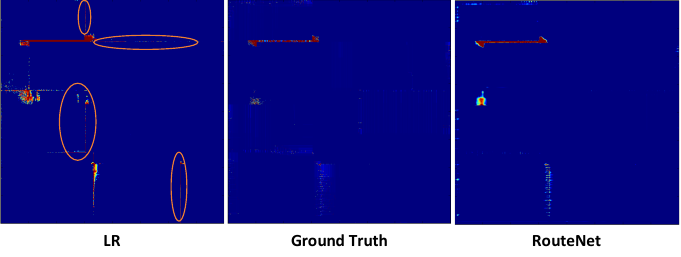}
  \caption{Visualization of hotspot detection results.}
  \label{resultGloRout}
\end{figure}

As  Table \ref{tbl:t4_route} shows, global routing is a much better hotspot detector than trial routing, although both methods have similar accuracy in overall \#DRV prediction. LR demonstrates better accuracy than global routing. SVM, however, is inferior to global routing even with our best effort on hyperparameter tuning. RouteNet is superior to all methods and improves global routing accuracy  by 50\%. Figure \ref{resultGloRout} provides an illustration of hotspot detection results. The result of RouteNet is closer to the ground truth. Orange circles indicate grid cells wrongly recognized as hotspots with high confidence by LR. These grid cells typically locate at the edges of macros.  LR exaggerates the influence of macro on them.

\subsection{Ablation Studies}

\begin{table}
  \centering
  \caption{Hotspot Detection for FCN Variations and Other Alternative Methods}
  \label{tbl:t5}
  \setlength{\tabcolsep}{1.2mm}
 \resizebox{\linewidth}{!}{%
    \begin{tabular}{l | c | c  c  c c  c  c | c  c   c  c  c  c c c c}
 	\hline
	\multirow{3}{*}{Circuit Name} & \multirow{3}{*}{\shortstack{FPR \\ (\%)}} &   \multicolumn{12}{c}{TPR (\%)}  \\ \cline{3-14}
					& 		
					& \multirow{2}{*}{\shortstack{Infer\\seen}}   
					& \multirow{2}{*}{\shortstack{Less\\data}} &
					\multirow{2}{*}{\shortstack{No\\short}}  & \multirow{2}{*}{\shortstack{Less\\conv}} & 
					\multirow{2}{*}{\shortstack{No\\pool}}    & \multirow{2}{*}{\shortstack{\textbf{Route}\\\textbf{Net}}} & \multirow{2}{*}{LR} &  \multirow{2}{*}{\shortstack{5$\times$5  \\ LR}} &  \multirow{2}{*}{\shortstack{9$\times$9  \\ LR}} & \multirow{2}{*}{SVM} & \multirow{2}{*}{\shortstack{5$\times$5  \\ SVM}}  & \multirow{2}{*}{\shortstack{9$\times$9\\SVM}}	\\
					& 	 &    &    &   &   &   &   &  &	  \\
	\hline
	des\_perf		& 0.54	&	77 & 71  &	  71	& 73	&   68   &	\textbf{74} & 54   &  58  & 58 &  42  &  47  &   29	\\
	edit\_dist      & 1.00	&   68 & 61  &    63    & 62   	&   55   &  \textbf{64} & 38   &  39  & 38 &  28  &  29  &   20 	\\
	fft             & 0.30	&   74 & 70	 &    68 	& 68    &   69   &  \textbf{71}	& 54   &  56  & 54 &  31  &  41  &   23	  \\
	matrix\_mult\_a & 0.21	&   51 & 46	 & 	  45 	& 45   	&   45   & 	\textbf{49}	& 34   &  36  & 35 &  12  &  32  & 	 9	  \\
	matrix\_mult\_b & 0.24	&   58 & 50  &    51 	& 50    &   50   &  \textbf{53}	& 41   &  44  & 42 &  20  &  39  &   16 	  \\
	\hline
	Average			& 0.46  &   66 & 60  &    60    & 60    &   57   &  \textbf{62} & 44   &  47  & 45 &  27  &  38  &    19 \\
	\hline
  \end{tabular}
  }
\end{table}

To further explore the hotspot detection problem, some variations of both FCN and alternative methods are evaluated. Results are shown in the Table \ref{tbl:t5}. These variant methods are briefly described as follows.
\begin{itemize}
\item Infer seen: Training and inference on different placements of the same circuit.
\item Less data: Trained on less data. Only placements from two circuits are used for training instead of four.
\item No short: The shortcut structure is removed from current FCN architecture.
\item Less conv: Three convolutional layers (with channels 64, 32, 32) in the middle of shortcut are removed, resulting in a shallower network.
\item No pool: Based on the shallow network above, the POOL layers are removed. TRANS layers are replaced by normal CONV layers.
\item $5\times5$ LR: Using window size of $5\times5$ grid cells to capture neighboring features of each grid cell in LR. Similarly, $9\times 9$ LR means $9\times9$ cells of window size.
\item $5\times5$ SVM: The same as the $5\times5$ LR above in feature extraction, but for SVM.
\end{itemize}

The effect of training and inference on different designs is explored by ``Infer seen'' and ``Less data'' in Table \ref{tbl:t5}. Difference in designs used for training and inference can be vital to the transferability of an algorithm. That is, if the distribution of DRC hotspots varies greatly among different designs, the pattern learned from training data may not be applicable to new ``unseen'' designs. Compared with original RouteNet, the better performance for ``Infer seen'' in Table \ref{tbl:t5} implies the existence of certain {\em pattern} unique to each design. But lower accuracy for ``Less data'' indicates that more training data from different designs can bridge such gap.

The FCN structure in RouteNet has both shallow and deep paths connecting input and output layers. In order to check the effect of such two-path structure, two variations ``No short'' and ``Less conv'' are tested. ``No short'' removes the shorter path and ``Less conv'' removes the longer path. As expected, Table \ref{tbl:t5} shows accuracy degradation for both variations. More interestingly, by removing POOL and TRANS structures in ``No pool'', which leads to a large reduction in receptive region, overall accuracy further degrades. It supports our claim on the importance of receptive region and the global information in hotspot detection.

Table \ref{tbl:t5} also shows how a larger receptive region affects other machine learning methods in hotspot detection. We tested several larger window sizes for feature extraction as $3\times3$, $5\times5$, $7\times7$, $9\times9$ grid cells. The $5\times5$ window size turns out to perform the best. Again, the large receptive region gives better results, but $5\times5$ is the upper limit in our experiment. An even larger window blurs the local information of the target grid cell. Compared with these alternative methods, RouteNet provides a better solution to obtain the benefit of large receptive region.

\section{Summary}

This work advanced the state of the art of routability prediction at two fronts. For overall routability forecast of mixed-size designs, RouteNet achieves similar accuracy as global routing but is several orders of magnitude faster. This largely solves the challenge of having both accurate and fast routability prediction of general designs. For DRC hotspot detection with consideration of macros, RouteNet also makes an important step forward by improving global router's accuracy by 50\%. 

Proposed in 2018, this work named RouteNet was the first CNN/FCN-based routability predictor that achieves sufficiently high accuracy and high speed. It attracts wide attention and becomes a common baseline in routability estimators. Some feature-extraction and model-design principles proposed in this work are widely adopted in later works.

%% file: txt/6_flowtuning.tex
\chapter{Design Flow Tuning}
\label{chapter:flow}

\section{Background}

 \begin{figure}[!b]
  \centering
    \includegraphics[width=0.78\columnwidth]{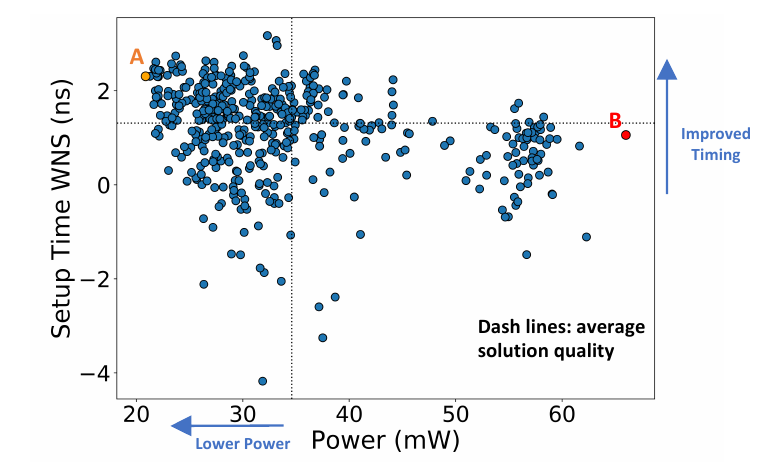}
  \caption{Solution quality variance in parameter space.}
  \label{space}
\end{figure}

The impact of parameter settings on overall design quality is phenomenal. Figure~\ref{space} plots the power and the worst negative setup time slack of design B22 from ITC'99~\cite{albrecht2005iwls}, when it is synthesized with different logic synthesis parameters. Changing logic synthesis parameters can result in $3\times$ difference in power and more than one clock cycle difference in slack. Therefore, high-quality and automatic design flow parameter tuning is highly desirable. 


Design space exploration (DSE), a problem similar to design flow parameter tuning, has been studied across various levels of abstractions \cite{liu2015supervised, mariani2012oscar, meng2016adaptive, tiwary2006generation, xydis2013meta, xydis2015spirit}. Active learning-based method is widely successful in DSE. This method builds an internal machine learning model to predict the design quality from design parameter space, and selects the next sampling point based on Gaussian process \cite{zuluaga2012smart, zuluaga2013active} or random forest model \cite{liu2013learning, meng2016adaptive}. Then, the flow result of the newly sampled parameter set is added into the dataset to re-train the machine learning model for the next sampling step. 
 
Despite their similarities, design flow parameter tuning is different from design space exploration. First, design flow parameter tuning often has a larger amount of prior data to learn from because similar parameters have been applied to previous designs multiple times already. Design space exploration, on the contrary, often has fewer prior data to learn from. This is because each design is different, and the impact of architecture decisions such as loop unrolling and pipelining can change significantly across different designs. Second, design flows such as synthesis and physical design flows often have an order of magnitude longer runtime and more parameters than those of design space exploration, including high-level synthesis design space exploration. 
This significantly reduces the number of sampling iterations and results in a smaller dataset for learning. 
Therefore, despite the similarity, automatic flow parameter tuning is more challenging from a time budget perspective. To apply the machine learning approach, we need to improve the efficiency of automatic flow parameter tuning with more advanced and customized learning techniques.

We work on the design flow parameter tuning problem, also named {\em parameter space exploration} to indicate both the similarity with and difference from conventional design space exploration. Synthesis or physical design parameters are tuned to optimize design quality after the complete synthesis and physical design flow. To collect data for experiments, we performed extensive synthesis and physical design runs with different synthesis parameters on many designs to build a dataset, where we notice the impact of parameters can be consistent for different designs. This allows transferring knowledge from known prior data.

\begin{figure}[!t]
  \centering
    \includegraphics[width=.7\columnwidth]{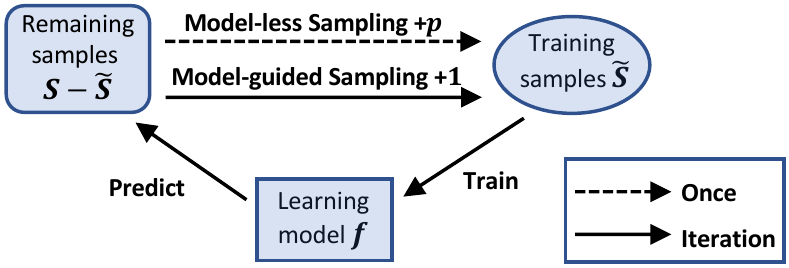}
  \caption{Iterative refinement framework.}
  \label{refinement}
\end{figure}

\textbf{Iterative Refinement Algorithm.} The iterative refinement framework has been adopted in many previous DSE works \cite{zuluaga2012smart, zuluaga2013active, mariani2012oscar}. It is illustrated in Figure \ref{refinement}. It divides the space exploration process into two phases: model construction and model refinement. At model construction phase, $p$ samples are selected and designers run design flow to build initial model $f$. Such a sampling process is referred to as {\em model-less sampling}. 

During model refinement phase, according to $f$'s prediction, it iteratively selects the most promising sample $s$ to run through design flow. Then $f$ is refined by the new completed set $\widetilde{S}$, which is augmented by $s$ at each iteration.
We name the model-based sample selection method at this phase as {\em model-guided sampling}.

\section{Methodology}

\subsection{Problem Formulation}

We refer to the parameters in logic synthesis or physical design scripts as {\em parameters} or {\em features}. Each parameter combination is also referred to as a {\em sample} or a parameter vector. A parameter combination $d$ consists of $c$ features and each feature has $n_i$ options, where $i \in [1, c]$. Continuous features can be discretized into categorical data. We use $S$ to denote the whole parameter space and $|S| = \prod^c_{i=1}{n_i}$. The parameter space grows exponentially when $c$ increases. We evaluate the design objectives after we complete the whole synthesis and physical design flow. Due to the large parameter space, the limited computation resources and the allowed execution time, only a small subset $\widetilde{S}$ of samples can complete design flow and be evaluated. The process of selecting samples to form $\widetilde{S}$ is referred to as {\em sampling}. The number of trials allowed is denoted as budget $b$, $|\widetilde{S}| <= b$.

For each single design objective such as power $P$, the goal of design flow parameter tuning framework $F$ is to find the sample with lowest $P$ with no more than $b$ samples. Assume learning model $f$ is used during exploration, 

\begin{gather*}
\widetilde{S} = F(S, b, f)  \\
F^{*} = \argmin_{F}{(\min{P[\widetilde{S}] - \min{P[S]}})}  
\end{gather*}
An alternative formulation is to minimize the number of samples $b$ while achieving power no higher than $P$.

For multiple design objectives, the goal of $F$ is to derive an approximated parameter set for Pareto-optimal samples, namely, Pareto frontier. The quality of Pareto frontier is measured by average distance from reference set (ADRS). A lower ADRS means the parameter set is closer to the actual Pareto set.

Assume two of the objectives are power $P$ and delay $D$. Given actual ground-truth Pareto frontier $T \subset S$ and approximate frontier $\Lambda \subset \widetilde{S}$, we have: 

\begin{gather*}
ADRS(T, \Lambda) = \frac {1}{|T|} \sum_{\tau\in T}\min_{\lambda \in \Lambda} \delta(\tau, \lambda)  \\
\delta(\tau = (P_{\tau}, D_{\tau}), \lambda= (P_{\lambda}, D_{\lambda})) = \max{(0, \frac{P_{\lambda} - P_{\tau}}{P_{\tau}}, \frac{D_{\lambda} - D_{\tau}}{D_{\tau}})}
\end{gather*}

\begin{algorithm}[!t]
\caption{FIST Framework}
\textbf{Input}: Parameter space $S$, budget $b$, completed samples $\widetilde{S}=\emptyset$, 
\\\hspace*{15mm}feature importance $I \in \mathbb{R}^c$, clustering refinement threshold $\theta$ \\  
\textbf{Clustering}: 
\begin{algorithmic}[1]
\State \mbox{Identify more important features $\iota = I > median(I)$,} where $\iota \in {\{0, 1\}}^c$
\State Build $m$ clusters $S_i$ ($i \in [1, m]$) as partition of $S$, s.t. $\forall s_i \in S_i$, $s_i[\iota]$ are the same \label{alg:last-step1}
\end{algorithmic}
\textbf{Model-less Sampling}: 
\begin{algorithmic}[1]
\setcounterref{ALG@line}{alg:last-step1}
\State Randomly select $p$ clusters $S_j$ ($j \in [1, p]$) from $m$ $S_i$ ($i \in [1, m]$)
\State Randomly select one $s_j$ from each $S_j$
\State Run and add $s_j$ ($j \in [1, k]$) to $\widetilde{S}$
\State $\forall s \in S_j$, label $s$ with $s_j$ and add $s$ to $\widetilde{S}_{approx}$ \label{alg:last-step2}
\end{algorithmic}
\textbf{Model-guided Sampling (Refinement)}
\begin{algorithmic}[1]
\setcounterref{ALG@line}{alg:last-step2}
\For{each int $i \in [1,b-p]$}
\State Initialize $f_i$, its depth depends on $i$
\If{$i <= \theta$} // Exploration and approximation
\State Train $f_i$ with $\widetilde{S}_{approx}$
\State Pick $s_a$ in $S - \widetilde{S}_{approx}$ based on $f_i$, run $s_a$, add to $\widetilde{S}$
\State Identify $S_a$ s.t. $s_a \in S_a$
\State $\forall s \in S_a$, label $s$ with $s_a$ and add $s$ to $\widetilde{S}_{approx}$
\Else  ~~// Exploitation
\State Train $f_i$ with $\widetilde{S}$
\State Pick $s$ in $S - \widetilde{S}$ based on $f_i$, run $s$, add to $\widetilde{S}$
\EndIf
\EndFor
\end{algorithmic}
\textbf{Output}: Completed samples $\widetilde{S}$ 
\label{alg7}
\end{algorithm}

\begin{figure}[t]
  \centering
    \includegraphics[width=.7\columnwidth]{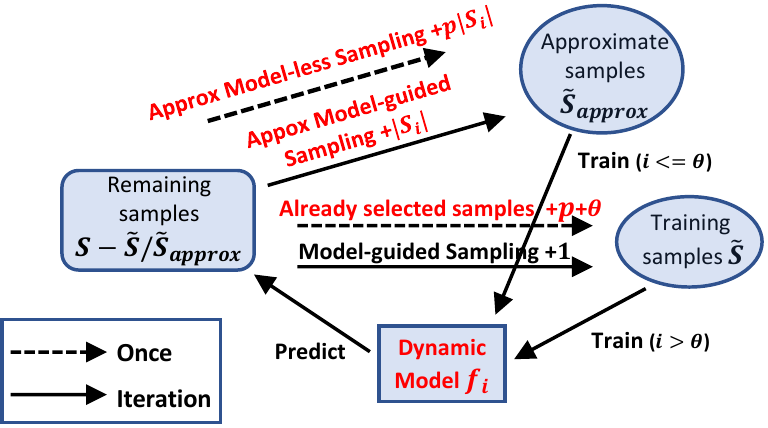}
  \caption{FIST framework.}
  \label{FISTimgs}
\end{figure}

\textbf{Method Overview.} Figure \ref{FISTimgs} and Algorithm \ref{alg7} illustrate our algorithm FIST. The major innovative strategies include: 1. Sampling by clustering; 2. Approximate samples; 3. Dynamic model. The ``approximate samples'' strategy is actually incorporated in ``sampling by clustering''.

\subsection{Clustering by Similarity in Important Features}

For a specific design, samples with the same values on some features will result in similar solution qualities, especially for the ``important'' features. The ``importance'' here means the extent to which each feature can influence the final solution quality. When evaluating the influence of each feature, the values of all other features are controlled to be the same. In Algorithm \ref{alg8}, samples with the same value for all other features except the evaluated one form measurement subgroups $S_k$. In this way, the summation of the solution quality variation $\sigma^2_k$ within each measurement subgroup reflects the importance of this evaluated parameter. A feature importance vector $I \in \mathbb{R}^c$ is generated. 

\begin{algorithm}[!bt]
\caption{Feature Importance Evaluation}
\textbf{Input}: Parameter space $S'$ with labels $L$ from prior designs
\begin{algorithmic}[1]
\For{each int $q \in [1, c]$}
\State $S_q =$ ($S'$ with the $q$th feature removed from all $s' \in S'$) 
\State Build $n$ measurement subgroups $S_k$ ($k \in [1, n]$) as a partition of $S_q$, s.t. $\forall s_k \in S_k$, $s_k$ are the same
\State $I[q] \propto \sum_{k=1}^n \sigma^2_k$, $\sigma^2_k$ is variance of $L$ in $S_k$
\EndFor
\end{algorithmic}
\label{alg8}
\textbf{Output}: Feature importance $I\in \mathbb{R}^c$
\end{algorithm}

For example, a parameter space $S'$ consists of two features, each with two options $\{0, 1\}$, then $S' = \{[0, 0], [0, 1], [1, 0], [1, 1]\}$. Assume the corresponding labels on solution quality $L = \{1, 2, 3, 4\}$. When measuring the first feature, $S_q$ is constructed by removing the first feature from $S'$, $S_q = \{[0], [1], [0], [1]\}$. Then $L$ for these two subgroups are $\{1, 3\}$ and $\{2, 4\}$. $I[1] = \sigma^2(1, 3) + \sigma^2(2, 4) = 2$. Similarly, for the second feature, $I[2] = \sigma^2(1, 2) + \sigma^2(3, 4) = 0.5$. Thus, the feature importance vector $I = [2, 0.5]$ and the binary vector indicating if each feature is important is $\iota = I > median(I) = [1, 0]$. In this case, the first feature is important, which means it has a stronger impact on solution quality $L$. FIST learns and transfers feature importance from prior data because such important parameters can be quite consistent among different designs, but notice that it is completely different from assuming certain universally good parameter settings across different designs ever exist.

Clustering is then performed with such prior knowledge on feature importance (line 1-2 in Algorithm \ref{alg7}). More important features $\iota \in {\{0, 1\}} ^c$ are first identified, then samples with the exact same values for $\iota$ are grouped into the same cluster $S_i$. In this way, the final solution qualities for the samples in the same cluster are closer. The sampling strategies in FIST use one sample to partially represent samples from the whole cluster, which makes sampling much more efficient.

\subsection{Model-less Sampling Based on Clusters}

The model-less sampling aims at exploring the whole parameter space with a limited number of samples $p < b$. During feature importance based clustering, the number of clusters $m$ is set to be greater than $p$ to retain in-cluster similarity. The value of $m$ can be easily adjusted by modifying the number of important features $\iota$. Sampling from different clusters avoids wasting budget on similar samples. As the red samples in Figure \ref{cluster} shows, only a subset of clusters are selected and one sample from each selected cluster is randomly chosen to represent its cluster. We complete the design flow of selected samples and put them into $\widetilde{S}$, which is the training data for constructing the machine learning model.

\subsection{Approximate Samples}

In order to enhance the machine learning model training with limited sampling that costs expensive runtime, we increase the sampling dataset in an approximate manner. 
If a cluster $S_j$ has only one sample $s_j \in S_j $ with known label $l(s_j)$, we apply this label to the rest of the samples in $S_j$ as training data. 
Although the design flow has not been run for $S_j-\{s_j\}$, their actual labels should be similar to $l(s_j)$, as they belong to the same cluster. By using $S_j-\{s_j\}$ as approximate samples, the entire cluster $S_j$ is included in set $\widetilde{S}_{approx}$. This process is indicated in Figure \ref{FISTimgs} and line 6 in Algorithm~\ref{alg7}. The usage of $\widetilde{S}_{approx}$ and approximate samples is shown in lines 10-13. This is partially inspired by the ``pseudo labeling'' \cite{lee2015icml} commonly used in semi-supervised learning. 


\subsection{Model-guided Sampling by Clustering}

We strive to balance ``exploration'' and ``exploitation'' in the model-guided sampling process. ``Exploration'' only acquires new knowledge from unexplored clusters while ``exploitation'' also makes use of promising explored clusters in sampling. At the beginning of model refinement phase, exploration is emphasized, because the number of completed samples $|\widetilde{S}|$ is relatively small and many clusters have not been explored. Thus, FIST identifies a new sample $s_a$ from unexplored clusters $S - \widetilde{S}_{approx}$ in line 11 of Algorithm~\ref{alg7}. Also, it adds whole cluster $S_a$ to approximate samples set $\widetilde{S}_{approx}$.

\begin{figure}[!t]
  \centering
  \includegraphics[width=.8\columnwidth]{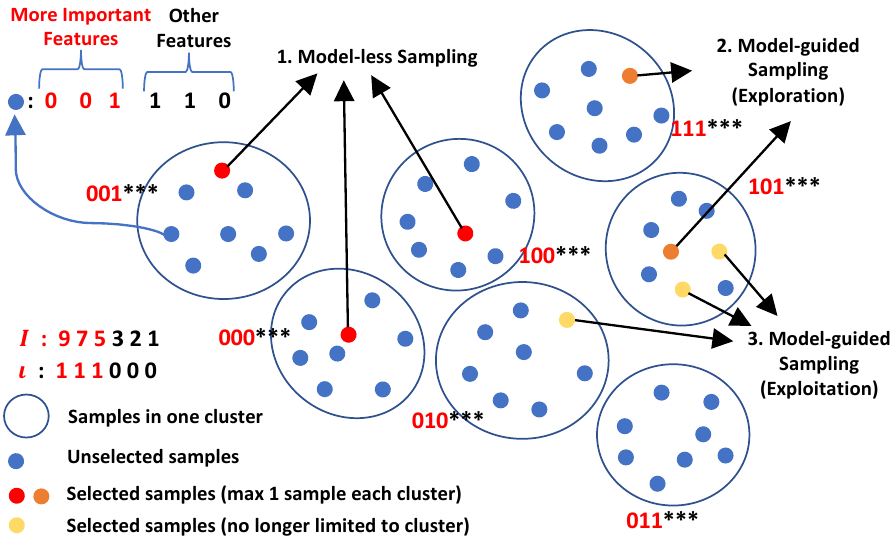}
  \caption{An example of sampling by clustering.}
  \label{cluster}
\end{figure}

After $\theta$ iterations, the emphasis is shifted to exploit explored clusters. Now neither cluster information nor approximate samples are considered anymore. The model is simply trained with completed samples $\widetilde{S}$ and the new selected sample in $S - \widetilde{S}$ is often from previously explored clusters. This is shown in line 16 of Algorithm~\ref{alg7} and yellow samples in Figure \ref{cluster}.

\subsection{Dynamic Tree Depth}

The bias-variance trade-off in machine learning indicates that an appropriate model complexity depends on training data size. The model refinement stage starts with $p$ samples and ends with $b$ completed samples. Assuming $b>2*p$, the number of training data at least doubles during model-guided sampling. Thus, it is rational to vary the model complexity accordingly. We choose to change the maximum tree depth through the model refinement process, as shown in line 8 in Algorithm~\ref{alg7}. Initially, we use relatively shallow trees, which result in a less complex model, then increase the maximum tree depth to the optimal depth.

\section{Evaluation}

\subsection{Experiment Setup}


Nine different designs from ITC'99~\cite{corno2000rt} are synthesized with a commercial synthesis tool in 45nm NanGate Library~\cite{NanGate} and then placed and routed by Cadence Encounter v14.28. Their post-synthesis gate number ranges from 167 to 76842. Both slack and power are measured by Encounter. When each design is tested, all other designs are utilized as ``known'' designs to evaluate feature importance. The design objectives are ``Power'', ``Setup Time WNS'' and ``Hold Time WNS'', where WNS means the worst negative slack. For each design, we choose nine synthesis parameters for tuning, and exhaustively collect all 1728 samples in the parameter space.   Synthesis parameters include: set\_max\_fanout, set\_max\_transition, set\_max\_capacitance, high\_fanout\_net\_threshold, set\_max\_area, insert\_clock\_gating, leakage\_power\allowbreak \_optimization, dynamic\_power\_optimization and compile\_type. Any non-numeric parameters are represented by multiple artificial features with one-hot encoding. 

We compare our method to prior arts in two ways. First, we evaluate the quality of samples with a fixed sampling budget $b$. In this case, $p = \frac{b-10}{2}$ samples are selected for model-less sampling. Second, we evaluate the number of iterations performed to reach a required design flow quality. In this case, we set $p=40$ for model-less sampling. We set the maximum tree depths of our dynamic models to be 3 and 10 for initial and final stages, respectively. The cluster refinement threshold 
 $\theta$ is set to 10 iterations. To reduce randomness, all single-objective and multi-objective results are obtained by taking an average of 500 and 1000 trials, respectively. 

\begin{table}[!b]
  \centering
  \caption{Methods Notation}
  \label{tbl:t3}
  \begin{tabular}{l | c }
  \hline
  Denotation           & Methods      \\
  \hline
  baseline\_RF         &  random sampling \& Random Forest model  \\
  baseline             &  random sampling \& XGBoost (same below)  \\
  dyn                  &  dynamic-depth tree model   \\
  mless                 &  model-less sampling by clustering     \\
  ref                  &  model-guided sampling in refinement by clustering \\
  rand                 &  feature importance assigned randomly    \\
  \hline
  \end{tabular}
\end{table}

\subsection{Flow Tuning Performance}
We first evaluate different methods by targeting three single objectives separately. Compared with exploring the Pareto frontier, such a simpler task is a more straightforward evaluation of space exploration algorithms. Denotations for different strategies are defined in Table \ref{tbl:t3}. FIST method can also be denoted as `mless\_ref\_dyn', which adopts all strategies including XGBoost, dynamic model and cluster sampling in both model-less sampling and refinement.


\begin{figure}[!tb]
  \centering
    \includegraphics[width=.7\columnwidth]{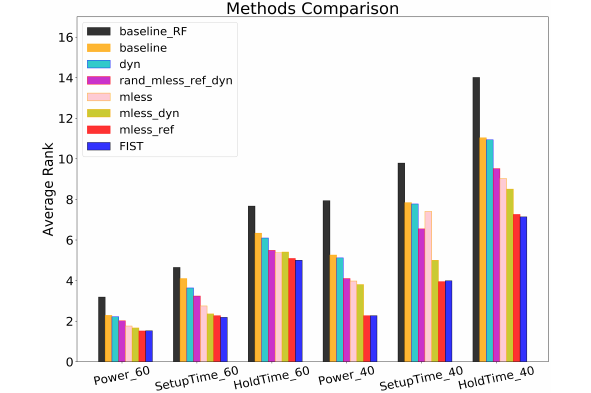}
    \vspace{-1mm}
  \caption{Best solution rank with the same sample cost.}
  \label{single_rank}
\vspace{2mm}
  \centering
    \includegraphics[width=.7\columnwidth]{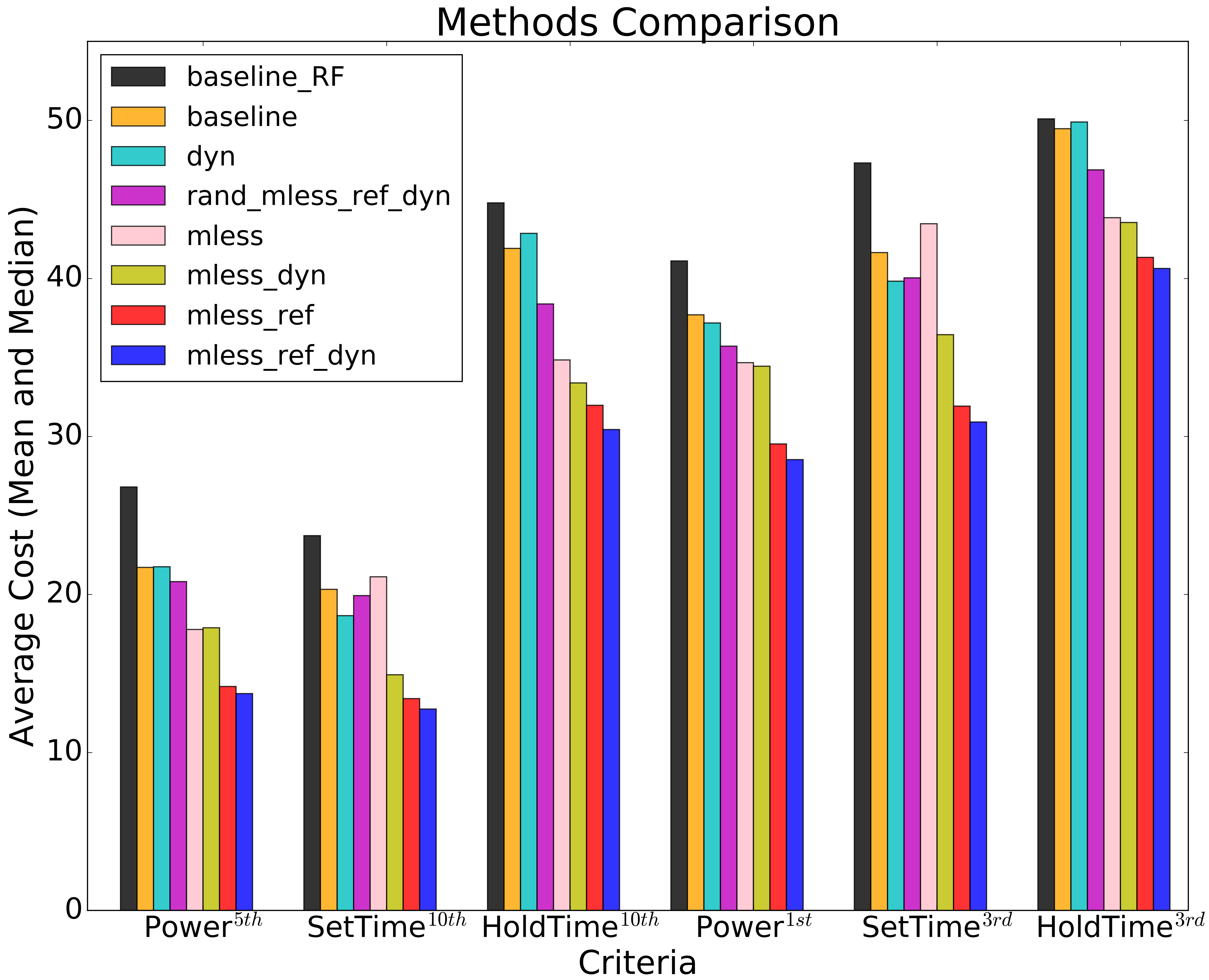}
    \vspace{-1mm}
  \caption{Sample cost to reach the same solution rank.}
  \label{single_cost}
\end{figure}

The quality of selections is measured by their rank in the whole parameter space. Figure \ref{single_rank} shows the best rank of explored results for three objectives given a fixed budget. For example, ``Power\_60'' means the algorithm attempts to minimize power with 60 samples for refinement. On average, FIST achieves 53\% reduction in ranking compared with the original framework \textit{baseline\_RF}. XGBoost outperforms Random Forest as the learning model and all other strategies contribute to a better ranking. Among these strategies, the contribution of cluster sampling is higher than the dynamic model.

Another method of note is ``rand\_mless\_ref\_dyn'', where feature importance is randomly assigned. Though worse than the FIST method, it still outperforms ``baseline''. On one hand, it indicates clustering sampling method itself benefits parameter tuning even without feature information; on the other hand, it proves the effectiveness of using important features and learning from other designs.

Two other popular methods are also compared with FIST in Table \ref{tbl:t4} for reference. TED sampling is proposed in \cite{liu2013learning} to replace random sampling in ``baseline\_RF'', but it fails to improve performance except on ``HoldTime''. We analyzed such unsupervised TED sampling method with our clustering strategy. On average the top 30 samples from TED falls into only 14 clusters, leaving the other 58 clusters empty. That is, under the view of supervised clustering, such unsupervised method cannot pick the most representative samples in parameter space. The genetic learning method \cite{ziegler2016scalable}, which is originally applied to primitives, is also implemented for comparison. As \cite{liu2015supervised} has concluded, the given budget is too limited for such genetic algorithms to accumulate a large enough population. 

\begin{table}[!htb]
  \centering
  \vspace{1mm}
  \caption{Rank Results with the Same Sample Cost}
  \vspace{-1mm}
  \label{tbl:t4}
  \begin{tabular}{l | c c c}
  \hline
  Methods           & Power\_{40} & SetupTime\_{40} & HoldTime\_{40}     \\
  \hline
  baseline\_RF                              & 8.0  & 9.8 & 14  \\
  baseline\_RF\_TED \cite{liu2013learning}  &  13  & 18  & 13  \\
  Genetic Algo \cite{ziegler2016scalable}  &  28  & 40  & 26 \\
  \hline
  Methods           & Power\_{60} & SetupTime\_{60} & HoldTime\_{60}     \\
  \hline
  baseline\_RF                              & 3.2  & 4.7 & 7.7  \\
  baseline\_RF\_TED \cite{liu2013learning}  & 7.9  & 10  & 7.2  \\
  Genetic Algo \cite{ziegler2016scalable}  &  23  & 15  & 19 \\
  \hline
  \end{tabular}
\end{table}

\begin{figure}[!t]
  \centering
    \includegraphics[width=.7\columnwidth]{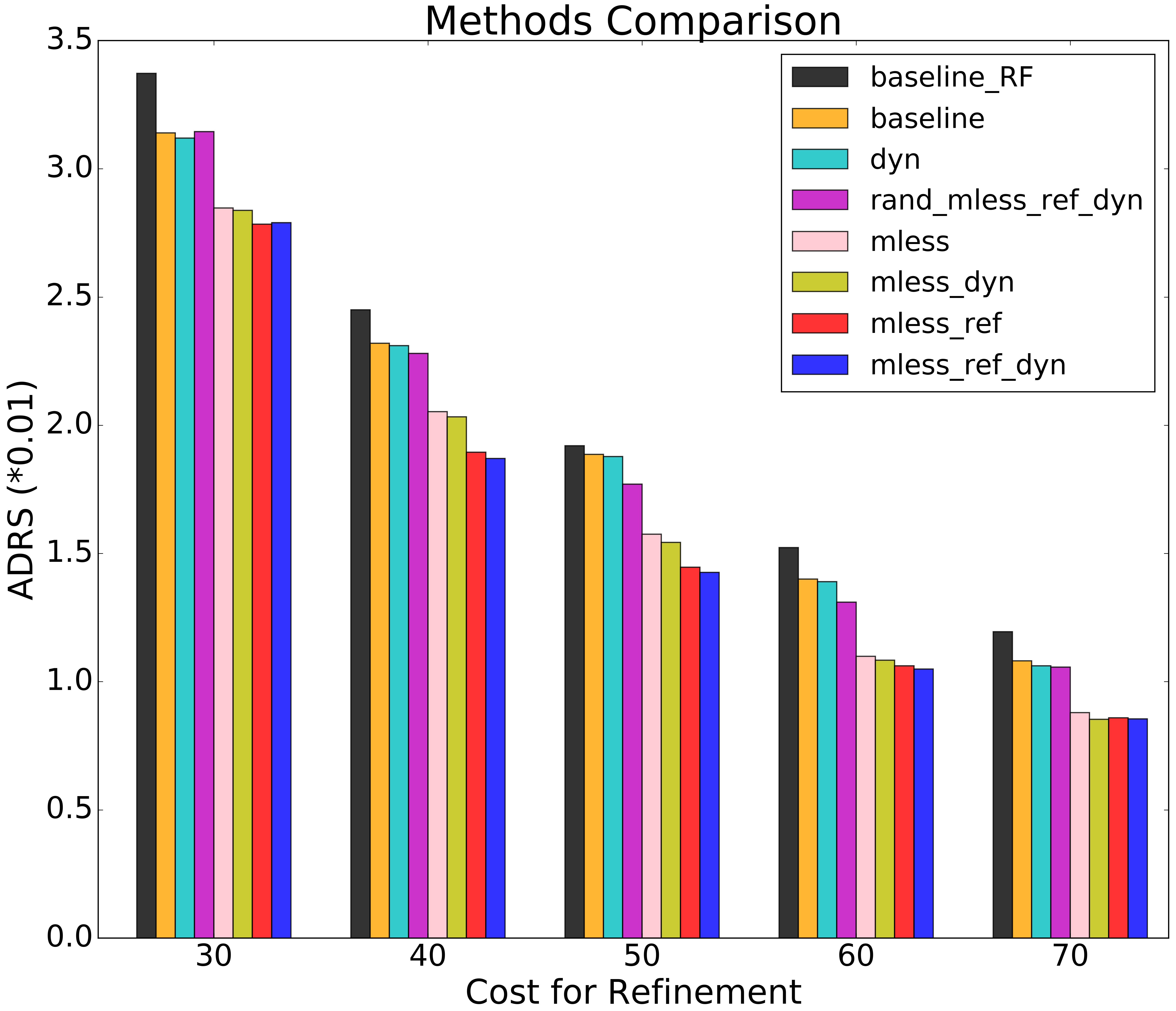}
    \vspace{-1mm}
  \caption{Best ADRS with the same sample cost.}
  \label{Pareto}
  \vspace{2mm}
  \centering
    \includegraphics[width=.7\columnwidth]{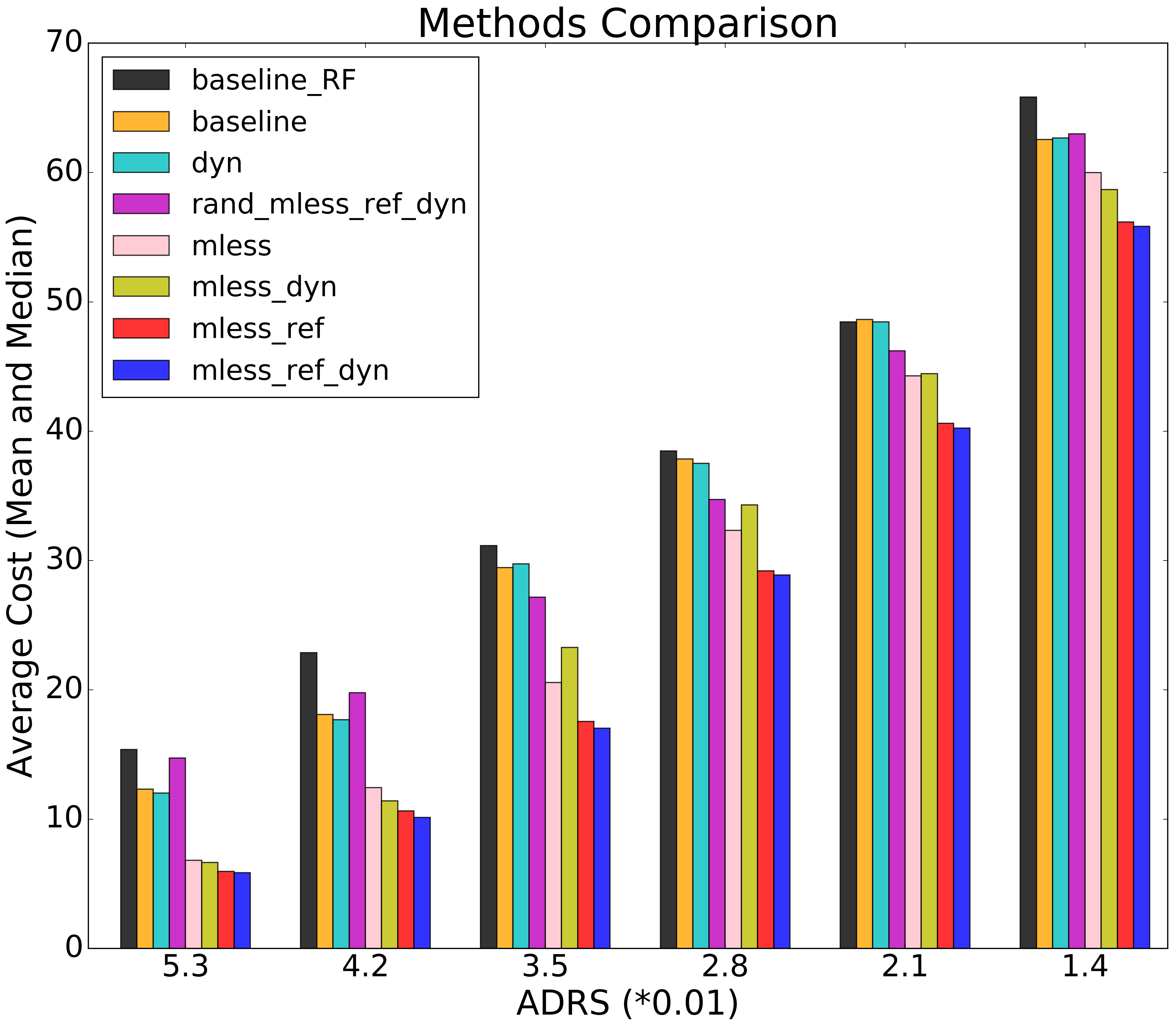}
    \vspace{-1mm}
  \caption{Sample cost to reach the same ADRS.}
  \label{Pareto2}
\end{figure}

Besides better sample quality under a fixed budget, we are also interested in reducing sample cost while reaching the same quality. The cost here refers to the number of samples synthesized in model refinement phase. Figure \ref{single_cost} indicates the cost to reach same solution ranks, where 35\% cost reduction is achieved by adopting all strategies. We can observe a similar trend for different strategies.


Besides single-objective optimization, designers are also concerned about optimizatizing multiple objectives in exploring Pareto frontier. The performance on Pareto frontier identification is evaluated with ADRS. ADRS can be measured because we know the quality of the whole design space after data collection. Figures \ref{Pareto} and \ref{Pareto2} show the performance in exploring Pareto frontier. Sample cost is fixed for Figure \ref{Pareto} and ADRS level is fixed for Figure \ref{Pareto2}. A range of cost levels and ADRS levels is covered. The effectiveness of all strategies is consistent on all cost or ADRS levels we have tested. On average 25\% improvement in ADRS and 37\% improvement in cost are achieved.


To further understand the effect of sampling by clustering, the similarity of samples by different sampling methods is shown in Table \ref{tbl:samp}. It compares random sampling, sampling within cluster, and sampling from different clusters. The standard deviations $\sigma$ of their solution quality are measured. The ``true'' and ``learn'' in parentheses indicate whether feature importance is ground-truth or learned from other designs. The in-cluster sampling gives much lower $\sigma$, which verifies that samples from the same cluster have much more similar  quality. This provides the rationale of using $\widetilde{S}_{approx}$ with approximate labels. The learned in-cluster $\sigma$ is only slightly higher than ground truth for timing, indicating that learned feature importance is close to the ground truth. 

However, $\sigma$ for cross-cluster sampling is not significantly higher than that for random sampling. That means simply sampling from different clusters does not lead to more representative samples. That is also why approximate samples  $\widetilde{S}_{approx}$ are necessary in such clustering strategy.

\begin{table}[!tb]
  \centering
  \caption{Standard Deviation of Samples}
  \label{tbl:samp}
  \begin{tabular}{l | c| c |c}
  \hline
                           & Power   & SetupTime & HoldTime   \\
  \hline
    Random sampling         &  3.35  &   0.377  &  0.288    \\  
  \hline
   In-cluster sampling (learn)    &  0.49  &   0.152  &  0.175    \\
   Cross-cluster sampling (learn) &  3.42  &   0.384  &  0.282    \\
   \hline
   In-cluster sampling (true)    &  0.54  &   0.135  &  0.146    \\
   Cross-cluster sampling (true) &  3.39  &   0.383  &  0.294    \\
  \hline
  \end{tabular}
\end{table}

\subsection{Experiment Setup on Industrial Designs}
We have developed a FIST-based automatic parameter tuning flow for industrial physical design flows based on commercial EDA tools.
The designs we experimented on are from a deep learning inference accelerator \cite{zimmer20190} implemented in 16nm FinFET technology: a 71K gate Processing Element (PE) 
and a 117K gate RISC-V microprocessor (RISC-V). The design objectives that FIST optimizes are `area' and `setup time TNS'. They are optimized under the condition that `hold time TNS' and `DRC violations' are met. The quality of FIST's parameter selections is compared with the quality of a set of parameter selections hand-tuned by experienced designers on these recently taped-out designs.

\begin{table}[!t]
  \centering
  \caption{Physical Design Parameters for Industrial Designs}
  \label{tbl:t_pd}
  \begin{tabular}{l | c }
  \hline
  Physical design parameter    & Settings      \\
  \hline
  postroute iterations                 &  0, 1, 2, 3, 4  \\
  cts.optimize.enable\_local\_skew     &  False, True   \\
  clock\_opt.hold.effort               &  low, medium, high    \\
  postroute (clock\_tran\_fix)         &  disable, enable      \\
  postroute (useful\_skew, timing\_opt)&  0, 1      \\
  useful\_skew (power\_opt)            &  0, 1      \\
  clock buffer max fanout             &  22, 36, 48, 96      \\
  target skew                          & 0.025, 0.05, 0.1, 0.3 \\
  setup uncertainty                    & -0.025, -0.05, -0.1 \\
  hvt cell swap enable during leakage optimization & 0, 1 \\
  extra hold uncertainty for SRAM macro & -0.025, -0.05, -0.1, -0.15 \\
  max util                             & 0.7, 0.78, 0.85 \\
  \multirow{2}{*}{hold uncertainty}    & -0.002, -0.005, -0.008\\
                                       & -0.01, -0.012 \\
  \hline
  \end{tabular}
\end{table}

Compared with experiments on ITC'99, this industrial experiment explores a much larger parameter space. Thirteen physical design parameters are tuned and each parameter provides 2 to 5 options. Details of the parameters are shown in Table~\ref{tbl:t_pd}. The whole parameter space consists of 1,382,400 samples. In this case, it is not possible to collect data exhaustively like in the ITC'99 experiment.
We limit the budget $b$ of FIST to be around 200, which is less than $0.02\%$ of parameter space. By comparison, the designer would hand-tune 30 parameter selections before settling on the final parameter selection.
We check whether FIST, the automatic parameter tuning method, provides better solutions. 

Though taking more trials than the hand-tuning process, parameter tuning with FIST can be fully automatic without any human knowledge. Hand-tuning parameters 200 times for multiple designs costs extra engineer time and is not likely to find a better solution than FIST. We set initial sampling number $p=100$ and cluster refinement threshold $\theta = 40$ based on the budget $b=200$. To leverage computer farms and prove the scalability of FIST, the ML model now selects around $10$ best samples at each iteration instead of just one. Then the design flows with selected parameters are completed on multiple machines in parallel.

\subsection{Performance on Industrial Designs}

\begin{figure}[!t]
  \centering
    \includegraphics[width=.9\columnwidth]{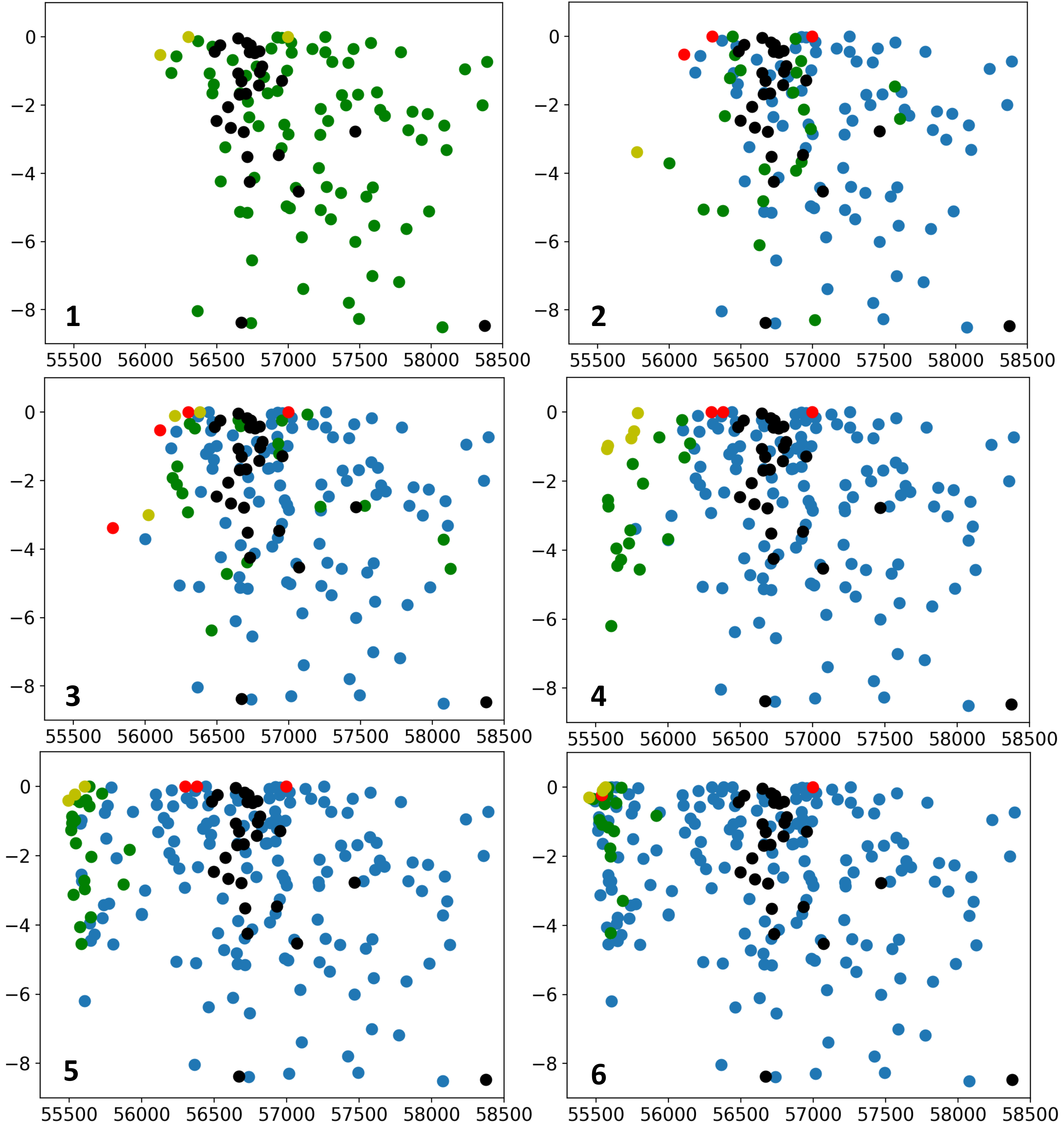}
  \caption{Parameter tuning process on PE. Area ($\si{\micro\meter}^2$) vs. setup TNS ($\si{\nano\second}$). Red and yellow are Pareto points. Black are  hand-tuned baselines.}
  \label{d1}
\end{figure}

\begin{figure}[!t]
  \centering
    \includegraphics[width=.9\columnwidth]{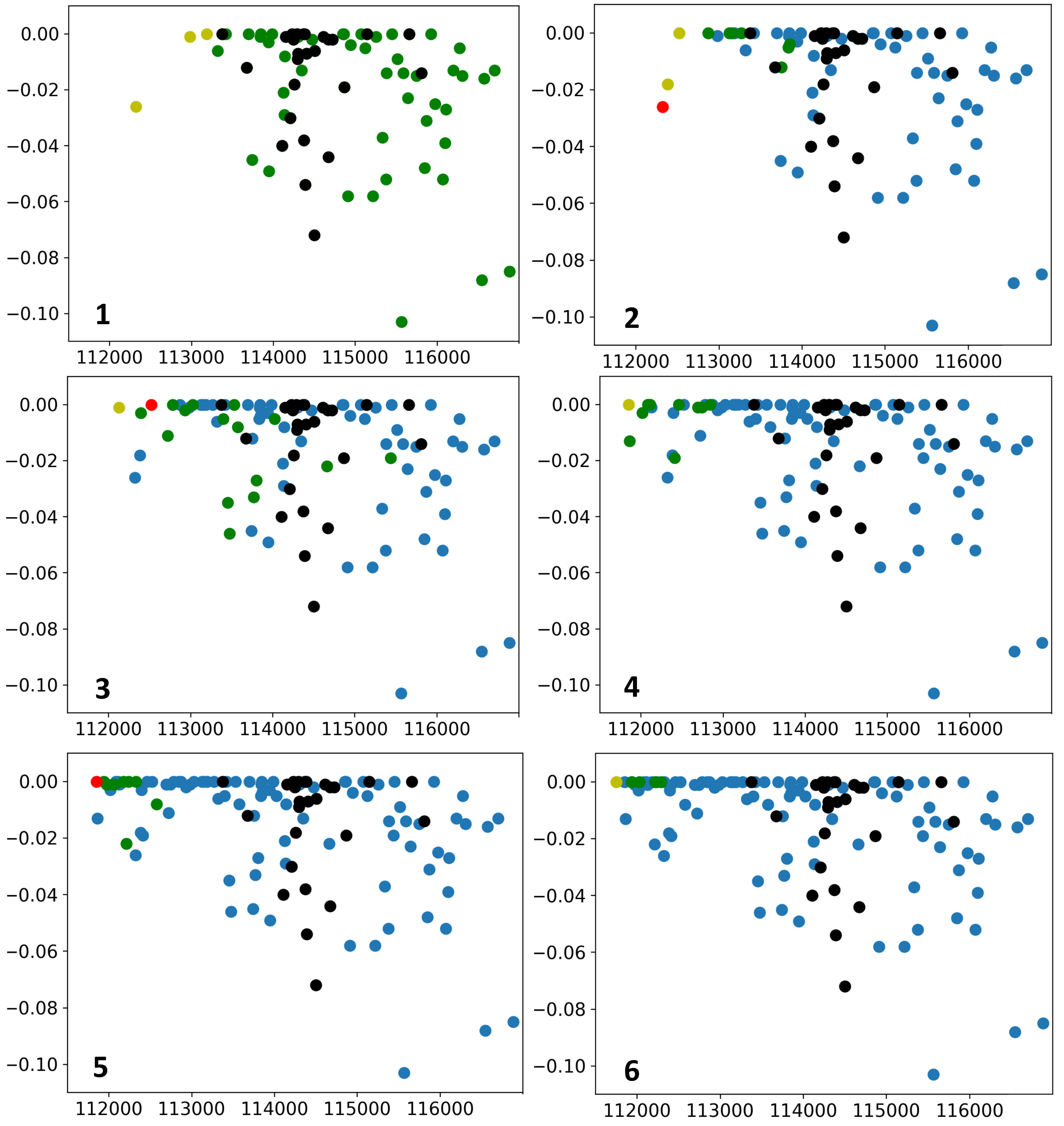}
  \caption{Parameter tuning process on RISC-V. Area ($\si{\micro\meter}^2$) vs. setup TNS ($\si{\nano\second}$). Red and yellow are Pareto points. Black are  hand-tuned baselines.}
  \label{d2}
  \vspace{-1mm}
\end{figure}

The qualities of parameter space exploration for PE and RISC-V are shown in Figure \ref{d1} and \ref{d2}, respectively. The x-axis shows area in $\si{\micro\meter}^2$ and y-axis shows setup time TNS in $\si{\nano\second}$. Points on the upper-left boundary of all the already explored samples are desirable Pareto points. We present six sequential stages during the tuning process, corresponding to six sub-graphs in Figure \ref{d1} and \ref{d2}. Subgraph 1 contains $p=100$ initial samples and each new subgraph adds around 20 new samples. In each subgraph ${sg}_i>1$, black points are $30$ parameter selections hand-tuned by the designers, green and yellow points are the $20$ new samples explored at that stage, blue and red points are the $100 + ({sg}_i-1)*20$ samples already explored in previous stages. Yellow and red points are Pareto points.

For PE, the best area of hand-tuned parameter selections (in black) is 56,483, while FIST finds a solution (in yellow) with the area of 55,453 with acceptable setup time closure. The improvement in area is 1.82\%. Similarly, in RISC-V, FIST reduced the best area from 113,375 (in black) to 111,751 (in yellow), improving the area by 1.43\%. Notice that such improvement is achieved by exploring less than $0.02\%$ of the parameter space.

Interestingly, the strategies of FIST can be clearly observed in different stages of this parameter tuning process. The hand-tuned solutions tend to aggregate into one cluster, which means they have similar design qualities. In comparison, the initial samples in subgraph 1 distribute much more sparsely. It is contributed by the cluster-based model-less sampling, which avoids selecting similar samples. After initial sampling, since cluster refinement threshold $\theta$ is set to 40, subgraph 2, 3 perform `exploration' and subgraph 4, 5, 6 perform `exploitation'. The `exploration' and `exploitation' show different effects. In subgraph 2, 3, new samples (green and yellow) slowly move towards the upper left direction, but still distribute sparsely, especially for PE. But in subgraph 4, 5, 6, when model exclusively performs `exploitation', new samples, which now concentrate around new Pareto points, generally have better quality. Notice that all 60 points at this stage outperform the hand-tuned solutions in `area'. By subgraph 6, new samples gradually converge at Pareto point, which means the best point that FIST can find is approximately reached.

\section{Summary}

Design flow parameter tuning is a daunting task and an efficient automatic approach is highly desirable. In this chapter, I present an efficient machine learning approach for automatic parameter tuning. We build a large dataset, from which we developed a clustering-based method to leverage prior data to improve sampling efficiency during exploration. We also introduce approximate sampling and dynamic modeling based on semi-supervised learning and bias-variance trade-off principles. Our approach either improves design quality significantly or requires much less sampling cost to achieve a given design performance compared with prior exploration methods. Finally, we validate our method on two more complicated industrial designs with a much larger parameter space. It improves the best hand-tuned solutions by experienced designers with reasonable budgets.

%% file: txt/7_conclusion.tex
\chapter{Conclusion}

In this dissertation, I study customized machine learning methods for intelligent circuit design and implementation, targeting multiple primary design objectives. For power, I present APOLLO, an automatically developed power modeling framework that unprecedentedly achieves high accuracy, fine temporal resolution, and low hardware implementation cost at the same time. For interconnect and timing, I present Net$^{2}$, providing much more accurate individual net size estimations and pre-placement timing reports than prior works/tools. Targeting IR drop, I present PowerNet, supporting accurate cross-design estimations for both vertorless and vector-based dynamic IR drop. For routability, I present RouteNet, the first CNN/FCN-based estimator supporting both coarse-grained and fine-grained routability estimations. Finally, besides targeting single design stage or single design objective, I present FIST for design flow tuning, optimizing the trade-off among power, performance, and area during both logic synthesis and physical design. My ultimate goal is to achieve superior circuit performance with less design cost, less engineering efforts, shorter turnaround time, and less computation cost.

ML methods have demonstrated great potential for hardware design. But severe challenges also exist and worth exploration in the future. First, it is challenging to get access to adequate training data with sufficient diversity, which limits the development of high-quality and generalized ML estimators. This may require collaborative training on decentralized private design data. Second, multiple security and reliability concerns regarding training data and ML model arise when ML for EDA becomes increasingly popular. Third, efficient deployment of ML models in existing design flows is still challenging. I expect ML models to better integrate with design flow or even completely replace some traditional design automation techniques in the future. 


%% file: dissertation.bbl
\begin{thebibliography}{100}

\bibitem{chen2020survey}
Yiran Chen, Yuan Xie, Linghao Song, Fan Chen, and Tianqi Tang.
\newblock A survey of accelerator architectures for deep neural networks.
\newblock {\em Engineering}, 2020.

\bibitem{moore1965cramming}
Gordon~E Moore et~al.
\newblock Cramming more components onto integrated circuits, 1965.

\bibitem{theis2017end}
Thomas~N Theis and H-S~Philip Wong.
\newblock The end of moore's law: A new beginning for information technology.
\newblock {\em Computing in Science \& Engineering}, 2017.

\bibitem{DesignCost}
IBS.
\newblock As chip design costs skyrocket, 3nm process node is in jeopardy.
\newblock \nolinkurl{https:
  //www.extremetech.com/computing/272096-3nm-process-node}, 2020.

\bibitem{xie2021apollo}
Zhiyao Xie, Xiaoqing Xu, Matt Walker, Joshua Knebel, Kumaraguru Palaniswamy,
  Nicolas Hebert, Jiang Hu, Huanrui Yang, Yiran Chen, and Shidhartha Das.
\newblock {APOLLO}: An automated power modeling framework for runtime power
  introspection in high-volume commercial microprocessors.
\newblock In {\em International Symposium on Microarchitecture (MICRO)}, 2021.

\bibitem{xie2021timing}
Zhiyao Xie, Rongjian Liang, Xiaoqing Xu, Jiang Hu, Chen-Chia Chang, Jingyu Pan,
  and Yiran Chen.
\newblock Pre-placement net length and timing estimation by customized graph
  neural network.
\newblock {\em IEEE Transactions on Computer-Aided Design of Integrated
  Circuits and Systems (TCAD)}, Accepted, 2022.

\bibitem{xie2021net}
Zhiyao Xie, Rongjian Liang, Xiaoqing Xu, Jiang Hu, Yixiao Duan, and Yiran Chen.
\newblock {Net$^\mathbf{2}$}: A graph attention network method customized for
  pre-placement net length estimation.
\newblock In {\em Asia and South Pacific Design Automation Conference
  (ASP-DAC)}, 2021.

\bibitem{xie2020fast}
Zhiyao Xie, Hai Li, Xiaoqing Xu, Jiang Hu, and Yiran Chen.
\newblock Fast {IR} drop estimation with machine learning.
\newblock In {\em International Conference on Computer-Aided Design (ICCAD)},
  2020.

\bibitem{xie2020powernet}
Zhiyao Xie, Haoxing Ren, Brucek Khailany, Ye~Sheng, Santosh Santosh, Jiang Hu,
  and Yiran Chen.
\newblock {PowerNet}: Transferable dynamic {IR} drop estimation via maximum
  convolutional neural network.
\newblock In {\em Asia and South Pacific Design Automation Conference
  (ASP-DAC)}, 2020.

\bibitem{xie2018routenet}
Zhiyao Xie, Yu-Hung Huang, Guan-Qi Fang, Haoxing Ren, Shao-Yun Fang, Yiran
  Chen, and Jiang Hu.
\newblock {RouteNet}: Routability prediction for mixed-size designs using
  convolutional neural network.
\newblock In {\em International Conference On Computer Aided Design (ICCAD)},
  2018.

\bibitem{xie2020fist}
Zhiyao Xie, Guan-Qi Fang, Yu-Hung Huang, Haoxing Ren, Yanqing Zhang, Brucek
  Khailany, Shao-Yun Fang, Jiang Hu, Yiran Chen, and Erick~Carvajal Barboza.
\newblock {FIST}: A feature-importance sampling and tree-based method for
  automatic design flow parameter tuning.
\newblock In {\em Asia and South Pacific Design Automation Conference
  (ASP-DAC)}, 2020.

\bibitem{haj2016fine}
Jawad Haj-Yihia, Ahmad Yasin, Yosi~Ben Asher, and Avi Mendelson.
\newblock {Fine-grain power breakdown of modern out-of-order cores and its
  implications on skylake-based systems}.
\newblock {\em ACM Transactions on Architecture and Code Optimization (TACO)},
  2016.

\bibitem{jacobson2011abstraction}
Hans Jacobson, Alper Buyuktosunoglu, Pradip Bose, Emrah Acar, and Richard
  Eickemeyer.
\newblock {Abstraction and microarchitecture scaling in early-stage power
  modeling}.
\newblock In {\em International Symposium on High-Performance Computer
  Architecture (HPCA)}, 2011.

\bibitem{zhou2019primal}
Yuan Zhou, Haoxing Ren, Yanqing Zhang, Ben Keller, Brucek Khailany, and Zhiru
  Zhang.
\newblock {PRIMAL}: Power inference using machine learning.
\newblock In {\em Design Automation Conference (DAC)}, 2019.

\bibitem{kim2019simmani}
Donggyu Kim, Jerry Zhao, Jonathan Bachrach, and Krste Asanovi{\'c}.
\newblock Simmani: Runtime power modeling for arbitrary {RTL} with automatic
  signal selection.
\newblock In {\em International Symposium on Microarchitecture (MICRO)}, 2019.

\bibitem{huang2012accurate}
Wei Huang, Charles Lefurgy, William Kuk, Alper Buyuktosunoglu, Michael Floyd,
  Karthick Rajamani, Malcolm Allen-Ware, and Bishop Brock.
\newblock {Accurate fine-grained processor power proxies}.
\newblock In {\em International Symposium on Microarchitecture (MICRO)}, 2012.

\bibitem{zoni2018powerprobe}
Davide Zoni, Luca Cremona, and William Fornaciari.
\newblock Powerprobe: Run-time power modeling through automatic {RTL}
  instrumentation.
\newblock In {\em Design, Automation \& Test in Europe Conference \& Exhibition
  (DATE)}, 2018.

\bibitem{godycki2014enabling}
Waclaw Godycki, Christopher Torng, Ivan Bukreyev, Alyssa Apsel, and Christopher
  Batten.
\newblock Enabling realistic fine-grain voltage scaling with reconfigurable
  power distribution networks.
\newblock In {\em International Symposium on Microarchitecture (MICRO)}, 2014.

\bibitem{kasture2015rubik}
Harshad Kasture, Davide~B Bartolini, Nathan Beckmann, and Daniel Sanchez.
\newblock Rubik: Fast analytical power management for latency-critical systems.
\newblock In {\em International Symposium on Microarchitecture (MICRO)}, 2015.

\bibitem{hsu2015adrenaline}
Chang-Hong Hsu, Yunqi Zhang, Michael~A Laurenzano, David Meisner, Thomas
  Wenisch, Jason Mars, Lingjia Tang, and Ronald~G Dreslinski.
\newblock Adrenaline: Pinpointing and reining in tail queries with quick
  voltage boosting.
\newblock In {\em International Symposium on High-Performance Computer
  Architecture (HPCA)}, 2015.

\bibitem{mair2017adaptiveclocking}
Hugh Mair, Ericbill Wang, Alice Wang, Ping Kao, Yuwen Tsai, Sumanth
  Gururajarao, Rolf Lagerquist, Jin Son, Gordon Gammie, Gordon Lin, et~al.
\newblock A 10nm {FinFET} 2.8{GHz} tri-gear deca-core {CPU} complex with
  optimized power-delivery network for mobile {SoC} performance.
\newblock In {\em International Solid-State Circuits Conference (ISSCC)}, 2017.

\bibitem{bowmanJSSCAdaptiveClocking}
Keith~A Bowman, Sarthak Raina, J~Todd Bridges, Daniel~J Yingling, Hoan~H
  Nguyen, Brad~R Appel, Yesh~N Kolla, Jihoon Jeong, Francois~I Atallah, and
  David~W Hansquine.
\newblock A 16 nm all-digital auto-calibrating adaptive clock distribution for
  supply voltage droop tolerance across a wide operating range.
\newblock {\em IEEE Journal of Solid-State Circuits (JSSC)}, 2016.

\bibitem{reddi2009voltage}
Vijay~Janapa Reddi, Meeta~S Gupta, Glenn Holloway, Gu-Yeon Wei, Michael~D
  Smith, and David Brooks.
\newblock Voltage emergency prediction: Using signatures to reduce operating
  margins.
\newblock In {\em International Symposium on High-Performance Computer
  Architecture (HPCA)}, 2009.

\bibitem{webel2020proactive}
T~Webel, PM~Lobo, T~Strach, PB~Parashurama, S~Purushotham, R~Bertran, and
  A~Buyuktosunoglu.
\newblock {Proactive power management in IBM z15}.
\newblock {\em IBM Journal of Research and Development}, 2020.

\bibitem{grochowski2002}
Ed~Grochowski, David Ayers, and Vivek Tiwari.
\newblock Microarchitectural simulation and control of di/dt-induced power
  supply voltage variation.
\newblock In {\em International Symposium on High-Performance Computer
  Architecture (HPCA)}, 2002.

\bibitem{powerpro}
Siemens.
\newblock {PowerPro® RTL Low-Power}.
\newblock \nolinkurl{https://eda.sw.siemens.com/en-US/ic/powerpro/}, 2021.

\bibitem{emulation-platform}
Cadence.
\newblock {Palladium® Z1 Enterprise Emulation Platform}.
\newblock
  \nolinkurl{https://www.cadence.com/en_US/home/tools/system-design-and-verification/acceleration-and-emulation/palladium-z1.html},
  2021.

\bibitem{JosephISLPED01}
R.~Joseph and M.~Martonosi.
\newblock Run-time power estimation in high performance microprocessors.
\newblock In {\em International Symposium on Low Power Electronics and Design
  (ISLPED)}, 2001.

\bibitem{bircher2007complete}
W~Lloyd Bircher and Lizy~K John.
\newblock {Complete system power estimation: A trickle-down approach based on
  performance events}.
\newblock In {\em International Symposium on Performance Analysis of Systems
  and Software (ISPASS)}, 2007.

\bibitem{walker2016accurate}
Matthew~J Walker, Stephan Diestelhorst, Andreas Hansson, Anup~K Das, Sheng
  Yang, Bashir~M Al-Hashimi, and Geoff~V Merrett.
\newblock Accurate and stable run-time power modeling for mobile and embedded
  {CPUs}.
\newblock {\em IEEE Transactions on Computer-Aided Design of Integrated
  Circuits and Systems (TCAD)}, 2016.

\bibitem{najem2016design}
Mohamad Najem, Pascal Benoit, Mohamad El~Ahmad, Gilles Sassatelli, and Lionel
  Torres.
\newblock A design-time method for building cost-effective run-time power
  monitoring.
\newblock {\em IEEE Transactions on Computer-Aided Design of Integrated
  Circuits and Systems (TCAD)}, 2017.

\bibitem{cremona2020automatic}
Luca Cremona, William Fornaciari, and Davide Zoni.
\newblock Automatic identification and hardware implementation of a
  resource-constrained power model for embedded systems.
\newblock {\em Elsevier Sustainable Computing: Informatics and Systems}, 2020.

\bibitem{pagliari2018all}
Daniele~Jahier Pagliari, Valentino Peluso, Yukai Chen, Andrea Calimera, Enrico
  Macii, and Massimo Poncino.
\newblock All-digital embedded meters for on-line power estimation.
\newblock In {\em Design, Automation \& Test in Europe Conference \& Exhibition
  (DATE)}, 2018.

\bibitem{zoni2018powertap}
Davide Zoni, Luca Cremona, Alessandro Cilardo, Mirko Gagliardi, and William
  Fornaciari.
\newblock {PowerTap: All-digital power meter modeling for run-time power
  monitoring}.
\newblock {\em Elsevier Microprocessors and Microsystems (MICPRO)}, 2018.

\bibitem{Fusion2}
Shekhar Kapoor and Mark Richards.
\newblock Getting better results faster with the singular {RTL-to-GDSII}
  product.
\newblock
  \nolinkurl{https://www.synopsys.com/implementation-and-signoff/resources/articles/unified-data-engines-fusion-compiler.html},
  2021.

\bibitem{ispatial}
Cadence.
\newblock Cadence digital full flow optimized to deliver improved quality of
  results with up to 3x faster throughput.
\newblock
  \nolinkurl{https://www.cadence.com/en_US/home/company/newsroom/press-releases/pr/2020/cadence-digital-full-flow-optimized-to-deliver-improved-quality-.html},
  2020.

\bibitem{Genus}
Cadence.
\newblock Cadence {Genus} user guide.
\newblock
  \nolinkurl{https://www.cadence.com/content/dam/cadence-www/global/en_US/documents/tools/digital-design-signoff/genus-synthesis-solution-ds.pdf},
  2019.

\bibitem{Innovus}
Cadence.
\newblock Innovus implementation system.
\newblock
  \nolinkurl{https://www.cadence.com/en_US/home/tools/digital-design-and-signoff/soc-implementation-and-floorplanning/innovus-implementation-system.html},
  2021.

\bibitem{bodapati2001prelayout}
Srinivas Bodapati and Farid~N Najm.
\newblock Prelayout estimation of individual wire lengths.
\newblock {\em IEEE Transactions on Very Large Scale Integration Systems
  (TVLSI)}, 2001.

\bibitem{fathi2009pre}
Bahareh Fathi, Laleh Behjat, and Logan~M Rakai.
\newblock A pre-placement net length estimation technique for mixed-size
  circuits.
\newblock In {\em International Workshop on System Level Interconnect
  Prediction (SLIP)}, 2009.

\bibitem{liu2012neural}
Qiang Liu, Jianguo Ma, and Qijun Zhang.
\newblock Neural network based pre-placement wirelength estimation.
\newblock In {\em International Conference on Field-Programmable Technology
  (FPT)}, 2012.

\bibitem{hyun2019accurate}
Daijoon Hyun, Yuepeng Fan, and Youngsoo Shin.
\newblock Accurate wirelength prediction for placement-aware synthesis through
  machine learning.
\newblock In {\em Design, Automation \& Test in Europe Conference \& Exhibition
  (DATE)}, 2019.

\bibitem{yamato2012fast}
Yuta Yamato, Tomokazu Yoneda, Kazumi Hatayama, and Michiko Inoue.
\newblock A fast and accurate per-cell dynamic {IR}-drop estimation method for
  at-speed scan test pattern validation.
\newblock In {\em International Test Conference (ITC)}, 2012.

\bibitem{ye2014chip}
Fangming Ye, Farshad Firouzi, Yang Yang, Krishnendu Chakrabarty, and Mehdi~B
  Tahoori.
\newblock On-chip voltage-droop prediction using support-vector machines.
\newblock In {\em VLSI Test Symposium (VTS)}, 2014.

\bibitem{lin2018ir}
Shih-Yao Lin, Yen-Chun Fang, Yu-Ching Li, Yu-Cheng Liu, Tsung-Shan Yang,
  Shang-Chien Lin, Chien-Mo Li, and Eric Jia-Wei Fang.
\newblock {IR} drop prediction of {ECO}-revised circuits using machine
  learning.
\newblock In {\em VLSI Test Symposium (VTS)}, 2018.

\bibitem{fang2018machine}
Yen-Chun Fang, Heng-Yi Lin, Min-Yan Sui, Chien-Mo Li, and Eric Jia-Wei Fang.
\newblock Machine-learning-based dynamic {IR} drop prediction for {ECO}.
\newblock In {\em International Conference On Computer Aided Design (ICCAD)},
  2018.

\bibitem{GLARE}
Yaoguang Wei, Cliff Sze, Natarajan Viswanathan, Zhuo Li, Charles~J. Alpert,
  Lakshmi Reddy, Andrew~D. Huber, Gustavo~E. Tellez, Douglas Keller, and
  Sachin~S. Sapatnekar.
\newblock Glare: Global and local wiring aware routability evaluation.
\newblock In {\em \textit{Design Automation Conference} (\textit{DAC})}, 2012.

\bibitem{LouISPD01}
Jinan Lou, Shankar Krishnamoorthy, and Henry~S. Sheng.
\newblock Estimating routing congestion using probabilistic analysis.
\newblock In {\em \textit{ACM International Symposium on Physical Design}
  (\textit{ISPD})}, 2001.

\bibitem{WestraISPD04}
Jurjen Westra, Chris Bartels, and Patrick Groeneveld.
\newblock Probabilistic congestion prediction.
\newblock In {\em \textit{ACM International Symposium on Physical Design}
  (\textit{ISPD})}, 2004.

\bibitem{spindler2007fast}
Peter Spindler and Frank~M Johannes.
\newblock Fast and accurate routing demand estimation for efficient
  routability-driven placement.
\newblock In {\em Design, Automation \& Test in Europe Conference \& Exhibition
  (DATE)}, 2007.

\bibitem{chan2017routability}
Wei-Ting~J Chan, Pei-Hsin Ho, Andrew~B Kahng, and Prashant Saxena.
\newblock Routability optimization for industrial designs at sub-14nm process
  nodes using machine learning.
\newblock In {\em International Symposium on Physical Design (ISPD)}, 2017.

\bibitem{chan2016beol}
Wei-Ting~J Chan, Yang Du, Andrew~B Kahng, Siddhartha Nath, and Kambiz Samadi.
\newblock Beol stack-aware routability prediction from placement using data
  mining techniques.
\newblock In {\em International Conference on Computer Design (ICCD)}, 2016.

\bibitem{zhou2015accurate}
Quan Zhou, Xueyan Wang, Zhongdong Qi, Zhuwei Chen, Qiang Zhou, and Yici Cai.
\newblock An accurate detailed routing routability prediction model in
  placement.
\newblock In {\em Asia Symposium on Quality Electronic Design (ASQED)}, 2015.

\bibitem{ziegler2016synthesis}
Matthew~M Ziegler, Hung-Yi Liu, George Gristede, Bruce Owens, Ricardo
  Nigaglioni, and Luca~P Carloni.
\newblock A synthesis-parameter tuning system for autonomous design-space
  exploration.
\newblock In {\em Design, Automation \& Test in Europe Conference \& Exhibition
  (\textit{DATE})}, 2016.

\bibitem{ziegler2016scalable}
Matthew~M Ziegler, Hung-Yi Liu, and Luca~P Carloni.
\newblock Scalable auto-tuning of synthesis parameters for optimizing
  high-performance processors.
\newblock In {\em International Symposium on Low Power Electronics and Design
  (ISLPED)}, 2016.

\bibitem{papamichael2015nautilus}
Michael~K Papamichael, Peter Milder, and James~C Hoe.
\newblock Nautilus: Fast automated {IP} design space search using guided
  genetic algorithms.
\newblock In {\em Design Automation Conference (\textit{DAC})}, 2015.

\bibitem{chen2016xgboost}
Tianqi Chen and Carlos Guestrin.
\newblock {Xgboost: A scalable tree boosting system}.
\newblock In {\em ACM SIGKDD International Conference on Knowledge Discovery
  and Data Mining (KDD)}, 2016.

\bibitem{huang2021machine}
Guyue Huang, Jingbo Hu, Yifan He, Jialong Liu, Mingyuan Ma, Zhaoyang Shen,
  Juejian Wu, Yuanfan Xu, Hengrui Zhang, Kai Zhong, et~al.
\newblock Machine learning for electronic design automation: A survey.
\newblock {\em ACM Transactions on Design Automation of Electronic Systems
  (TODAES)}, 2021.

\bibitem{rapp2021mlcad}
Martin Rapp, Hussam Amrouch, Yibo Lin, Bei Yu, David~Z Pan, Marilyn Wolf, and
  J{\"o}rg Henkel.
\newblock {MLCAD}: A survey of research in machine learning for {CAD} keynote
  paper.
\newblock {\em IEEE Transactions on Computer-Aided Design of Integrated
  Circuits and Systems (TCAD)}, 2021.

\bibitem{dai2018fast}
Steve Dai, Yuan Zhou, Hang Zhang, Ecenur Ustun, Evangeline~FY Young, and Zhiru
  Zhang.
\newblock Fast and accurate estimation of quality of results in high-level
  synthesis with machine learning.
\newblock In {\em Symposium on Field-Programmable Custom Computing Machines
  (FCCM)}, 2018.

\bibitem{makrani2019pyramid}
Hosein~Mohammadi Makrani, Farnoud Farahmand, Hossein Sayadi, Sara Bondi, Sai
  Manoj~Pudukotai Dinakarrao, Houman Homayoun, and Setareh Rafatirad.
\newblock Pyramid: Machine learning framework to estimate the optimal timing
  and resource usage of a high-level synthesis design.
\newblock In {\em International Conference on Field-Programmable Logic and
  Applications (FPL)}, 2019.

\bibitem{liu2013learning}
Hung-Yi Liu and Luca~P Carloni.
\newblock On learning-based methods for design-space exploration with
  high-level synthesis.
\newblock In {\em Design automation conference (DAC)}, 2013.

\bibitem{liu2016efficient}
Dong Liu and Benjamin~Carrion Schafer.
\newblock Efficient and reliable high-level synthesis design space explorer for
  {FPGAs}.
\newblock In {\em International Conference on Field Programmable Logic and
  Applications (FPL)}, 2016.

\bibitem{zhang2020grannite}
Yanqing Zhang, Haoxing Ren, and Brucek Khailany.
\newblock {GRANNITE}: Graph neural network inference for transferable power
  estimation.
\newblock In {\em Design Automation Conference (DAC)}, 2020.

\bibitem{yu2018developing}
Cunxi Yu, Houping Xiao, and Giovanni De~Micheli.
\newblock Developing synthesis flows without human knowledge.
\newblock In {\em Design Automation Conference (DAC)}, 2018.

\bibitem{neto2019lsoracle}
Walter~Lau Neto, Max Austin, Scott Temple, Luca Amaru, Xifan Tang, and
  Pierre-Emmanuel Gaillardon.
\newblock {LSOracle}: A logic synthesis framework driven by artificial
  intelligence.
\newblock In {\em International Conference On Computer Aided Design (ICCAD)},
  2019.

\bibitem{hosny2020drills}
Abdelrahman Hosny, Soheil Hashemi, Mohamed Shalan, and Sherief Reda.
\newblock Drills: Deep reinforcement learning for logic synthesis.
\newblock In {\em Asia and South Pacific Design Automation Conference
  (ASP-DAC)}, 2020.

\bibitem{neto2021slap}
Walter~Lau Neto, Matheus~T Moreira, Yingjie Li, Luca Amar{\`u}, Cunxi Yu, and
  Pierre-Emmanuel Gaillardon.
\newblock {SLAP}: A supervised learning approach for priority cuts technology
  mapping.
\newblock In {\em Design Automation Conference (DAC)}, 2021.

\bibitem{barboza2019machine}
Erick~Carvajal Barboza, Nishchal Shukla, Yiran Chen, and Jiang Hu.
\newblock Machine learning-based pre-routing timing prediction with reduced
  pessimism.
\newblock In {\em Design Automation Conference (DAC)}, 2019.

\bibitem{kahng2015si}
Andrew~B Kahng, Mulong Luo, and Siddhartha Nath.
\newblock {SI} for free: machine learning of interconnect coupling delay and
  transition effects.
\newblock In {\em International Workshop on System Level Interconnect
  Prediction (SLIP)}, 2015.

\bibitem{mirhoseini2021graph}
Azalia Mirhoseini, Anna Goldie, Mustafa Yazgan, Joe~Wenjie Jiang, Ebrahim
  Songhori, Shen Wang, Young-Joon Lee, Eric Johnson, Omkar Pathak, Azade Nazi,
  et~al.
\newblock A graph placement methodology for fast chip design.
\newblock {\em Nature}, 2021.

\bibitem{chen2020pros}
Jingsong Chen, Jian Kuang, Guowei Zhao, Dennis J-H Huang, and Evangeline~FY
  Young.
\newblock {PROS}: A plug-in for routability optimization applied in the
  state-of-the-art commercial {EDA} tool using deep learning.
\newblock In {\em International Conference On Computer Aided Design (ICCAD)},
  2020.

\bibitem{huang2019routability}
Yu-Hung Huang, Zhiyao Xie, Guan-Qi Fang, Tao-Chun Yu, Haoxing Ren, Shao-Yun
  Fang, Yiran Chen, and Jiang Hu.
\newblock Routability-driven macro placement with embedded cnn-based prediction
  model.
\newblock In {\em Design, Automation \& Test in Europe Conference \& Exhibition
  (DATE)}, 2019.

\bibitem{chang2021auto}
Chen-Chia Chang, Jingyu Pan, Tunhou Zhang, Zhiyao Xie, Jiang Hu, Weiyi Qi,
  Chunwei Lin, Rongjian Liang, Joydeep Mitra, Elias Fallon, and Yiran Chen.
\newblock Automatic routability predictor development using neural architecture
  search.
\newblock In {\em International Conference On Computer Aided Design (ICCAD)},
  2021.

\bibitem{pan2022towards}
Jingyu Pan, Chen-Chia Chang, Zhiyao Xie, Ang Li, Minxue Tang, Tunhou Zhang,
  Jiang Hu, and Yiran Chen.
\newblock Towards collaborative intelligence: Routability estimation based on
  decentralized private data.
\newblock In {\em Design Automation Conference (DAC)}, 2022.

\bibitem{ho2019incpird}
Chia-Tung Ho and Andrew~B Kahng.
\newblock {IncPIRD}: Fast learning-based prediction of incremental {IR} drop.
\newblock In {\em International Conference on Computer-Aided Design (ICCAD)},
  2019.

\bibitem{zhou2020gridnet}
Han Zhou, Wentian Jin, and Sheldon X-D Tan.
\newblock {GridNet}: Fast data-driven {EM}-induced {IR} drop prediction and
  localized fixing for on-chip power grid networks.
\newblock In {\em International Conference On Computer Aided Design (ICCAD)},
  2020.

\bibitem{lu2019gan}
Yi-Chen Lu, Jeehyun Lee, Anthony Agnesina, Kambiz Samadi, and Sung~Kyu Lim.
\newblock {GAN-CTS}: A generative adversarial framework for clock tree
  prediction and optimization.
\newblock In {\em International Conference On Computer Aided Design (ICCAD)},
  2019.

\bibitem{liang2020routing}
Rongjian Liang, Zhiyao Xie, Jinwook Jung, Vishnavi Chauha, Yiran Chen, Jiang
  Hu, Hua Xiang, and Gi-Joon Nam.
\newblock Routing-free crosstalk prediction.
\newblock In {\em International Conference on Computer Aided Design (ICCAD)},
  2020.

\bibitem{lu2020tp}
Yi-Chen Lu, Sai Surya~Kiran Pentapati, Lingjun Zhu, Kambiz Samadi, and Sung~Kyu
  Lim.
\newblock {TP-GNN}: A graph neural network framework for tier partitioning in
  monolithic 3d {ICs}.
\newblock In {\em Design Automation Conference (DAC)}, 2020.

\bibitem{katz2011learning}
Yoav Katz, Michal Rimon, Avi Ziv, and Gai Shaked.
\newblock Learning microarchitectural behaviors to improve stimuli generation
  quality.
\newblock In {\em Design Automation Conference (DAC)}, 2011.

\bibitem{fine2003coverage}
Shai Fine and Avi Ziv.
\newblock Coverage directed test generation for functional verification using
  bayesian networks.
\newblock In {\em Design Automation Conference (DAC)}, 2003.

\bibitem{ma2019high}
Yuzhe Ma, Haoxing Ren, Brucek Khailany, Harbinder Sikka, Lijuan Luo,
  Karthikeyan Natarajan, and Bei Yu.
\newblock High performance graph convolutional networks with applications in
  testability analysis.
\newblock In {\em Design Automation Conference (DAC)}, 2019.

\bibitem{yang2018layout}
Haoyu Yang, Jing Su, Yi~Zou, Yuzhe Ma, Bei Yu, and Evangeline~FY Young.
\newblock Layout hotspot detection with feature tensor generation and deep
  biased learning.
\newblock {\em IEEE Transactions on Computer-Aided Design of Integrated
  Circuits and Systems (TCAD)}, 2018.

\bibitem{yang2019gan}
Haoyu Yang, Shuhe Li, Zihao Deng, Yuzhe Ma, Bei Yu, and Evangeline~FY Young.
\newblock {GAN-OPC}: Mask optimization with lithography-guided generative
  adversarial nets.
\newblock {\em IEEE Transactions on Computer-Aided Design of Integrated
  Circuits and Systems (TCAD)}, 2019.

\bibitem{ye2019lithogan}
Wei Ye, Mohamed~Baker Alawieh, Yibo Lin, and David~Z Pan.
\newblock {LithoGAN}: End-to-end lithography modeling with generative
  adversarial networks.
\newblock In {\em Design Automation Conference (DAC)}, 2019.

\bibitem{kwon2019learning}
Jihye Kwon, Matthew~M Ziegler, and Luca~P Carloni.
\newblock A learning-based recommender system for autotuning design fiows of
  industrial high-performance processors.
\newblock In {\em Design Automation Conference (DAC)}, 2019.

\bibitem{kunal2020general}
Kishor Kunal, Jitesh Poojary, Tonmoy Dhar, Meghna Madhusudan, Ramesh Harjani,
  and Sachin~S Sapatnekar.
\newblock A general approach for identifying hierarchical symmetry constraints
  for analog circuit layout.
\newblock In {\em International Conference On Computer Aided Design (ICCAD)},
  2020.

\bibitem{li2016analog}
Hao Li, Fanshu Jiao, and Alex Doboli.
\newblock Analog circuit topological feature extraction with unsupervised
  learning of new sub-structures.
\newblock In {\em Design, Automation \& Test in Europe Conference \& Exhibition
  (DATE)}, 2016.

\bibitem{wang2020gcn}
Hanrui Wang, Kuan Wang, Jiacheng Yang, Linxiao Shen, Nan Sun, Hae-Seung Lee,
  and Song Han.
\newblock {GCN-RL} circuit designer: Transferable transistor sizing with graph
  neural networks and reinforcement learning.
\newblock In {\em Design Automation Conference (DAC)}, 2020.

\bibitem{hakhamaneshi2019bagnet}
Kourosh Hakhamaneshi, Nick Werblun, Pieter Abbeel, and Vladimir Stojanovi{\'c}.
\newblock {BagNet}: Berkeley analog generator with layout optimizer boosted
  with deep neural networks.
\newblock In {\em International Conference on Computer-Aided Design (ICCAD)},
  2019.

\bibitem{ren2020paragraph}
Haoxing Ren, George~F Kokai, Walker~J Turner, and Ting-Sheng Ku.
\newblock {ParaGraph}: Layout parasitics and device parameter prediction using
  graph neural networks.
\newblock In {\em Design Automation Conference (DAC)}, 2020.

\bibitem{kunal2020gana}
Kishor Kunal, Tonmoy Dhar, Meghna Madhusudan, Jitesh Poojary, Arvind Sharma,
  Wenbin Xu, Steven~M Burns, Jiang Hu, Ramesh Harjani, and Sachin~S Sapatnekar.
\newblock {GANA}: Graph convolutional network based automated netlist
  annotation for analog circuits.
\newblock In {\em Design, Automation \& Test in Europe Conference \& Exhibition
  (DATE)}, 2020.

\bibitem{liu2020towards}
Mingjie Liu, Keren Zhu, Jiaqi Gu, Linxiao Shen, Xiyuan Tang, Nan Sun, and
  David~Z Pan.
\newblock Towards decrypting the art of analog layout: Placement quality
  prediction via transfer learning.
\newblock In {\em Design, Automation \& Test in Europe Conference \& Exhibition
  (DATE)}, 2020.

\bibitem{li2020customized}
Yaguang Li, Yishuang Lin, Meghna Madhusudan, Arvind Sharma, Wenbin Xu, Sachin~S
  Sapatnekar, Ramesh Harjani, and Jiang Hu.
\newblock A customized graph neural network model for guiding analog {IC}
  placement.
\newblock In {\em International Conference On Computer Aided Design (ICCAD)},
  2020.

\bibitem{xu2019wellgan}
Biying Xu, Yibo Lin, Xiyuan Tang, Shaolan Li, Linxiao Shen, Nan Sun, and
  David~Z Pan.
\newblock Wellgan: Generative-adversarial-network-guided well generation for
  analog/mixed-signal circuit layout.
\newblock In {\em Design Automation Conference (DAC)}, 2019.

\bibitem{liu2020closing}
Mingjie Liu, Keren Zhu, Xiyuan Tang, Biying Xu, Wei Shi, Nan Sun, and David~Z
  Pan.
\newblock Closing the design loop: Bayesian optimization assisted hierarchical
  analog layout synthesis.
\newblock In {\em Design Automation Conference (DAC)}, 2020.

\bibitem{zhu2019geniusroute}
Keren Zhu, Mingjie Liu, Yibo Lin, Biying Xu, Shaolan Li, Xiyuan Tang, Nan Sun,
  and David~Z Pan.
\newblock Geniusroute: A new analog routing paradigm using generative neural
  network guidance.
\newblock In {\em International Conference on Computer-Aided Design (ICCAD)},
  2019.

\bibitem{stratigopoulos2008error}
Haralampos-G Stratigopoulos and Yiorgos Makris.
\newblock Error moderation in low-cost machine-learning-based analog/{RF}
  testing.
\newblock {\em IEEE Transactions on Computer-Aided Design of Integrated
  Circuits and Systems (TCAD)}, 2008.

\bibitem{ICC2}
Synopsys.
\newblock {IC Compiler II} for physical implementation.
\newblock
  \nolinkurl{https://www.synopsys.com/implementation-and-signoff/physical-implementation/ic-compiler.html},
  2021.

\bibitem{dsoai}
Synopsys.
\newblock {DSO.ai}: {AI}-driven design applications.
\newblock
  \nolinkurl{https://www.synopsys.com/implementation-and-signoff/ml-ai-design/dso-ai.html},
  2021.

\bibitem{cerebrus}
Cadence.
\newblock {Cadence Cerebrus} intelligent chip explorer.
\newblock
  \nolinkurl{https://www.cadence.com/en_US/home/tools/digital-design-and-signoff/soc-implementation-and-floorplanning/cerebrus-intelligent-chip-explorer.html},
  2021.

\bibitem{brooks2000wattch}
David Brooks, Vivek Tiwari, and Margaret Martonosi.
\newblock Wattch: A framework for architectural-level power analysis and
  optimizations.
\newblock {\em ACM SIGARCH Computer Architecture News}, 2000.

\bibitem{lee2006accurate}
Benjamin~C Lee and David~M Brooks.
\newblock Accurate and efficient regression modeling for microarchitectural
  performance and power prediction.
\newblock {\em ACM SIGOPS Operating Systems Review}, 2006.

\bibitem{li2009mcpat}
Sheng Li, Jung~Ho Ahn, Richard~D Strong, Jay~B Brockman, Dean~M Tullsen, and
  Norman~P Jouppi.
\newblock {McPAT: an integrated power, area, and timing modeling framework for
  multicore and manycore architectures}.
\newblock In {\em International Symposium on Microarchitecture (MICRO)}, 2009.

\bibitem{rethinagiri2014system}
Santhosh~Kumar Rethinagiri, Oscar Palomar, Rabie Ben~Atitallah, Smail Niar,
  Osman Unsal, and Adrian~Cristal Kestelman.
\newblock System-level power estimation tool for embedded processor based
  platforms.
\newblock In {\em Workshop on Rapid Simulation and Performance Evaluation:
  Methods and Tools (RAPIDO)}, 2014.

\bibitem{bogliolo2000regression}
Alessandro Bogliolo, Luca Benini, and Giovanni De~Micheli.
\newblock Regression-based {RTL} power modeling.
\newblock {\em ACM Transactions on Design Automation of Electronic Systems
  (TODAES)}, 2000.

\bibitem{shao2014aladdin}
Yakun~Sophia Shao, Brandon Reagen, Gu-Yeon Wei, and David Brooks.
\newblock Aladdin: A pre-{RTL}, power-performance accelerator simulator
  enabling large design space exploration of customized architectures.
\newblock In {\em International Symposium on Computer Architecture (ISCA)},
  2014.

\bibitem{brooks2003new}
David Brooks, Pradip Bose, Viji Srinivasan, Michael~K Gschwind, Philip~G Emma,
  and Michael~G Rosenfield.
\newblock New methodology for early-stage, microarchitecture-level
  power-performance analysis of microprocessors.
\newblock {\em IBM Journal of Research and Development}, 2003.

\bibitem{lee2015dynamic}
Dongwook Lee, Lizy~K John, and Andreas Gerstlauer.
\newblock Dynamic power and performance back-annotation for fast and accurate
  functional hardware simulation.
\newblock In {\em Design, Automation \& Test in Europe Conference \& Exhibition
  (DATE)}, 2015.

\bibitem{kumar2019learning}
Ajay Krishna~Ananda Kumar and Andreas Gerstlauer.
\newblock Learning-based {CPU} power modeling.
\newblock In {\em Workshop on Machine Learning for CAD (MLCAD)}, 2019.

\bibitem{ye2000design}
Wu~Ye, Narayanan Vijaykrishnan, Mahmut Kandemir, and Mary~Jane Irwin.
\newblock {The design and use of SimplePower: A cycle-accurate energy
  estimation tool}.
\newblock In {\em Design Automation Conference (DAC)}, 2000.

\bibitem{wu1998cycle}
Qing Wu, Qinru Qiu, Massoud Pedram, and Chih-Shun Ding.
\newblock Cycle-accurate macro-models for {RT-level} power analysis.
\newblock {\em IEEE Transactions on Very Large Scale Integration Systems
  (TVLSI)}, 1998.

\bibitem{coburn2005power}
Joel Coburn, Srivaths Ravi, and Anand Raghunathan.
\newblock {Power emulation: a new paradigm for power estimation}.
\newblock In {\em Design Automation Conference (DAC)}, 2005.

\bibitem{yang2015early}
Jianlei Yang, Liwei Ma, Kang Zhao, Yici Cai, and Tin-Fook Ngai.
\newblock {Early stage real-time SoC power estimation using RTL
  instrumentation}.
\newblock In {\em Asia and South Pacific Design Automation Conference
  (ASP-DAC)}, 2015.

\bibitem{sunwoo2010presto}
Dam Sunwoo, Gene~Y Wu, Nikhil~A Patil, and Derek Chiou.
\newblock {PrEsto: An FPGA-accelerated power estimation methodology for complex
  systems}.
\newblock In {\em International Conference on Field Programmable Logic and
  Applications (FPL)}, 2010.

\bibitem{bellosa2000benefits}
Frank Bellosa.
\newblock {The benefits of event: driven energy accounting in power-sensitive
  systems}.
\newblock In {\em ACM SIGOPS European Workshop (EW)}, 2000.

\bibitem{bertran2010decomposable}
Ramon Bertran, Marc Gonzalez, Xavier Martorell, Nacho Navarro, and Eduard
  Ayguade.
\newblock Decomposable and responsive power models for multicore processors
  using performance counters.
\newblock In {\em ACM International Conference on Supercomputing (ICS)}, 2010.

\bibitem{gilberto2005power}
C~Gilberto and M~Margaret.
\newblock Power prediction for intel xscale processors using performance
  monitoring unit events.
\newblock In {\em International Symposium on Low Power Electronics and Design
  (ISLPED)}, 2005.

\bibitem{goel2010portable}
Bhavishya Goel, Sally~A McKee, Roberto Gioiosa, Karan Singh, Major Bhadauria,
  and Marco Cesati.
\newblock Portable, scalable, per-core power estimation for intelligent
  resource management.
\newblock In {\em International Conference on Green Computing (IGCC)}, 2010.

\bibitem{isci2003runtime}
Canturk Isci and Margaret Martonosi.
\newblock Runtime power monitoring in high-end processors: Methodology and
  empirical data.
\newblock In {\em International Symposium on Microarchitecture (MICRO)}, 2003.

\bibitem{oboril2015high}
Fabian Oboril, Jos Ewert, and Mehdi~B Tahoori.
\newblock High-resolution online power monitoring for modern microprocessors.
\newblock In {\em Design, Automation \& Test in Europe Conference \& Exhibition
  (DATE)}, 2015.

\bibitem{pricopi2013power}
Mihai Pricopi, Thannirmalai~Somu Muthukaruppan, Vanchinathan Venkataramani,
  Tulika Mitra, and Sanjay Vishin.
\newblock Power-performance modeling on asymmetric multi-cores.
\newblock In {\em International Conference on Compilers, Architecture and
  Synthesis for Embedded Systems (CASES)}, 2013.

\bibitem{rodrigues2013study}
Rance Rodrigues, Arunachalam Annamalai, Israel Koren, and Sandip Kundu.
\newblock A study on the use of performance counters to estimate power in
  microprocessors.
\newblock {\em IEEE Transactions on Circuits and Systems II: Express Briefs
  (TCAS-II)}, 2013.

\bibitem{sagi2020lightweight}
Mark Sagi, Nguyen Anh~Vu Doan, Martin Rapp, Thomas Wild, J{\"o}rg Henkel, and
  Andreas Herkersdorf.
\newblock A lightweight nonlinear methodology to accurately model multicore
  processor power.
\newblock {\em IEEE Transactions on Computer-Aided Design of Integrated
  Circuits and Systems (TCAD)}, 2020.

\bibitem{singh2009real}
Karan Singh, Major Bhadauria, and Sally~A McKee.
\newblock Real time power estimation and thread scheduling via performance
  counters.
\newblock {\em ACM SIGARCH Computer Architecture News}, 2009.

\bibitem{kalyanam2017power}
Vijay~Kiran Kalyanam, Peter~G Sassone, and Jacob~A Abraham.
\newblock Power prediction of embedded scalar and vector processor: Challenges
  and solutions.
\newblock In {\em International Symposium on Quality Electronic Design
  (ISQED)}, 2017.

\bibitem{binkert2011gem5}
Nathan Binkert, Bradford Beckmann, Gabriel Black, Steven~K Reinhardt, Ali
  Saidi, Arkaprava Basu, Joel Hestness, Derek~R Hower, Tushar Krishna, Somayeh
  Sardashti, et~al.
\newblock The gem5 simulator.
\newblock {\em ACM SIGARCH Computer Architecture News}, 2011.

\bibitem{binkert2006m5}
Nathan~L Binkert, Ronald~G Dreslinski, Lisa~R Hsu, Kevin~T Lim, Ali~G Saidi,
  and Steven~K Reinhardt.
\newblock The {M5} simulator: Modeling networked systems.
\newblock {\em IEEE Micro}, 2006.

\bibitem{xi2015quantifying}
Sam~Likun Xi, Hans Jacobson, Pradip Bose, Gu-Yeon Wei, and David Brooks.
\newblock Quantifying sources of error in {McPAT} and potential impacts on
  architectural studies.
\newblock In {\em International Symposium on High-Performance Computer
  Architecture (HPCA)}, 2015.

\bibitem{lee2015powertrain}
Wooseok Lee, Youngchun Kim, Jee~Ho Ryoo, Dam Sunwoo, Andreas Gerstlauer, and
  Lizy~K John.
\newblock {PowerTrain: A learning-based calibration of McPAT power models}.
\newblock In {\em International Symposium on Low Power Electronics and Design
  (ISLPED)}, 2015.

\bibitem{kipf2017semi}
Thomas~N. Kipf and Max Welling.
\newblock Semi-supervised classification with graph convolutional networks.
\newblock In {\em International Conference on Learning Representations (ICLR)},
  2017.

\bibitem{bhagavatula2012low}
Srikar Bhagavatula and Byunghoo Jung.
\newblock {A low power real-time on-chip power sensor in 45-nm SOI}.
\newblock {\em IEEE Transactions on Circuits and Systems I: Regular Papers
  (TCAS-I)}, 2012.

\bibitem{bhagavatula2013power}
Srikar Bhagavatula and Byunghoo Jung.
\newblock A power sensor with 80ns response time for power management in
  microprocessors.
\newblock In {\em Custom Integrated Circuits Conference (CICC)}, 2013.

\bibitem{kalyanam2020proactive}
Vijay~Kiran Kalyanam, Eric Mahurin, Keith Bowman, and Jacob Abraham.
\newblock A proactive voltage-droop-mitigation system in a 7nm {Hexagon™}
  processor.
\newblock In {\em IEEE Symposium on VLSI Circuits (VLSI)}, 2020.

\bibitem{hadjilambrou2019gest}
Zacharias Hadjilambrou, Shidhartha Das, Paul~N Whatmough, David Bull, and
  Yiannakis Sazeides.
\newblock {GeST: An automatic framework for generating CPU stress-tests}.
\newblock In {\em International Symposium on Performance Analysis of Systems
  and Software (ISPASS)}, 2019.

\bibitem{tibshirani1996regression}
Robert Tibshirani.
\newblock Regression shrinkage and selection via the lasso.
\newblock {\em Journal of the Royal Statistical Society: Series B
  (Methodological)}, 1996.

\bibitem{zhang2010nearly}
Cun-Hui Zhang.
\newblock Nearly unbiased variable selection under minimax concave penalty.
\newblock {\em The Annals of statistics}, 2010.

\bibitem{parikh2014proximal}
Neal Parikh and Stephen Boyd.
\newblock Proximal algorithms.
\newblock {\em Foundations and Trends in optimization}, 2014.

\bibitem{wright2015coordinate}
Stephen~J Wright.
\newblock Coordinate descent algorithms.
\newblock {\em Mathematical Programming}, 2015.

\bibitem{wen2018survey}
Fei Wen, Lei Chu, Peilin Liu, and Robert~C Qiu.
\newblock A survey on nonconvex regularization-based sparse and low-rank
  recovery in signal processing, statistics, and machine learning.
\newblock {\em IEEE Access}, 2018.

\bibitem{hoerl1970ridge}
Arthur~E Hoerl and Robert~W Kennard.
\newblock Ridge regression: applications to nonorthogonal problems.
\newblock {\em Technometrics}, 1970.

\bibitem{catapult-hls}
Siemens.
\newblock Catapult® high-level synthesis.
\newblock
  \nolinkurl{https://eda.sw.siemens.com/en-US/ic/ic-design/high-level-synthesis-and-verification-platform/},
  2021.

\bibitem{design-compilier}
Synopsys.
\newblock {Design Compiler® RTL Synthesis}.
\newblock
  \nolinkurl{https://www.synopsys.com/implementation-and-signoff/rtl-synthesis-test/design-compiler-nxt.html},
  2021.

\bibitem{arm-neoverse-n1-trm}
Arm.
\newblock {Arm Neoverse N1 Core Technical Reference Manual}.
\newblock \nolinkurl{https://developer.arm.com/documentation/100616/0301},
  2021.

\bibitem{arm-neoverse-n1}
Andrea Pellegrini, Nigel Stephens, Magnus Bruce, Yasuo Ishii, Joseph Pusdesris,
  Abhishek Raja, Chris Abernathy, Jinson Koppanalil, Tushar Ringe, Ashok
  Tummala, et~al.
\newblock {The Arm Neoverse N1 Platform: Building Blocks for the Next-Gen
  Cloud-to-Edge Infrastructure SoC}.
\newblock {\em IEEE Micro}, 2020.

\bibitem{christy20208}
Robert Christy, Stuart Riches, Sujil Kottekkat, Prasanth Gopinath, Ketan
  Sawant, Anitha Kona, and Rob Harrison.
\newblock {A 3GHz Arm Neoverse N1 CPU in 7nm FinFET for infrastructure
  applications}.
\newblock In {\em International Solid-State Circuits Conference (ISSCC)}, 2020.

\bibitem{arm-cortex-a77-trm}
Arm.
\newblock {Arm Cortex-A77 Core Technical Reference Manual}.
\newblock
  \nolinkurl{https://developer.arm.com/documentation/101111/latest/preface},
  2021.

\bibitem{vcs}
Synopsys.
\newblock {VCS® functional verification solution}.
\newblock
  \nolinkurl{https://www.synopsys.com/verification/simulation/vcs.html}, 2021.

\bibitem{paszke2019pytorch}
Adam Paszke, Sam Gross, Francisco Massa, Adam Lerer, James Bradbury, Gregory
  Chanan, Trevor Killeen, Zeming Lin, Natalia Gimelshein, Luca Antiga, et~al.
\newblock {PyTorch}: An imperative style, high-performance deep learning
  library.
\newblock {\em International Conference on Neural Information Processing
  Systems (NeurIPS)}, 2019.

\bibitem{scikit-learn}
Fabian Pedregosa, Ga{\"e}l Varoquaux, Alexandre Gramfort, Vincent Michel,
  Bertrand Thirion, Olivier Grisel, Mathieu Blondel, Peter Prettenhofer, Ron
  Weiss, Vincent Dubourg, et~al.
\newblock {Scikit-learn: Machine learning in Python}.
\newblock {\em Journal of Machine Learning Research (JMLR)}, 2011.

\bibitem{harris2020array}
Charles~R. Harris, K.~Jarrod Millman, St{'{e}}fan~J. van~der Walt, Ralf
  Gommers, Pauli Virtanen, David Cournapeau, et~al.
\newblock Array programming with {NumPy}.
\newblock {\em Nature}, 2020.

\bibitem{nagelkerke1991note}
Nico~JD Nagelkerke et~al.
\newblock A note on a general definition of the coefficient of determination.
\newblock {\em Biometrika}, 1991.

\bibitem{xuegong2000introduction}
Zhang Xuegong.
\newblock Introduction to statistical learning theory and support vector
  machines.
\newblock {\em Acta Automatica Sinica}, 2000.

\bibitem{lee1988thirteen}
Joseph Lee~Rodgers and W~Alan Nicewander.
\newblock Thirteen ways to look at the correlation coefficient.
\newblock {\em The American Statistician}, 1988.

\bibitem{MagenSLIP04}
Nir Magen, Avinoam Kolodny, Uri Weiser, and Nachum Shamir.
\newblock Interconnect-power dissipation in a microprocessor.
\newblock In {\em International Workshop on System Level Interconnect
  Prediction (SLIP)}, 2004.

\bibitem{PedramDAC91}
Massoud Pedram and Narasimha Bhat.
\newblock Layout driven technology mapping.
\newblock In {\em Design Automation Conference (DAC)}, 1991.

\bibitem{PedramICCAD91}
Massoud Pedram and Narasimha~B Bhat.
\newblock Layout driven logic restructuring/decomposition.
\newblock In {\em International Conference On Computer Aided Design (ICCAD)},
  1991.

\bibitem{hu2003wire}
Bo~Hu and Malgorzata Marek-Sadowska.
\newblock Wire length prediction based clustering and its application in
  placement.
\newblock In {\em Design Automation Conference (DAC)}, 2003.

\bibitem{kahng2005intrinsic}
Andrew~B Kahng and Sherief Reda.
\newblock Intrinsic shortest path length: A new, accurate a priori wirelength
  estimator.
\newblock In {\em International Conference On Computer Aided Design (ICCAD)},
  2005.

\bibitem{liu2004pre}
Qinghua Liu and Malgorzata Marek-Sadowska.
\newblock Pre-layout wire length and congestion estimation.
\newblock In {\em Design Automation Conference (DAC)}, 2004.

\bibitem{velivckovic2017graph}
Petar Veli{\v{c}}kovi{\'c} et~al.
\newblock Graph attention networks.
\newblock In {\em International Conference on Learning Representations (ICLR)},
  2017.

\bibitem{kipf2016semi}
Thomas~N Kipf and Max Welling.
\newblock Semi-supervised classification with graph convolutional networks.
\newblock In {\em International Conference on Learning Representations (ICLR)},
  2017.

\bibitem{hamilton2017inductive}
Will Hamilton, Zhitao Ying, and Jure Leskovec.
\newblock Inductive representation learning on large graphs.
\newblock In {\em International Conference on Neural Information Processing
  Systems (NeurIPS)}, 2017.

\bibitem{zhang2019circuit}
Guo Zhang, Hao He, and Dina Katabi.
\newblock {Circuit-GNN}: Graph neural networks for distributed circuit design.
\newblock In {\em International Conference on Machine Learning (ICML)}, 2019.

\bibitem{kahng2013learning}
Andrew~B Kahng, Seokhyeong Kang, Hyein Lee, Siddhartha Nath, and Jyoti
  Wadhwani.
\newblock Learning-based approximation of interconnect delay and slew in
  signoff timing tools.
\newblock In {\em International Workshop on System Level Interconnect
  Prediction (SLIP)}, 2013.

\bibitem{han2014deep}
Seung-Soo Han, Andrew~B Kahng, Siddhartha Nath, and Ashok~S Vydyanathan.
\newblock A deep learning methodology to proliferate golden signoff timing.
\newblock In {\em Design, Automation \& Test in Europe Conference \& Exhibition
  (DATE)}, 2014.

\bibitem{karypis1999multilevel}
George Karypis, Rajat Aggarwal, Vipin Kumar, and Shashi Shekhar.
\newblock Multilevel hypergraph partitioning: Applications in {VLSI} domain.
\newblock {\em IEEE Transactions on Very Large Scale Integration Systems
  (TVLSI)}, 1999.

\bibitem{he2016deep}
Kaiming He, Xiangyu Zhang, Shaoqing Ren, and Jian Sun.
\newblock Deep residual learning for image recognition.
\newblock In {\em Conference on Computer Vision and Pattern Recognition
  (CVPR)}, 2016.

\bibitem{ronneberger2015u}
Olaf Ronneberger, Philipp Fischer, and Thomas Brox.
\newblock {U-Net}: Convolutional networks for biomedical image segmentation.
\newblock In {\em International Conference on Medical Image Computing and
  Computer Assisted Intervention (MICCAI)}, 2015.

\bibitem{breiman2001random}
Leo Breiman.
\newblock Random forests.
\newblock {\em Machine learning}, 2001.

\bibitem{NanGate}
Si2.
\newblock {NanGate} 45nm open cell library.
\newblock \nolinkurl{https://si2.org/open-cell-library/}, 2018.

\bibitem{brglez1989combinational}
Franc Brglez, David Bryan, and Krzysztof Kozminski.
\newblock Combinational profiles of sequential benchmark circuits.
\newblock In {\em International Symposium on Circuits and Systems (ISCAS)},
  1989.

\bibitem{corno2000rt}
Fulvio Corno, Matteo~Sonza Reorda, and Giovanni Squillero.
\newblock {RT-level ITC'99 benchmarks and first ATPG results}.
\newblock {\em IEEE Design \& Test of computers}, 2000.

\bibitem{possignolo2017anubis}
Rafael~T Possignolo, Nursultan Kabylkas, and Jose Renau.
\newblock {ANUBIS}: A new benchmark for incremental synthesis.
\newblock In {\em International Workshop on Logic and Synthesis (IWLS)}, 2017.

\bibitem{albrecht2005iwls}
Christoph Albrecht.
\newblock {IWLS} 2005 benchmarks.
\newblock \nolinkurl{http://iwls.org/iwls2005/benchmarks.html}, 2005.

\bibitem{Fey/Lenssen/2019}
Matthias Fey and Jan~E. Lenssen.
\newblock Fast graph representation learning with {PyTorch Geometric}.
\newblock In {\em International Conference on Learning Representations Workshop
  (ICLR-W)}, 2019.

\bibitem{chang2003statistical}
Hongliang Chang and Sachin~S Sapatnekar.
\newblock Statistical timing analysis considering spatial correlations using a
  single {PERT}-like traversal.
\newblock In {\em International Conference On Computer Aided Design (ICCAD)},
  2003.

\bibitem{tehranipoor2010power}
Mohammad Tehranipoor and Kenneth~M Butler.
\newblock Power supply noise: A survey on effects and research.
\newblock {\em IEEE Design \& Test of Computers}, 2010.

\bibitem{chen1997power}
Howard~H Chen and David~D Ling.
\newblock Power supply noise analysis methodology for deep-submicron {VLSI}
  chip design.
\newblock In {\em Design Automation Conference (DAC)}, 1997.

\bibitem{Redhawk}
ANSYS.
\newblock {ANSYS RedHawk website}.
\newblock
  \nolinkurl{https://www.ansys.com/products/semiconductors/ansys-redhawk},
  2018.

\bibitem{nithin2010dynamic}
SK~Nithin, Gowrysankar Shanmugam, and Sreeram Chandrasekar.
\newblock Dynamic voltage ({IR}) drop analysis and design closure: Issues and
  challenges.
\newblock In {\em International Symposium on Quality Electronic Design
  (ISQED)}, 2010.

\bibitem{zhao2002hierarchical}
Min Zhao, Rajendran~V Panda, Sachin~S Sapatnekar, and David Blaauw.
\newblock Hierarchical analysis of power distribution networks.
\newblock {\em IEEE Transactions on Computer-Aided Design of Integrated
  Circuits and Systems (TCAD)}, 2002.

\bibitem{qian2005power}
Haifeng Qian, Sani~R Nassif, and Sachin~S Sapatnekar.
\newblock Power grid analysis using random walks.
\newblock {\em IEEE Transactions on Computer-Aided Design of Integrated
  Circuits and Systems (TCAD)}, 2005.

\bibitem{chen2001efficient}
Tsung-Hao Chen and Charlie Chung-Ping Chen.
\newblock Efficient large-scale power grid analysis based on preconditioned
  krylov-subspace iterative methods.
\newblock In {\em Design Automation Conference (DAC)}, 2001.

\bibitem{zhuo2008power}
Cheng Zhuo, Jiang Hu, Min Zhao, and Kangsheng Chen.
\newblock Power grid analysis and optimization using algebraic multigrid.
\newblock {\em IEEE Transactions on Computer-Aided Design of Integrated
  Circuits and Systems (TCAD)}, 2008.

\bibitem{dhotre2017identification}
Harshad Dhotre, Stephan Eggersgl{\"u}{\ss}, and Rolf Drechsler.
\newblock Identification of efficient clustering techniques for test power
  activity on the layout.
\newblock In {\em Asian Test Symposium (ATS)}, 2017.

\bibitem{batch_norm}
Sergey Ioffe and Christian Szegedy.
\newblock Batch normalization: accelerating deep network training by reducing
  internal covariate shift.
\newblock In {\em \textit{International Conference on Machine Learning}
  (\textit{ICML})}, 2015.

\bibitem{Adam}
Diederik~P. Kingma and Jimmy Ba.
\newblock Adam: A method for stochastic optimization.
\newblock In {\em \textit{International Conference for Learning
  Representations} (\textit{ICLR})}, 2015.

\bibitem{srivastava2014dropout}
Nitish Srivastava, Geoffrey Hinton, Alex Krizhevsky, Ilya Sutskever, and Ruslan
  Salakhutdinov.
\newblock Dropout: a simple way to prevent neural networks from overfitting.
\newblock {\em The Journal of Machine Learning Research (JMLR)}, 2014.

\bibitem{paszke2017automatic}
Adam Paszke, Sam Gross, Soumith Chintala, Gregory Chanan, Edward Yang, Zachary
  DeVito, Zeming Lin, Alban Desmaison, Luca Antiga, and Adam Lerer.
\newblock Automatic differentiation in {PyTorch}.
\newblock In {\em \textit{Advances in Neural Information Processing Systems
  Workshops} (\textit{NeurIPS-W})}, 2017.

\bibitem{kendall1948rank}
Maurice~George Kendall.
\newblock Rank correlation methods.
\newblock 1948.

\bibitem{tabrizi2019eh}
Aysa~Fakheri Tabrizi, Nima~Karimpour Darav, Logan Rakai, Ismail Bustany, Andrew
  Kennings, and Laleh Behjat.
\newblock Eh? predictor: A deep learning framework to identify detailed routing
  short violations from a placed netlist.
\newblock {\em IEEE Transactions on Computer-Aided Design of Integrated
  Circuits and Systems (TCAD)}, 2019.

\bibitem{tabrizi2018machine}
Aysa~Fakheri Tabrizi, Logan Rakai, Nima~Karimpour Darav, Ismail Bustany, Laleh
  Behjat, Shuchang Xu, and Andrew Kennings.
\newblock A machine learning framework to identify detailed routing short
  violations from a placed netlist.
\newblock In {\em Design Automation Conference (DAC)}, 2018.

\bibitem{ISPD2015}
Ismail~S. Bustany, David Chinnery, Joseph~R. Shinnerl, and Vladimir Yutsis.
\newblock {ISPD} 2015 benchmarks with fence regions and routing blockages for
  detailed-routing-driven placement.
\newblock In {\em \textit{ACM International Symposium on Physical Design}
  (\textit{ISPD})}, 2015.

\bibitem{macroPlace}
Chien-Hsiung Chiou, Chin-Hao Chang, Szu-To Chen, and Yao-Wen Chang.
\newblock Circular-contour-based obstacle-aware macro placement.
\newblock In {\em \textit{IEEE Asia and South Pacific Design Automation
  Conference} (\textit{ASP-DAC})}, 2016.

\bibitem{scikitlearn}
F.~Pedregosa, G.~Varoquaux, A.~Gramfort, V.~Michel, B.~Thirion, O.~Grisel,
  M.~Blondel, P.~Prettenhofer, R.~Weiss, V.~Dubourg, J.~Vanderplas, A.~Passos,
  D.~Cournapeau, M.~Brucher, M.~Perrot, and E.~Duchesnay.
\newblock Scikit-learn: Machine learning in {P}ython.
\newblock {\em Journal of Machine Learning Research (\textit{JMLR})}, 2011.

\bibitem{liu2015supervised}
Hung-Yi Liu.
\newblock {\em Supervised Design-Space Exploration}.
\newblock Columbia University, 2015.

\bibitem{mariani2012oscar}
Giovanni Mariani, Gianluca Palermo, Vittorio Zaccaria, and Cristina Silvano.
\newblock {OSCAR}: An optimization methodology exploiting spatial correlation
  in multicore design spaces.
\newblock {\em IEEE Transactions on Computer-Aided Design of Integrated
  Circuits and Systems (TCAD)}, 2012.

\bibitem{meng2016adaptive}
Pingfan Meng, Alric Althoff, Quentin Gautier, and Ryan Kastner.
\newblock Adaptive threshold non-pareto elimination: Re-thinking machine
  learning for system level design space exploration on {FPGA}s.
\newblock In {\em Design, Automation \& Test in Europe Conference \& Exhibition
  (\textit{DATE})}, 2016.

\bibitem{tiwary2006generation}
Saurabh~K Tiwary, Pragati~K Tiwary, and Rob~A Rutenbar.
\newblock Generation of yield-aware {Pareto} surfaces for hierarchical circuit
  design space exploration.
\newblock In {\em Design Automation Conference (\textit{DAC})}, 2006.

\bibitem{xydis2013meta}
Sotirios Xydis, Gianluca Palermo, Vittorio Zaccaria, and Cristina Silvano.
\newblock A meta-model assisted coprocessor synthesis framework for
  compiler/architecture parameters customization.
\newblock In {\em Design, Automation \& Test in Europe Conference \& Exhibition
  (\textit{DATE})}, 2013.

\bibitem{xydis2015spirit}
Sotirios Xydis, Gianluca Palermo, Vittorio Zaccaria, and Cristina Silvano.
\newblock {SPIRIT}: spectral-aware pareto iterative refinement optimization for
  supervised high-level synthesis.
\newblock {\em IEEE Transactions on Computer-Aided Design of Integrated
  Circuits and Systems (TCAD)}, 2015.

\bibitem{zuluaga2012smart}
Marcela Zuluaga, Andreas Krause, Peter Milder, and Markus P{\"u}schel.
\newblock {"Smart"} design space sampling to predict pareto-optimal solutions.
\newblock In {\em International Conference on Languages, Compilers, Tools and
  Theory for Embedded Systems (LCTES)}, 2012.

\bibitem{zuluaga2013active}
Marcela Zuluaga, Guillaume Sergent, Andreas Krause, and Markus P{\"u}schel.
\newblock Active learning for multi-objective optimization.
\newblock In {\em International Conference on Machine Learning
  (\textit{ICML})}, 2013.

\bibitem{lee2015icml}
Dong-Hyun Lee.
\newblock Pseudo-label: The simple and efficient semi-supervised learning
  method for deep neural networks.
\newblock In {\em International Conference on Machine Learning Workshop
  (ICML-W)}, 2015.

\bibitem{zimmer20190}
Brian Zimmer, Rangharajan Venkatesan, Yakun~Sophia Shao, Jason Clemons, Matthew
  Fojtik, Nan Jiang, Ben Keller, Alicia Klinefelter, Nathaniel Pinckney,
  Priyanka Raina, et~al.
\newblock A 0.11 {pJ/Op}, 0.32-128 {TOPS}, scalable multi-chip-module-based
  deep neural network accelerator with ground-reference signaling in 16nm.
\newblock In {\em IEEE Symposium on VLSI Circuits (\textit{VLSI})}, 2019.

\end{thebibliography}
